\documentclass[10pt]{article} 

\usepackage[accepted]{tmlr}

\usepackage{amsmath,amsfonts,bm}

\def\eqref#1{equation~\ref{#1}}

\def\1{\bm{1}}

\DeclareMathAlphabet{\mathsfit}{\encodingdefault}{\sfdefault}{m}{sl}
\SetMathAlphabet{\mathsfit}{bold}{\encodingdefault}{\sfdefault}{bx}{n}

\usepackage[hidelinks]{hyperref}
\usepackage{url}

\usepackage{pifont}
\usepackage{dsfont}
\usepackage{amsthm}
\usepackage{comment}
\usepackage{amssymb}
\usepackage{aliascnt}
\usepackage{pgfplots}
\usepackage{booktabs}
\usepackage{multirow}
\usepackage{enumitem}
\usepackage{mathtools}
\usepackage{subcaption}
\usepackage[table,svgnames]{xcolor}
\usepackage[capitalize,noabbrev,nameinlink]{cleveref}

\newcommand{\cmark}{\ding{51}} 
\newcommand{\xmark}{\ding{55}} 

\definecolor{tomato}{rgb}{1.0, 0.39, 0.28}
\definecolor{mediumseagreen}{rgb}{0.24, 0.7, 0.44}
\definecolor{dodgerblue}{rgb}{0.12, 0.56, 1.0}

\theoremstyle{plain}

\newaliascnt{proposition}{theorem}
\newtheorem{proposition}[proposition]{Proposition}
\aliascntresetthe{proposition}

\newaliascnt{lemma}{theorem}
\newtheorem{lemma}[lemma]{Lemma}
\aliascntresetthe{lemma}

\newaliascnt{corollary}{theorem}
\newtheorem{corollary}[corollary]{Corollary}
\aliascntresetthe{corollary}

\theoremstyle{definition}
\newaliascnt{definition}{theorem}

\aliascntresetthe{definition}

\newaliascnt{assumption}{theorem}

\aliascntresetthe{assumption}

\theoremstyle{remark}
\newaliascnt{remark}{theorem}
\newtheorem{remark}[remark]{Remark}
\aliascntresetthe{remark}

\usetikzlibrary{calc}

\newcommand{\mapx}[1]{0.2569 + (log10(#1) + 3) / 4 * (0.9378 - 0.2569)}
\newcommand{\mapy}[1]{0.3812 + (log10(#1) + 2) / 4 * (0.9653 - 0.3812)}

\newcommand{\circled}[2][black]{\tikz[baseline=(c.base)]\node[draw=#1, fill=#1, text=white, circle, inner sep=1pt, font=\scriptsize\bfseries] (c) {#2};}

\newenvironment{labeledmap}[2][]{%
  \begin{tikzpicture}
    \node[anchor=south west, inner sep=0] (map) {\includegraphics[#1]{#2}};
    \begin{scope}[x={(map.south east)}, y={(map.north west)}]
}{%
    \end{scope}
  \end{tikzpicture}}

\newcommand{\maprect}[6][black]{%
  \pgfmathsetmacro{\xa}{\mapx{#3}}\pgfmathsetmacro{\xb}{\mapx{#4}}%
  \pgfmathsetmacro{\ya}{\mapy{#5}}\pgfmathsetmacro{\yb}{\mapy{#6}}%
  \draw[#1, thick] (\xa, \ya) rectangle (\xb, \yb);
  \node[draw=#1, fill=#1, text=white, circle, inner sep=1pt, font=\scriptsize\bfseries, anchor=south west] at (\xa, \yb) {#2};}

\makeatletter
\newcommand{\defregion}[3]{\@namedef{region@color@#1}{#2}\@namedef{region@box@#1}{#3}}
\defregion{1}{Orange}{0.1/10.0/40/100}
\defregion{2}{LimeGreen}{0.001/0.1/0.01/1.0}
\defregion{3}{DeepSkyBlue}{0.5/10.0/1/100}
\defregion{4}{Purple}{0.001/0.01/2/100}
\defregion{5}{RoyalBlue}{0.01/0.04/15/100}
\newcommand{\regicon}[1]{\circled[\@nameuse{region@color@#1}]{#1}}
\newcommand{\regrect}[2][]{%
  \if\relax\detokenize{#1}\relax
    \edef\region@box{\@nameuse{region@box@#2}}%
  \else
    \edef\region@box{#1}%
  \fi
  \expandafter\region@rect\region@box\@nil{#2}}
\def\region@rect#1/#2/#3/#4\@nil#5{\maprect[\@nameuse{region@color@#5}]{#5}{#1}{#2}{#3}{#4}}
\makeatother
\newcommand{\regone}{\regicon{1}}
\newcommand{\regtwo}{\regicon{2}}
\newcommand{\regthree}{\regicon{3}}
\newcommand{\regfour}{\regicon{4}}
\newcommand{\regfive}{\regicon{5}}

\newcommand{\rowlabel}[1]{\begin{minipage}[c]{0.05\linewidth}\centering\rotatebox{90}{#1}\end{minipage}}

\title{Stochastic Signed Distance Processes}

\author{\name Hiroki Sakuma \email sakuma.h.2b63@m.isct.ac.jp\\
  \addr Department of Systems and Control Engineering\\Institute of Science Tokyo
  \AND
  \name Masatoshi Okutomi \email mxo@ctrl.titech.ac.jp\\
  \addr Department of Systems and Control Engineering\\Institute of Science Tokyo
}

\def\month{11}  
\def\year{2026} 
\def\openreview{\url{https://openreview.net/forum?id=JSftvBX0lg}} 

\begin{document}

\maketitle

\begin{abstract}
Multi-view surface reconstruction is a core problem in computer vision.
One prominent line of work represents the surface implicitly as a signed distance field (SDF), optimizing it based on the photometric loss between rendered and observed pixel colors. 
These approaches typically employ SDF-based volume rendering to obtain a differentiable relaxation of discontinuous visibility along rays, thereby reducing reliance on silhouette supervision.
In this paper, we reformulate SDF-based volume rendering as probabilistic surface rendering, where each pixel color is modeled as a mixture distribution induced by the random first ray--surface intersection.
To this end, we introduce \textit{Stochastic Signed Distance Processes} (SSDP), which model the SDF along each ray as a stochastic process, inducing a first-passage-time distribution for each ray.
We then derive the first-passage probability for each sampling interval based on Bayesian filtering, together with its practical approximation for parallel rendering.
We further show that NeuS, an existing SDF-based volume rendering method, arises as a special case of our formulation.
Experiments on the DTU and MobileBrick datasets demonstrate that our method outperforms baselines in both surface reconstruction and uncertainty quantification, supporting the effectiveness of our first-passage formulation.
Our code is available at \url{https://github.com/skmhrk1209/SSDP}.
\end{abstract}

\section{Introduction}
\label{sec:introduction}

Reconstructing a surface from multi-view images is a core problem in computer vision.
Its applications range from content creation for films and games to digital twins for robotics and autonomous driving.
Among many surface representations, signed distance fields (SDFs), a special case of level-set functions, have become a natural choice for this task.
They provide a continuous, resolution-independent implicit representation that defines the surface as the zero-level set, support direct distance queries without explicit nearest-point search, and naturally handle topology changes during optimization \citep{levelset}.

A direct way to optimize an implicit surface via rendering is to find the first intersection for each ray and compute the gradients of the photometric loss with respect to the intersection via implicit differentiation \citep{idr}.
While this approach is straightforward, it involves discontinuous visibility along rays: gradients are propagated only through intersecting rays.
Therefore, direct surface rendering often relies on additional visibility cues, such as silhouette supervision with foreground masks, to facilitate optimization.

To obtain a differentiable relaxation of discontinuous visibility along rays, NeuS \citep{neus} and VolSDF \citep{volsdf} developed SDF-based volume rendering, where SDF values are mapped to \textit{attenuation coefficients}, enabling direct application of volume rendering to the SDF and its optimization without silhouette supervision.
\textit{Objects as Volumes} \citep{oav} further gives a stochastic representation of opaque solids based on \textit{exponential transport} and discusses its relationship to NeuS and VolSDF. 
However, exponential transport is justified only when the \textit{hazard rate} at a certain time can be summarized by a local attenuation coefficient that is independent of the survival history. 
This assumption is valid for the Boolean--Poisson model \citep{boolean_poisson}, where particles are distributed independently. 
In contrast, general opaque solids exhibit spatially correlated geometry, so the first-passage-time density at a certain time can depend on the survival history and cannot be reduced to a local attenuation coefficient.
Since NeuS and VolSDF also map SDF values to local attenuation coefficients, their rendering formulations admit the same exponential-transport interpretation and are therefore subject to the same limitation.

To address the limitation of exponential transport for spatially correlated geometry, we reformulate SDF-based volume rendering as probabilistic surface rendering, where each pixel color is modeled as a mixture distribution induced by the random first ray--surface intersection.
To this end, we introduce \textit{Stochastic Signed Distance Processes} (SSDP), which model the SDF along each ray as a stochastic process, inducing a first-passage-time distribution for each ray.
Rather than specifying the hazard rate as a local attenuation coefficient, we directly derive the first-passage probability for each sampling interval, propagating the survival-conditioned distribution via Bayesian filtering.
Furthermore, for practical applications, we introduce a parallelizable approximation that eliminates the reliance on sequential Bayesian filtering, achieving optimization efficiency comparable to that of SDF-based volume rendering.

Our contributions are summarized as follows:
\begin{itemize}
\item We reformulate SDF-based volume rendering as probabilistic surface rendering, where each pixel color is modeled as a mixture distribution induced by the random first ray--surface intersection.
\item We introduce \textit{Stochastic Signed Distance Processes} (SSDP), which model the SDF along each ray as a stochastic process, inducing a first-passage-time distribution for each ray.
\item We derive the first-passage probability for each sampling interval based on Bayesian filtering, together with its practical approximation for parallel rendering. 
\item We show that NeuS arises as a special case of our formulation.
\item We demonstrate the effectiveness of our method in both surface reconstruction and uncertainty quantification on the DTU and MobileBrick datasets.
\end{itemize}

\section{Related Work}
\label{sec:related_work}

Recent progress in neural surface reconstruction has been driven by the widespread adoption of implicit 3D representations following NeRF \citep{nerf}.
Common implicit representations include signed distance fields (SDFs) \citep{deepsdf} and occupancy networks \citep{occnet}.
We focus on SDFs for their analytical tractability as continuous implicit scalar fields.

Two common approaches for rendering SDFs are surface rendering and SDF-based volume rendering.
Surface rendering is a direct way to optimize an SDF, where the ray--surface intersection for each ray is found via sphere tracing \citep{sdfdiff,dist} or root finding.
The gradients of the photometric loss with respect to the SDF are propagated via the implicitly differentiable intersection \citep{idr,dvr}.
However, discontinuous visibility yields very sparse gradients, making optimization prone to poor local minima.
Hence, it is common to employ additional silhouette supervision with foreground masks.

To tackle this issue, NeuS \citep{neus} and VolSDF \citep{volsdf} introduced SDF-based volume rendering, where SDF values are mapped to \textit{attenuation coefficients}, enabling direct application of volume rendering \citep{nerf} to the SDF and its optimization without silhouette supervision.
UNIS \citep{unis} systematically studies mapping functions from SDF values to attenuation coefficients.
\textit{Objects as Volumes} (OaV) \citep{oav} is closely related to our work because it also provides a stochastic foundation for SDF-based volume rendering methods such as NeuS and VolSDF. 
OaV builds on exponential transport and derives conditions under which stochastic opaque solids can be represented by exponential transport. 
A central assumption is that the binary indicator process along each ray, representing whether a point is inside or outside the solid, forms a continuous-time Markov chain. 
This assumption plays the role of the independent-collision structure in the Boolean--Poisson model \citep{boolean_poisson}, leading to an exponential-transport formulation in terms of a local attenuation coefficient.
While the resulting formulation is physically grounded and satisfies reciprocity and reversibility, this Markovian assumption does not generally hold for opaque solids with spatially correlated geometry \citep{non_exponential}.
In contrast, we model the SDF along each ray as a stochastic process and derive, for each sampling interval, the first-passage probability induced by the ray-wise stochastic dynamics.
Thus, OaV and SSDP generalize SDF-based volume rendering in different directions: OaV generalizes from the perspective of volumetric light transport, while SSDP generalizes from the perspective of first-passage times for stochastic processes.

In the context of uncertainty visualization, \cite{neg_abs_approx} derived the probability that an exponentially correlated Gaussian field first crosses an isovalue within each sampling interval along a ray.
Their resulting first-passage probability coincides with ours under the negative-absorbing approximation introduced in \cref{sec:method/first_passage_probability/negative_absorbing_approximation}, but the simplification of the survival event is left implicit in their derivation, as made precise in \cref{rem:negative_absorbing_approximation}.
We instead start from the full Bayesian formulation conditioned on the survival history, make explicit the single assumption under which the first-passage probability can be computed from the interval endpoints alone, and quantify the approximation error in \cref{sec:approximation_error_validation}.
Moreover, their scope is confined to visualizing the uncertainty of a given field, whereas we learn the stochastic process itself from multi-view images for surface reconstruction.

\section{Method}
\label{sec:method}

\begin{figure}[t]
\centering
\includegraphics[width=1.0\linewidth]{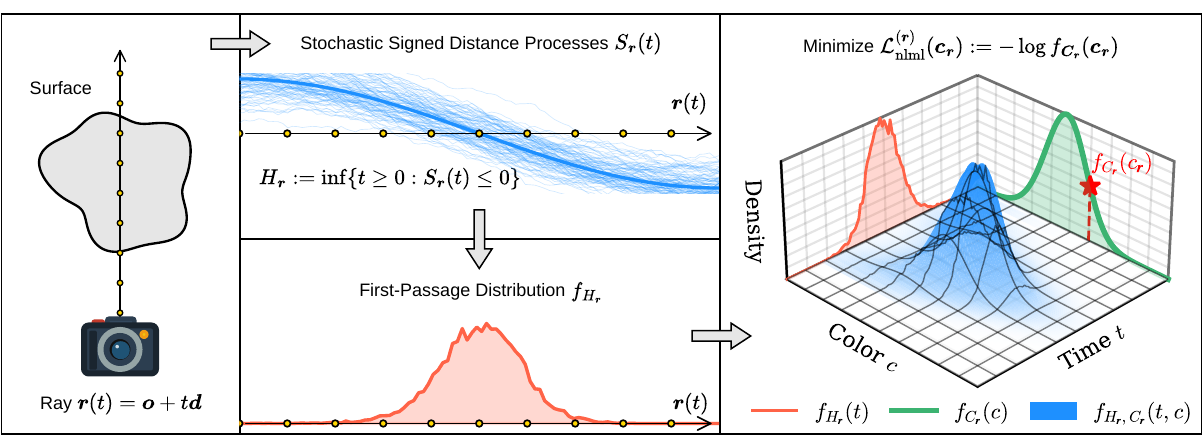}
\caption{Overview of our method. We model the SDF along each ray as a stochastic process $S_{\bm{r}}(t)$, which we call a Stochastic Signed Distance Process (SSDP), inducing a first-passage-time distribution $f_{H_{\bm{r}}}$. $S_{\bm{r}}(t)$ is optimized by minimizing the negative log marginal likelihood $\mathcal{L}_{\mathrm{nlml}}^{(\bm{r})}(\bm{c}_{\bm{r}}) \coloneq -\log{f_{\bm{C}_{\bm{r}}}(\bm{c}_{\bm{r}})}$ of the observed color $\bm{c}_{\bm{r}}$, where $f_{\bm{C}_{\bm{r}}}$ is obtained by marginalizing out the first-passage time $H_{\bm{r}}$ from the joint density $f_{H_{\bm{r}}, \bm{C}_{\bm{r}}}$.}
\label{fig:overview}
\end{figure}

Following prior works \citep{idr,neus}, we represent the geometry and the appearance of a scene by an SDF and a color field, respectively.
\cref{fig:overview} provides a high-level overview of the main components of our formulation: the ray-wise stochastic process, the first-passage-time distribution induced by this process, and the training objective based on this distribution.
We first revisit SDF-based volume rendering introduced in NeuS \citep{neus} (\cref{sec:method/preliminaries}) and then introduce probabilistic surface rendering, where each pixel color is modeled as a mixture distribution induced by the random first ray--surface intersection (\cref{sec:method/probabilistic_surface_rendering}).
To this end, we introduce Stochastic Signed Distance Processes (SSDP), which model the SDF along each ray as a stochastic process (\cref{sec:method/stochastic_signed_distance_processes}).
We then derive the first-passage probability for each sampling interval via Bayesian filtering, together with a practical approximation for parallel rendering (\cref{sec:method/first_passage_probability}).
We further show that NeuS arises as a special case of our formulation (\cref{sec:method/relationship_to_prior_works}).
We finally introduce a probabilistic photometric loss to optimize SSDPs (\cref{sec:method/loss_functions}).

\subsection{Preliminaries}
\label{sec:method/preliminaries}

\paragraph{Notation.}

For a random variable $X$, let $f_X$ and $F_X$ denote its probability density function (PDF) and cumulative distribution function (CDF), respectively.
For an event $\mathcal{A}$, let $P(\mathcal{A})$ denote its probability.
We use lowercase letters for realizations of random variables, e.g., $x$ for $X$.
When the underlying random variable $X$ is clear from the context, we use $P(\cdot \mid x)$ as shorthand for $P(\cdot \mid X = x)$.
Let $\varphi(\cdot; \mu, \sigma^{2})$ and $\Phi(\cdot; \mu, \sigma^{2})$ denote the PDF and CDF of the Gaussian distribution $\mathcal{N}(\mu, \sigma^{2})$ with mean $\mu$ and variance $\sigma^{2}$, respectively.
Similarly, let $\varphi_{+}(\cdot; \mu, \sigma^{2})$ and $\Phi_{+}(\cdot; \mu, \sigma^{2})$ denote the PDF and CDF of the truncated Gaussian distribution $\mathcal{N}_{+}(\mu, \sigma^{2})$ on $[0, \infty)$, respectively.
Let $\mathbb{E}[\cdot]$ and $\mathbb{V}[\cdot]$ denote the expectation and variance operators, respectively.

\paragraph{Signed distance field (SDF).}

An SDF is a surface representation defined by a continuous function $\mu: \mathbb{R}^{3} \rightarrow \mathbb{R}$ that maps a spatial position $\bm{x} \in \mathbb{R}^{3}$ to the signed Euclidean distance to the surface.
Let $\Omega \subset \mathbb{R}^3$ denote a closed set representing an object and $\partial \Omega$ denote its surface.
The SDF $\mu$ for $\Omega$ is defined as follows:
\begin{align}
\label{eq:sdf}
\mu(\bm{x}) \coloneq
\begin{cases}
- d(\bm{x}, \partial \Omega) & \text{if} \ \bm{x} \in \Omega , \\
\phantom{-} d(\bm{x}, \partial \Omega) & \text{otherwise} ,
\end{cases}
\end{align}
where $d(\bm{x}, \partial \Omega) \coloneq \inf_{\bm{y} \in \partial \Omega} \|\bm{x} - \bm{y}\|_{2}$ denotes the Euclidean distance from $\bm{x}$ to $\partial \Omega$.
The surface is represented by the zero-level set $\partial \Omega = \{\bm{x} \in \mathbb{R}^{3} \mid \mu(\bm{x}) = 0\}$.
A key property of SDFs is that they satisfy the Eikonal equation \citep{eikonal_equation} almost everywhere:
\begin{align}
\label{eq:eikonal_equation}
\|\nabla \mu(\bm{x})\|_{2} 
& \stackrel{\text{a.e.}}{=} 1 .
\end{align}

\paragraph{NeRF.}

NeRF \citep{nerf} represents a scene with neural density and radiance fields, both of which are parameterized by MLPs.
Given a ray $\bm{r}(t) = \bm{o} + t \bm{d}$ with an origin $\bm{o} \in \mathbb{R}^{3}$ and a direction $\bm{d} \in \mathbb{S}^{2}$, the \textit{expected} color along the ray is given by the following volume rendering:
\begin{align}
\label{eq:color_expectation_nerf_continuous}
\bar{\bm{c}}(\bm{r}) 
& = \int_{t_{n}}^{t_{f}} T(t) \sigma(\bm{r}(t)) \bm{c}(\bm{r}(t), \bm{d}) \, \mathrm{d} t , \quad
T(t) 
= \exp \left ( -\int_{t_{n}}^{t} \sigma(\bm{r}(s)) \, \mathrm{d} s \right ) ,
\end{align}
where $[t_{n}, t_{f}]$ denotes near and far bounds, $\sigma(\bm{r}(t))$ denotes the attenuation coefficient, $\bm{c}(\bm{r}(t), \bm{d})$ denotes the view-dependent color, and $T(t)$ denotes the accumulated transmittance along the ray, i.e., the probability that the ray has not terminated before $t$.
This integral is approximated with discrete samples $\{t_{i}\}_{i=0}^{N}$ as:
\begin{align}
\label{eq:color_expectation_nerf_discrete}
\bar{\bm{c}}(\bm{r}) 
& \approx \sum_{i=0}^{N-1} T_{i} o_{i} \bm{c}(\bm{r}(t_{i}), \bm{d}) , \quad
T_{i} 
= \prod_{j=0}^{i-1} (1 - o_{j}) ,
\end{align}
where $o_{i}$ denotes the \textit{opacity} for the $i$-th interval $[t_{i}, t_{i+1}]$:
\begin{align}
\label{eq:opacity_nerf}
o_{i} 
& = 1 - \exp \left (-\int_{t_{i}}^{t_{i+1}} \sigma(\bm{r}(t)) \, \mathrm{d} t \right ) \nonumber \\
& \approx 1 - \exp(-\sigma(\bm{r}(t_{i})) \Delta t_{i}) .
\end{align}

\paragraph{NeuS.}

NeuS \citep{neus} introduced a formulation for volume rendering of implicit surfaces represented as the zero-level set of an SDF.
The attenuation coefficient $\sigma$ in \cref{eq:color_expectation_nerf_continuous} is not directly parameterized by an MLP as in NeRF but derived from the SDF $\mu$ as follows:
\begin{align}
\label{eq:density_neus}
\sigma(t) 
& = \left [ -\frac{\mathrm{d}}{\mathrm{d} t} \log \varsigma_{s}(\mu(\bm{r}(t))) \right ]_{+} ,
\end{align}
where $[\cdot]_{+} \coloneq \max(\cdot, 0)$ and $\varsigma_{s}(x) \coloneq (1 + \exp(-x / s))^{-1}$ denotes the logistic sigmoid, which is the CDF of a logistic distribution with location $0$ and scale $s$.
Then, the opacity $o_{i}$ is given by assuming $\mu(\bm{r}(t))$ is monotonic with respect to $t$ within each interval $[t_{i}, t_{i+1}]$:
\begin{align}
\label{eq:opacity_neus}
o_{i} 
& = 1 - \exp \left (-\int_{t_{i}}^{t_{i+1}} \sigma(t) \, \mathrm{d} t \right ) \\
& = \left [ \frac{\varsigma_{s}(\mu(\bm{r}(t_{i}))) - \varsigma_{s}(\mu(\bm{r}(t_{i+1})))}{\varsigma_{s}(\mu(\bm{r}(t_{i})))} \right ]_{+} .
\end{align}

\subsection{Probabilistic Surface Rendering}
\label{sec:method/probabilistic_surface_rendering}

We first define general surface rendering from a probabilistic perspective.
Given a ray $\bm{r}(t) = \bm{o} + t \bm{d}$ with an origin $\bm{o} \in \mathbb{R}^{3}$ and a direction $\bm{d} \in \mathbb{S}^{2}$, we model the SDF along the ray as a stochastic process denoted by $S_{\bm{r}}(t)$, as explained in \cref{sec:method/stochastic_signed_distance_processes}.
Then, we define a random variable $H_{\bm{r}}$ for the first-passage time, i.e., the time when the ray first intersects the surface, with the convention that $\inf \varnothing = \infty$ when no such $t$ exists:
\begin{align}
\label{eq:first_passage_time}
H_{\bm{r}} 
& \coloneq \inf\{t \ge 0: S_{\bm{r}}(t) \le 0\} .
\end{align}
The marginal distribution of the pixel color
$\bm{C}_{\bm{r}} \in \mathbb{R}^{3}$ is given by the following mixture distribution:
\begin{align}
\label{eq:color_pdf_marginal_continuous}
f_{\bm{C}_{\bm{r}}}(\bm{c}_{\bm{r}}) 
& = \mathbb{E}_{H_{\bm{r}}}[f_{\bm{C}_{\bm{r}}\mid H_{\bm{r}}}(\bm{c}_{\bm{r}}\mid H_{\bm{r}})] .
\end{align}
Then, we approximate the above expectation with discrete samples $\{t_{i}\}_{i=0}^{N}$ along the ray, yielding:
\begin{align}
\label{eq:color_pdf_marginal_discrete}
f_{\bm{C}_{\bm{r}}}(\bm{c}_{\bm{r}}) 
& \approx \sum_{i=0}^{N-1} \underbrace{f_{\bm{C}_{\bm{r}} \mid H_{\bm{r}}}(\bm{c}_{\bm{r}} \mid t_{i}) P(t_{i} < H_{\bm{r}} \le t_{i+1})}_{\text{Foreground term}} 
+ \underbrace{f_{\bm{C}_{\bm{r}} \mid H_{\bm{r}}}(\bm{c}_{\bm{r}} \mid \infty) P(H_{\bm{r}} > t_{N})}_{\text{Background term}} ,
\end{align}
where $P(t_{i} < H_{\bm{r}} \le t_{i+1})$ represents the first-passage probability for the $i$-th interval, while $P(H_{\bm{r}} > t_{N})$ represents the probability that the ray does not intersect the surface over $(t_{0}, t_{N}]$.

For the foreground term in \cref{eq:color_pdf_marginal_discrete}, following prior works \citep{idr,neus}, which employ $L_{1}$ loss as a photometric loss, we assume $\bm{C}_{\bm{r}}^{(k)} \mid H_{\bm{r}} = t \sim \mathrm{Laplace}(\hat{\bm{c}}_{\bm{r}}^{(k)}(t), b)$, where $k \in \{\mathrm{R}, \mathrm{G}, \mathrm{B}\}$ denotes an RGB channel and $\mathrm{Laplace}(\hat{\bm{c}}_{\bm{r}}^{(k)}(t), b)$ denotes a Laplace distribution with a location $\hat{\bm{c}}_{\bm{r}}^{(k)}(t)$ and a homogeneous scale $b$.
Note that $\hat{\bm{c}}_{\bm{r}}^{(k)}(t) \in \mathbb{R}$ denotes the $k$-th component of $\hat{\bm{c}}_{\bm{r}}(t) \in \mathbb{R}^{3}$.
For simplicity, we assume conditional independence across the RGB channels, yielding:
\begin{align}
\label{eq:color_pdf_conditional}
f_{\bm{C}_{\bm{r}} \mid H_{\bm{r}}}(\bm{c}_{\bm{r}} \mid t) 
& = \prod_{k \in \{\mathrm{R}, \mathrm{G}, \mathrm{B}\}} \frac{1}{2 b} \exp \left ( -\frac{|\bm{c}_{\bm{r}}^{(k)} - \hat{\bm{c}}_{\bm{r}}^{(k)}(t)|}{b} \right ) .
\end{align}
We parameterize $\hat{\bm{c}}_{\bm{r}}(t)$ with a neural field as $\hat{\bm{c}}_{\bm{r}}(t) \coloneq \hat{\bm{c}}_{\bm{\psi}}(\bm{r}(t), \bm{d})$ while fixing $b$ as a constant value.
For the background term in \cref{eq:color_pdf_marginal_discrete}, following NeuS, we model the unbounded background by a separate NeRF.
For notational uniformity, we denote the volume-rendered color for unbounded space $H_{\bm{r}} > t_{N}$ by $\hat{\bm{c}}_{\bm{r}}(\infty)$, so that the background term can be written in the same form as \cref{eq:color_pdf_conditional}.

In the following sections, we focus on how to derive the first-passage probability $P(t_{i} < H_{\bm{r}} \le t_{i+1})$ based on the stochastic process $S_{\bm{r}}(t)$. 
We omit the subscript $\bm{r}$ when it does not cause ambiguity.

\subsection{Stochastic Signed Distance Processes}
\label{sec:method/stochastic_signed_distance_processes}

To make surface rendering probabilistic, we model the SDF along each ray as a stochastic process.
A natural alternative is to directly model the SDF as a random field over $\mathbb{R}^{3}$, but such a formulation typically couples all queried points through a dense covariance matrix, making rendering prohibitively expensive.
For example, the naive rendering cost for $M$ rays and $N$ samples per ray scales as $O((MN)^{3})$ for a dense Gaussian process.
On the other hand, our per-ray formulation leads to an $O(MN)$ rendering cost and enables parallel rendering across rays, which is crucial for practical applications.

Given a ray $\bm{r}(t)$, we define a \textit{Stochastic Signed Distance Process (SSDP)} as $S_{\bm{r}}(t) \coloneq \mu_{\bm{r}}(t) + R_{\bm{r}}(t)$, where $\mu_{\bm{r}}(t) \coloneq \mu_{\bm{\theta}}(\bm{r}(t))$ denotes the deterministic \textit{mean field} optimized with the Eikonal regularizer \citep{eikonal_loss} to approximate an SDF and $R_{\bm{r}}(t)$ denotes the stochastic \textit{residual process} modeled as a time-inhomogeneous Ornstein--Uhlenbeck (OU) process \citep{ou_process}:
\begin{align}
\label{eq:ou_sde}
\mathrm{d} R_{\bm{r}}(t) 
& = -\kappa_{\bm{r}}(t) R_{\bm{r}}(t) \, \mathrm{d} t + \tau_{\bm{r}}(t) \, \mathrm{d} W_{\bm{r}}(t) ,
\end{align}
where $\kappa_{\bm{r}}(t) > 0$, $\tau_{\bm{r}}(t) > 0$, and $W_{\bm{r}}(t)$ denotes the Wiener process.
We parameterize $\kappa_{\bm{r}}(t)$ and $\tau_{\bm{r}}(t)$ with a view-dependent neural field as $\kappa_{\bm{r}}(t) \coloneq \kappa_{\bm{\phi}}(\bm{r}(t), \bm{d})$ and $\tau_{\bm{r}}(t) \coloneq \tau_{\bm{\phi}}(\bm{r}(t), \bm{d})$ to model ray-wise stochastic dynamics.
The solution of the above stochastic differential equation for any $t \ge t_{i}$ is given by:
\begin{align}
\label{eq:ou_solution}
R_{\bm{r}}(t) 
& = \Psi_{\bm{r}}(t_{i}, t) (R_{\bm{r}}(t_{i}) + M_{\bm{r}}(t)) ,
\end{align}
where $\Psi_{\bm{r}}(s, t) \coloneq \exp(-\int_{s}^{t} \kappa_{\bm{r}}(u) \, \mathrm{d} u)$ and $M_{\bm{r}}(t) \coloneq \int_{t_{i}}^{t} \Psi_{\bm{r}}(s, t_{i}) \tau_{\bm{r}}(s) \, \mathrm{d} W_{\bm{r}}(s)$ is a continuous local martingale.
Taking the expectation of \cref{eq:ou_solution} yields $\mathbb{E}[R_{\bm{r}}(t)] = \Psi_{\bm{r}}(t_{0}, t) \mathbb{E}[R_{\bm{r}}(t_{0})]$.
Letting $R_{\bm{r}}(t_{0})$ be independent of $W_{\bm{r}}(t)$ and follow any zero-mean distribution yields $\mathbb{E}[S_{\bm{r}}(t)] = \mu_{\bm{r}}(t)$ for all $\bm{r}$ and $t$, indicating that $\mathbb{E}[S_{\bm{r}}(t)]$ is consistent across viewpoints.
This is essential to reconstruct the mean field $\mu_{\bm{\theta}}$ from multi-view observations.
In contrast, the stochastic residual $R_{\bm{r}}(t)$, and hence the distribution of $H_{\bm{r}}$, is ray-wise and depends on the ray $\bm{r}$.
The first-passage-time distribution thus represents the uncertainty of the first ray--surface intersection conditioned on each ray, rather than that of the 3D geometry itself shared across viewpoints.
Unless otherwise stated, we instantiate the initial distribution as $R_{\bm{r}}(t_{0}) \sim \mathcal{N}(0, \sigma_{0}^{2})$, where the initial variance $\sigma_{0}^{2}$ is parameterized by a single learnable scalar shared across rays.
Under this instantiation, $S_{\bm{r}}(t)$ is a Gaussian process, enabling the closed-form computations in \cref{sec:method/first_passage_probability}.

Here, the transition kernel of $S_{\bm{r}}(t)$ is given by:
\begin{align}
\label{eq:ou_transition_kernel_continuous}
S_{\bm{r}}(t) \mid S_{\bm{r}}(t_{i}) = s_{i} 
& \sim \mathcal{N}(\alpha_{i}(t) s_{i} + \beta_{i}(t), \gamma_{i}(t)) ,
\end{align}
where $\alpha_{i}(t) = \Psi_{\bm{r}}(t_{i}, t)$, $\beta_{i}(t) = \mu_{\bm{r}}(t) - \alpha_{i}(t) \mu_{\bm{r}}(t_{i})$ and $\gamma_{i}(t) = \int_{t_{i}}^{t} \Psi_{\bm{r}}(s, t)^{2} \tau_{\bm{r}}(s)^{2} \, \mathrm{d} s$.
In practice, we assume piecewise-constant coefficients
$\kappa_{\bm{r}}(t) \equiv \kappa_{i}$ and $\tau_{\bm{r}}(t) \equiv \tau_{i}$ on each interval $[t_{i}, t_{i+1})$. Letting $S_{i} \coloneq S_{\bm{r}}(t_{i})$, $\mu_{i} \coloneq \mu_{\bm{r}}(t_{i})$, and $\Delta t_{i} \coloneq t_{i+1} - t_{i}$, the transition kernel of $S_{i}$ is given by:
\begin{align}
\label{eq:ou_transition_kernel_discrete}
S_{i+1} \mid S_{i} = s_{i} 
& \sim \mathcal{N}(\alpha_{i} s_{i} + \beta_{i}, \gamma_{i}) ,
\end{align}
where $\alpha_{i} = \exp(-\kappa_{i} \Delta t_{i})$, $\beta_{i} = \mu_{i+1} - \alpha_{i} \mu_{i}$, and $\gamma_{i} = \tau_{i}^{2} (1 - \alpha_{i}^{2}) / 2 \kappa_{i}$.

\subsection{Derivation of First-Passage Probability $P(t_{i} < H \le t_{i+1})$}
\label{sec:method/first_passage_probability}

In this subsection, we derive the first-passage probability $P(t_{i} < H \le t_{i+1})$ for the $i$-th interval based on the SSDP $S(t)$.
We first consider the following events:
\begin{align}
\label{eq:survival_event}
\mathcal{A}_{i} 
& \coloneq \left \{ \inf_{t \in (0, t_{i}]} S(t) > 0 \right \}, \quad
\mathcal{B}_{i} 
\coloneq \left \{ \inf_{t \in (t_{i}, t_{i+1}]} S(t) \le 0 \right \} ,
\end{align}
where $\mathcal{A}_{i}$ denotes the event that the ray has not intersected the surface by $t_{i}$ and $\mathcal{B}_{i}$ denotes the event that the ray intersects the surface at least once in $(t_{i}, t_{i+1}]$.
Conditioning $P(t_{i} < H \le t_{i+1})$ on $\mathcal{A}_{i}$ yields:
\begin{align}
\label{eq:first_passage_probability}
P(t_{i} < H \le t_{i+1}) 
& = P(\mathcal{B}_{i} \mid \mathcal{A}_{i}) P(\mathcal{A}_{i}) , \\
\label{eq:survival_probability}
P(\mathcal{A}_{i}) 
& = P(\mathcal{A}_{0}) \prod_{j=0}^{i-1} (1 - P(\mathcal{B}_{j} \mid \mathcal{A}_{j})) .
\end{align}
Note that $P(\mathcal{B}_{i} \mid \mathcal{A}_{i})$ can be equated with the opacity $o_{i}$ in \cref{eq:opacity_nerf,eq:opacity_neus}.
Since $S(t)$ is Markovian, $P(\mathcal{B}_{i} \mid \mathcal{A}_{i})$ can be further decomposed into $\mathcal{A}_{i}$-dependent $f_{S_{i} \mid \mathcal{A}_{i}}$ and $\mathcal{A}_{i}$-independent $P(\mathcal{B}_{i} \mid S_{i})$: 
\begin{align}
\label{eq:opacity_ssdp}
P(\mathcal{B}_{i} \mid \mathcal{A}_{i}) 
& = \int_{0}^{\infty} P(\mathcal{B}_{i} \mid s_{i}, \mathcal{A}_{i}) f_{S_{i} \mid \mathcal{A}_{i}}(s_{i}) \, \mathrm{d} s_{i} \nonumber \\
& = \int_{0}^{\infty} P(\mathcal{B}_{i} \mid s_{i}) f_{S_{i} \mid \mathcal{A}_{i}}(s_{i}) \, \mathrm{d} s_{i} \nonumber \\
& = \mathbb{E}_{S_{i} \mid \mathcal{A}_{i}} [P(\mathcal{B}_{i} \mid S_{i})],
\end{align}
where $f_{S_{i} \mid \mathcal{A}_{i}}$ denotes the \textit{posterior} PDF of $S_{i}$ conditioned on $\mathcal{A}_{i}$, and $P(\mathcal{B}_{i} \mid S_{i})$ denotes the zero-crossing probability for the $i$-th interval conditioned on $S_{i}$.
Note that $P(\mathcal{B}_{i} \mid \mathcal{A}_{i})$ cannot be computed only from the marginal statistics of $S_{i}$ since the distribution of $S_{i}$ changes depending on $\mathcal{A}_{i}$.
Since the expectation in \cref{eq:opacity_ssdp} is taken over $s_{i} > 0$, we only consider $s_{i} > 0$ for $P(\mathcal{B}_{i} \mid s_{i})$ and omit this restriction for brevity.
In the following sections, we explain how to derive $P(\mathcal{B}_{i} \mid S_{i})$ and $f_{S_{i} \mid \mathcal{A}_{i}}$.
We summarize the symbols for mean and variance used in our formulation in \cref{tab:symbols} for reference.

\begin{table}[h]
\centering
\caption{Summary of symbols for mean and variance.}
\label{tab:symbols}
\begin{tabular}{cccccc}
\toprule
& $S_{i}$ & $S_{i} \mid S_{i-1}$ & $S_{i} \mid \mathcal{A}_{i-1}$ & $S_{i} \mid \mathcal{A}_{i}$ & $S_{i} \mid S_{i+1}, \mathcal{A}_{i-1}$ \\
\midrule
$\mathbb{E}[\cdot]$ & $\mu_{i}$ & $\mu_{i \mid i-1}$ & $\hat{\mu}_{i}$ & $\tilde{\mu}_{i}$ & $\mu_{i \mid i+1}$ \\
$\mathbb{V}[\cdot]$ & $\sigma_{i}^{2}$ & $\sigma_{i \mid i-1}^{2}$ & $\hat{\sigma}_{i}^{2}$ & $\tilde{\sigma}_{i}^{2}$ & $\sigma_{i \mid i+1}^{2}$ \\
\bottomrule
\end{tabular}
\end{table}

\subsubsection{Derivation of Zero-Crossing Probability $P(\mathcal{B}_{i} \mid S_{i})$}
\label{sec:method/first_passage_probability/conditional_zero_crossing_probability}

\begin{figure}[t]
\centering
\begin{minipage}[b]{0.49\linewidth}
\centering
\includegraphics[width=1.0\linewidth]{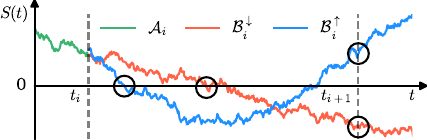}
\subcaption{$P(\mathcal{B}_{i} \mid s_{i}) = P(\mathcal{B}_{i}^{\downarrow} \mid s_{i}) + P(\mathcal{B}_{i}^{\uparrow} \mid s_{i})$ (\ref{eq:crossing_probability})}
\end{minipage}
\begin{minipage}[b]{0.49\linewidth}
\centering
\includegraphics[width=1.0\linewidth]{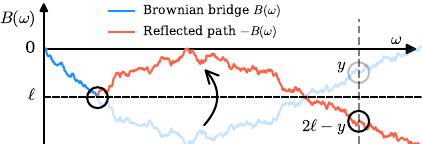}
\subcaption{Reflection principle for $P(\mathcal{B}_{i}^{\uparrow} \mid s_{i})$ (\ref{eq:crossing_probability_up_integrand})}
\end{minipage}
\caption{Illustration of the downward zero-crossing $\mathcal{B}_{i}$, which is decomposed into $\mathcal{B}_{i}^{\downarrow}$ and $\mathcal{B}_{i}^{\uparrow}$ based on the sign of $S_{i+1}$. $P(\mathcal{B}_{i}^{\uparrow} \mid s_{i})$ is computed by applying the reflection principle to the underlying Brownian bridge.}
\label{fig:zero_crossing}
\end{figure}

We first decompose $\mathcal{B}_{i}$ into $\mathcal{B}_{i}^{\downarrow}$ and $\mathcal{B}_{i}^{\uparrow}$ based on the sign of $S_{i+1}$ as illustrated in \cref{fig:zero_crossing}:
\begin{align}
\label{eq:crossing_event}
\mathcal{B}_{i}^{\downarrow} 
& \coloneq \{S_{i+1} \le 0\}, \quad 
\mathcal{B}_{i}^{\uparrow} 
\coloneq \left \{ \inf_{t \in (t_{i}, t_{i+1})} S(t) \le 0 \right \} \cap \{S_{i+1} > 0\} ,
\end{align}
where $\mathcal{B}_{i}^{\downarrow}$ denotes events where the ray is inside the surface at $t_{i+1}$ and $\mathcal{B}_{i}^{\uparrow}$ denotes events where the ray enters the surface at some $t \in (t_{i}, t_{i+1})$ and exits the surface by $t_{i+1}$.
$\mathcal{B}_{i}^{\downarrow} \cap \mathcal{B}_{i}^{\uparrow} = \varnothing$ yields:
\begin{align}
\label{eq:crossing_probability}
P(\mathcal{B}_{i} \mid s_{i}) 
& = P(\mathcal{B}_{i}^{\downarrow} \mid s_{i}) + P(\mathcal{B}_{i}^{\uparrow} \mid s_{i}) .
\end{align}
Here, $P(\mathcal{B}_{i}^{\downarrow} \mid S_{i})$ is given by the following transition kernel:
\begin{align}
\label{eq:crossing_probability_down}
P(\mathcal{B}_{i}^{\downarrow} \mid s_{i}) 
& = F_{S_{i+1} \mid S_{i}}(0 \mid s_{i})
= \Phi(0; \mu_{i+1 \mid i}, \sigma_{i+1 \mid i}^{2}) .
\end{align}
For $P(\mathcal{B}_{i}^{\uparrow} \mid S_{i})$, we first condition it on the endpoint $S_{i+1}$ and then integrate it out with the transition kernel:
\begin{align}
\label{eq:crossing_probability_up_integral}
P(\mathcal{B}_{i}^{\uparrow} \mid s_{i}) 
& = \int_{0}^{\infty} P(\mathcal{B}_{i}^{\uparrow} \mid s_{i}, s_{i+1}) f_{S_{i+1} \mid S_{i}}(s_{i+1} \mid s_{i}) \, \mathrm{d} s_{i+1} .
\end{align}
$P(\mathcal{B}_{i}^{\uparrow} \mid s_{i}, s_{i+1})$ is given by the following proposition.
Please refer to \cref{sec:crossing_probability_up_integrand_proof} for the proof.
\begin{proposition}
\label{prop:crossing_probability_up_integrand}
Let $\Theta(t) \coloneq \int_{t_{i}}^{t} \Psi(s, t_{i})^{2} \tau(s)^{2} \mathrm{d} s$ denote the quadratic variation of $M(t)$.
By the Dambis--Dubins--Schwarz theorem \citep{dds}, $M(t)$ can be viewed as a time-changed Brownian motion $B(\Theta(t))$.
We approximate $\Psi(t_{i}, t)^{-1} \mu(t)$ by its linear interpolation over $[t_{i}, t_{i+1}]$; since the resulting interpolation error scales as $O(\Delta t_{i}^{2})$, this approximation becomes accurate for sufficiently small $\Delta t_{i}$.
We apply the reflection principle to the Brownian bridge induced by $B$, yielding:
\begin{align}
\label{eq:crossing_probability_up_integrand}
P(\mathcal{B}_{i}^{\uparrow} \mid s_{i}, s_{i+1}) 
& \approx \mathds{1}(s_{i+1} > 0) \exp \left ( - \frac{2 s_{i} s_{i+1}}{\Omega_{i} \Psi_{i}} \right ) ,
\end{align}
where $\Omega_{i} \coloneqq \Theta(t_{i+1})$ and $\Psi_{i} \coloneq \Psi(t_{i}, t_{i+1})$.
\end{proposition}
A detailed analysis of the interpolation error against a Monte Carlo reference is given in \cref{sec:interpolation_error_validation}.

Therefore, $P(\mathcal{B}_{i}^{\uparrow} \mid S_{i})$ is given by:
\begin{align}
\label{eq:crossing_probability_up}
P(\mathcal{B}_{i}^{\uparrow} \mid s_{i})
& \approx (1 - \Phi(\xi_{i} \sigma_{i+1 \mid i}^{2}; \mu_{i+1 \mid i}, \sigma_{i+1 \mid i}^{2})) \exp \left ( - \xi_{i} \mu_{i+1 \mid i} + \frac{1}{2} \xi_{i}^{2} \sigma_{i+1 \mid i}^{2} \right ) ,
\end{align}
where $\xi_{i} \coloneq 2 \Omega_{i}^{-1} \Psi_{i}^{-1} s_{i}$.
Finally, $P(\mathcal{B}_{i} \mid S_{i})$ is given by \cref{eq:crossing_probability,eq:crossing_probability_down,eq:crossing_probability_up}.

\subsubsection{Bayesian Filtering for Posterior $f_{S_{i} \mid \mathcal{A}_{i}}$}
\label{sec:method/first_passage_probability/bayesian_filtering}

Since $f_{S_{i} \mid \mathcal{A}_{i}}$ cannot be obtained in closed form, we approximate it sequentially via Bayesian filtering.
The \textit{prior} $f_{S_{i} \mid \mathcal{A}_{i-1}}$ can be obtained from the \textit{posterior} $f_{S_{i-1} \mid \mathcal{A}_{i-1}}$ via the transition kernel, but not in closed form.
Therefore, we approximate it by a Gaussian distribution via moment matching. 
Given the posterior moments $\tilde{\mu}_{i-1} \coloneq \mathbb{E}[S_{i-1} \mid \mathcal{A}_{i-1}]$ and  $\tilde{\sigma}_{i-1}^{2} \coloneq \mathbb{V}[S_{i-1} \mid \mathcal{A}_{i-1}]$, we propagate them via the transition kernel:
\begin{align}
\label{eq:prior_mean}
\hat{\mu}_{i} 
& \coloneq \mathbb{E}[S_{i} \mid \mathcal{A}_{i-1}] 
= \alpha_{i-1} \tilde{\mu}_{i-1} + \beta_{i-1} , \\
\label{eq:prior_variance}
\hat{\sigma}_{i}^{2} 
& \coloneq \mathbb{V}[S_{i} \mid \mathcal{A}_{i-1}] 
= \alpha_{i-1}^{2} \tilde{\sigma}_{i-1}^{2} + \gamma_{i-1} ,
\end{align}
where $S_{i} \mid \mathcal{A}_{i-1} \approx \mathcal{N}(\hat{\mu}_{i}, \hat{\sigma}_{i}^{2})$.
Next, given an \textit{observation} $\mathcal{B}_{i-1}^{c}$, which corresponds to no zero-crossings in $(t_{i-1}, t_{i}]$, the posterior $f_{S_{i} \mid \mathcal{A}_{i}}$ is given via Bayesian updating:
\begin{align}
\label{eq:posterior_pdf}
f_{S_{i} \mid \mathcal{A}_{i}}(s_{i}) 
& = f_{S_{i} \mid \mathcal{A}_{i-1}, \mathcal{B}_{i-1}^{c}}(s_{i}) \nonumber \\
& = \frac{f_{S_{i} \mid \mathcal{A}_{i-1}}(s_{i}) P(\mathcal{B}_{i-1}^{c} \mid s_{i}, \mathcal{A}_{i-1})}{P(\mathcal{B}_{i-1}^{c} \mid \mathcal{A}_{i-1})} ,
\end{align}
where $\mathcal{A}_{i} = \mathcal{A}_{i-1} \cap \mathcal{B}_{i-1}^{c}$.
Here, $P(\mathcal{B}_{i-1}^{\downarrow} \mid S_{i}, \mathcal{A}_{i-1})$ is given by:
\begin{align}
\label{eq:crossing_probability_posterior_down}
P(\mathcal{B}_{i-1}^{\downarrow} \mid s_{i}, \mathcal{A}_{i-1}) 
& = \mathds{1}(s_{i} \le 0) .
\end{align}
$P(\mathcal{B}_{i-1}^{\uparrow} \mid S_{i}, \mathcal{A}_{i-1})$ is given by the following corollary. 
Please refer to \cref{sec:crossing_probability_posterior_up_proof} for the proof.
\begin{corollary}
\label{coro:crossing_probability_posterior_up}
For tractability, we replace the intractable posterior $f_{S_{i-1} \mid \mathcal{A}_{i-1}}$ with a truncated Gaussian PDF $\varphi_{+}(\cdot; \hat{\mu}_{i-1}, \hat{\sigma}_{i-1}^{2})$ obtained by Bayesian updating of the prior $f_{S_{i-1} \mid \mathcal{A}_{i-2}}(\cdot) \approx \varphi(\cdot; \hat{\mu}_{i-1}, \hat{\sigma}_{i-1}^{2})$ with an observation $\{S_{i-1} > 0\}$.
Then, $P(\mathcal{B}_{i-1}^{\uparrow} \mid S_{i}, \mathcal{A}_{i-1})$ is approximated by:
\begin{align}
\label{eq:crossing_probability_posterior_up}
P(\mathcal{B}_{i-1}^{\uparrow} \mid s_{i}, \mathcal{A}_{i-1}) 
& \approx \mathds{1}(s_{i} > 0) G(s_{i}) ,
\end{align}
where 
\begin{align}
\label{eq:snis_weight}
G(s_{i}) 
& \coloneq \frac{1 - \Phi(\xi_{i-1} \sigma_{i-1 \mid i}^{2}; \mu_{i-1 \mid i}, \sigma_{i-1 \mid i}^{2})}{1 - \Phi(0; \mu_{i-1 \mid i}, \sigma_{i-1 \mid i}^{2})} \exp(-\xi_{i-1} \mu_{i-1 \mid i} + \frac{1}{2} \xi_{i-1}^{2} \sigma_{i-1 \mid i}^{2}) ,
\end{align}
and $\xi_{i-1} \coloneq 2 \Omega_{i-1}^{-1} \Psi_{i-1}^{-1} s_{i}$.
\end{corollary}
Therefore, substituting \cref{eq:crossing_probability_posterior_down,eq:crossing_probability_posterior_up} for \cref{eq:posterior_pdf} yields:
\begin{align}
\label{eq:posterior_pdf_snis}
f_{S_{i} \mid \mathcal{A}_{i}}(s_{i})
& \approx \frac{\varphi_{+}(s_{i}; \hat{\mu}_{i}, \hat{\sigma}_{i}^{2}) W(s_{i})}{Z_{i}} , \\
\label{eq:norm_constant_snis}
Z_{i} 
& \coloneq \mathbb{E}_{s_{i} \sim \mathcal{N}_{+}(\hat{\mu}_{i}, \hat{\sigma}_{i}^{2})}[W(s_{i})] ,
\end{align}
where $W(s_{i}) \coloneq 1 - G(s_{i})$.
Finally, from \cref{eq:opacity_ssdp,eq:crossing_probability,eq:posterior_pdf_snis}, $P(\mathcal{B}_{i} \mid \mathcal{A}_{i})$ can be estimated by self-normalized importance sampling:
\begin{align}
\label{eq:opacity_ssdp_snis}
P(\mathcal{B}_{i} \mid \mathcal{A}_{i}) 
& \approx \frac{\mathbb{E}_{s_{i} \sim \mathcal{N}_{+}(\hat{\mu}_{i}, \hat{\sigma}_{i}^{2})}[P(\mathcal{B}_{i} \mid s_{i}) W(s_{i})]}{Z_{i}} .
\end{align}
In the same way, the posterior mean and variance are given by:
\begin{align}
\label{eq:posterior_mean_snis}
\tilde{\mu}_{i} 
& \coloneq \mathbb{E}[S_{i} \mid \mathcal{A}_{i}]
\approx \frac{\mathbb{E}_{s_{i} \sim \mathcal{N}_{+}(\hat{\mu}_{i}, \hat{\sigma}_{i}^{2})}[s_{i} W(s_{i})]}{Z_{i}} , \\
\label{eq:posterior_variance_snis}
\tilde{\sigma}_{i}^{2} 
& \coloneq \mathbb{V}[S_{i} \mid \mathcal{A}_{i}]
\approx \frac{\mathbb{E}_{s_{i} \sim \mathcal{N}_{+}(\hat{\mu}_{i}, \hat{\sigma}_{i}^{2})}[s_{i}^{2} W(s_{i})]}{Z_{i}} - \tilde{\mu}_{i}^{2} .
\end{align}
In practice, we approximate \cref{eq:norm_constant_snis,eq:opacity_ssdp_snis,eq:posterior_mean_snis,eq:posterior_variance_snis} by Gauss--Legendre quadrature, where the quadrature nodes are transformed by the quantile function $\Phi_{+}^{-1}(\hat{\mu}_{i}, \hat{\sigma}_{i}^{2})$.

\subsubsection{Negative-Absorbing Approximation}
\label{sec:method/first_passage_probability/negative_absorbing_approximation}

The full Bayesian formulation provides a principled way to propagate the survival-conditioned distribution $f_{S_i \mid \mathcal{A}_i}$ along each ray. 
This provides a more faithful first-passage-time distribution for each ray that can account for multiple downward zero-crossings.
The main drawback is computational inefficiency; the computation of $f_{S_{i} \mid \mathcal{A}_{i}}$ has a sequential dependence across samples along each ray and thus cannot be parallelized over $i$.
For practicality, we introduce a \textit{negative-absorbing approximation}, which removes the possibility of sign reversion of $S(t)$.
As a result, Bayesian filtering can be removed, enabling parallel computation over $i$.
Then, we assume $S(t)$ has no sign reversion along each ray: once $S(t)$ becomes negative at $t = t^{*}$, it never returns to a positive value for any $t > t^{*}$. 
Although this assumption is not exact for objects with finite volume, where a downward zero-crossing may be followed by an upward zero-crossing, this mismatch mainly affects cases where the ray later undergoes another downward zero-crossing, as in rays intersecting multiple parts of a single object or multiple objects.
Empirically, we found this approximation stable in our experiments while enabling fully parallel rendering.
We provide an ablation study of this approximation in \cref{sec:experiments}.
Under this assumption, $\mathcal{A}_{i}$ and $\mathcal{B}_{i}$ can be simplified as follows:
\begin{align}
\label{eq:survival_event_approx}
\mathcal{A}_{i} 
& \equiv \{S_{i} > 0\} , \quad
\mathcal{B}_{i} 
\equiv \mathcal{B}_{i}^{\downarrow} 
= \{S_{i+1} \le 0\} .
\end{align}
Therefore, $P(\mathcal{B}_{i} \mid \mathcal{A}_{i}) \eqqcolon o^{\mathrm{SSDP}}_{i}$ reduces to:
\begin{align}
\label{eq:opacity_ssdp_approx}
o^{\mathrm{SSDP}}_{i}
& = P(S_{i+1} \le 0 \mid S_{i} > 0) \nonumber \\
& = \frac{P(S_{i} > 0,\ S_{i+1} \le 0)}{P(S_{i} > 0)} \nonumber \\
& = \frac{F_{S_{i+1}}(0) - F_{S_{i}, S_{i+1}}(0, 0)}{1 - F_{S_{i}}(0)} ,
\end{align}
where $F_{S_{i}, S_{i+1}}$ denotes the bivariate CDF of $(S_{i}, S_{i+1})$.
\begin{remark}
\label{rem:negative_absorbing_approximation}
\cite{neg_abs_approx} derived the probability that an exponentially correlated Gaussian field---whose restriction to a ray has the Markov correlation structure of an OU process---first crosses an isovalue within each sampling interval along the ray.
Their first-passage probability coincides with \cref{eq:first_passage_probability} under the negative-absorbing approximation. 
Their derivation, however, factorizes the joint probability of the events at the sampled points as if these events formed a Markov chain. 
Since the Markov property yields conditional independence given \textit{values} but not given \textit{events}, the resulting expression is not exact even for their own target. 
In our notation, this implicit step corresponds precisely to the simplification $\mathcal{A}_{i} \equiv \{S_{i} > 0\}$ in \cref{eq:survival_event_approx}; likewise, crossings that revert within an interval, $\mathcal{B}_{i}^{\uparrow}$, have no counterpart in their formulation, which corresponds to the other simplification $\mathcal{B}_{i} \equiv \mathcal{B}_{i}^{\downarrow} = \{S_{i+1} \le 0\}$.
We instead start from the full Bayesian formulation and reach \cref{eq:opacity_ssdp_approx} through these two simplifications, derived jointly from the single explicit assumption that $S(t)$ never reverts its sign---under which both are exact---thereby making explicit what is
discarded in computing the first-passage probability from the interval endpoints alone.
\end{remark}
A detailed analysis of each simplification in \cref{eq:survival_event_approx} and its failure cases is given in \cref{sec:approximation_error_validation}.

\subsection{Relationship to Prior Works}
\label{sec:method/relationship_to_prior_works}

In this subsection, we clarify the relationship between our formulation and two closely related formulations, NeuS \citep{neus} and \textit{Objects as Volumes} (OaV) \citep{oav}.
We first show that NeuS arises as a special case of our formulation in the perfectly correlated limit of the OU process.
We then discuss the condition under which OaV reduces to NeuS and show that this case is also covered by our formulation.

\paragraph{NeuS.}

Here, we show the theoretical connection between SSDP and NeuS.
First, we consider the perfectly correlated limit of the OU process, i.e., $\alpha_{i} \rightarrow 1$ and $\gamma_{i} \rightarrow 0$.
This limit collapses the ray-wise stochastic dynamics, freezing the residual process at its initial value and yielding $S_{i} = \mu_{i} + \varepsilon$, where $\varepsilon \coloneq R_{\bm{r}}(t_{0})$ denotes a ray-wise residual shared across $i$.
To recover NeuS, we instantiate the initial distribution as $\varepsilon \sim \mathrm{Logistic}(0, s)$, i.e., the logistic distribution whose CDF is $\varsigma_{s}$.
Under this shared-residual limit, the bivariate CDF value $F_{S_{i}, S_{i+1}}(0, 0)$ reduces to $\min(F_{S_{i}}(0), F_{S_{i+1}}(0))$.
Then, \cref{eq:opacity_ssdp_approx} reduces to:
\begin{align}
\label{eq:opacity_ssdp_approx_neus}
o^{\mathrm{NeuS}}_{i}
& = \frac{F_{S_{i+1}}(0) - F_{S_{i}, S_{i+1}}(0, 0)}{1 - F_{S_{i}}(0)} \nonumber \\
& = \frac{F_{S_{i+1}}(0) - \min(F_{S_{i}}(0), F_{S_{i+1}}(0))}{1 - F_{S_{i}}(0)} \nonumber \\
& = \left [ \frac{F_{S_{i+1}}(0) - F_{S_{i}}(0)}{1 - F_{S_{i}}(0)} \right ]_{+} \nonumber \\
& = \left [ \frac{\varsigma_{s}(\mu_{i}) - \varsigma_{s}(\mu_{i+1})}{\varsigma_{s}(\mu_{i})} \right ]_{+} .
\end{align}
This is equivalent to NeuS's opacity formulation (\ref{eq:opacity_neus}), indicating that NeuS arises as a special case of our formulation.
Importantly, this derivation reveals that SSDP is a natural generalization of NeuS with the ray-wise covariance structure of the stochastic process.

\paragraph{Objects as Volumes (OaV).}

OaV gives a stochastic interpretation of opaque solids as volumes. 
In its implicit-surface form, OaV builds the attenuation coefficient $\sigma$ from the marginal distribution of $S(t)$, rather than from the transition kernel.
With the notation of this paper, $\sigma$ is given by:
\begin{align}
\label{eq:density_oav}
\sigma(\bm{r}(t)) 
& = \frac{f_{S(t)}(0)}{1 - F_{S(t)}(0)} \|\nabla \mu(\bm{r}(t))\| A(\bm{r}(t), \bm{d}) ,
\end{align}
where $A(\bm{r}(t), \bm{d})$ denotes the projected-area term induced by the distribution $\mathcal{D}_{\bm{N}(\bm{x})}$ over normals $\bm{N}(\bm{x})$:
\begin{align}
\label{eq:projected_area_oav}
A(\bm{x}, \bm{d}) 
& = \mathbb{E}_{\tilde{\bm{n}} \sim \mathcal{D}_{\bm{N}(\bm{x})}}[|\tilde{\bm{n}} \cdot \bm{d}|]
\end{align}
Note that $\mathcal{D}_{\bm{N}(\bm{x})}$ is not derived from $S(t)$, but is a separate model introduced by OaV.
Choosing the Dirac delta distribution as $\mathcal{D}_{\bm{N}(\bm{x})}$ makes \cref{eq:density_oav} reduce to:
\begin{align}
\label{eq:density_oav_dirac}
\sigma(\bm{r}(t)) 
& = \frac{f_{S(t)}(0)}{1 - F_{S(t)}(0)} |\nabla \mu(\bm{r}(t)) \cdot \bm{d}|
\end{align}
To derive the opacity $o_{i}$ using endpoint CDF values $(F_{S_{i}}, F_{S_{i+1}})$, we adopt the assumption from NeuS that $\mu(\bm{r}(t))$ is monotonic with respect to $t$ within each interval $[t_{i}, t_{i+1}]$. 
In this case, the opacity $o_{i}$ is given by:
\begin{align}
\label{eq:oav_opacity_dirac}
o^{\mathrm{OaV}}_{i} & = 
\begin{dcases}
\frac{F_{S_{i+1}}(0) - F_{S_{i}}(0)}{1 - F_{S_{i}}(0)} & \text{if} \ \mu_{i} \ge \mu_{i+1} , \\
\frac{F_{S_{i}}(0) - F_{S_{i+1}}(0)}{1 - F_{S_{i+1}}(0)} & \text{otherwise} .
\end{dcases}
\end{align}
Thus, $o^{\mathrm{OaV}}_{i}$ is equivalent to $o^{\mathrm{NeuS}}_{i}$ when $\mu_{i} \ge \mu_{i+1}$, but behaves differently otherwise:
OaV assigns nonzero opacity even when $\mu_{i} < \mu_{i+1}$, treating both entering and exiting directions as attenuation.
This property makes the rendering model satisfy \textit{reciprocity}, as described by \cite{oav}.
In contrast, our formulation does not enforce reciprocity; instead, it derives the opacity from the first-passage perspective, as the conditional probability $P(\mathcal{B}_{i} \mid \mathcal{A}_{i})$ of the downward zero-crossing event $\mathcal{B}_{i}$ given the survival event $\mathcal{A}_{i}$.

\subsection{Loss Functions}
\label{sec:method/loss_functions}

Given the color PDF $f_{\bm{C}_{\bm{r}}}$, we minimize the negative log marginal likelihood (NLML) $\mathcal{L}_{\mathrm{nlml}}^{(\bm{r})}$ along with the Eikonal regularizer $\mathcal{L}_{\mathrm{reg}}^{(\bm{r})}$ \citep{eikonal_loss} to make the mean field $\mu_{\bm{\theta}}$ satisfy the Eikonal equation (\ref{eq:eikonal_equation}):
\begin{align}
\label{eq:total_loss}
& \mathcal{L}(\bm{\theta}, \bm{\phi}, \bm{\psi}) \coloneq \sum_{(\bm{r}, \bm{c}_{\bm{r}}) \in \mathcal{B}} \left ( \mathcal{L}_{\mathrm{nlml}}^{(\bm{r})}(\bm{c}_{\bm{r}}; \bm{\theta}, \bm{\phi}, \bm{\psi}) + \lambda \mathcal{L}_{\mathrm{reg}}^{(\bm{r})}(\bm{\theta}) \right ) , \\
\label{eq:nlml_loss}
& \mathcal{L}_{\mathrm{nlml}}^{(\bm{r})}(\bm{c}_{\bm{r}}; \bm{\theta}, \bm{\phi}, \bm{\psi}) \coloneq -\log{f_{\bm{C}_{\bm{r}}}(\bm{c}_{\bm{r}}; \bm{\theta}, \bm{\phi}, \bm{\psi})} , \\
\label{eq:eikonal_regularizer}
& \mathcal{L}_{\mathrm{reg}}^{(\bm{r})}(\bm{\theta}) \coloneq \frac{1}{N} \sum_{i=0}^{N-1} (\|\nabla \mu_{\bm{\theta}}(\bm{x}) |_{\bm{x}=\bm{r}(t_{i})} \|_{2} - 1)^{2} ,
\end{align}
where $\mathcal{B}$ denotes a batch of rays, and $\bm{\theta}$, $\bm{\phi}$, and $\bm{\psi}$ denote the parameters of the mean field $\mu_{\bm{\theta}}$, the OU network $(\kappa_{\bm{\phi}}, \tau_{\bm{\phi}})$, and the color field $\hat{\bm{c}}_{\bm{\psi}}$, respectively.
Note that prior works \citep{idr,neus} employ the following \textit{expectation} loss $\mathcal{L}_{\mathrm{exp}}^{(\bm{r})}$ between the expected and observed colors instead of $\mathcal{L}_{\mathrm{nlml}}^{(\bm{r})}$:
\begin{align}
\label{eq:exp_loss}
\mathcal{L}_{\mathrm{exp}}^{(\bm{r})}(\bm{c}_{\bm{r}}; \bm{\theta}, \bm{\phi}, \bm{\psi}) 
& \coloneq \|\mathbb{E}[\bm{C}_{\bm{r}}] - \bm{c}_{\bm{r}}\|_{1} .
\end{align}
We provide an ablation study of the photometric losses $\mathcal{L}_{\mathrm{nlml}}^{(\bm{r})}$ and $\mathcal{L}_{\mathrm{exp}}^{(\bm{r})}$ in \cref{sec:experiments/evaluation_results}.

\section{Experiments}
\label{sec:experiments}

\subsection{Implementation Details}
\label{sec:experiments/implementation_details}

\paragraph{Baselines.}

To verify the effectiveness of our formulation, we compare our method with several baselines that employ SDF-based volume rendering.
Note that the purpose of our comparisons is not to demonstrate state-of-the-art reconstruction accuracy at the system level.
Therefore, we focus on apples-to-apples comparisons, where all components except the rendering core are kept identical across methods.
A natural baseline would be NeuS \citep{neus} since its rendering core can be recovered as a special case of our formulation, as shown in \cref{sec:method/relationship_to_prior_works}.
However, NeuS relies on a renderer-coupled sampling scheme, which prevents us from replacing only the rendering core with our formulation while keeping the rest unchanged.
Therefore, we adopt NeuS-Facto \citep{nerfstudio} as a baseline, which decouples sampling from rendering with the model-agnostic \textit{proposal network} introduced in Mip-NeRF 360 \citep{mipnerf360}.
We summarize the differences between NeuS and NeuS-Facto in \cref{sec:neus_vs_neus_facto}.
We instantiate our method, denoted \textit{SSDP-Facto}, by replacing only the NeuS-based rendering core of NeuS-Facto with our formulation.
We further design analogous variants of \textit{Objects as Volumes} (OaV) \citep{oav} and UNIS \citep{unis}, denoted \textit{OaV-Facto} and \textit{UNIS-Facto}, by replacing the NeuS-based rendering core of NeuS-Facto with their formulations, respectively.
Among the case studies C0--C5 of UNIS, we instantiate C1--C3, where the opacity of each interval can be computed exactly from the SDF values at the two endpoints as in NeuS, as required to replace only the rendering core.
Note that NeuS is equivalent to C0, and VolSDF \citep{volsdf} is covered by C1, which modifies the density mapping of VolSDF to be first-order unbiased.
We then compare SSDP-Facto with the three controlled baselines: NeuS-Facto, OaV-Facto, and UNIS-Facto (C1--C3).
We also report NeuS and OaV as references; since they employ their own sampling schemes, comparisons with them are not apples-to-apples with respect to the rendering core.
No such reference is reported for UNIS, whose official implementation is not publicly available.
All models except OaV are trained under the same configuration on Nerfstudio \citep{nerfstudio} to mitigate implementation-level differences.
Since OaV and its dependencies are not implemented on Nerfstudio, we train and evaluate OaV$^{*}$ with the official implementation as is, which therefore additionally involves implementation-level differences.

\paragraph{SSDP.}

We implement the OU network $(\kappa_{\bm{\phi}}, \tau_{\bm{\phi}})$ as a 2-layer MLP with 256 hidden units per layer and two scalar outputs, followed by Softplus to enforce positivity. 
The OU network takes as input the concatenated feature vectors from the mean field $\mu_{\bm{\theta}}$ at the two endpoints of each interval $[t_i, t_{i+1})$.
Please refer to \cref{sec:network_architecture} for details.
We use 32 quadrature nodes to approximate \cref{eq:norm_constant_snis,eq:opacity_ssdp_snis,eq:posterior_mean_snis,eq:posterior_variance_snis}. 
We set the scale $b$ in \cref{eq:color_pdf_conditional} to $1$.
The number of samples $N$ per ray and the loss weight $\lambda$ for the Eikonal term are set to $48$ and $0.1$.
All other networks and hyperparameters are kept identical to NeuS-Facto.

\subsection{Datasets}
\label{sec:experiments/datasets}

\paragraph{DTU.}

We evaluate our method mainly on the DTU dataset \citep{dtu}, which is widely used for benchmarking surface reconstruction methods.
We use 15 scenes following prior works \citep{idr,neus}, each of which includes 49 or 64 views with a resolution of $1600 \times 1200$.

\paragraph{MobileBrick.}

We evaluate our method additionally on the MobileBrick dataset \citep{mobilebrick}, which focuses on objects with detailed geometry made of LEGO bricks, providing precise ground-truth meshes.
We use 18 scenes from the test split, each of which includes 43 to 134 views with a resolution of $1920 \times 1440$.

\subsection{Evaluation Metrics}
\label{sec:experiments/evaluation_metrics}

\paragraph{Surface reconstruction.}

For the DTU dataset, we report Chamfer distance (CD) between the ground-truth point cloud and a point cloud randomly sampled from the reconstructed mesh, following the official evaluation protocol.
For the MobileBrick dataset, we additionally report $F_{1}$ scores at thresholds of 2.5\,mm and 5\,mm, following the official evaluation protocol.
We extract a mesh from the mean field $\mu_{\bm{\theta}}$ using Marching Cubes \citep{marching_cubes} with a grid resolution of $512^{3}$.
Note that many prior works apply filtering to the reconstructed mesh before evaluation: even when ground-truth foreground masks are not available during training, mesh vertices outside the visual hull based on the foreground masks are removed.
This process underestimates the penalty for spurious surfaces, making it difficult to evaluate a trade-off between \textit{false positives} and \textit{false negatives}. 
Therefore, we additionally evaluate the reconstructed mesh filtered with the union of viewing frustums instead of the visual hull.
This is a conservative filter for cases where there are vertices that are not included in the field of view of any camera.
We refer to the evaluation protocols with viewing-frustum and visual-hull filtering as the \textit{unmasked} and \textit{masked} protocols, respectively.

\paragraph{Uncertainty quantification.}
Along with the surface reconstruction metrics, we report representative evaluation metrics for uncertainty quantification, namely continuous ranked probability score (CRPS), mean absolute error (MAE), expected calibration error (ECE), and sharpness, to evaluate the learned first-passage-time distribution for each ray.
We compute these metrics on a subset of rays that intersect the ground-truth mesh, using the first ray--mesh intersection as the observed first-passage time, and hence for the conditional first-passage time $H_{\bm{r}} \mid H_{\bm{r}} \le t_{N}$.
Evaluating whether each ray actually intersects a surface, e.g., via the Brier score for $P(H_{\bm{r}} \le t_{N})$, would require a per-ray hit/miss label; however, since the foreground masks for the DTU and MobileBrick datasets cover only the target object, no label defines whether reconstructing actual surfaces outside the foreground region, such as the table, should be penalized.
The definitions of these metrics are given in \cref{tab:uncertainty_metrics}.
Note that while CRPS can be interpreted independently as a proper scoring rule, neither ECE nor sharpness possesses this property, requiring them to be interpreted jointly.
For the DTU dataset, since the ground-truth mesh is not provided for each scene, we extract a mesh from the ground-truth point cloud by Poisson surface reconstruction \citep{poisson_surface_reconstruction}, filtering out vertices outside the visual hull based on the foreground masks.
While the extracted meshes inevitably contain errors, they are common to all models and thus preserve the relative comparison.
The evaluation on MobileBrick is free from such reconstruction errors, as its precise ground-truth meshes are provided.

\subsection{Evaluation Results}
\label{sec:experiments/evaluation_results}

\begin{figure}[t]

\centering

\begin{minipage}[b]{0.245\linewidth}
\centering
\includegraphics[width=1.0\linewidth]{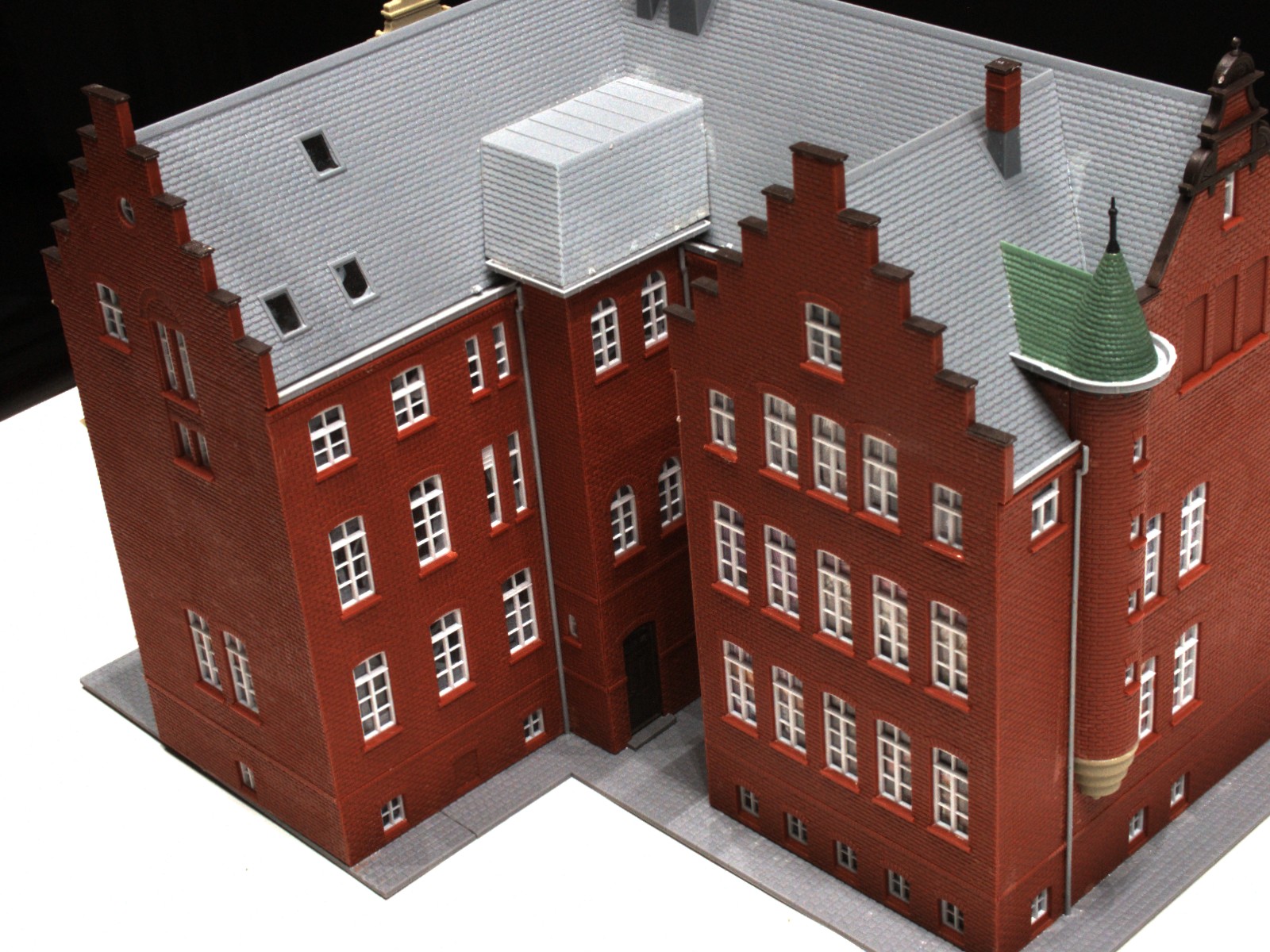}
\end{minipage}
\begin{minipage}[b]{0.245\linewidth}
\centering
\includegraphics[width=1.0\linewidth]{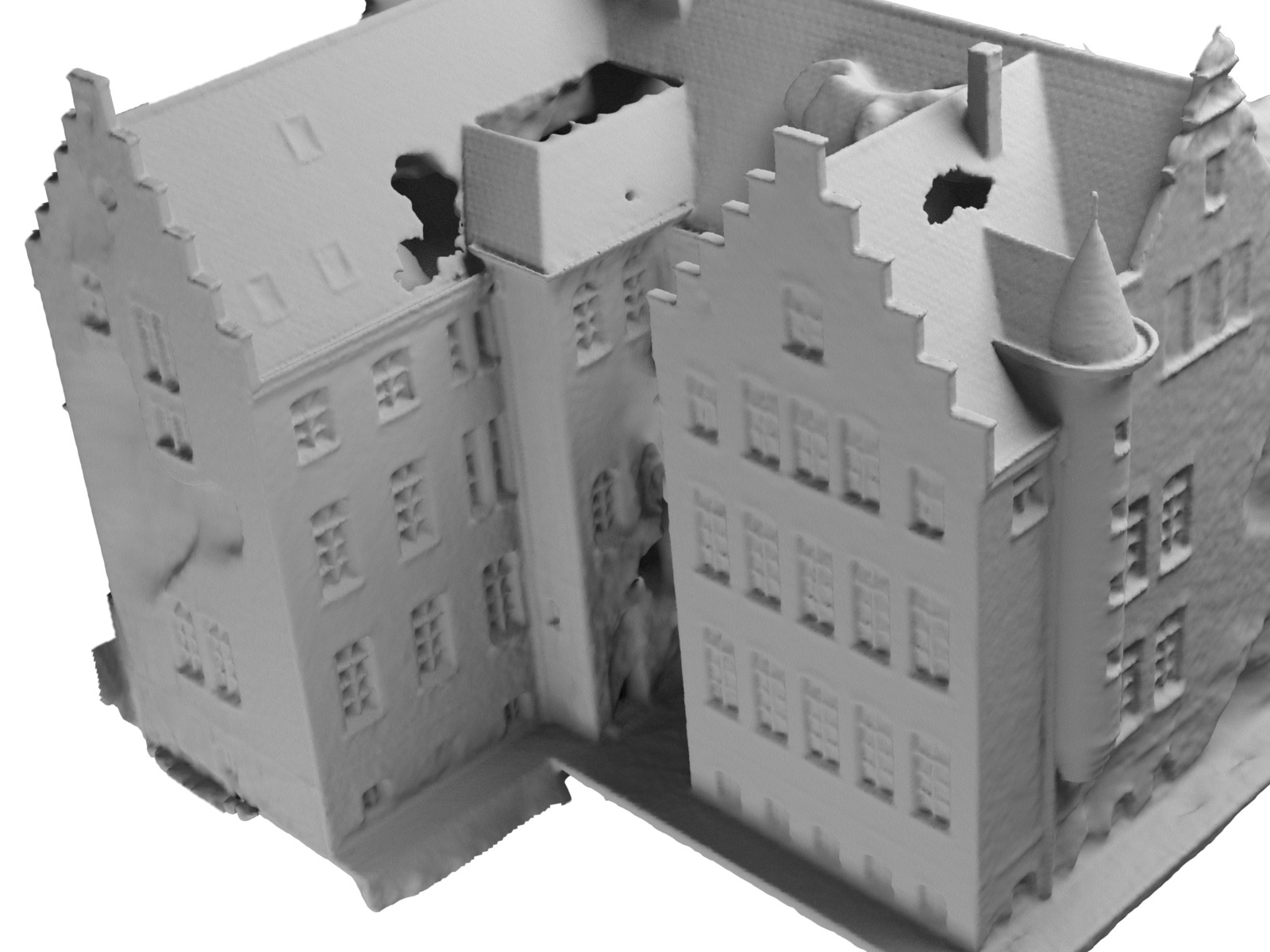}
\end{minipage}
\begin{minipage}[b]{0.245\linewidth}
\centering
\includegraphics[width=1.0\linewidth]{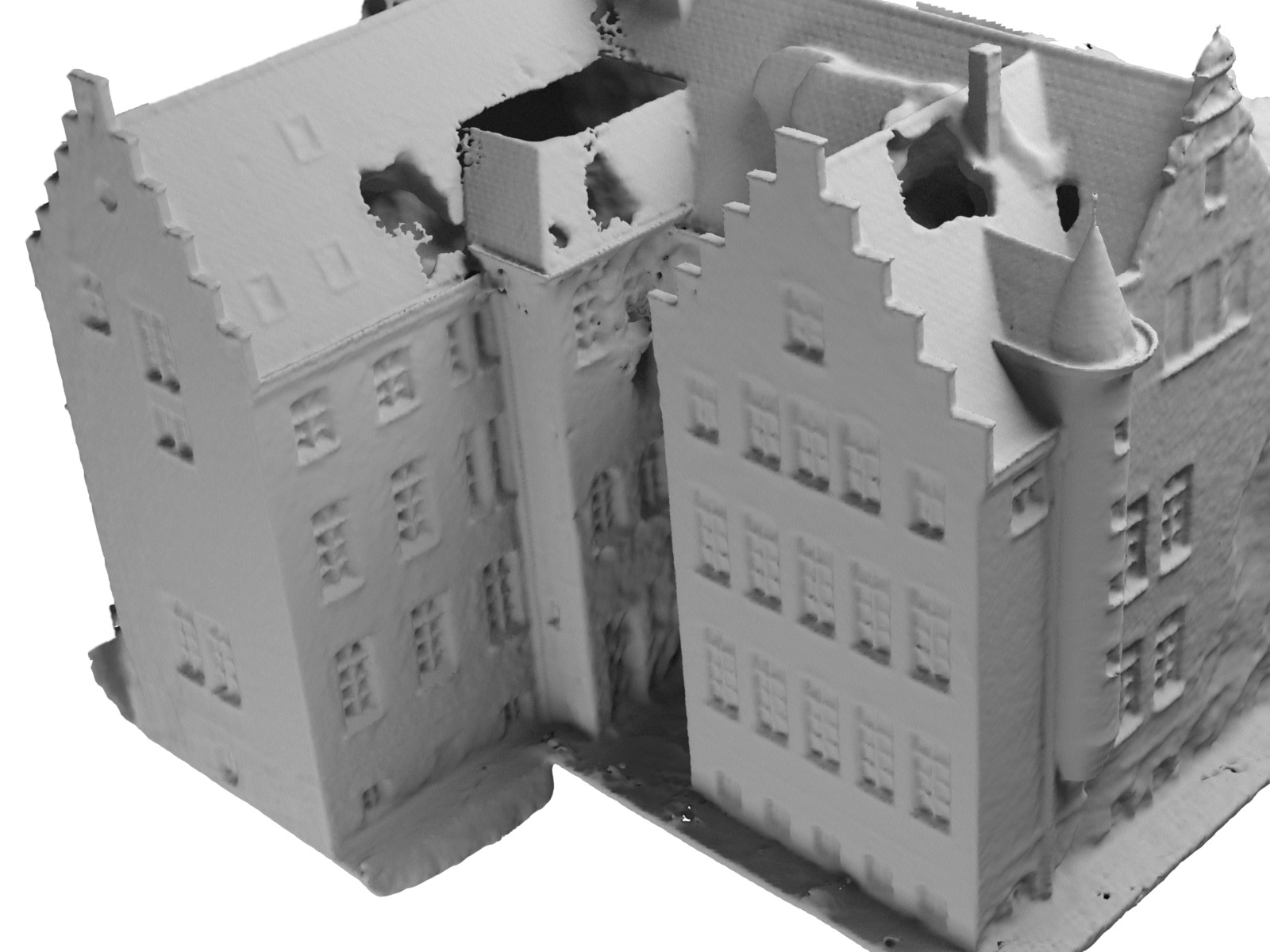}
\end{minipage}
\begin{minipage}[b]{0.245\linewidth}
\centering
\includegraphics[width=1.0\linewidth]{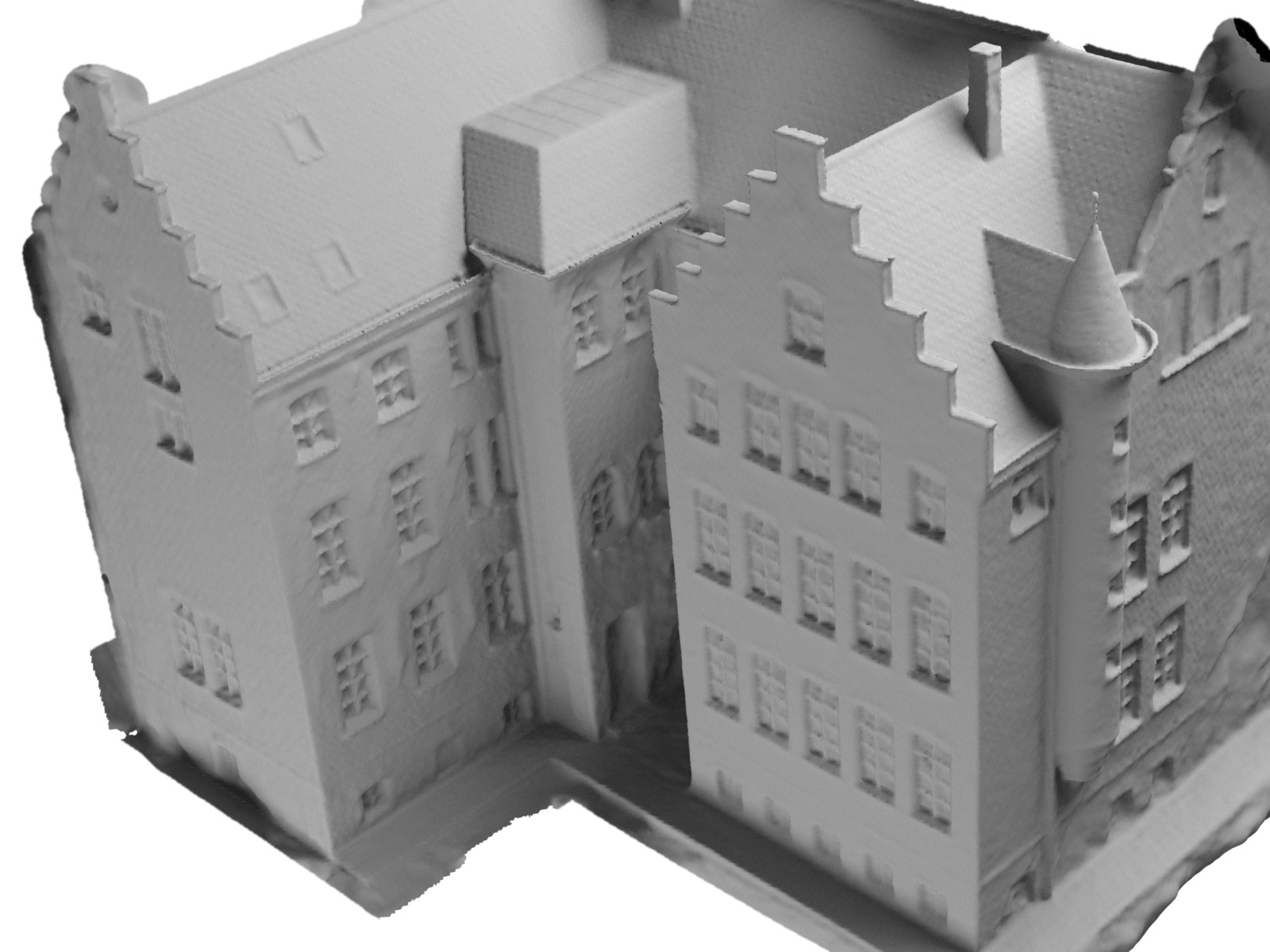}
\end{minipage}

\begin{minipage}[b]{0.245\linewidth}
\centering
\includegraphics[width=1.0\linewidth]{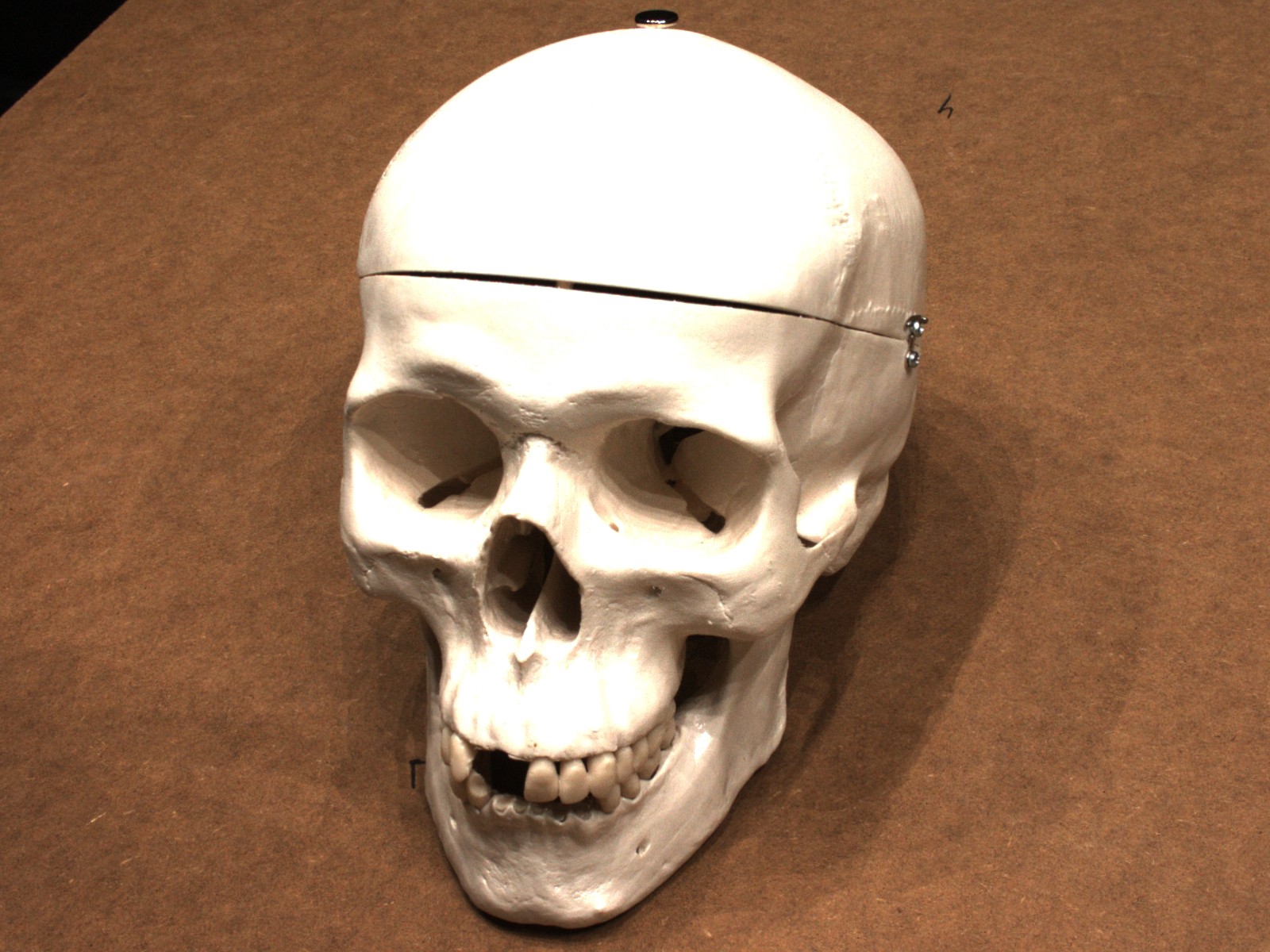}
\subcaption{Ground Truth}
\end{minipage}
\begin{minipage}[b]{0.245\linewidth}
\centering
\includegraphics[width=1.0\linewidth]{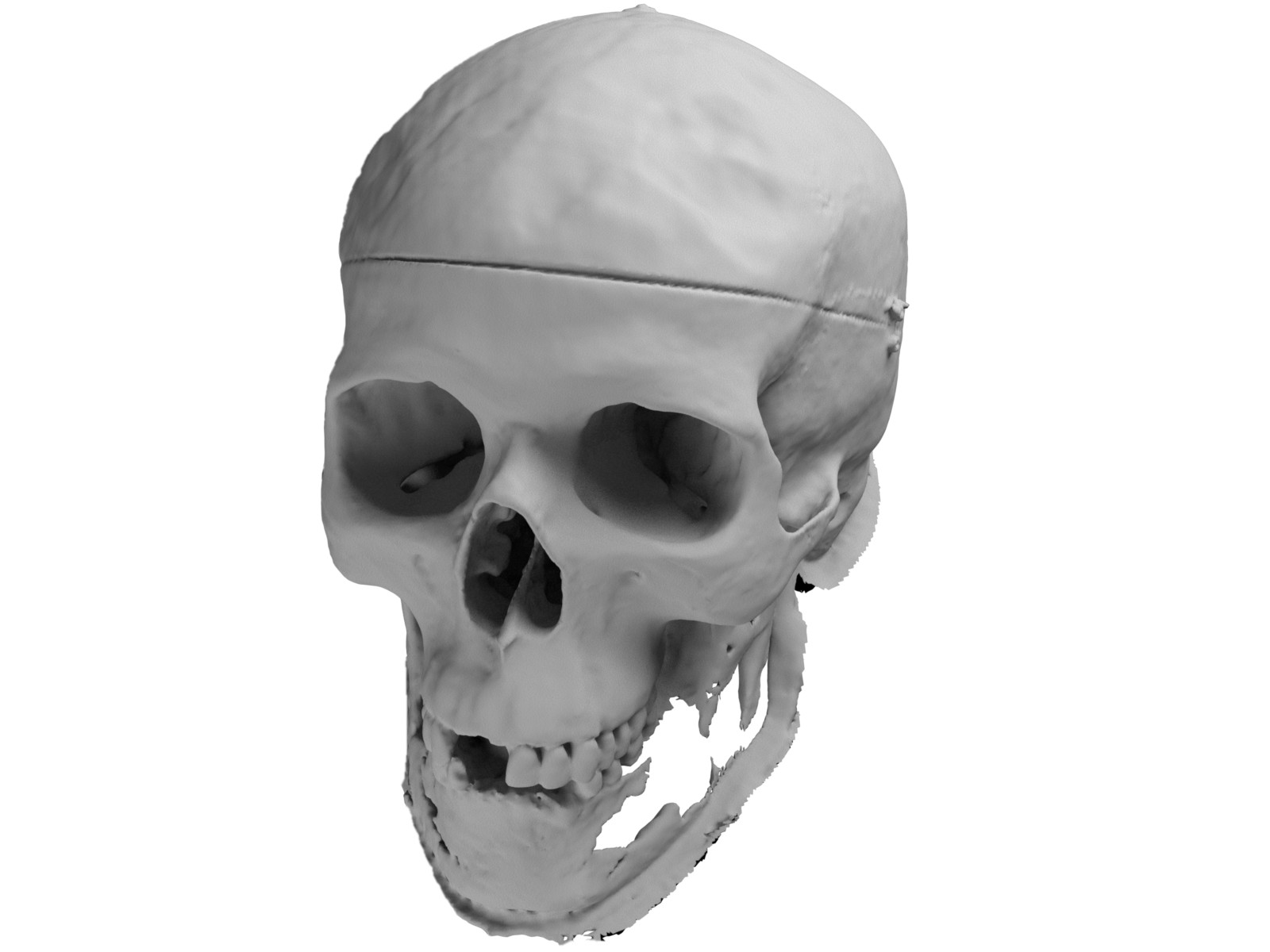}
\subcaption{NeuS-Facto}
\end{minipage}
\begin{minipage}[b]{0.245\linewidth}
\centering
\includegraphics[width=1.0\linewidth]{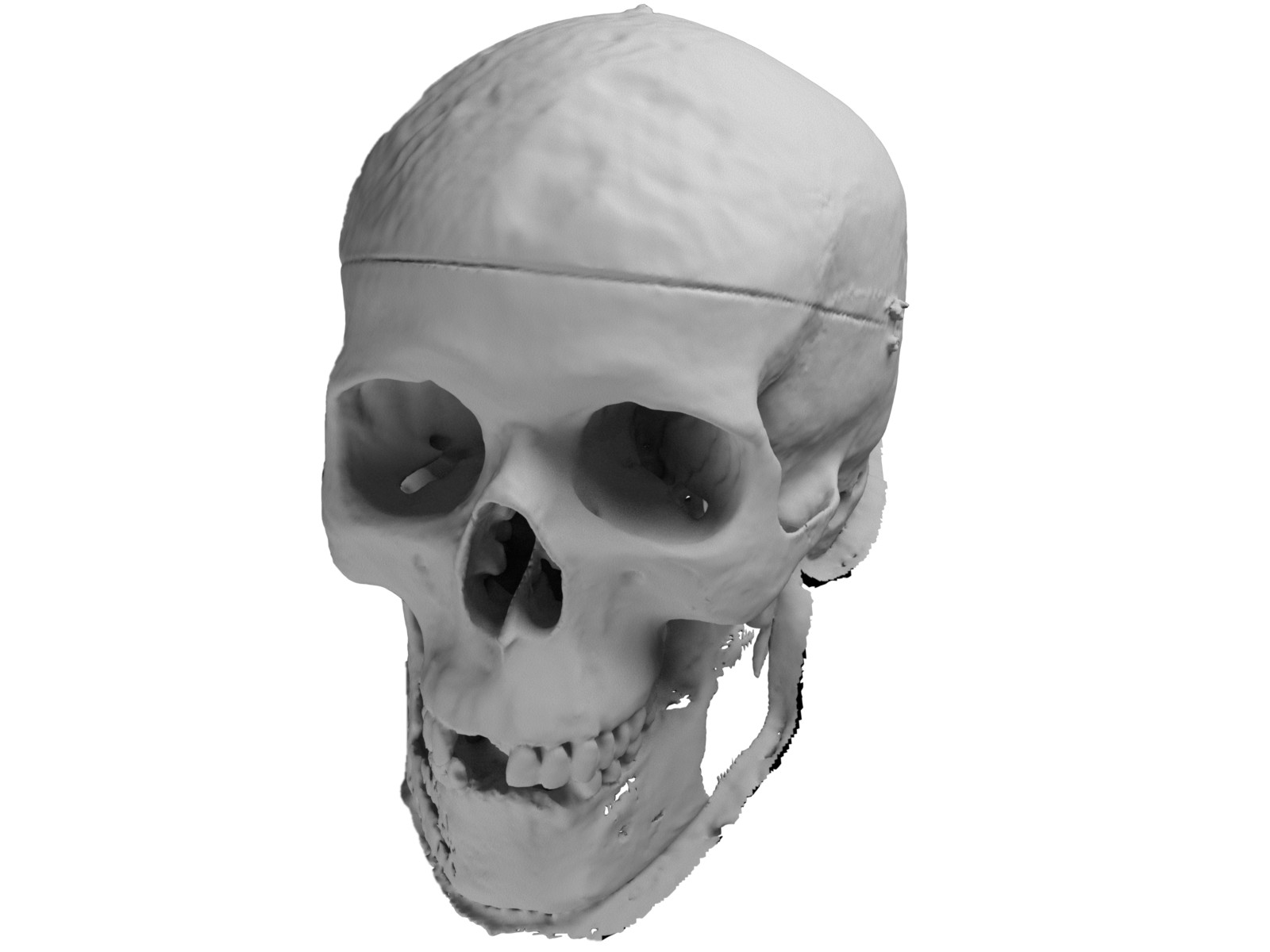}
\subcaption{OaV-Facto}
\end{minipage}
\begin{minipage}[b]{0.245\linewidth}
\centering
\includegraphics[width=1.0\linewidth]{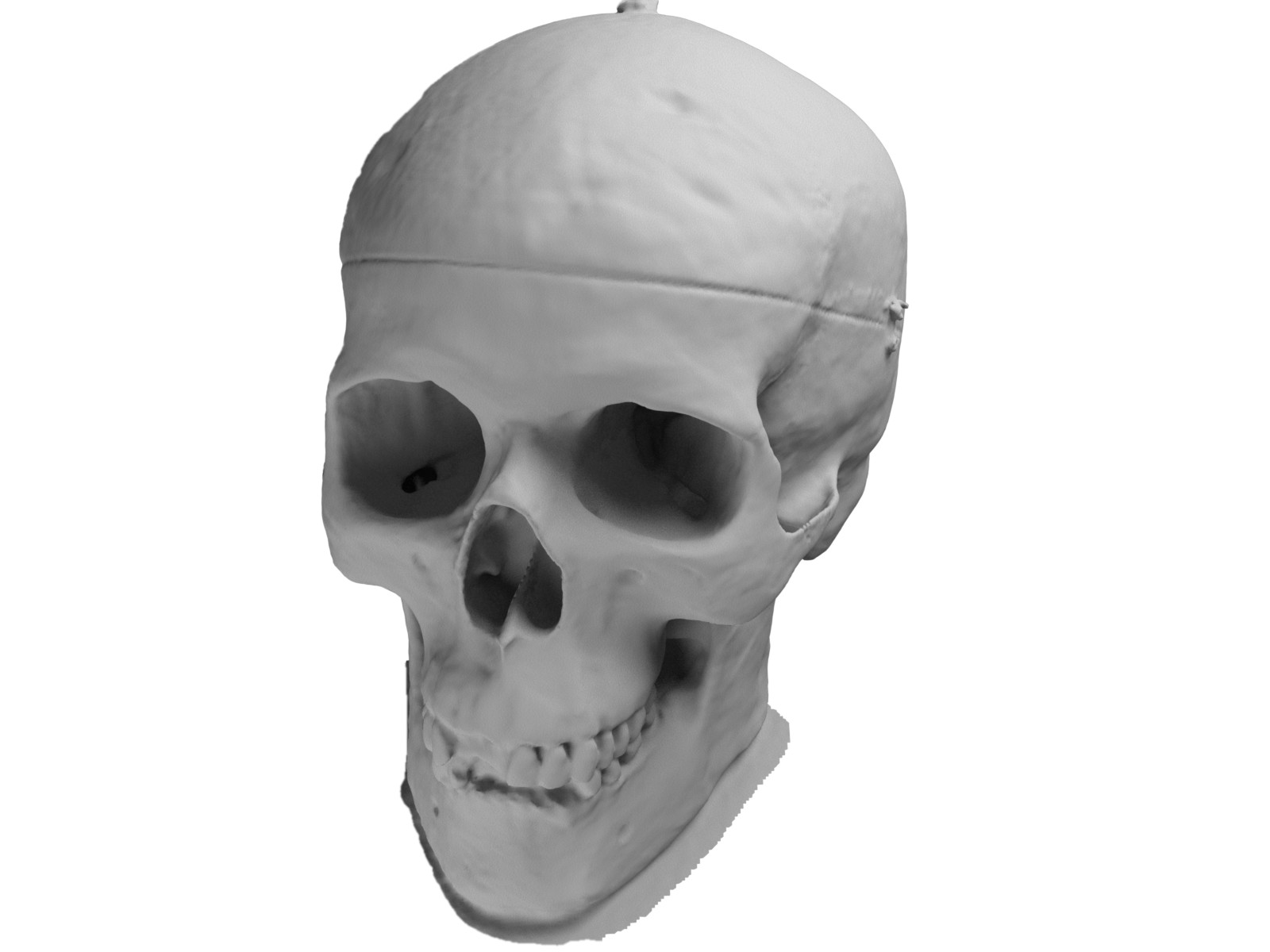}
\subcaption{SSDP-Facto}
\end{minipage}

\caption{Visualization examples on the DTU dataset.}
\label{fig:dtu_visualization_examples_subset}
\end{figure}

\paragraph{Comparisons on the DTU dataset.}
In their official implementations, NeuS and OaV sample approximately four times as many total rays during training as NeuS-Facto, making a fair comparison of their rendering cores difficult. 
To address this discrepancy, we linearly scale the number of training iterations and learning rate schedules to align the total number of sampled rays, allowing for a fair comparison of the models within each ray-budget group.
As shown in \cref{tab:dtu_benchmark}, SSDP-Facto consistently outperforms the baselines across all metrics in both ray-budget groups. 
For the surface reconstruction metrics, SSDP-Facto achieves the best performance among all models including the original NeuS and OaV under both the masked and unmasked protocols. 
Furthermore, the small gap between the unmasked and masked Chamfer distances indicates that SSDP-Facto effectively mitigates spurious surfaces outside the visual hull. 
For the uncertainty quantification metrics, SSDP-Facto also achieves the best performance among all models, except for ECE. 
Notably, for CRPS, SSDP-Facto yields a substantial improvement of approximately 1\,mm over NeuS-Facto, demonstrating the improved distributional quality of the learned first-passage-time distribution for each ray.
As a complementary analysis, the ablation study of the photometric losses in \cref{tab:dtu_ablation_photometric_losses} shows that the performance gains of SSDP-Facto over NeuS-Facto cannot be attributed to the NLML loss alone, but rather to its combination with our rendering formulation, as explained later in this section.
Per-scene results with the mean and standard deviation over scenes are provided in \cref{tab:dtu_per_scene_41m,tab:dtu_per_scene_154m}.
Visualization examples for a subset and all scenes are shown in \cref{fig:dtu_visualization_examples_subset,fig:dtu_visualization_examples_all}, respectively.

\begin{table}[t]
\centering
\caption{Evaluation results on the DTU dataset. \textbf{Bold} indicates the best results per metric within each ray-budget group. $^{*}$Reproduced with the official implementation. $^{\dagger}$Total number of rays sampled during training. $^{\ddagger}$Training time measured on a single NVIDIA H100. $^{\S}$Peak GPU memory allocated during training.}
\label{tab:dtu_benchmark}
{
\setlength{\tabcolsep}{2.0pt}
\begin{tabular}{l@{\hspace{-10pt}}rrrrrrrrr}
\toprule
& & & & \multicolumn{2}{c}{Surface Reconstruction} & \multicolumn{4}{c}{Uncertainty Quantification} \\
\cmidrule(lr){5-6} \cmidrule(lr){7-10}
& & & & \multicolumn{1}{c}{Unmasked} & \multicolumn{1}{c}{Masked} & & & & \\
\cmidrule(lr){5-5} \cmidrule(lr){6-6}
Model & Rays$^{\dagger}$ [M] & Time$^{\ddagger}$ [h] & Mem.$^{\S}$ [GB] & CD $\downarrow$ & CD $\downarrow$ & CRPS $\downarrow$ & MAE $\downarrow$ & ECE $\downarrow$ & Sharp. $\downarrow$ \\
\midrule
NeuS & 41 & 0.65 & 36.84 & 1.20 & 1.11 & 2.68 & 2.99 & 0.29 & 1.89 \\
\arrayrulecolor{black!25}
\midrule
\arrayrulecolor{black}
NeuS-Facto & 41 & 0.34 & 15.17 & 1.31 & 1.19 & 3.10 & 3.67 & 0.29 & 2.64 \\
UNIS-Facto (C1) & 41 & 0.34 & 15.17 & 1.34 & 1.23 & 3.03 & 3.53 & 0.28 & 2.31 \\
UNIS-Facto (C2) & 41 & 0.34 & 15.17 & 1.40 & 1.30 & 3.37 & 3.68 & 0.24 & 5.70 \\
UNIS-Facto (C3) & 41 & 0.34 & 15.17 & 1.32 & 1.22 & 3.11 & 3.63 & 0.29 & 2.28 \\
OaV-Facto & 41 & 0.35 & 15.17 & 1.49 & 1.33 & 4.35 & 5.55 & 0.34 & 4.14 \\
\rowcolor{DodgerBlue!10} 
SSDP-Facto & 41 & 0.38 & 15.18 & \textbf{1.02} & \textbf{0.96} & \textbf{1.98} & \textbf{2.28} & \textbf{0.15} & \textbf{1.47} \\
\midrule
NeuS & 154 & 2.62 & 36.84 & 0.98 & 0.89 & 2.50 & 2.77 & 0.31 & 1.84 \\
OaV$^{*}$ & 154 & --- & --- & 1.53 & 1.07 & 1.99 & 2.43 & \textbf{0.10} & 2.32 \\
\arrayrulecolor{black!25}
\midrule
\arrayrulecolor{black}
NeuS-Facto & 154 & 1.35 & 15.17 & 1.13 & 1.02 & 2.85 & 3.32 & 0.31 & 2.36 \\
UNIS-Facto (C1) & 154 & 1.35 & 15.17 & 1.12 & 1.05 & 2.79 & 3.22 & 0.30 & 2.03 \\
UNIS-Facto (C2) & 154 & 1.35 & 15.17 & 1.19 & 1.11 & 2.96 & 3.16 & 0.25 & 5.21 \\
UNIS-Facto (C3) & 154 & 1.35 & 15.17 & 1.16 & 1.03 & 2.85 & 3.28 & 0.30 & 2.02 \\
OaV-Facto & 154 & 1.44 & 15.17 & 1.25 & 1.10 & 4.62 & 5.79 & 0.36 & 4.03 \\
\rowcolor{DodgerBlue!10} 
SSDP-Facto & 154 & 1.55 & 15.18 & \textbf{0.88} & \textbf{0.84} & \textbf{1.86} & \textbf{2.17} & 0.17 & \textbf{1.49} \\
\bottomrule
\end{tabular}
}
\end{table}

\begin{figure}[t]

\centering

\begin{minipage}[b]{0.245\linewidth}
\centering
\includegraphics[width=1.0\linewidth]{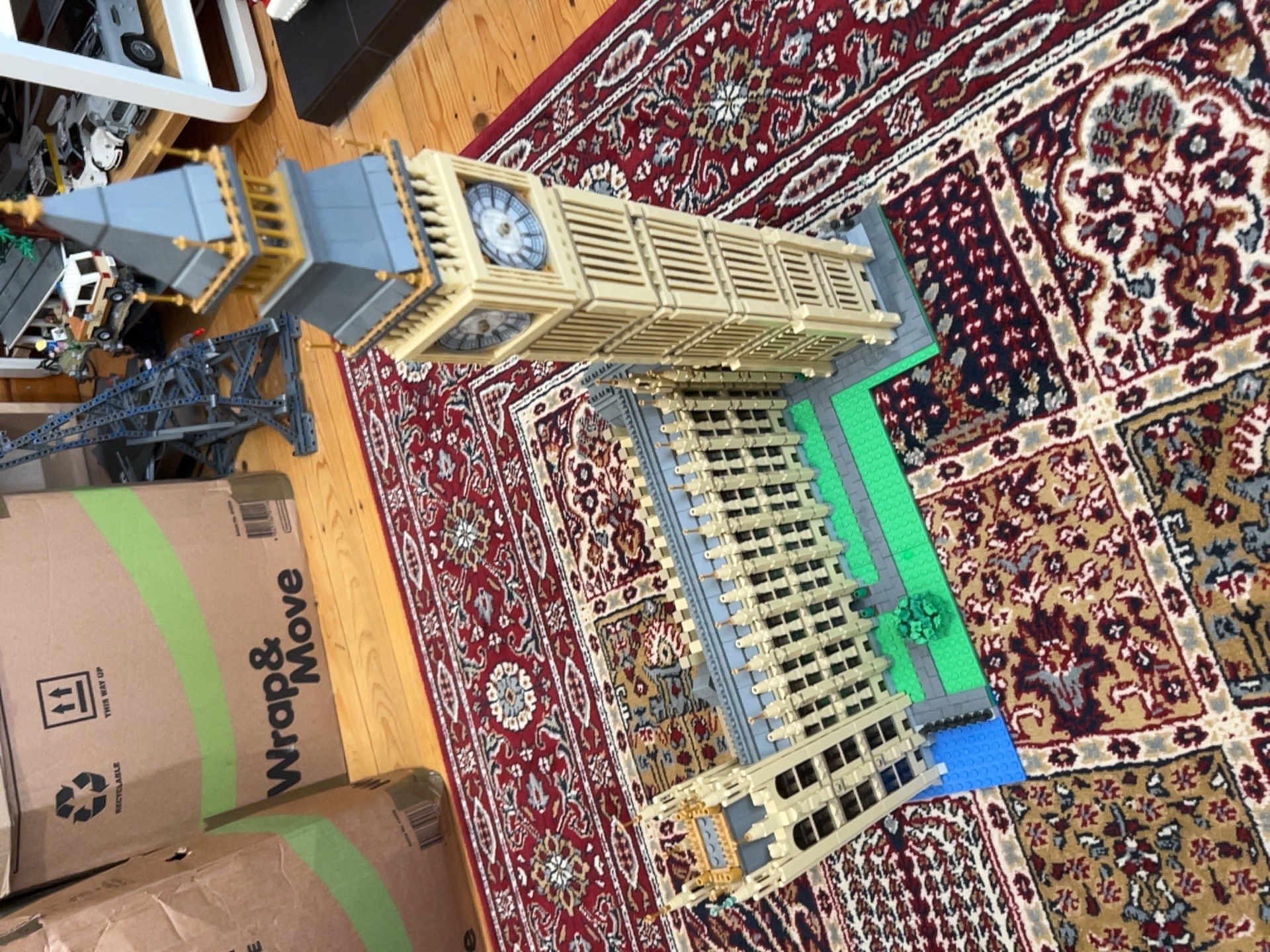}
\end{minipage}
\begin{minipage}[b]{0.245\linewidth}
\centering
\includegraphics[width=1.0\linewidth]{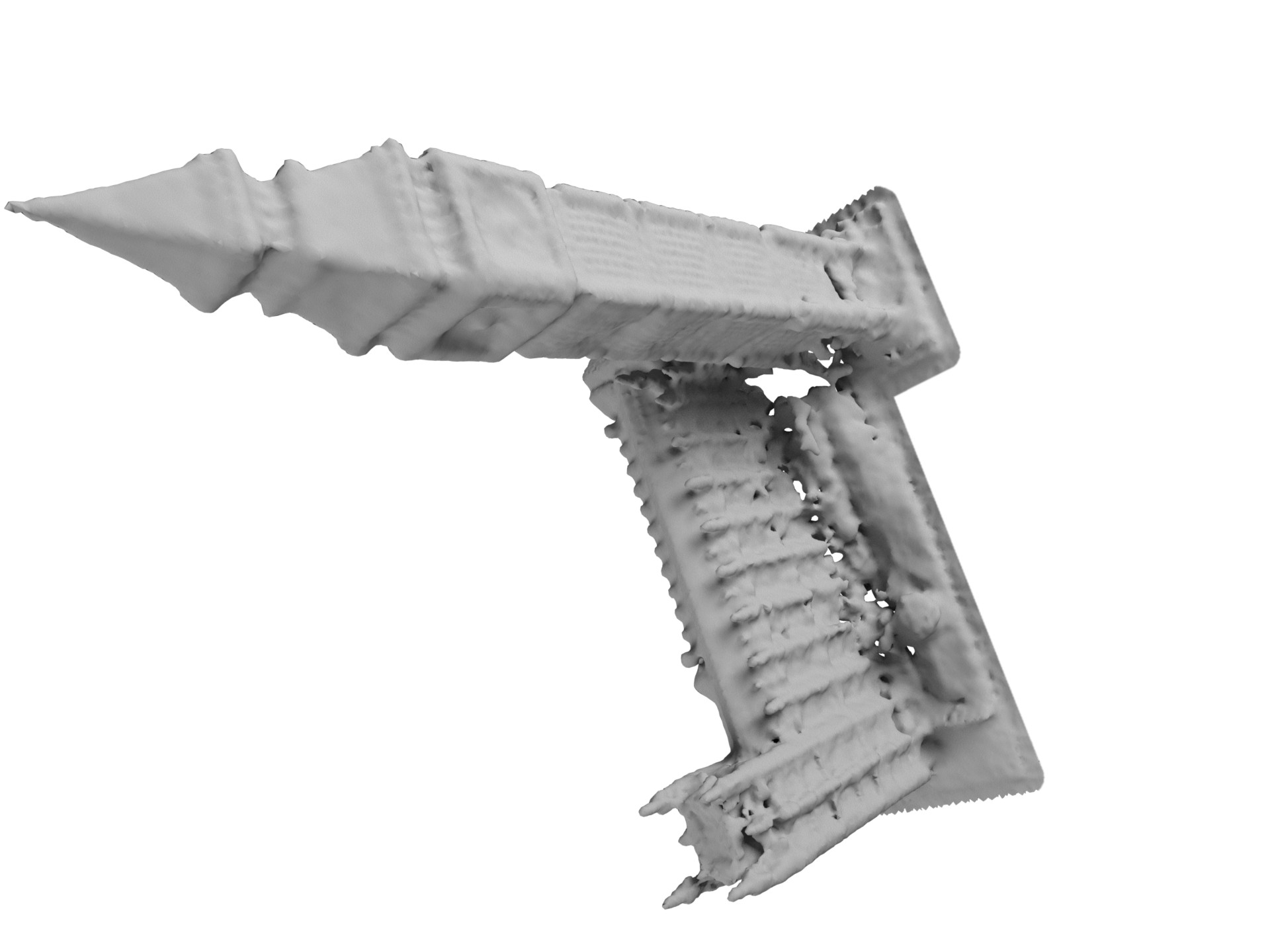}
\end{minipage}
\begin{minipage}[b]{0.245\linewidth}
\centering
\includegraphics[width=1.0\linewidth]{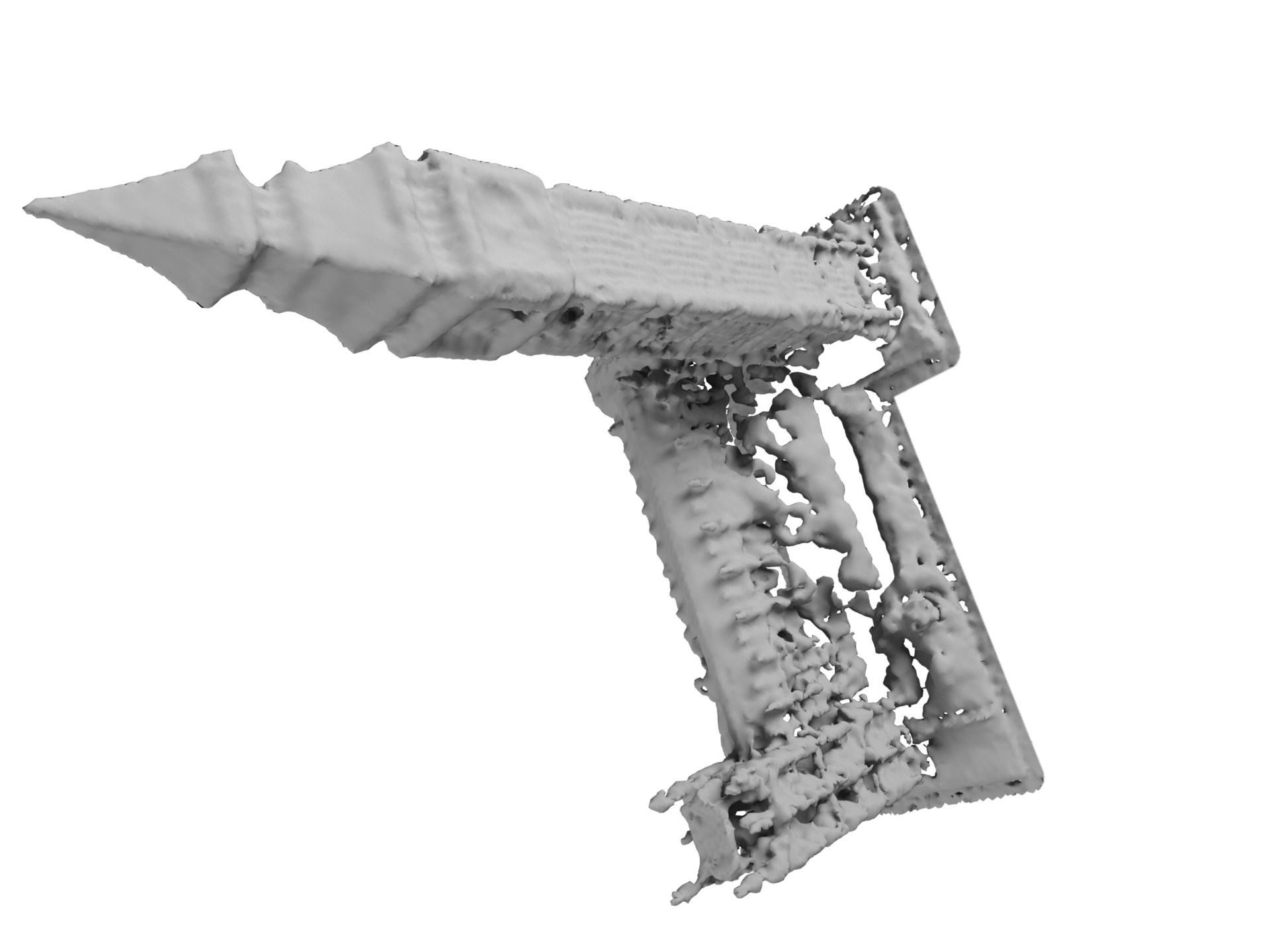}
\end{minipage}
\begin{minipage}[b]{0.245\linewidth}
\centering
\includegraphics[width=1.0\linewidth]{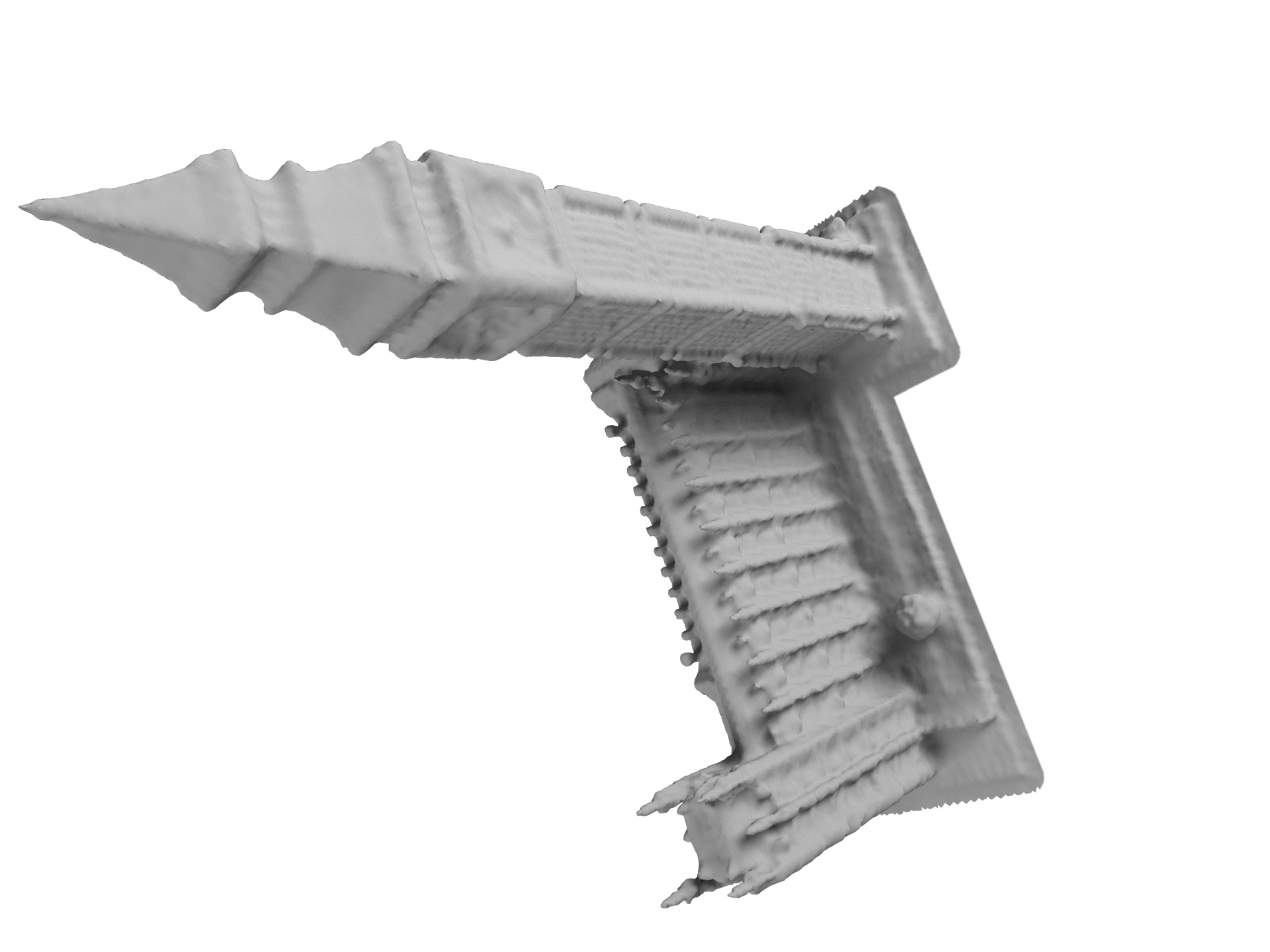}
\end{minipage}

\begin{minipage}[b]{0.245\linewidth}
\centering
\includegraphics[width=1.0\linewidth]{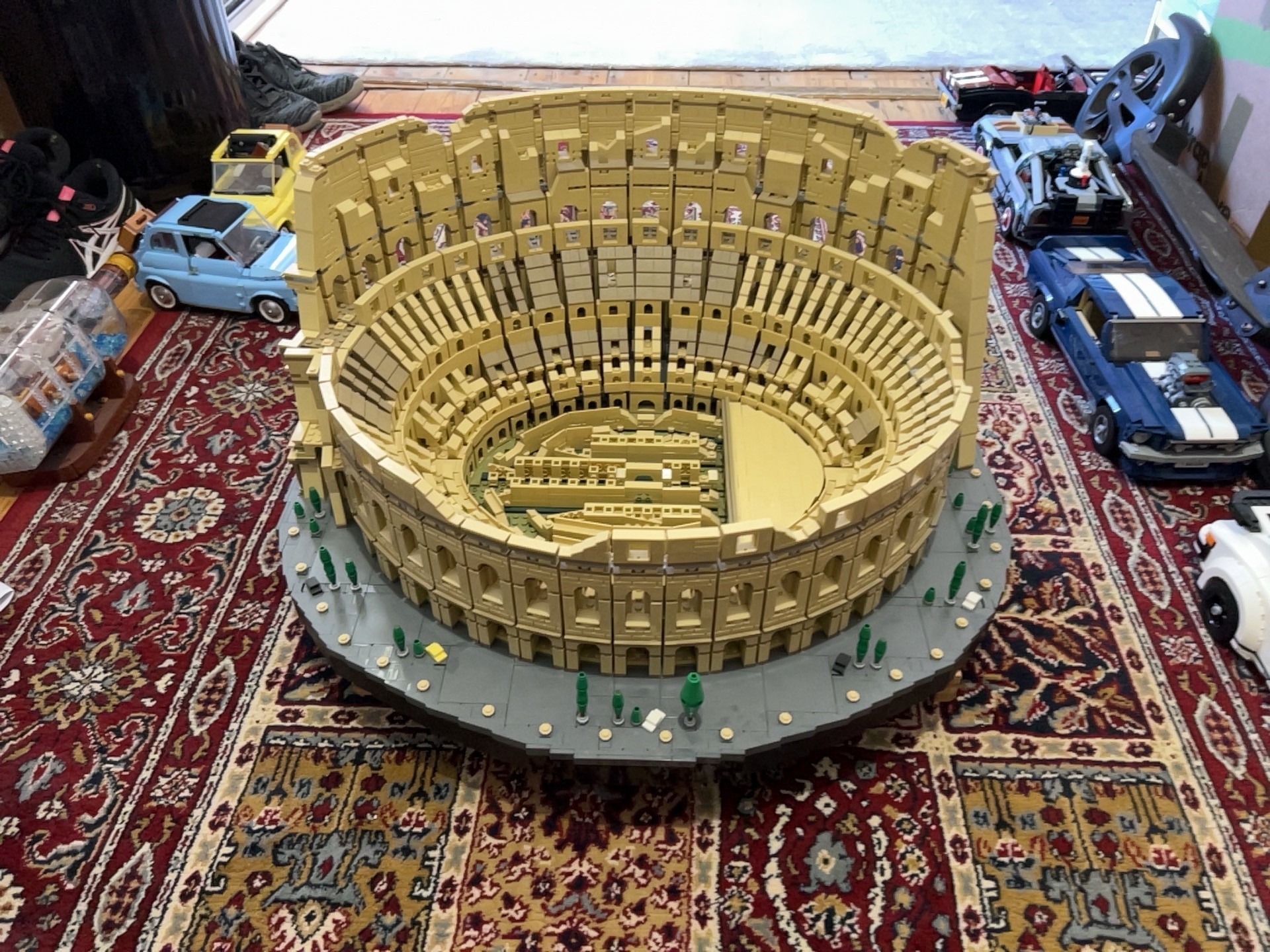}
\subcaption{Ground Truth}
\end{minipage}
\begin{minipage}[b]{0.245\linewidth}
\centering
\includegraphics[width=1.0\linewidth]{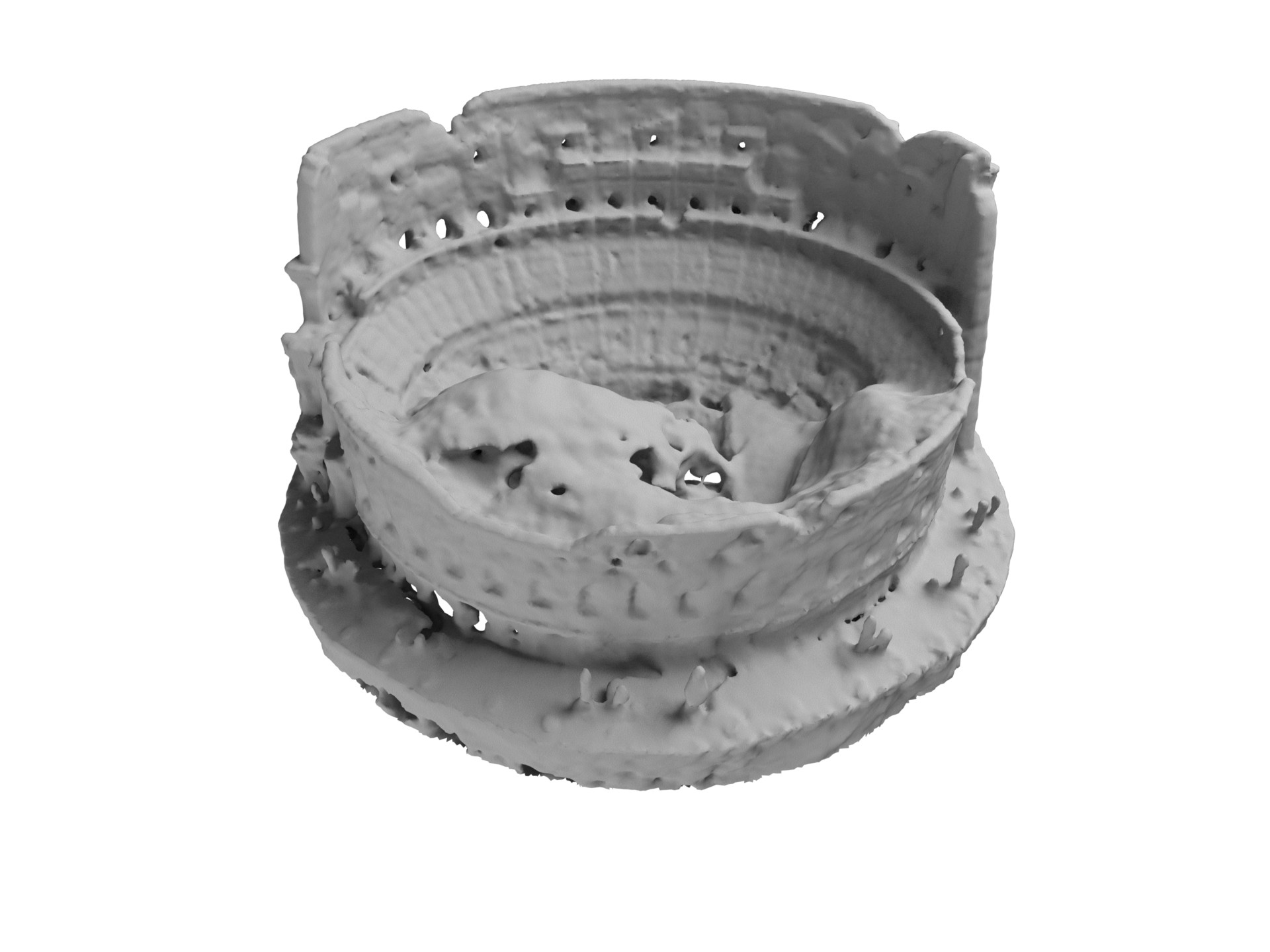}
\subcaption{NeuS-Facto}
\end{minipage}
\begin{minipage}[b]{0.245\linewidth}
\centering
\includegraphics[width=1.0\linewidth]{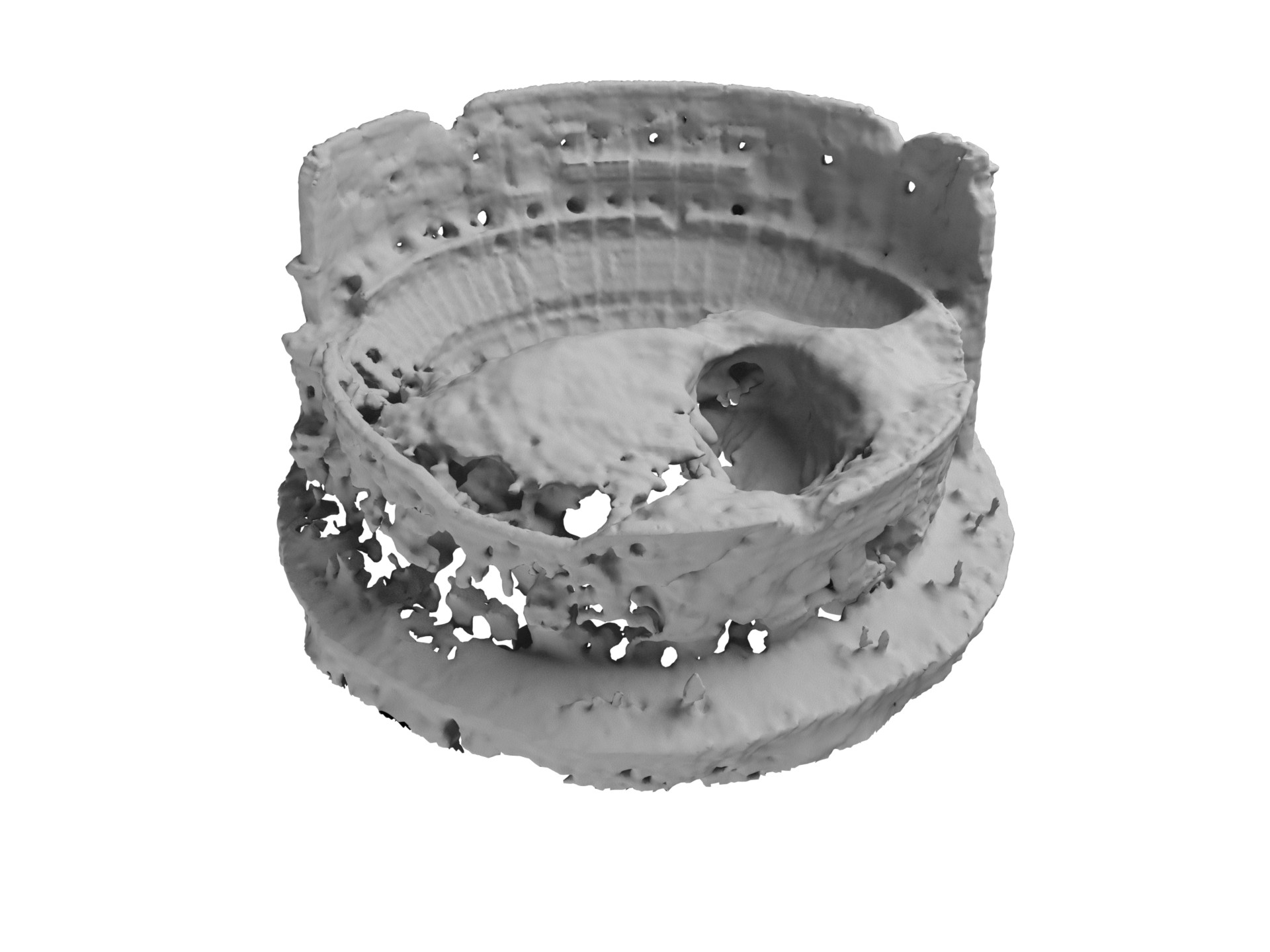}
\subcaption{OaV-Facto}
\end{minipage}
\begin{minipage}[b]{0.245\linewidth}
\centering
\includegraphics[width=1.0\linewidth]{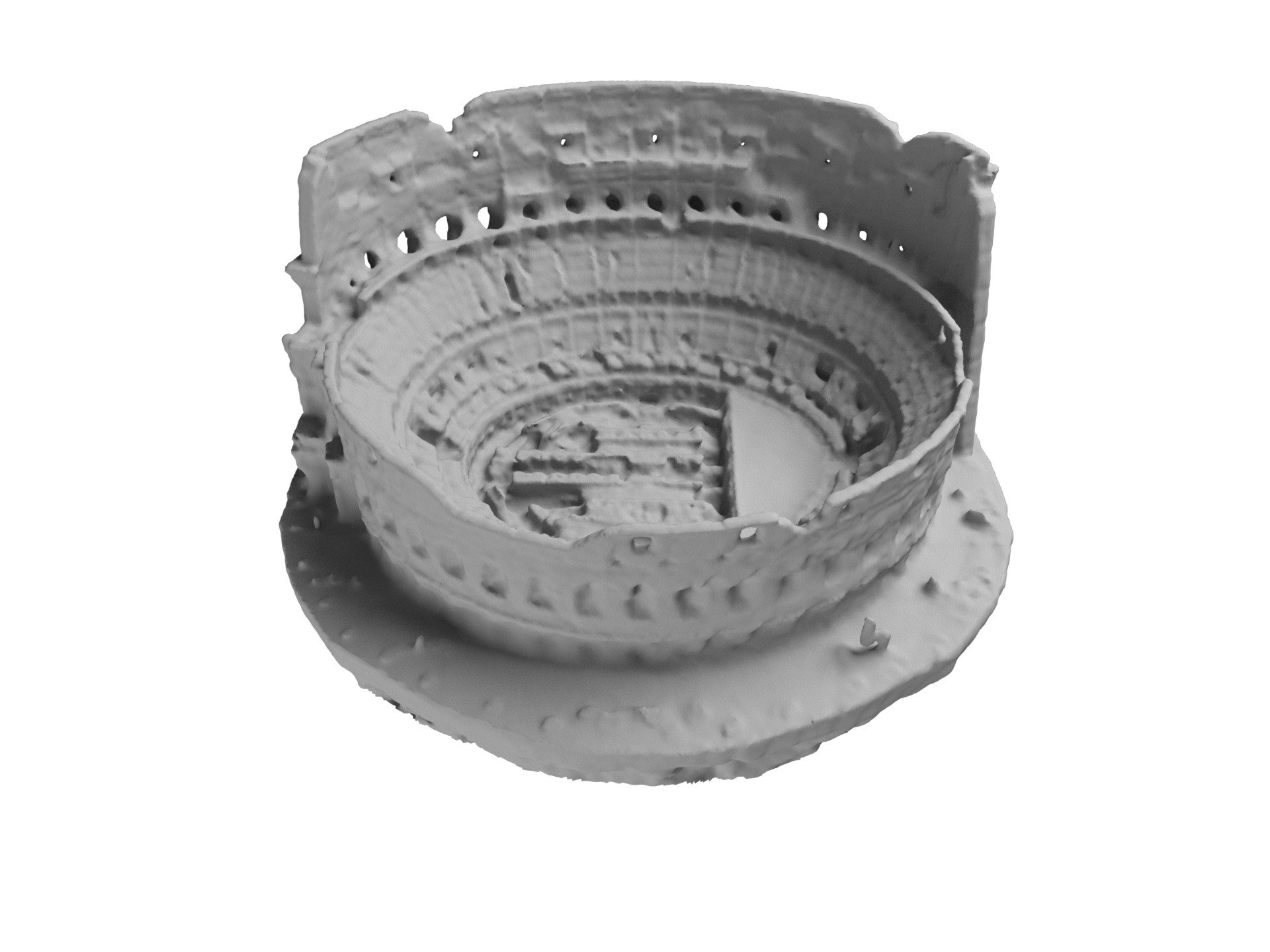}
\subcaption{SSDP-Facto}
\end{minipage}

\caption{Visualization examples on the MobileBrick dataset.}
\label{fig:lego_visualization_examples_subset}
\end{figure}

\paragraph{Comparisons on the MobileBrick dataset.}
To verify the broader applicability of our method, we also conduct comparisons on the MobileBrick dataset.
As summarized in \cref{tab:lego_benchmark}, we observe a consistent trend with the DTU dataset, where SSDP-Facto outperforms the baselines across all metrics except ECE, demonstrating that the effectiveness of our method is not limited to a specific dataset but is broadly applicable to diverse scenes.
Per-scene results with the mean and standard deviation over scenes are provided in \cref{tab:lego_per_scene}.
Visualization examples for a subset and all scenes are shown in \cref{fig:lego_visualization_examples_subset,fig:lego_visualization_examples_all}, respectively.

\begin{table}[t]
\centering
\caption{Evaluation results on the MobileBrick dataset. \textbf{Bold} indicates the best results per metric.}
\label{tab:lego_benchmark}
{
\setlength{\tabcolsep}{3.0pt}
\begin{tabular}{lrrrrrrrrrr}
\toprule
& \multicolumn{6}{c}{Surface Reconstruction} & \multicolumn{4}{c}{Uncertainty Quantification} \\
\cmidrule(lr){2-7} \cmidrule(lr){8-11}
& \multicolumn{3}{c}{Unmasked} & \multicolumn{3}{c}{Masked} & & & & \\
\cmidrule(lr){2-4} \cmidrule(lr){5-7}
Model & CD $\downarrow$ & F1@2.5 $\uparrow$ & F1@5 $\uparrow$ & CD $\downarrow$ & F1@2.5 $\uparrow$ & F1@5 $\uparrow$ & CRPS $\downarrow$ & MAE $\downarrow$ & ECE $\downarrow$ & Sharp. $\downarrow$ \\
\midrule
NeuS & 6.28 & 0.68 & 0.85 & 6.10 & 0.69 & 0.86 & 4.62 & 5.16 & 0.28 & 2.77 \\
\arrayrulecolor{black!25}
\midrule
\arrayrulecolor{black}
NeuS-Facto & 5.64 & 0.70 & 0.87 & 5.53 & 0.70 & 0.88 & 4.46 & 5.61 & 0.24 & 4.77 \\
UNIS-Facto (C1) & 6.34 & 0.67 & 0.86 & 6.02 & 0.68 & 0.87 & 4.78 & 5.90 & 0.22 & 4.55 \\
UNIS-Facto (C2) & 5.27 & 0.68 & 0.87 & 5.42 & 0.68 & 0.87 & 4.29 & 5.47 & \textbf{0.19} & 10.50 \\
UNIS-Facto (C3) & 6.23 & 0.69 & 0.86 & 5.90 & 0.69 & 0.87 & 4.81 & 5.88 & 0.22 & 4.14 \\
OaV-Facto & 7.09 & 0.65 & 0.84 & 6.58 & 0.66 & 0.85 & 5.12 & 6.35 & 0.25 & 4.17 \\
\rowcolor{DodgerBlue!10} 
SSDP-Facto & \textbf{4.97} & \textbf{0.72} & \textbf{0.89} & \textbf{4.90} & \textbf{0.72} & \textbf{0.89} & \textbf{4.17} & \textbf{4.56} & 0.23 & \textbf{1.80} \\
\bottomrule
\end{tabular}
}
\end{table}

\paragraph{Ablation study of the negative-absorbing approximation.}
To investigate the impact of the negative-absorbing approximation introduced in \cref{sec:method/first_passage_probability/negative_absorbing_approximation}, we compare the performance with and without this approximation on the DTU dataset. 
As shown in \cref{tab:dtu_ablation_negative_absorbing_approx}, the negative-absorbing approximation achieves an approximately twofold speedup in training time by eliminating the reliance on Bayesian filtering.
Interestingly, it yields a slight improvement in the surface reconstruction metrics, suggesting that it serves as an effective inductive bias in practice.
Conversely, the full Bayesian formulation demonstrates better performance in all uncertainty quantification metrics except sharpness, indicating that propagating the survival-conditioned distribution via Bayesian filtering improves the distributional quality of the learned first-passage-time distribution for each ray.
However, this improvement is marginal---for instance, only about 0.1\,mm for CRPS---and does not justify the approximately doubled training time.
Therefore, we adopt the negative-absorbing approximation as our default configuration for practical applications.
\cref{sec:approximation_error_validation} further analyzes, on synthetic scenes that deliberately violate each assumption in \cref{eq:survival_event_approx}, when and why the approximation error arises during training and how it affects the learned geometry.

\begin{table}[t]
\centering
\caption{Ablation study of the negative-absorbing approximation (\cref{sec:method/first_passage_probability/negative_absorbing_approximation}) on the DTU dataset. $^{\ddagger}$Training time measured on a single NVIDIA H100. $^{\S}$Peak GPU memory allocated during training.}
\label{tab:dtu_ablation_negative_absorbing_approx}
{
\setlength{\tabcolsep}{3pt}
\begin{tabular}{crrrrrrrr}
\toprule
& & & \multicolumn{2}{c}{Surface Reconstruction} & \multicolumn{4}{c}{Uncertainty Quantification} \\
\cmidrule(lr){4-5} \cmidrule(lr){6-9}
& & & \multicolumn{1}{c}{Unmasked} & \multicolumn{1}{c}{Masked} & & & & \\
\cmidrule(lr){4-4} \cmidrule(lr){5-5}
Neg.-Abs. Approx. & Time$^{\ddagger}$ [h] & Mem.$^{\S}$ [GB] & CD $\downarrow$ & CD $\downarrow$ & CRPS $\downarrow$ & MAE $\downarrow$ & ECE $\downarrow$ & Sharp. $\downarrow$ \\
\midrule
\xmark & 0.66 & 15.18 & 1.04 & 1.00 & \textbf{1.89} & \textbf{2.19} & \textbf{0.10} & 1.57 \\
\cmark & 0.38 & 15.18 & \textbf{1.02} & \textbf{0.96} & 1.98 & 2.28 & 0.15 & \textbf{1.47} \\
\bottomrule
\end{tabular}
}
\end{table}

\paragraph{Ablation study of the photometric losses.}
To verify the effectiveness of the NLML loss $\mathcal{L}_{\mathrm{nlml}}$ (\ref{eq:nlml_loss}) and assess whether the superiority of SSDP-Facto shown in \cref{tab:dtu_benchmark} can be attributed to the NLML loss alone, we compare it with the conventional expectation loss $\mathcal{L}_{\mathrm{exp}}$ (\ref{eq:exp_loss}) commonly used in prior works \citep{idr,neus}.
As demonstrated in \cref{tab:dtu_ablation_photometric_losses}, applying the NLML loss to NeuS-Facto improves surface reconstruction, but degrades uncertainty quantification, except for sharpness.
In contrast, applying the NLML loss to SSDP-Facto yields substantial improvements across all metrics except ECE.
Furthermore, when both models are trained with the NLML loss, SSDP-Facto outperforms NeuS-Facto across all metrics.
These results indicate that the performance gains of SSDP-Facto over NeuS-Facto cannot be attributed to the NLML loss alone, but rather to its combination with our rendering formulation.

\begin{table}[t]
\centering
\caption{Ablation study of the photometric losses (\cref{sec:method/loss_functions}) on the DTU dataset.}
\label{tab:dtu_ablation_photometric_losses}
{
\begin{tabular}{lccrrrrrr}
\toprule
& & & \multicolumn{2}{c}{Surface Reconstruction} & \multicolumn{4}{c}{Uncertainty Quantification} \\
\cmidrule(lr){4-5} \cmidrule(lr){6-9}
& & & \multicolumn{1}{c}{Unmasked} & \multicolumn{1}{c}{Masked} & & & & \\
\cmidrule(lr){4-4} \cmidrule(lr){5-5}
Model & $\mathcal{L}_{\mathrm{exp}}$ & $\mathcal{L}_{\mathrm{nlml}}$ & CD $\downarrow$ & CD $\downarrow$ & CRPS $\downarrow$ & MAE $\downarrow$ & ECE $\downarrow$ & Sharp. $\downarrow$ \\
\midrule
\multirow{2}{*}{NeuS-Facto} & \cmark & \xmark & 1.31 & 1.19 & 3.10 & 3.67 & 0.29 & 2.64 \\
& \xmark & \cmark & 1.21 & 1.13 & 3.64 & 3.70 & 0.34 & 1.57 \\
\midrule
\multirow{2}{*}{SSDP-Facto} & \cmark & \xmark & 1.49 & 1.44 & 2.48 & 3.27 & \textbf{0.13} & 3.95 \\
& \xmark & \cmark & \textbf{1.02} & \textbf{0.96} & \textbf{1.98} & \textbf{2.28} & 0.15 & \textbf{1.47} \\
\bottomrule
\end{tabular}
}
\end{table}

\section{Conclusion}
\label{sec:conclusion}

In this paper, we introduced Stochastic Signed Distance Processes (SSDP), which model the SDF along each ray as a stochastic process.
Building on SSDP, we derived the first-passage probability for each sampling interval based on Bayesian filtering, together with its practical approximation for parallel rendering.
We also showed that NeuS arises as a special case of our formulation, clarifying the connection between existing SDF-based volume rendering and our probabilistic surface rendering.
Our experiments highlight that our method improves both surface reconstruction and uncertainty quantification over controlled baselines on the DTU and MobileBrick datasets.
The ablation study of the negative-absorbing approximation further shows that it provides a practical accuracy--efficiency trade-off, while the full Bayesian formulation improves overall performance in uncertainty quantification at a higher computational cost.
In addition, the ablation study of the photometric losses shows that the NLML loss improves overall performance in both surface reconstruction and uncertainty quantification when combined with our rendering formulation, indicating that the performance gains over the baseline cannot be attributed to the NLML loss alone.

A limitation of our method is the reliance on the negative-absorbing approximation for parallel rendering.
While this approximation performed comparably to the full Bayesian formulation in our experiments, and the analysis in \cref{sec:approximation_error_validation} on synthetic scenes that deliberately violate each assumption in \cref{eq:survival_event_approx} shows that the approximation error arises only in the early stage of training and does not lead to a large difference in the learned geometry, the full Bayesian formulation may be preferable in scenes where rays frequently exhibit multiple downward zero-crossings through surfaces thinner than the sampling interval.

Another limitation is that SSDP is formulated as a ray-wise stochastic process rather than a random field over $\mathbb{R}^{3}$.
To obtain the uncertainty of the 3D geometry itself shared across viewpoints, the SDF should ideally be modeled as a single random field whose posterior distribution is inferred from all observations.
However, this requires the joint distribution of the first-passage times over all rays, which is intractable because the Markov property of SSDP holds only along each individual ray.
Since SSDP marginalizes out the first-passage time of each ray independently, the learned variance is that of each ray rather than that of the shared 3D geometry, even if the ray-wise processes are defined as restrictions of a single random field.
One promising direction for future work is therefore to overcome this limitation by approximately incorporating correlations across rays.
Once the uncertainty of the shared 3D geometry becomes available, the next viewpoint to observe could be selected as the one expected to reduce this uncertainty the most.
Such active surface reconstruction would lead to accurate geometry with fewer views and lower training cost.

\subsubsection*{Broader Impact Statement}
This statement discusses the dual-use, privacy, and environmental aspects of our work.
Regarding dual use, as with 3D reconstruction in general, the same reconstruction capability that supports content creation and digital twins can be misused to reconstruct people or private spaces from images without consent.
Regarding privacy, the outputs of our method can themselves constitute personal data: the learned SDF, and any mesh extracted from it, can carry personally identifiable information, such as the geometry of a face, while the color field and the background NeRF effectively retain the appearance of the scene, including its surroundings, and allow it to be re-rendered from arbitrary viewpoints.
Reconstructed meshes and model checkpoints should therefore be handled under the same data-protection policies as the source images.
Since our method is agnostic to scene content, it offers no technical mechanism against such misuse; mitigation is therefore procedural: images of people or private spaces should be captured and used only with consent and in compliance with applicable regulations.
Our experiments involve no personal data, as the DTU and MobileBrick datasets contain only inanimate objects.
Regarding environmental impact, our method is a per-scene optimization on a single GPU and involves no large-scale pretraining, so its footprint is orders of magnitude smaller than that of training recent large-scale models.

\subsubsection*{Acknowledgments}
This work was supported by JST BOOST, Japan Grant Number JPMJBS2430.

\bibliography{main}
\bibliographystyle{tmlr}

\appendix

\section{Proof of \cref{prop:crossing_probability_up_integrand}}
\label{sec:crossing_probability_up_integrand_proof}

We restate the solution (\ref{eq:ou_solution}) of the stochastic differential equation for the OU process (\ref{eq:ou_sde}) as follows:
\begin{align*}
R(t) 
& = \Psi(t_{i}, t) (R(t_{i}) + M(t)) ,
\end{align*}
where 
\begin{align*}
\Psi(s, t) 
& \coloneq \exp(-\int_{s}^{t} \kappa(u) \, \mathrm{d} u)
\end{align*}
and 
\begin{align*}
M(t) 
& \coloneq \int_{t_{i}}^{t} \Psi(s, t_{i}) \tau(s) \, \mathrm{d} W(s)
\end{align*}
is an It\^o integral and hence a continuous local martingale.
Throughout this section, we only consider $t \ge t_{i}$ and omit this restriction for brevity.
By the Dambis--Dubins--Schwarz theorem \citep{dds} applied to $M(t)$, there exists a Wiener process $B$ such that, up to indistinguishability:
\begin{align*}
M(t) 
& = B(\Theta(t)) ,
\end{align*}
where 
\begin{align*}
\Theta(t) 
& \coloneq \int_{t_{i}}^{t} \Psi(s, t_{i})^{2} \tau(s)^{2} \, \mathrm{d} s
\end{align*}
denotes the quadratic variation of $M(t)$.
Therefore, $S(t)$ can be rewritten as:
\begin{align*}
S(t) 
& = R(t) + \mu(t) \\
& = \Psi(t_{i}, t) X(t) ,
\end{align*}
where 
\begin{align*}
X(t) 
& \coloneq R(t_{i}) + \eta(t) + B(\Theta(t))
\end{align*}
and
\begin{align*}
\eta(t) 
& \coloneq \Psi(t_{i}, t)^{-1} \mu(t) .
\end{align*}
Note that $\Psi(t_{i}, t) > 0$ ensures $S(t)$ and $X(t)$ have the same zero-crossings on $(t_{i}, t_{i+1})$:
\begin{align*}
\inf_{t \in (t_{i}, t_{i+1})} S(t) \le 0 
& \iff \inf_{t \in (t_{i}, t_{i+1})} X(t) \le 0 .
\end{align*}
Here, switching the \textit{clock} from $t$ to $\omega \coloneq \Theta(t)$ yields:
\begin{align*}
Y(\omega) 
& \coloneq X(\Theta^{-1}(\omega)) \\
& = R(t_{i}) + \eta(\Theta^{-1}(\omega)) + B(\omega) ,
\end{align*}
where $\Theta(t)$ is strictly increasing and thus its inverse $\Theta^{-1}(\omega)$ can be defined.
Given $a > 0$ and $b > 0$, conditioning on $S(t_{i}) = a$ and $S(t_{i+1}) = b$ yields:
\begin{align*}
\{S(t_{i}) = a, S(t_{i+1}) = b\} 
& \iff \{X(t_{i}) = a, X(t_{i+1}) = b'\} \\
& \iff \{Y(0) = a, Y(\Omega_{i}) = b'\} ,
\end{align*}
where $\Omega_{i} \coloneq \Theta(t_{i+1})$, $\Psi_{i} \coloneq \Psi(t_{i}, t_{i+1})$, and $b' \coloneq \Psi_{i}^{-1} b$.
Under this conditioning, $B(\Omega_{i})$ is given by:
\begin{align}
\label{eq:endpoint_condition}
B(\Omega_{i}) 
& = (b' - a) - (\eta(t_{i+1}) - \eta(t_{i})) \nonumber \\
& \eqqcolon c .
\end{align}
Therefore, $Y(\omega)$ can be rewritten as:
\begin{align}
\label{eq:non_linear_brownian_bridge}
Y({\omega}) \mid \{Y(0) = a, Y(\Omega_{i}) = b'\} 
& = R(t_{i}) + \eta(\Theta^{-1}(\omega)) + B_{c}(\omega) ,
\end{align}
where 
\begin{align}
\label{eq:generalized_brownian_bridge}
B_{c}({\omega})  
& \coloneq B(\omega) \mid \{B(\Omega_{i}) = c\} \nonumber \\
& \stackrel{d}{=} B_{0}(\omega) + \frac{\omega}{\Omega_{i}} c
\end{align}
denotes a generalized Brownian bridge from $0$ to $c$ and
\begin{align}
\label{eq:standardized_brownian_bridge}
B_{0}({\omega})  
& \coloneq B(\omega) \mid \{B(\Omega_{i}) = 0\} \nonumber \\
& \stackrel{d}{=} B(\omega) - \frac{\omega}{\Omega_{i}} B(\Omega_{i})
\end{align}
denotes the standardized Brownian bridge from $0$ to $0$.
Combining  \cref{eq:endpoint_condition,eq:non_linear_brownian_bridge,eq:generalized_brownian_bridge,eq:standardized_brownian_bridge} yields:
\begin{align*}
Y({\omega}) \mid \{Y(0) = a, Y(\Omega_{i}) = b'\} 
& \stackrel{d}{=} a + B_{0}(\omega) + \frac{\omega}{\Omega_{i}} (b' - a) + \delta(\omega) \\
& \stackrel{d}{=} a + B_{c'}(\omega) + \delta(\omega) ,
\end{align*}
where $c' \coloneq b' - a$ and 
\begin{align*}
\delta(\omega) 
& = h(\omega) - ((1 - \omega / \Omega_{i}) \ h(0) + (\omega / \Omega_{i}) \ h(\Omega_{i}))
\end{align*}
denotes the residual between $h(\omega) \coloneq \eta(\Theta^{-1}(\omega))$ and its linear interpolation.
Assuming that $h(\omega)$ is twice continuously differentiable on $[0,\Omega_{i}]$, we have $\sup_{\zeta \in [0,\Omega_i]} |h''(\zeta)| < \infty$.
Therefore, applying the Lagrange form of the linear interpolation error to $h(\omega)$ yields:
\begin{align*}
|\delta(\omega)| 
& \le \frac{\Omega_{i}^{2}}{8} \sup_{\zeta \in [0, \Omega_{i}]}|h''(\zeta)| 
= O(\Omega_{i}^{2}) .
\end{align*}
Since the integrand $\Psi(s, t_{i})^{2} \tau(s)^{2}$ in $\Theta(t_{i+1})$ is bounded on $s \in [t_{i}, t_{i+1}]$, we have $\Omega_{i} \coloneq \Theta(t_{i+1}) = O(\Delta t_{i})$ and hence $|\delta(\omega)| = O(\Delta t_{i}^{2})$, where $\Delta t_{i} \coloneq t_{i+1} - t_{i}$.
Neglecting the higher-order residual $\delta(\omega)$ yields:
\begin{align*}
Y(\omega) \mid \{Y(0) = a, Y(\Omega_{i}) = b'\} 
& \stackrel{d}{\approx} a + B_{c'}(\omega) \\
& \stackrel{d}{\approx} a + B(\omega) \mid \{B(\Omega_{i}) = c'\} .
\end{align*}
From the above, $P(\mathcal{B}_{i}^{\uparrow} \mid S(t_{i}) = a, S(t_{i+1}) = b)$ is given by:
\begin{align*}
P(\mathcal{B}_{i}^{\uparrow} \mid S(t_{i}) = a, S(t_{i+1}) = b) 
& = P \left ( \left \{ \inf_{t \in (t_{i}, t_{i+1})} S(t) \le 0 \right \} \ \middle | \ S(t_{i}) = a, S(t_{i+1}) = b \right ) \\
& = P \left ( \left \{ \inf_{t \in (t_{i}, t_{i+1})} X(t) \le 0 \right \} \ \middle | \ X(t_{i}) = a, X(t_{i+1}) = b' \right ) \\
& = P \left ( \left \{ \inf_{\omega \in (0, \Omega_{i})} Y(\omega) \le 0 \right \} \ \middle | \ Y(0) = a, Y(\Omega_{i}) = b' \right ) \\
& \approx P \left ( \left \{ \inf_{\omega \in (0, \Omega_{i})} B(\omega) \le -a \right \} \ \middle | \ B(\Omega_{i}) = c' \right ) .
\end{align*}
Finally, applying the reflection principle, which is stated in \cref{lemma:reflection_principle}, to $B(\omega)$, yields:
\begin{align*}
P(\mathcal{B}_{i}^{\uparrow} \mid S(t_{i}) = a, S(t_{i+1}) = b)
& \approx P \left ( \left \{ \inf_{\omega \in (0, \Omega_{i})} B(\omega) \le -a \right \} \ \middle | \ B(\Omega_{i}) = c' \right ) \\
& \approx \frac{P \left ( \left \{ \inf_{\omega \in (0, \Omega_{i})} B(\omega) \le -a \right \}, B(\Omega_{i}) \in \mathrm{d} c' \right )}{P \left ( B(\Omega_{i}) \in \mathrm{d} c' \right )} \\
& \approx \frac{\varphi(-2a - c'; 0, \Omega_{i})}{\varphi(c'; 0, \Omega_{i})} \\
& \approx \exp \left ( - \frac{2 a b'}{\Omega_{i}} \right ) \\
& \approx \exp \left ( - \frac{2 a b}{\Omega_{i} \Psi_{i}} \right ) .
\end{align*}

\begin{lemma}[Reflection Principle]
\label{lemma:reflection_principle}
Given a crossing level $\ell < 0$ and an endpoint $y > \ell$, the joint probability that a Wiener process $W_{t}$ crosses $\ell$ downward at $t \in (0, T)$ and takes $y$ at $t = T$ is given by:
\begin{align*}
P \left ( \left \{\inf_{t \in (0, T)} W_{t} \le \ell \right \}, W_{T} \in \mathrm{d} y \right ) 
& = P(W_{T} \in \mathrm{d} (2\ell - y)) \\
& = f_{W_{T}}(2\ell - y) \, \mathrm{d} y \\
& = \varphi(2\ell - y; 0, T) \, \mathrm{d} y ,
\end{align*}
where $f_{W_{t}}(x) \coloneq \varphi(x; 0, t)$ denotes the Gaussian PDF for $W_{t}$.
\end{lemma}

\section{Proof of \cref{coro:crossing_probability_posterior_up}}
\label{sec:crossing_probability_posterior_up_proof}

By conditioning on $S_{i-1}$, $P(\mathcal{B}_{i-1}^{\uparrow} \mid s_{i}, \mathcal{A}_{i-1})$ can be rewritten as:
\begin{align}
\label{eq:appendix/a2/01}
P(\mathcal{B}_{i-1}^{\uparrow} \mid s_{i}, \mathcal{A}_{i-1}) 
& = \int_{0}^{\infty} P(\mathcal{B}_{i-1}^{\uparrow} \mid s_{i-1}, s_{i}) f_{S_{i-1} \mid S_{i}, \mathcal{A}_{i-1}}(s_{i-1} \mid s_{i}) \, \mathrm{d} s_{i-1} .
\end{align}
$P(\mathcal{B}_{i-1}^{\uparrow} \mid s_{i-1}, s_{i})$ is given by \cref{prop:crossing_probability_up_integrand}:
\begin{align}
\label{eq:appendix/a2/02}
P(\mathcal{B}_{i-1}^{\uparrow} \mid s_{i-1}, s_{i})
& \approx \mathds{1}(s_{i} > 0) \exp \left ( - \frac{2 s_{i-1} s_{i}}{\Omega_{i-1} \Psi_{i-1}} \right ) .
\end{align}
$f_{S_{i-1} \mid S_{i}, \mathcal{A}_{i-1}}(s_{i-1} \mid s_{i})$ can be rewritten based on Bayes' theorem:
\begin{align}
\label{eq:appendix/a2/03}
f_{S_{i-1} \mid S_{i}, \mathcal{A}_{i-1}}(s_{i-1} \mid s_{i}) 
& = \frac{f_{S_{i} \mid S_{i-1}}(s_{i} \mid s_{i-1}) f_{S_{i-1} \mid \mathcal{A}_{i-1}}(s_{i-1})}{\int_{0}^{\infty} f_{S_{i} \mid S_{i-1}}(s_{i} \mid s_{i-1}) f_{S_{i-1} \mid \mathcal{A}_{i-1}}(s_{i-1}) \, \mathrm{d} s_{i-1}} .
\end{align}
We replace the intractable posterior $f_{S_{i-1} \mid \mathcal{A}_{i-1}}$ with a truncated Gaussian PDF $\varphi_{+}(\cdot; \hat{\mu}_{i-1}, \hat{\sigma}_{i-1}^{2})$ obtained by Bayesian updating of the prior $f_{S_{i-1} \mid \mathcal{A}_{i-2}}(\cdot) \approx \varphi(\cdot; \hat{\mu}_{i-1}, \hat{\sigma}_{i-1}^{2})$ with an observation $\{S_{i-1} > 0\}$, yielding:
\begin{align}
\label{eq:appendix/a2/04}
& f_{S_{i} \mid S_{i-1}}(s_{i} \mid s_{i-1}) f_{S_{i-1} \mid \mathcal{A}_{i-1}}(s_{i-1}) \nonumber \\
\approx & \ \varphi(s_{i}; \alpha_{i-1} s_{i-1} + \beta_{i-1}, \gamma_{i-1}) \varphi_{+}(s_{i-1}; \hat{\mu}_{i-1}, \hat{\sigma}_{i-1}^{2}) \nonumber \\
\propto & \ \mathds{1}(s_{i-1} > 0) \varphi(s_{i}; \alpha_{i-1} s_{i-1} + \beta_{i-1}, \gamma_{i-1}) \varphi(s_{i-1}; \hat{\mu}_{i-1}, \hat{\sigma}_{i-1}^{2}) \nonumber \\
\propto & \ \mathds{1}(s_{i-1} > 0) \varphi(s_{i}; \alpha_{i-1} \hat{\mu}_{i-1} + \beta_{i-1}, \alpha_{i-1}^{2} \hat{\sigma}_{i-1}^{2} + \gamma_{i-1}) \varphi(s_{i-1}; \mu_{i-1 \mid i}, \sigma_{i-1 \mid i}^{2}) ,
\end{align}
where $\sigma_{i-1 \mid i}^{2} \coloneq (1 / \hat{\sigma}_{i-1}^{2} + \alpha_{i-1}^{2} / \gamma_{i-1})^{-1}$ and $\mu_{i-1 \mid i} \coloneq \sigma_{i-1 \mid i}^{2}(\hat{\mu}_{i-1} / \hat{\sigma}_{i-1}^{2} + \alpha_{i-1}(s_{i} - \beta_{i-1}) / \gamma_{i-1})$.

Substituting \cref{eq:appendix/a2/04} for \cref{eq:appendix/a2/03} yields:
\begin{align}
\label{eq:appendix/a2/05}
f_{S_{i-1} \mid S_{i}, \mathcal{A}_{i-1}}(s_{i-1} \mid s_{i}) 
& \approx \frac{\mathds{1}(s_{i-1} > 0) \varphi(s_{i-1}; \mu_{i-1 \mid i}, \sigma_{i-1 \mid i}^{2})}{\int_{0}^{\infty} \mathds{1}(s_{i-1} > 0) \varphi(s_{i-1}; \mu_{i-1 \mid i}, \sigma_{i-1 \mid i}^{2}) \, \mathrm{d} s_{i-1}} \nonumber \\
& \approx \varphi_{+}(s_{i-1}; \mu_{i-1 \mid i}, \sigma_{i-1 \mid i}^{2}) .
\end{align}
Substituting \cref{eq:appendix/a2/02,eq:appendix/a2/05} for  \cref{eq:appendix/a2/01} yields:
\begin{align*}
P(\mathcal{B}_{i-1}^{\uparrow} \mid s_{i}, \mathcal{A}_{i-1}) 
& = \int_{0}^{\infty} P(\mathcal{B}_{i-1}^{\uparrow} \mid s_{i-1}, s_{i}) f_{S_{i-1} \mid S_{i}, \mathcal{A}_{i-1}}(s_{i-1} \mid s_{i}) \, \mathrm{d} s_{i-1} \\
& \approx \mathds{1}(s_{i} > 0) \int_{0}^{\infty} \exp \left ( - \frac{2 s_{i-1} s_{i}}{\Omega_{i-1} \Psi_{i-1}} \right ) \varphi_{+}(s_{i-1}; \mu_{i-1 \mid i}, \sigma_{i-1 \mid i}^{2}) \, \mathrm{d} s_{i-1} \\
& \approx \mathds{1}(s_{i} > 0) \int_{0}^{\infty} \exp(- \xi_{i-1} s_{i-1}) \varphi_{+}(s_{i-1}; \mu_{i-1 \mid i}, \sigma_{i-1 \mid i}^{2}) \, \mathrm{d} s_{i-1} \\
& \approx \frac{\mathds{1}(s_{i} > 0)}{1 - \Phi(0; \mu_{i-1 \mid i}, \sigma_{i-1 \mid i}^{2})} \int_{0}^{\infty} \exp(- \xi_{i-1} s_{i-1}) \varphi(s_{i-1}; \mu_{i-1 \mid i}, \sigma_{i-1 \mid i}^{2}) \, \mathrm{d} s_{i-1} \\
& \approx \mathds{1}(s_{i} > 0) \frac{1 - \Phi(0; \mu_{i-1 \mid i} - \xi_{i-1} \sigma_{i-1 \mid i}^{2}, \sigma_{i-1 \mid i}^{2})}{1 - \Phi(0; \mu_{i-1 \mid i}, \sigma_{i-1 \mid i}^{2})} \exp \left ( - \xi_{i-1} \mu_{i-1 \mid i} + \frac{1}{2} \xi_{i-1}^{2} \sigma_{i-1 \mid i}^{2} \right ) \\
& \approx \mathds{1}(s_{i} > 0) \frac{1 - \Phi(\xi_{i-1} \sigma_{i-1 \mid i}^{2}; \mu_{i-1 \mid i}, \sigma_{i-1 \mid i}^{2})}{1 - \Phi(0; \mu_{i-1 \mid i}, \sigma_{i-1 \mid i}^{2})} \exp \left ( - \xi_{i-1} \mu_{i-1 \mid i} + \frac{1}{2} \xi_{i-1}^{2} \sigma_{i-1 \mid i}^{2} \right ) ,
\end{align*}
where $\xi_{i-1} \coloneq 2 \Omega_{i-1}^{-1} \Psi_{i-1}^{-1} s_{i}$.

\section{Network Architecture}
\label{sec:network_architecture}

\begin{figure}[h]
\centering
\includegraphics[width=1.0\linewidth]{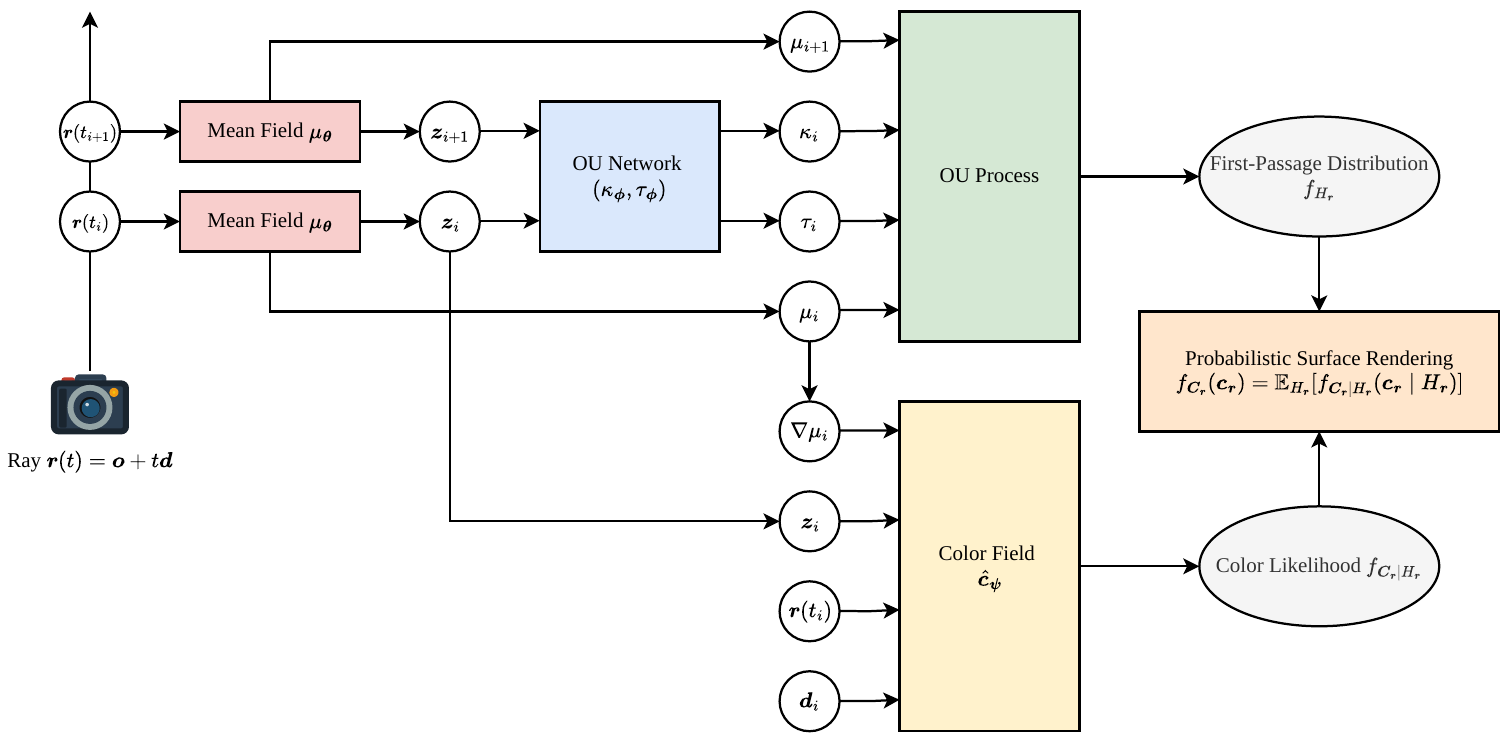}
\caption{Network architecture of our method. The OU network $(\kappa_{\bm{\phi}}, \tau_{\bm{\phi}})$ takes as input the concatenated feature vectors $\boldsymbol{z}_{i} \oplus \boldsymbol{z}_{i+1}$ produced by the mean field $\mu_{\bm{\theta}}$ at the two endpoints of each interval $[t_i, t_{i+1})$. The mean field $\mu_{\bm{\theta}}$ and the color field $\hat{\bm{c}}_{\bm{\psi}}$ are identical to those in NeuS-Facto.}
\label{fig:network_architecture}
\end{figure}

\clearpage

\section{Differences between NeuS and NeuS-Facto}
\label{sec:neus_vs_neus_facto}

\begin{table}[h]
\centering
\caption{Differences between NeuS \citep{neus} and NeuS-Facto \citep{nerfstudio}.}
\label{tab:neus_vs_neus_facto}
\begin{tabular}{lrr}
\toprule
Parameter & NeuS & NeuS-Facto \\
\midrule
Sampling algorithm & Hierarchical Sampling & Proposal Network \\
Number of samples per ray $N$ & 128 & 48 \\
Number of layers of the mean field $\mu_{\bm{\theta}}$ & 8 & 2 \\
Number of layers of the color field $\hat{\bm{c}}_{\bm{\psi}}$ & 4 & 2 \\
Multi-resolution hash encoding & \xmark & \cmark \\
\bottomrule
\end{tabular}
\end{table}

\section{Evaluation Metrics for Uncertainty Quantification}
\label{sec:uncertainty_metrics}

\begin{table}[h]
\centering
\caption{Evaluation metrics for uncertainty quantification used in our experiments. 
We compute these metrics on a subset $\mathcal{R}$ of rays that intersect the ground-truth mesh, using the first ray--mesh intersection as the observed first-passage time $h_{\bm{r}}$, and hence for the conditional first-passage time $\widetilde{H}_{\bm{r}} \coloneq H_{\bm{r}} \mid H_{\bm{r}} \le t_{N}$. $B = 100$ denotes the number of calibration bins.}
\label{tab:uncertainty_metrics}
\begin{tabular}{ll}
\toprule
Metric & Definition \\
\midrule
Mean Absolute Error (MAE) & $\displaystyle \frac{1}{|\mathcal{R}|} \sum_{\bm{r} \in \mathcal{R}} |\mathbb{E}[\widetilde{H}_{\bm{r}}] - h_{\bm{r}}|$ \\
Continuous Ranked Probability Score (CRPS) & $\displaystyle \frac{1}{|\mathcal{R}|} \sum_{\bm{r} \in \mathcal{R}} \int (F_{\widetilde{H}_{\bm{r}}}(t) - \mathds{1}(h_{\bm{r}} \le t))^{2} \, \mathrm{d} t$ \\
Expected Calibration Error (ECE) & $\displaystyle \frac{1}{B} \sum_{i=1}^{B} \left | \left ( \frac{1}{|\mathcal{R}|} \sum_{\bm{r} \in \mathcal{R}} \mathds{1} \left ( F_{\widetilde{H}_{\bm{r}}}(h_{\bm{r}}) \le \frac{i}{B} \right ) \right ) - \frac{i}{B} \right |$ \\
Sharpness & $\displaystyle \frac{1}{|\mathcal{R}|} \sum_{\bm{r} \in \mathcal{R}} \sqrt{\mathbb{V}[\widetilde{H}_{\bm{r}}]}$ \\
\bottomrule
\end{tabular}
\end{table}

\section{Trade-Off between Training Time and Performance}
\label{sec:time_vs_performance}

\begin{figure}[h]
\centering
\begin{minipage}[b]{0.49\linewidth}
\begin{tikzpicture}
\begin{axis}[
width=1.0\linewidth,
height=0.75\linewidth,
xlabel={Average training time [h]},
ylabel={CD [mm]},
xmin=0,
grid=both,
legend pos=north east,
legend cell align=left,
tick align=inside,
thick,
mark size=2pt,
]
\addplot[
  color=tomato,
  mark=*,
] coordinates {
  (0.65, 1.20)
  (1.32, 1.06)
  (2.62, 0.98)
};
\addlegendentry{NeuS}
\addplot[
  color=mediumseagreen,
  mark=square*,
] coordinates {
  (0.34, 1.31)
  (0.70, 1.22)
  (1.35, 1.13)
};
\addlegendentry{NeuS-Facto}
\addplot[
  color=dodgerblue,
  mark=diamond*,
  mark size=3pt,
] coordinates {
  (0.38, 1.02)
  (0.82, 0.93)
  (1.55, 0.88)
};
\addlegendentry{SSDP-Facto}
\end{axis}
\end{tikzpicture}
\subcaption{CD $\downarrow$}
\end{minipage}
\begin{minipage}[b]{0.49\linewidth}
\begin{tikzpicture}
\begin{axis}[
width=1.0\linewidth,
height=0.75\linewidth,
xlabel={Average training time [h]},
ylabel={CRPS [mm]},
xmin=0,
grid=both,
legend pos=north east,
legend cell align=left,
tick align=inside,
thick,
mark size=2pt,
]
\addplot[
  color=tomato,
  mark=*,
] coordinates {
  (0.65, 2.68)
  (1.32, 2.60)
  (2.62, 2.50)
};
\addlegendentry{NeuS}
\addplot[
  color=mediumseagreen,
  mark=square*,
] coordinates {
  (0.34, 3.10)
  (0.70, 2.95)
  (1.35, 2.85)
};
\addlegendentry{NeuS-Facto}
\addplot[
  color=dodgerblue,
  mark=diamond*,
  mark size=3pt,
] coordinates {
  (0.38, 1.98)
  (0.82, 1.91)
  (1.55, 1.86)
};
\addlegendentry{SSDP-Facto}
\end{axis}
\end{tikzpicture}
\subcaption{CRPS $\downarrow$}
\end{minipage}
\caption{Trade-off between training time and unmasked Chamfer distance (mm) / CRPS (mm) on the DTU dataset. Training time is measured on a single NVIDIA H100.}
\label{fig:dtu_time_vs_performance}
\end{figure}

\section{Interpolation Error Validation of \cref{prop:crossing_probability_up_integrand}}
\label{sec:interpolation_error_validation}

\cref{prop:crossing_probability_up_integrand} gives a closed-form expression for $P(\mathcal{B}_{i}^{\uparrow} \mid S_{i}, S_{i+1})$, the probability that the ray enters the surface at some $t \in (t_{i}, t_{i+1})$ and exits by $t_{i+1}$.
The proof in \cref{sec:crossing_probability_up_integrand_proof} consists of three steps:
(i) the martingale part $M(t)$ of the residual process $R(t)$ is transformed into a Brownian motion by switching the clock to its quadratic variation $\omega = \Theta(t)$ (Dambis--Dubins--Schwarz theorem);
(ii) $h(\omega) \coloneq \eta(\Theta^{-1}(\omega))$ with $\eta(t) \coloneq \Psi(t_{i}, t)^{-1} \mu(t)$ is replaced by its linear interpolation over $[0, \Omega_{i}]$;
(iii) the reflection principle is applied to the resulting Brownian bridge.
Only step (ii) is an approximation, and its interpolation error is $O(\Omega_{i}^{2}) = O(\Delta t_{i}^{2})$.
In this experiment, we directly compare the marginal upward zero-crossing probability $P(\mathcal{B}_{i}^{\uparrow})$ with its Monte Carlo estimate, and measure the approximation error as a function of the shape of the mean field $\mu(t)$, the OU parameters $(\kappa_{i}, \tau_{i})$, and the sampling interval $\Delta t_{i}$.

We consider the $i$-th interval $[t_{i}, t_{i+1}]$ and parameterize the mean field as $\mu(t) = \mu_{i} + a_{i} (t - t_{i}) + b_{i} (t - t_{i})^{2}$, where the quadratic term is included so that the curvature of the mean field, which the linear interpolation of $h(\omega)$ in step (ii) does not capture, can be varied independently.
We marginalize \cref{eq:crossing_probability_up} over $S_{i} \sim \mathcal{N}_{+}(\mu_{i}, \sigma_{i}^{2})$, which corresponds to the conditioning in \cref{eq:opacity_ssdp_snis}, yielding:
\begin{align}
\label{eq:marginal_crossing_probability_up}
P(\mathcal{B}_{i}^{\uparrow})
& = \mathbb{E}_{s_{i} \sim \mathcal{N}_{+}(\mu_{i}, \sigma_{i}^{2})}[P(\mathcal{B}_{i}^{\uparrow} \mid s_{i})] .
\end{align}
We compare \cref{eq:marginal_crossing_probability_up} with its Monte Carlo estimate, varying $\mu_{i}$, $\sigma_{i}$, $a_{i}$, $b_{i}$, $\kappa_{i}$, $\tau_{i}$, and $\Delta t_{i}$.
However, not all of these parameters are independent.
Let $\sigma_{\infty}^{2} \coloneq \tau_{i}^{2} / 2 \kappa_{i}$ denote the stationary variance of the residual process $R(t)$, in terms of which the transition variance in \cref{eq:ou_transition_kernel_discrete} is written as $\gamma_{i} = \sigma_{\infty}^{2} (1 - e^{-2 \kappa_{i} \Delta t_{i}})$.
We define the following dimensionless quantities:
\begin{align}
\bar{\mu}_{i} \coloneq \frac{\mu_{i}}{\sigma_{\infty}} , \quad
\bar{\sigma}_{i} \coloneq \frac{\sigma_{i}}{\sigma_{\infty}} , \quad
\bar{a}_{i} \coloneq \frac{a_{i}}{\kappa_{i} \sigma_{\infty}} , \quad
\bar{b}_{i} \coloneq \frac{b_{i}}{\kappa_{i}^{2} \sigma_{\infty}} , \quad
\Delta \bar{t}_{i} \coloneq \kappa_{i} \Delta t_{i} .
\end{align}
Measuring time in units of the correlation length $1 / \kappa_{i}$ and amplitude in units of $\sigma_{\infty}$ reduces the residual process $R(t)$ to a standard OU process, and \cref{eq:marginal_crossing_probability_up} is determined by these normalized quantities alone, as stated in the following lemma.

\begin{lemma}
\label{lemma:standard_ou_process}
Let $\bar{t} \coloneq \kappa_{i} (t - t_{i})$ and $\bar{S}(\bar{t}) \coloneq S(t_{i} + \bar{t} / \kappa_{i}) / \sigma_{\infty}$.
Then, the following hold.
\begin{enumerate}[label=(\alph*)]
\item $\bar{S}(\bar{t}) = \bar{\mu}(\bar{t}) + \bar{R}(\bar{t})$, where $\bar{\mu}(\bar{t}) = \bar{\mu}_{i} + \bar{a}_{i} \bar{t} + \bar{b}_{i} \bar{t}^{2}$, $\mathrm{d} \bar{R}(\bar{t}) = -\bar{R}(\bar{t}) \, \mathrm{d} \bar{t} + \sqrt{2} \, \mathrm{d} \bar{W}(\bar{t})$ for a Wiener process $\bar{W}$, and $\bar{S}(0) \sim \mathcal{N}_{+}(\bar{\mu}_{i}, \bar{\sigma}_{i}^{2})$.
That is, $\bar{S}$ is an SSDP with $(\kappa, \tau) = (1, \sqrt{2})$ on the interval $[0, \Delta \bar{t}_{i}]$.
\item The exact probability is invariant: $P(\mathcal{B}_{i}^{\uparrow}) = P(\bar{\mathcal{B}}_{i}^{\uparrow})$, where $\bar{\mathcal{B}}_{i}^{\uparrow}$ denotes the event $\mathcal{B}_{i}^{\uparrow}$ for $\bar{S}$.
\item The approximate probability is invariant: \cref{eq:marginal_crossing_probability_up} for $\bar{S}$ equals \cref{eq:marginal_crossing_probability_up} for $S$.
\end{enumerate}
\end{lemma}

\begin{proof}
All three claims follow from one operation, $(\mathcal{T} f)(\bar{t}) \coloneq f(t_{i} + \bar{t} / \kappa_{i}) / \sigma_{\infty}$ for a function $f$ on $[t_{i}, t_{i+1}]$, which substitutes $t = t_{i} + \bar{t} / \kappa_{i}$ and divides by $\sigma_{\infty}$.
\begin{enumerate}[label=(\alph*)]
\item Applying $\mathcal{T}$ to $S$ gives $\bar{S}(\bar{t}) = \bar{\mu}(\bar{t}) + \bar{R}(\bar{t})$ with $\bar{\mu}(\bar{t})$, $\bar{S}(0)$, and the drift term as stated.
For the diffusion term, the scaling and shift properties of the Wiener process imply that $\bar{W}(\bar{t}) \coloneq \sqrt{\kappa_{i}} \, (W(t_{i} + \bar{t} / \kappa_{i}) - W(t_{i}))$ is again a Wiener process, so the diffusion coefficient becomes $\tau_{i} / (\sigma_{\infty} \sqrt{\kappa_{i}}) = \sqrt{2}$.
\item Applying $\mathcal{T}$ to a sample path of $S(t)$ gives a sample path of $\bar{S}(\bar{t})$.
Since $\mathcal{T}$ consists of the monotonic bijection $t \mapsto \bar{t}$ from $[t_{i}, t_{i+1}]$ onto $[0, \Delta \bar{t}_{i}]$ and division by a positive constant $\sigma_{\infty}$, and $\mathcal{B}_{i}^{\uparrow}$ is defined by the signs of the path, the path belongs to $\bar{\mathcal{B}}_{i}^{\uparrow}$ if and only if the original belongs to $\mathcal{B}_{i}^{\uparrow}$.
\item Applying $\mathcal{T}$ to the ingredients of \cref{eq:crossing_probability_up} divides $\mu_{i+1 \mid i}$ and $\sigma_{i+1 \mid i}$ by $\sigma_{\infty}$, and multiplies $\xi_{i}$ by $\sigma_{\infty}$.
Every term in \cref{eq:crossing_probability_up} is a dimensionless combination of these, namely $\xi_{i} \mu_{i+1 \mid i}$, $\xi_{i} \sigma_{i+1 \mid i}$, and $\Phi \left ( \xi_{i} \sigma_{i+1 \mid i}^{2}; \mu_{i+1 \mid i}, \sigma_{i+1 \mid i}^{2} \right ) = \Phi \left ( \left ( \xi_{i} \sigma_{i+1 \mid i}^{2} - \mu_{i+1 \mid i} \right ) / \sigma_{i+1 \mid i}; 0, 1 \right )$, yielding $P(\mathcal{B}_{i}^{\uparrow} \mid S_{i}) = P(\bar{\mathcal{B}}_{i}^{\uparrow} \mid \bar{S}_{i})$, where $\bar{S}_{i} = S_{i} / \sigma_{\infty}$.
Since $\bar{S}_{i} = \bar{S}(0) \sim \mathcal{N}_{+}(\bar{\mu}_{i}, \bar{\sigma}_{i}^{2})$ is given by \cref{lemma:standard_ou_process} (a), taking the expectation over $S_{i}$ on both sides gives the claim. \qedhere
\end{enumerate}
\end{proof}

Therefore, instead of varying $\mu_{i}$, $\sigma_{i}$, $a_{i}$, $b_{i}$, $\kappa_{i}$, $\tau_{i}$, and $\Delta t_{i}$ independently, we sweep the normalized quantities $\bar{\mu}_{i}$, $\bar{\sigma}_{i}$, $\bar{a}_{i}$, $\bar{b}_{i}$, and $\Delta \bar{t}_{i}$ in the standard form $(\kappa_{i}, \tau_{i}) = (1, \sqrt{2})$.
In the standard form, the quadratic variation that determines the order of the interpolation error reduces to $\bar{\Omega}_{i} = e^{2 \Delta \bar{t}_{i}} - 1$, which is approximately $2 \Delta \bar{t}_{i}$ for $\Delta \bar{t}_{i} \ll 1$.
The order of the interpolation error is $O(\bar{\Omega}_{i}^{2}) = O(\Delta \bar{t}_{i}^{2})$ as shown in \cref{sec:crossing_probability_up_integrand_proof}.
How it translates into the error of $P(\mathcal{B}_{i}^{\uparrow})$ is measured by the following experiment.

\paragraph{Experimental design.}
The shape of the mean field is restricted to the situation where a surviving path approaches the surface ($\bar{\mu}_{i} > 0$, $\bar{a}_{i} \le 0$, $\bar{b}_{i} \ge 0$), and we vary $\bar{\mu}_{i}, \bar{\sigma}_{i} \in \{0.1, 1, 10\}$, $\bar{a}_{i} \in \{0, -0.1, -1, -10\}$, and $\bar{b}_{i} \in \{0, 0.1, 1, 10\}$.
$\Delta \bar{t}_{i}$ takes $11$ values logarithmically spaced from $0.1$ to $10$.
For each combination of $\bar{\mu}_{i}$, $\bar{\sigma}_{i}$, $\bar{a}_{i}$, $\bar{b}_{i}$, and $\Delta \bar{t}_{i}$, we sample paths of the SSDP in the standard form sequentially from the transition kernel in \cref{eq:ou_transition_kernel_discrete} at the endpoints of $n = 1000$ equal subdivision intervals, and detect $\mathcal{B}_{i}^{\uparrow}$ from the signs at these endpoints only.
However, since sign-change detection misses zero-crossings within a subdivision interval, with a leading error of $O(n^{-1/2})$ \citep{discrete_killing_error}, we remove this bias by Richardson extrapolation from three nested thinnings of the same paths ($n$, $n/2$, and $n/4$ subdivision intervals), which cancels the $O(n^{-1/2})$ and $O(n^{-1})$ terms.
We use $m = 10^{8}$ paths per combination.
The expectation in \cref{eq:marginal_crossing_probability_up} is computed by Gauss--Legendre quadrature in the CDF space with $1000$ nodes.

\begin{figure}[t]
\centering
\begin{minipage}[b]{0.49\linewidth}
\centering
\includegraphics[width=1.0\linewidth]{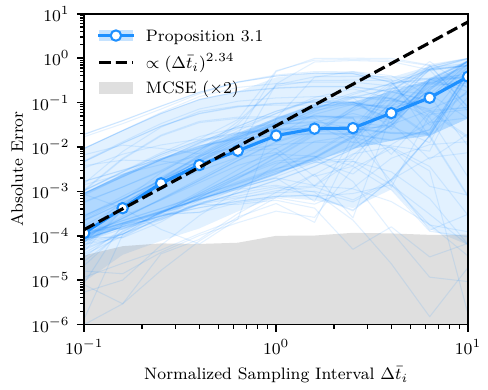}
\subcaption{Absolute Error}
\end{minipage}
\begin{minipage}[b]{0.49\linewidth}
\centering
\includegraphics[width=1.0\linewidth]{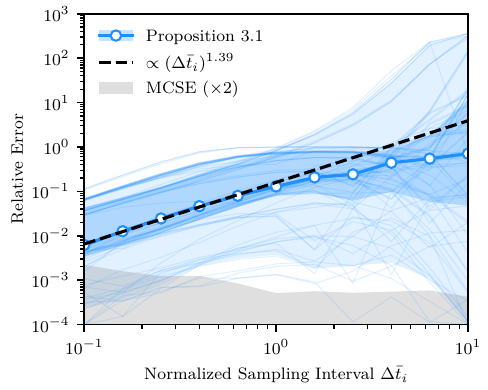}
\subcaption{Relative Error}
\end{minipage}
\caption{Absolute (a) and relative (b) errors of $P(\mathcal{B}_{i}^{\uparrow})$ (\ref{eq:marginal_crossing_probability_up}) against the Monte Carlo reference as functions of the normalized sampling interval $\Delta \bar{t}_{i}$. Each thin line is one combination of $(\bar{\mu}_{i}, \bar{\sigma}_{i}, \bar{a}_{i}, \bar{b}_{i})$, the thick line is the median across combinations, and the dark and light blue bands are the 25th--75th and 5th--95th percentile ranges, respectively. The gray band is twice the Monte Carlo standard error (MCSE). The dashed line is a power law fitted by least squares to the median for $\Delta \bar{t}_{i} < 1$, with the fitted exponent shown in the legend. Combinations whose relative MCSE exceeds $1\%$ at any $\Delta \bar{t}_{i}$ are excluded.}
\label{fig:interpolation_error_validation/crossing_probability_up_errors}
\end{figure}

\paragraph{Evaluation results.}
\cref{fig:interpolation_error_validation/crossing_probability_up_errors} shows the absolute and relative errors of $P(\mathcal{B}_{i}^{\uparrow})$.
The absolute error of $P(\mathcal{B}_{i}^{\uparrow})$ decreases as a power law in $\Delta \bar{t}_{i}$ for $\Delta \bar{t}_{i} < 1$, and the exponent fitted to the median is $2.34$, i.e., at least as fast as the $O(\bar{\Omega}_{i}^{2}) = O(\Delta \bar{t}_{i}^{2})$ interpolation error.
The relative error also decreases as a power law in $\Delta \bar{t}_{i}$ for $\Delta \bar{t}_{i} < 1$, with a fitted exponent of $1.39$.
This rules out the possibility that the decrease of the absolute error merely reflects that of $P(\mathcal{B}_{i}^{\uparrow})$ itself, which vanishes as $\Delta \bar{t}_{i} \to 0$, and shows that the approximation error vanishes faster than the probability.
The power-law regime extends to $\Delta \bar{t}_{i} \approx 1$, i.e., the approximation error decreases roughly as $O(\Delta \bar{t}_{i}^{2})$ as long as the sampling interval $\Delta t_{i}$ is shorter than the correlation length $1 / \kappa_{i}$.

\clearpage

\section{Failure Cases of the Negative-Absorbing Approximation}
\label{sec:approximation_error_validation}

\begin{figure}[t]
\centering

\rowlabel{$w = 0.05$}%
\begin{subfigure}[c]{0.47\linewidth}
\centering
\includegraphics[width=0.49\linewidth]{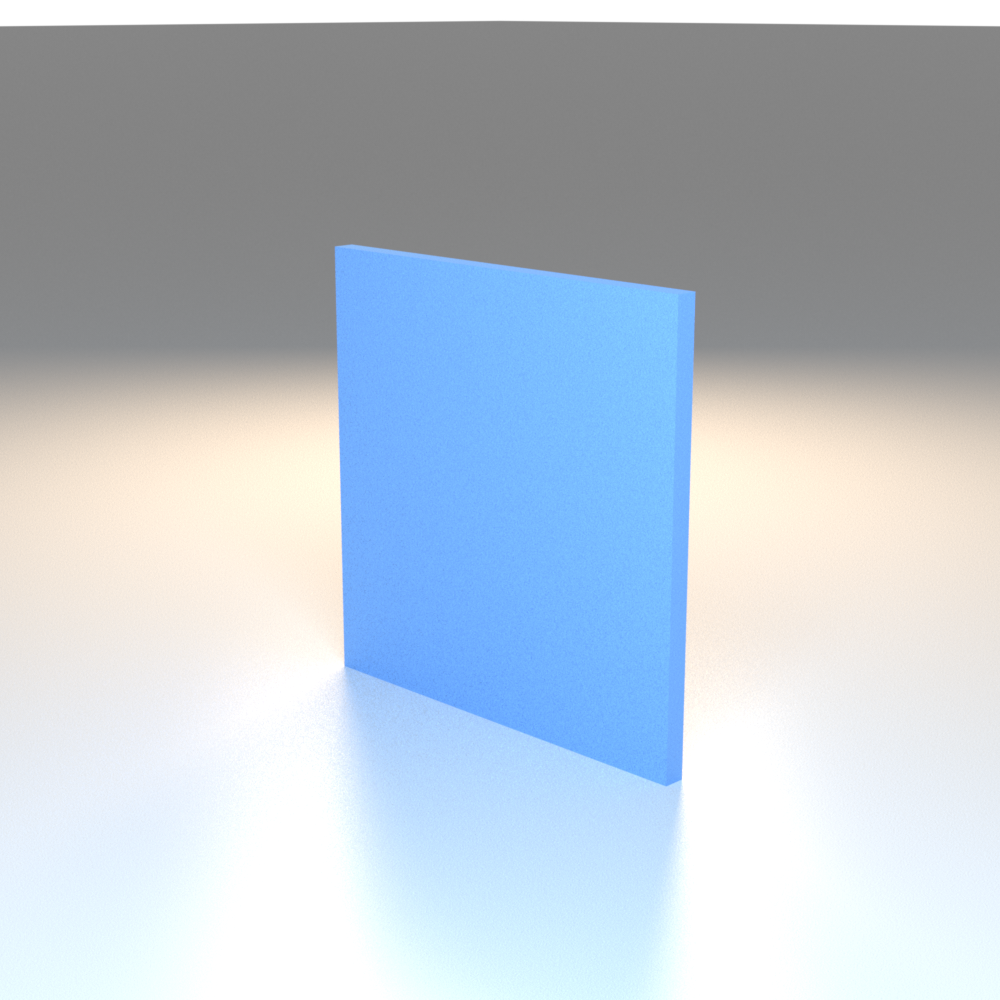}\hfill
\includegraphics[width=0.49\linewidth]{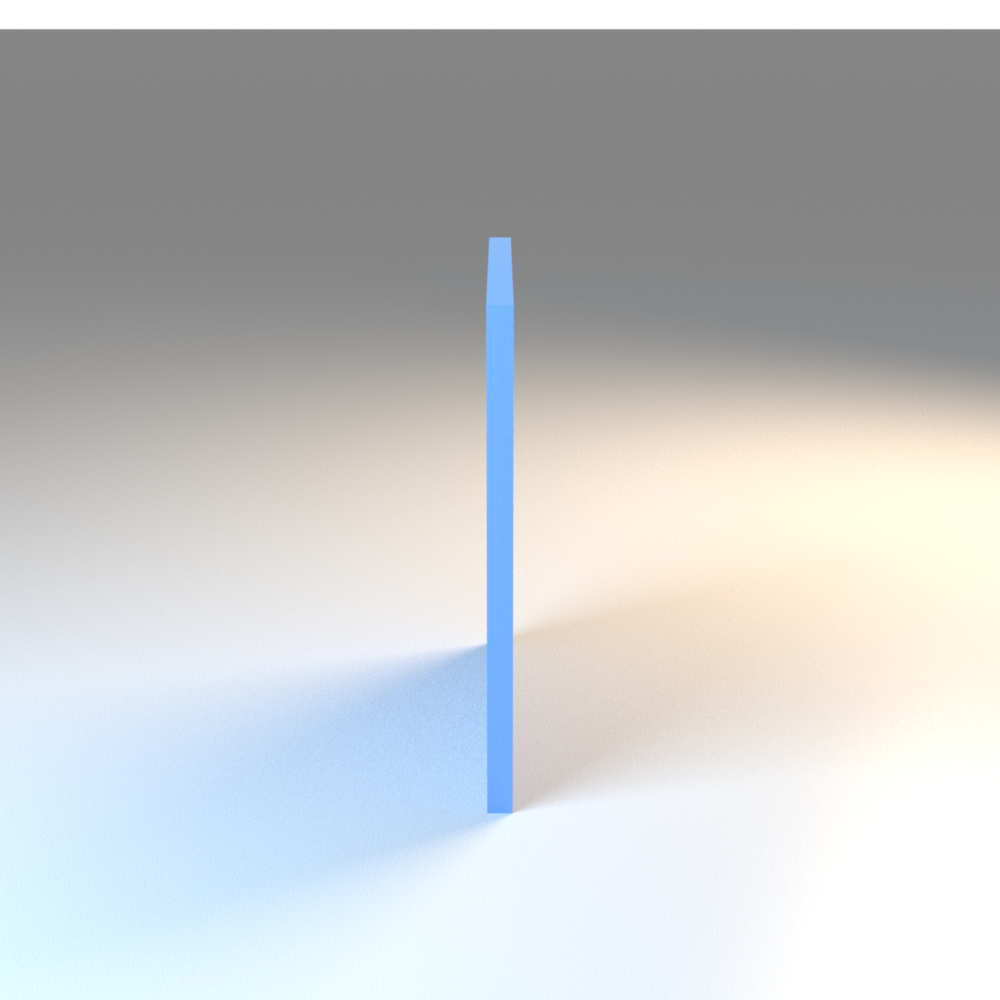}
\end{subfigure}\hfill\vrule\hfill
\begin{subfigure}[c]{0.47\linewidth}
\centering
\includegraphics[width=0.49\linewidth]{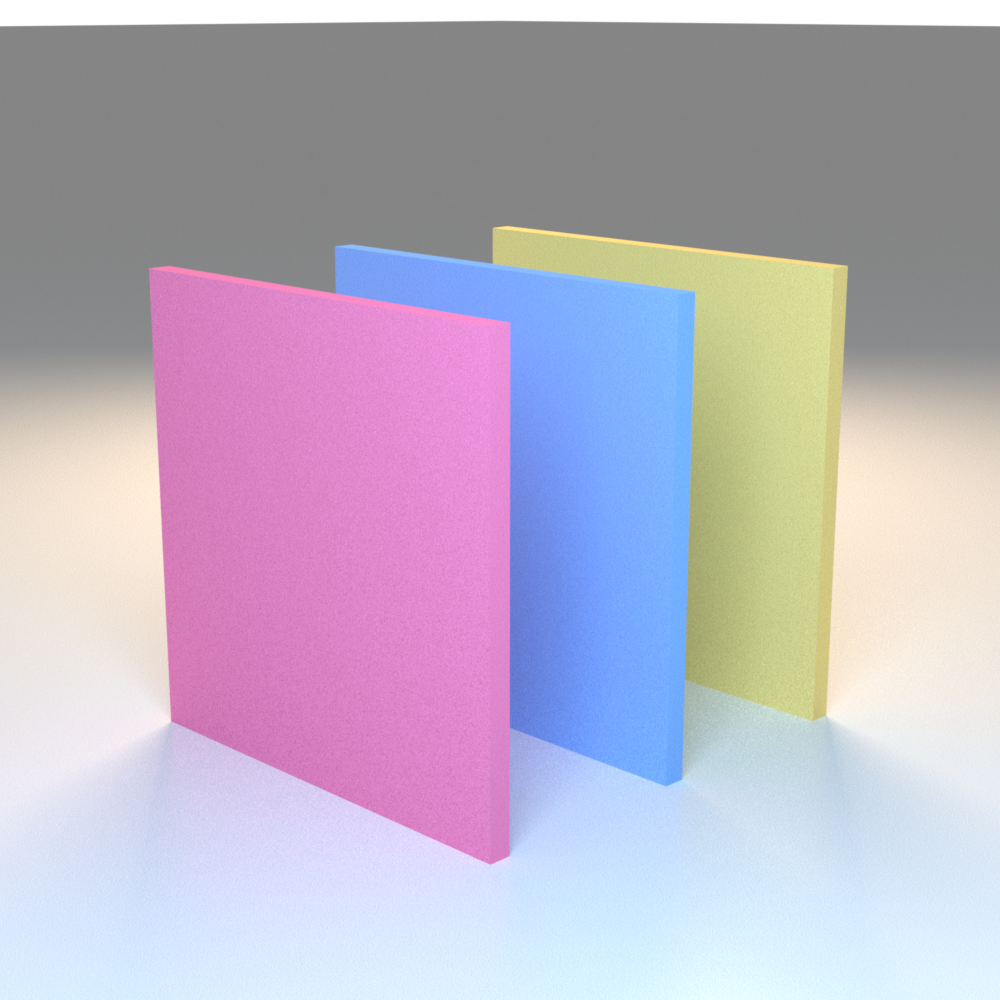}\hfill
\includegraphics[width=0.49\linewidth]{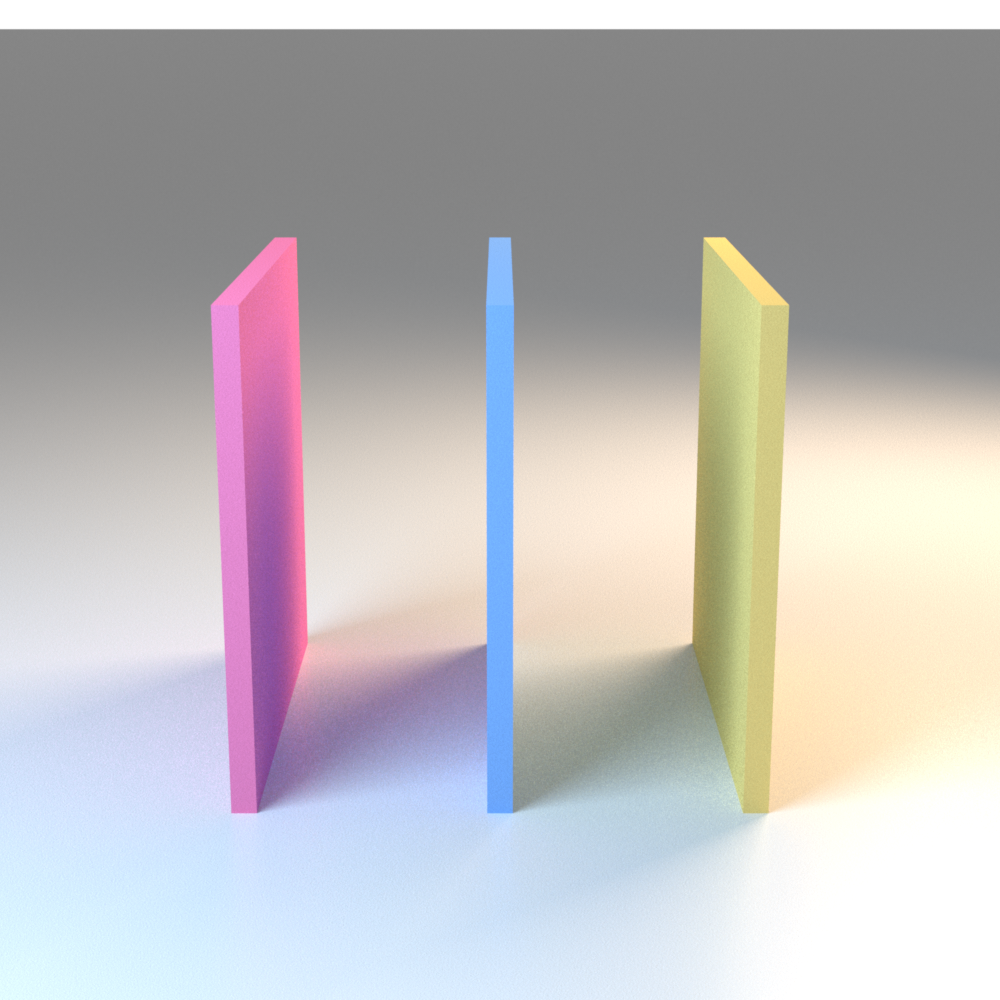}
\end{subfigure}

\vspace{1pt}

\rowlabel{$w = 0.025$}%
\begin{subfigure}[c]{0.47\linewidth}
\centering
\includegraphics[width=0.49\linewidth]{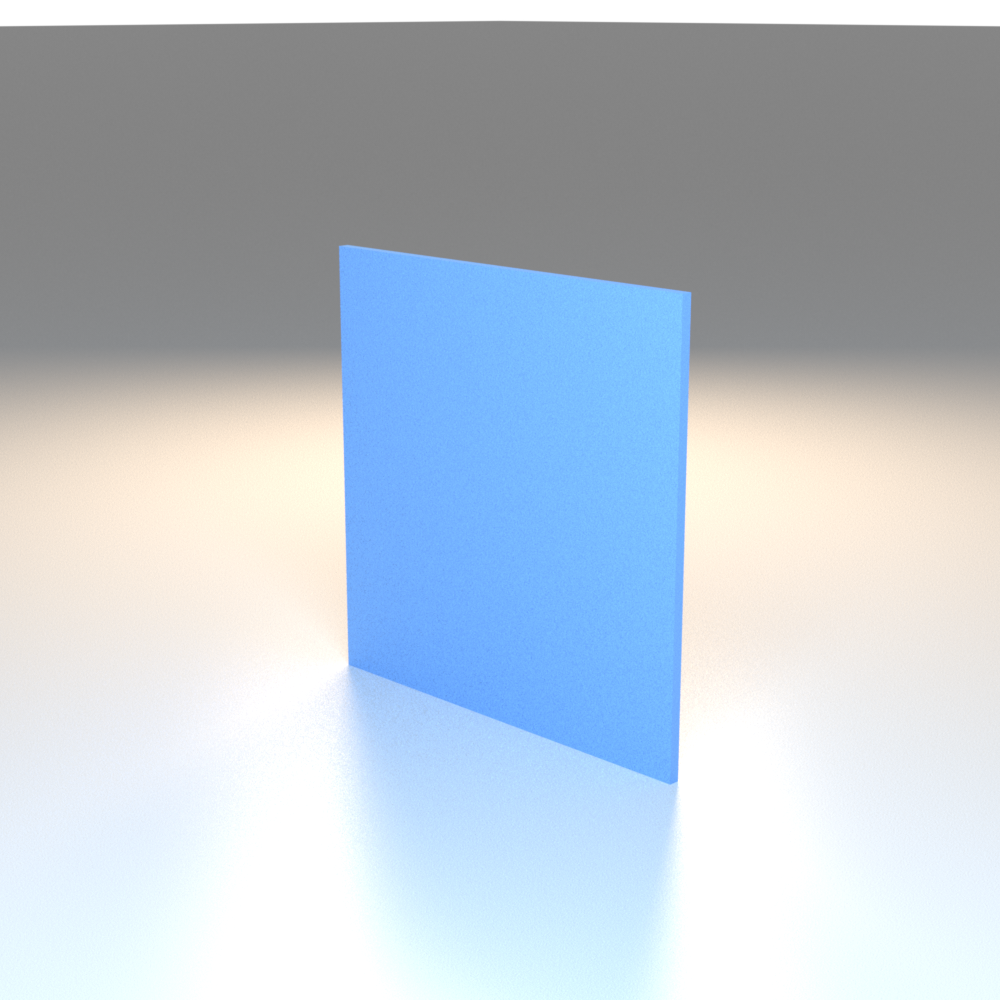}\hfill
\includegraphics[width=0.49\linewidth]{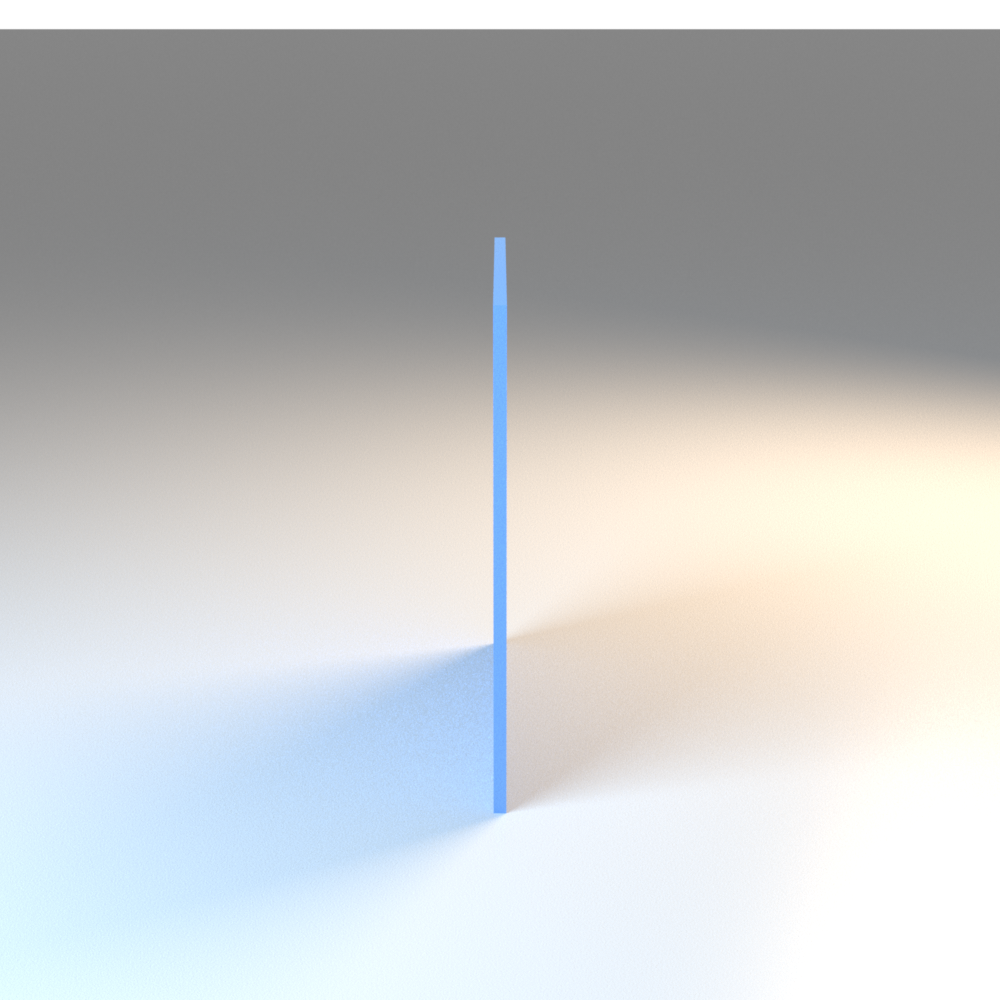}
\caption*{$K = 1$}
\end{subfigure}\hfill\vrule\hfill
\begin{subfigure}[c]{0.47\linewidth}
\centering
\includegraphics[width=0.49\linewidth]{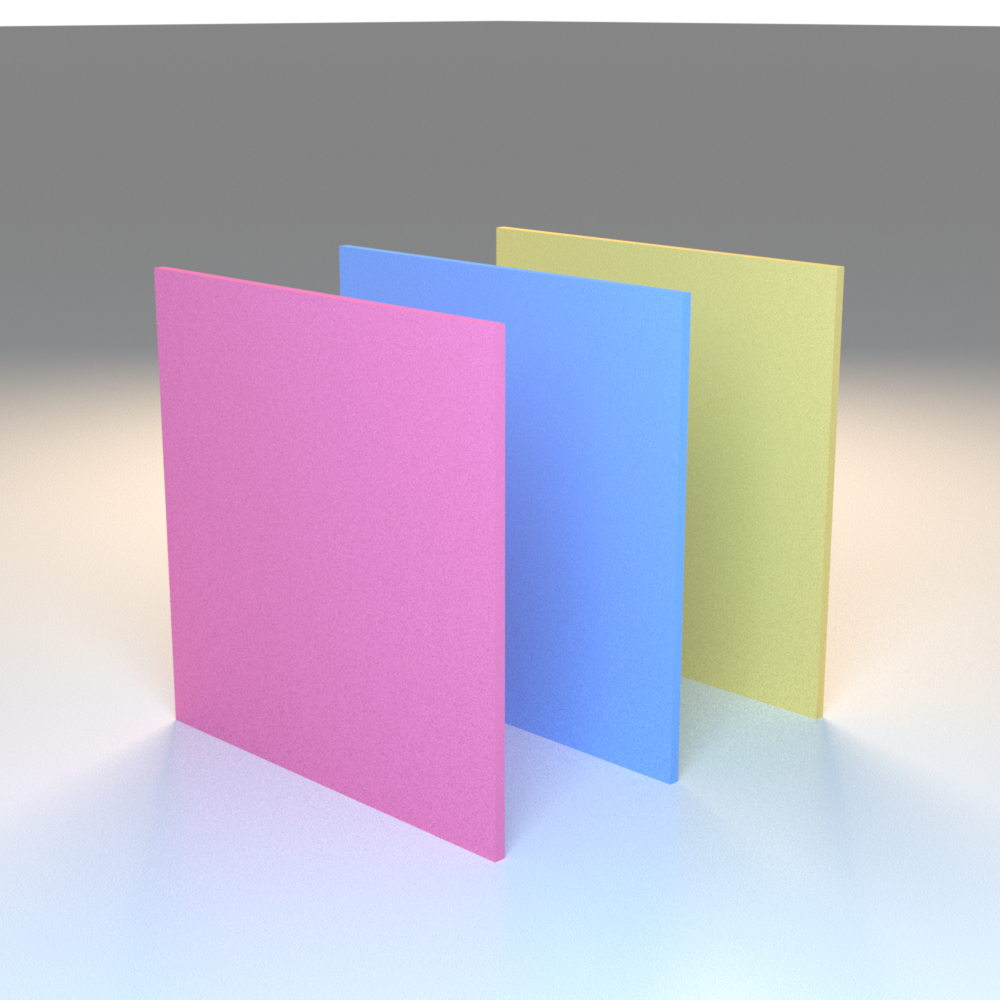}\hfill
\includegraphics[width=0.49\linewidth]{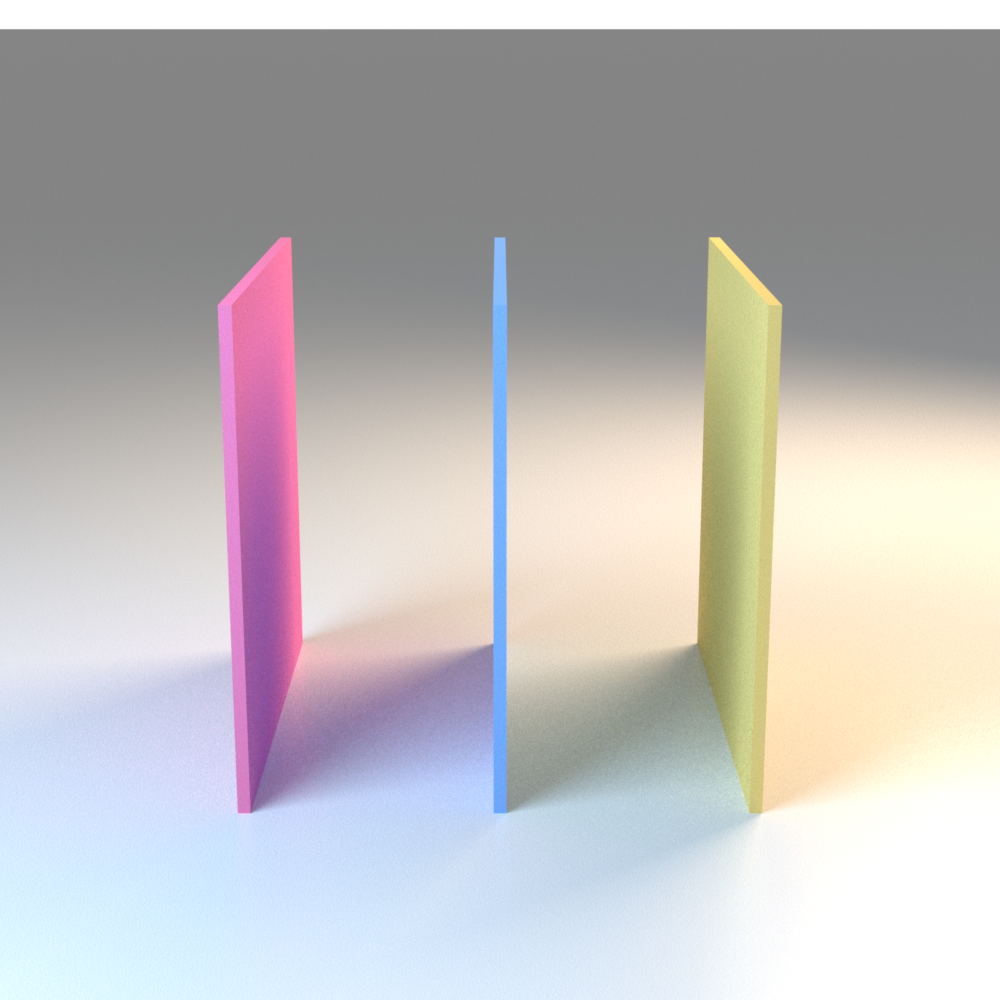}
\caption*{$K = 3$}
\end{subfigure}
\caption{Synthetic scenes, each of which consists of $K$ parallel slabs of thickness $w$ stacked along the $x$-axis, rendered with the path tracer of Mitsuba 3 \citep{mitsuba3}.}
\label{fig:approximation_error_validation/scenes}
\end{figure}

The negative-absorbing approximation (\cref{sec:method/first_passage_probability/negative_absorbing_approximation}) assumes that $S(t)$ never returns to a positive value once it becomes negative, and consists of the two simplifications in \cref{eq:survival_event_approx}: the survival event is replaced by the sign at the left endpoint, $\mathcal{A}_{i} \equiv \{S_{i} > 0\}$, and the zero-crossing event is replaced by the sign at the right endpoint, $\mathcal{B}_{i} \equiv \mathcal{B}_{i}^{\downarrow} = \{S_{i+1} \le 0\}$, i.e., $\mathcal{B}_{i}^{\uparrow}$ is dropped.
As noted in \cref{sec:method/first_passage_probability/negative_absorbing_approximation}, the assumption is violated whenever the ray exits an object of finite volume, i.e., whenever a downward zero-crossing is followed by an upward one, and the two simplifications are affected by this violation at different scales.
If the ray later undergoes another downward zero-crossing, as when it intersects multiple surfaces, $\{S_{i} > 0\}$ no longer represents the survival event at the later zero-crossing, which affects the former simplification.
If the upward zero-crossing occurs within the same sampling interval as the downward one, as when the surface is thinner than the sampling interval, the zero-crossing is not captured by $\{S_{i+1} \le 0\}$, which affects the latter.

In this section, we quantify the error of each simplification separately against a Monte Carlo (MC) reference on synthetic scenes where the number and the thickness of the surfaces are controlled (\cref{sec:approximation_error_validation/survival,sec:approximation_error_validation/crossing}), and then examine their combination, i.e., the negative-absorbing approximation itself (\cref{sec:approximation_error_validation/negative_absorbing}).

\paragraph{Synthetic scenes.}
Each scene consists of $K$ parallel slabs of thickness $w$ stacked along the $x$-axis, as shown in \cref{fig:approximation_error_validation/scenes}.
We use $K \in \{1, 3\}$ to control the number of surfaces a ray intersects and $w \in \{0.05, 0.025\}$ to control their thickness.
The slabs with a lateral extent of $[-0.5, 0.5]^{2}$ are centered at $x = 0$ for $K = 1$ and at $x \in \{-0.5, 0, 0.5\}$ for $K = 3$.
We render each scene with the path tracer of Mitsuba 3 \citep{mitsuba3} from $64$ views at a resolution of $1000 \times 1000$.
We train the SSDP-Facto variants described below on each scene, but replace the proposal network with a uniform sampler with $N = 48$ samples along each ray, yielding a sampling interval of $\Delta t_{i} \approx 0.04$.
The two thicknesses straddle this interval: a slab with $w = 0.05$ always contains a sample, whereas a slab with $w = 0.025$ can fall entirely between two consecutive samples.

\paragraph{SSDP variants.}
To isolate the two simplifications $\mathcal{A}_{i} \equiv \{S_{i} > 0\}$ and $\mathcal{B}_{i} \equiv \{S_{i+1} \le 0\}$, we switch each of them on and off independently, yielding the four SSDP variants in \cref{tab:approximation_error_validation/ssdp_variants}.
We refer to each variant by whether it keeps the survival conditioning ($\mathcal{A}$ \cmark / \xmark) and the upward zero-crossing ($\mathcal{B}^{\uparrow}$ \cmark / \xmark).
($\mathcal{A}$ \cmark, $\mathcal{B}^{\uparrow}$ \cmark) denotes the full Bayesian formulation (\cref{sec:method/first_passage_probability/bayesian_filtering}), and ($\mathcal{A}$ \xmark, $\mathcal{B}^{\uparrow}$ \xmark) denotes the negative-absorbing approximation (\cref{sec:method/first_passage_probability/negative_absorbing_approximation}).
The other two are designed for this analysis: ($\mathcal{A}$ \cmark, $\mathcal{B}^{\uparrow}$ \xmark) denotes the Bayesian formulation with $P(\mathcal{B}_{i}^{\uparrow} \mid S_{i})$ dropped, and ($\mathcal{A}$ \xmark, $\mathcal{B}^{\uparrow}$ \cmark) replaces the survival-conditioned distribution that $S_{i} \mid \mathcal{A}_{i}$ follows by the truncated Gaussian distribution $\mathcal{N}_{+}(\mu_{i}, \sigma_{i}^{2})$ that $S_{i} \mid S_{i} > 0$ follows, as in the negative-absorbing approximation, while $P(\mathcal{B}_{i}^{\uparrow} \mid S_{i})$ is kept.

\begin{table}[h]
\centering
\caption{SSDP variants with the conditioning on $\mathcal{A}$ and the upward zero-crossing $\mathcal{B}^{\uparrow}$ switched on and off.}
\label{tab:approximation_error_validation/ssdp_variants}
\begin{tabular}{lcc}
\toprule
& $\mathcal{B}^{\uparrow}$ \cmark & $\mathcal{B}^{\uparrow}$ \xmark \\
\midrule
$\mathcal{A}$ \cmark & Full Bayesian formulation & \cref{sec:method/first_passage_probability/bayesian_filtering} without $P(\mathcal{B}_{i}^{\uparrow} \mid S_{i})$ \\
$\mathcal{A}$ \xmark & \cref{sec:method/first_passage_probability/negative_absorbing_approximation} with $P(\mathcal{B}_{i}^{\uparrow} \mid S_{i})$ & Negative-absorbing approximation \\
\bottomrule
\end{tabular}
\end{table}

\paragraph{Evaluation metrics.}
In \cref{sec:experiments/evaluation_metrics}, the first-passage-time distribution for a ray is evaluated by the CRPS against a single observation.
Here, the target is not a single observation but the exact first-passage-time distribution for the same process, which is estimated by MC.
Let $F$ and $G$ denote the CDFs of the first-passage time $H$ along a ray given by an SSDP variant and by the MC reference, respectively, both conditioned on $H \le t_{N}$ as in \cref{sec:experiments/evaluation_metrics}.
Taking the expectation of the CRPS of $F$ over the observation $h \sim G$ yields:
\begin{align}
\label{eq:crps_decomposition}
\mathbb{E}_{h \sim G} \left [ \int (F(t) - \mathds{1}(h \le t))^{2} \, \mathrm{d} t \right ]
& = \int (F(t) - G(t))^{2} \, \mathrm{d} t + \int G(t) (1 - G(t)) \, \mathrm{d} t .
\end{align}
The second term does not depend on $F$; it is the expected CRPS of the exact distribution $F = G$ itself and hence the minimum attainable.
The first term, which is called the Cram\'{e}r distance, is the excess over this minimum and vanishes if and only if $F = G$.
We therefore report the Cram\'{e}r distance between $F$ and $G$ as the evaluation metric for all experiments in this section.
The MC reference is computed by the subdivision and extrapolation procedure in \cref{sec:interpolation_error_validation}.

\paragraph{Experimental design.}
To see for which OU parameters each simplification fails, independently of training, we take the analytic SDF of each scene as the mean field $\mu(t)$, fix $\kappa_{i} \equiv \kappa$ and $\tau_{i} \equiv \tau$, and evaluate the Cram\'{e}r distance between the first-passage-time distribution given by each SSDP variant and the MC reference along the $x$-axis, on a grid of $(\kappa, \tau) \in [10^{-2}, 10^{2}] \times [10^{-3}, 10^{1}]$ spaced logarithmically with $11$ points each, with $10^{5}$ MC paths per grid point, which we call an \textit{error map}.
To see whether training passes through the region where each simplification fails and whether the error persists, we train each SSDP variant for $10^{4}$ iterations and, every $100$ iterations, compute the Cram\'{e}r distance between the first-passage-time distribution given by each SSDP variant and the MC reference, with $10^{3}$ paths per ray, over the rays from the training views that intersect the ground-truth mesh.
Note that this measures not the consistency of the first-passage-time distribution with ground-truth geometry, as in \cref{sec:experiments/evaluation_metrics}, but the consistency of the first-passage-time distribution with its own learned process, i.e., the training dynamics of the approximation error.
To connect the two, we overlay on each analytic error map the learned $(\kappa, \tau)$ at each checkpoint, summarized by the median over the intervals of the rays parallel to the $x$-axis and the mean over these rays, as a \textit{trajectory}; since the learned OU parameters vary along and across the rays, the trajectory should be read only as a rough indication of the region that training passes through, not as the parameters of every interval.

\subsection{Failure Cases of the Survival Approximation $\mathcal{A}_{i} \equiv \{S_{i} > 0\}$}
\label{sec:approximation_error_validation/survival}

\begin{figure}[t]
\centering

\rowlabel{$K = 1$}
\hfill\rule[-2.75cm]{0.5pt}{5.75cm}\hfill
\begin{minipage}[c]{0.31\linewidth}
\centering
\begin{labeledmap}[width=1.0\linewidth]{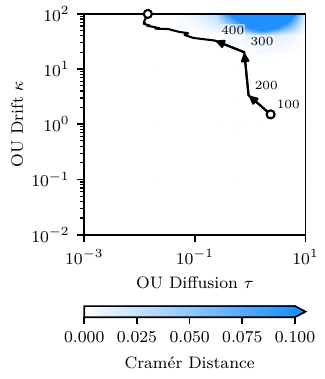}
\regrect{1}
\end{labeledmap}
\end{minipage}
\begin{minipage}[c]{0.31\linewidth}
\centering
\begin{labeledmap}[width=1.0\linewidth]{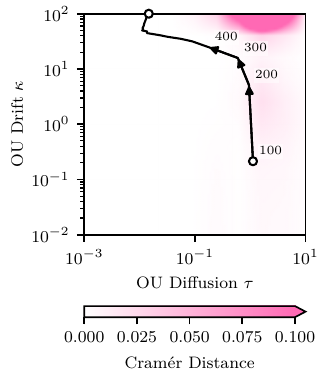}
\regrect{1}
\end{labeledmap}
\end{minipage}
\begin{minipage}[c]{0.31\linewidth}
\centering
\includegraphics[width=1.0\linewidth]{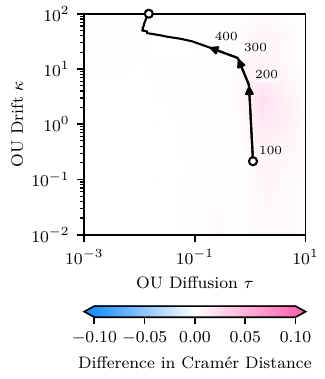}
\end{minipage}

\rowlabel{$K = 3$}
\hfill\rule[-2.5cm]{0.5pt}{5.75cm}\hfill
\begin{minipage}[c]{0.31\linewidth}
\centering
\begin{labeledmap}[width=1.0\linewidth]{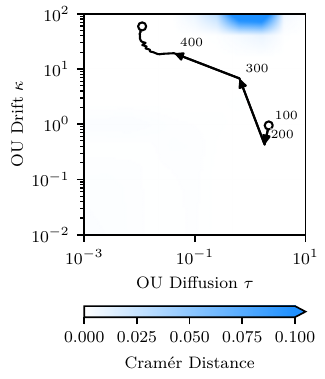}
\regrect{1}
\end{labeledmap}
\subcaption{$\mathcal{A}$ \cmark, $\mathcal{B}^{\uparrow}$ \cmark}
\end{minipage}
\begin{minipage}[c]{0.31\linewidth}
\centering
\begin{labeledmap}[width=1.0\linewidth]{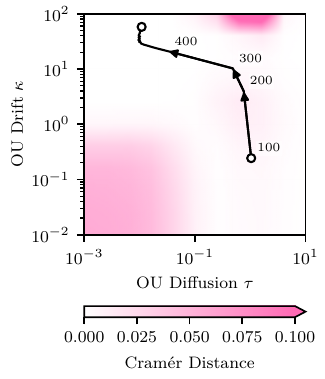}
\regrect{1}
\regrect{2}
\end{labeledmap}
\subcaption{$\mathcal{A}$ \xmark, $\mathcal{B}^{\uparrow}$ \cmark}
\end{minipage}
\begin{minipage}[c]{0.31\linewidth}
\centering
\begin{labeledmap}[width=1.0\linewidth]{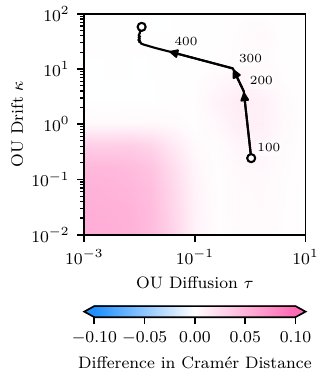}
\regrect{2}
\end{labeledmap}
\subcaption{Difference (b) $-$ (a)}
\end{minipage}

\caption{Error maps for the survival approximation $\mathcal{A}_{i} \equiv \{S_{i} > 0\}$ on the scenes $K \in \{1, 3\}$: the Cram\'{e}r distance between the first-passage-time distribution given by each variant and the MC reference over the OU parameters $(\kappa, \tau)$, and the difference between the two distances. The line is the training trajectory of the OU parameters learned by the corresponding variant. The numbered regions are explained in the text.}
\label{fig:approximation_error_validation/survival/k1_vs_k3/analytic}
\end{figure}

\begin{figure}[t]
\centering
\begin{minipage}[b]{0.49\linewidth}
\centering
\includegraphics[width=1.0\linewidth]{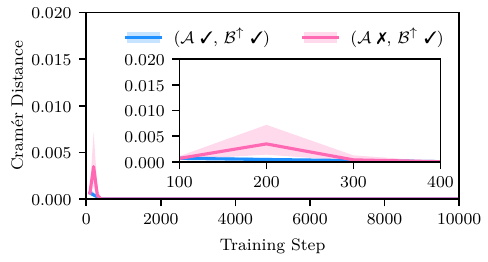}
\subcaption{$K = 1$}
\end{minipage}
\begin{minipage}[b]{0.49\linewidth}
\centering
\includegraphics[width=1.0\linewidth]{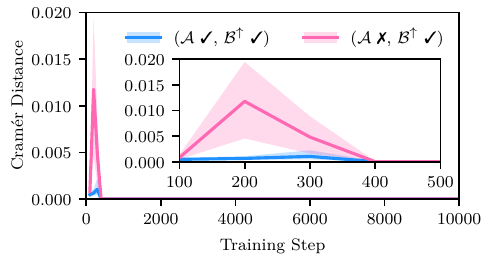}
\subcaption{$K = 3$}
\end{minipage}
\caption{Training dynamics of the approximation error for the survival approximation $\mathcal{A}_{i} \equiv \{S_{i} > 0\}$ on the scenes $K \in \{1, 3\}$: the Cram\'{e}r distance between the first-passage-time distribution given by each variant and the MC reference of the process learned by that variant up to each checkpoint, with the median and the interquartile range over the rays.}
\label{fig:approximation_error_validation/survival/k1_vs_k3/learned}
\end{figure}

To isolate the survival approximation, we compare ($\mathcal{A}$ \cmark, $\mathcal{B}^{\uparrow}$ \cmark) and ($\mathcal{A}$ \xmark, $\mathcal{B}^{\uparrow}$ \cmark) on the scenes $(K, w) \in \{(1, 0.05), (3, 0.05)\}$, which differ only in the number of surfaces the ray intersects.
\cref{fig:approximation_error_validation/survival/k1_vs_k3/analytic} shows the error maps, on which we mark the regions where the approximation error appears and explain each of them below.
\begin{itemize}[leftmargin=*]
\item[{\regone}] The error shared by both variants at large $\kappa$ and $\tau$.
It is the interpolation error of \cref{prop:crossing_probability_up_integrand} at $\kappa \Delta t_{i} > 1$ shown in \cref{sec:interpolation_error_validation}, which affects $P(\mathcal{B}_{i}^{\uparrow} \mid S_{i})$ and hence every variant that keeps $\mathcal{B}^{\uparrow}$.
Here, it is confined to large $\tau$: since a sample lies inside the slab, the zero-crossing at the slab is captured by $\mathcal{B}_{i}^{\downarrow}$, and $\mathcal{B}^{\uparrow}$ occurs only away from the slab, where the mean field is positive and only the noise can drive the process to zero.
It cancels in the difference.
\item[{\regtwo}] The error due to dropping the conditioning on $\mathcal{A}$.
It appears only for $K = 3$ and where both $\kappa$ and $\tau$ are small, i.e., near the perfectly correlated limit in \cref{sec:method/relationship_to_prior_works}, where the residual from the mean field is shared along the ray and thus carries the survival history to the later zero-crossings.
It vanishes as the residual decorrelates along the ray, since the survival history then carries no information even in the exact formulation.
The residual decorrelates either by increasing $\kappa$, which shortens the correlation length $1 / \kappa$ so that the residual cannot retain its survival history, or by increasing $\tau$, which drowns the survival history in the noise injected at every interval.
For $K = 1$, no such error appears, since there is no later zero-crossing that the survival history could affect.
\end{itemize}
The training trajectories show that training increases $\kappa$ before decreasing $\tau$ and thus does not pass through \regtwo.
Accordingly, \cref{fig:approximation_error_validation/survival/k1_vs_k3/learned} shows that the approximation error of both variants is confined to the first few hundred iterations and remains negligible for the rest of training.

\newpage

\subsection{Failure Cases of the Crossing Approximation $\mathcal{B}_{i} \equiv \{S_{i+1} \le 0\}$}
\label{sec:approximation_error_validation/crossing}

\begin{figure}[t]
\centering

\rowlabel{$w = 0.05$}
\hfill\rule[-2.75cm]{0.5pt}{5.75cm}\hfill
\begin{minipage}[c]{0.31\linewidth}
\centering
\begin{labeledmap}[width=1.0\linewidth]{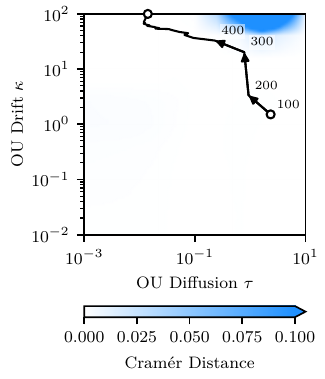}
\regrect{1}
\end{labeledmap}
\end{minipage}
\begin{minipage}[c]{0.31\linewidth}
\centering
\begin{labeledmap}[width=1.0\linewidth]{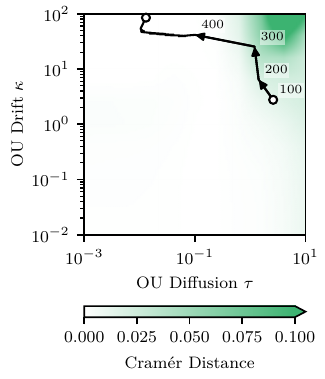}
\regrect{3}
\end{labeledmap}
\end{minipage}
\begin{minipage}[c]{0.31\linewidth}
\centering
\begin{labeledmap}[width=1.0\linewidth]{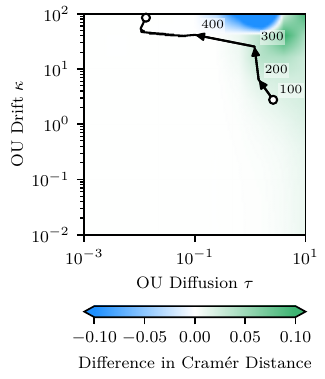}
\regrect{1}
\regrect{3}
\end{labeledmap}
\end{minipage}

\rowlabel{$w = 0.025$}
\hfill\rule[-2.5cm]{0.5pt}{5.75cm}\hfill
\begin{minipage}[c]{0.31\linewidth}
\centering
\begin{labeledmap}[width=1.0\linewidth]{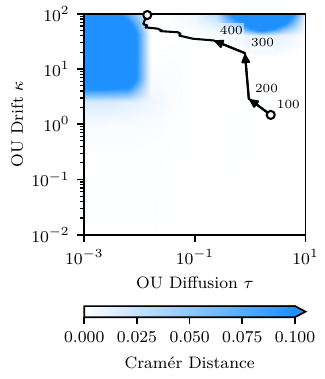}
\regrect{1}
\regrect{4}
\end{labeledmap}
\subcaption{$\mathcal{A}$ \cmark, $\mathcal{B}^{\uparrow}$ \cmark}
\end{minipage}
\begin{minipage}[c]{0.31\linewidth}
\centering
\begin{labeledmap}[width=1.0\linewidth]{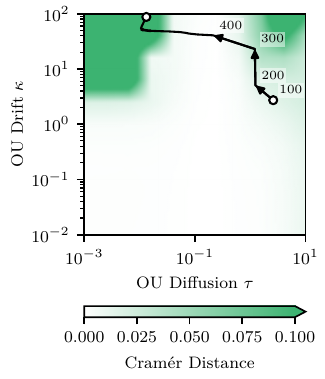}
\regrect{3}
\regrect{4}
\regrect{5}
\end{labeledmap}
\subcaption{$\mathcal{A}$ \cmark, $\mathcal{B}^{\uparrow}$ \xmark}
\end{minipage}
\begin{minipage}[c]{0.31\linewidth}
\centering
\begin{labeledmap}[width=1.0\linewidth]{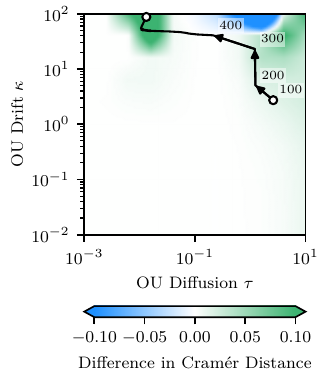}
\regrect{1}
\regrect{3}
\regrect{5}
\end{labeledmap}
\subcaption{Difference (b) $-$ (a)}
\end{minipage}

\caption{Error maps for the crossing approximation $\mathcal{B}_{i} \equiv \{S_{i+1} \le 0\}$ on the scenes $w \in \{0.05, 0.025\}$, with the thinner slab between two consecutive samples: the Cram\'{e}r distance between the first-passage-time distribution given by each variant and the MC reference over the OU parameters $(\kappa, \tau)$, and the difference between the two distances. The line is the training trajectory of the OU parameters learned by the corresponding variant. The numbered regions are explained in the text.}
\label{fig:approximation_error_validation/crossing/w0050_vs_w0025/analytic}
\end{figure}

\begin{figure}[t]
\centering

\rowlabel{No sample in the slab}
\hfill\rule[-2.75cm]{0.5pt}{5.75cm}\hfill
\begin{minipage}[c]{0.31\linewidth}
\centering
\begin{labeledmap}[width=1.0\linewidth]{figures/ray/approx-B/BF-UP_vs_BF/W-0.025/phase-0.5/analytic_first_passage_pmf_cramer_distance_map_BF-UP_vs_MC.pdf}
\regrect{1}
\regrect{4}
\end{labeledmap}
\end{minipage}
\begin{minipage}[c]{0.31\linewidth}
\centering
\begin{labeledmap}[width=1.0\linewidth]{figures/ray/approx-B/BF-UP_vs_BF/W-0.025/phase-0.5/analytic_first_passage_pmf_cramer_distance_map_BF_vs_MC.pdf}
\regrect{3}
\regrect{4}
\regrect{5}
\end{labeledmap}
\end{minipage}
\begin{minipage}[c]{0.31\linewidth}
\centering
\begin{labeledmap}[width=1.0\linewidth]{figures/ray/approx-B/BF-UP_vs_BF/W-0.025/phase-0.5/analytic_first_passage_pmf_cramer_difference_map_BF-UP_vs_BF.pdf}
\regrect{1}
\regrect{3}
\regrect{5}
\end{labeledmap}
\end{minipage}

\rowlabel{A sample in the slab}
\hfill\rule[-2.5cm]{0.5pt}{5.75cm}\hfill
\begin{minipage}[c]{0.31\linewidth}
\centering
\begin{labeledmap}[width=1.0\linewidth]{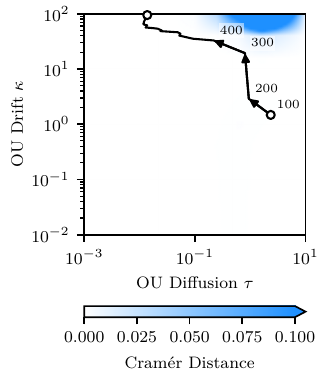}
\regrect{1}
\end{labeledmap}
\subcaption{$\mathcal{A}$ \cmark, $\mathcal{B}^{\uparrow}$ \cmark}
\end{minipage}
\begin{minipage}[c]{0.31\linewidth}
\centering
\begin{labeledmap}[width=1.0\linewidth]{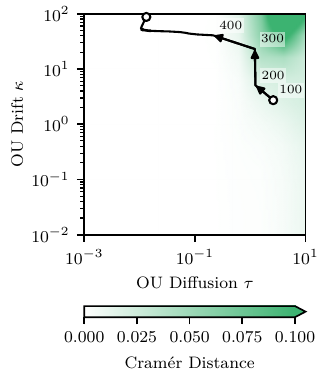}
\regrect{3}
\end{labeledmap}
\subcaption{$\mathcal{A}$ \cmark, $\mathcal{B}^{\uparrow}$ \xmark}
\end{minipage}
\begin{minipage}[c]{0.31\linewidth}
\centering
\begin{labeledmap}[width=1.0\linewidth]{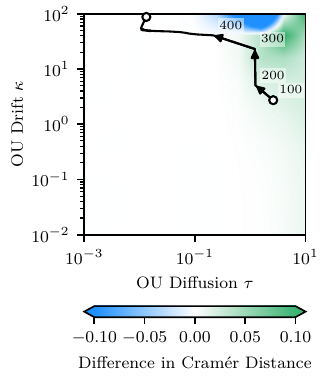}
\regrect{1}
\regrect{3}
\end{labeledmap}
\subcaption{Difference (b) $-$ (a)}
\end{minipage}

\caption{Error maps for the crossing approximation $\mathcal{B}_{i} \equiv \{S_{i+1} \le 0\}$ on the scene $w = 0.025$ without (top) and with (bottom) a sample inside the slab: the Cram\'{e}r distance between the first-passage-time distribution given by each variant and the MC reference over the OU parameters $(\kappa, \tau)$, and the difference between the two distances. The line is the training trajectory of the OU parameters learned by the corresponding variant. The numbered regions are those of \cref{fig:approximation_error_validation/crossing/w0050_vs_w0025/analytic}.}
\label{fig:approximation_error_validation/crossing/phase_05_vs_00/analytic}
\end{figure}

To isolate the crossing approximation, we compare ($\mathcal{A}$ \cmark, $\mathcal{B}^{\uparrow}$ \cmark) and ($\mathcal{A}$ \cmark, $\mathcal{B}^{\uparrow}$ \xmark) on the scenes $(K, w) \in \{(1, 0.05), (1, 0.025)\}$, which differ only in the thickness of the slab, with the samples placed so that the thinner slab falls entirely between two consecutive samples.
\cref{fig:approximation_error_validation/crossing/w0050_vs_w0025/analytic} shows the error maps.
Comparing the maps for the two thicknesses, \regone{} and \regthree{} appear in both and are thus not caused by the thickness, whereas \regfour{} and \regfive{} appear only for the thinner slab.
\begin{itemize}[leftmargin=*]
\item[{\regone}] The error of ($\mathcal{A}$ \cmark, $\mathcal{B}^{\uparrow}$ \cmark) at large $\kappa$ and $\tau$, which is the same as in \cref{sec:approximation_error_validation/survival}.
Here it does not cancel in the difference, since ($\mathcal{A}$ \cmark, $\mathcal{B}^{\uparrow}$ \xmark) drops $\mathcal{B}^{\uparrow}$ and is therefore free of the interpolation error of \cref{prop:crossing_probability_up_integrand}.
\item[{\regthree}] The error due to dropping $\mathcal{B}^{\uparrow}$ away from the slab.
At large $\tau$, the noise makes the process cross zero even away from the slab, inside intervals whose endpoints are both positive.
($\mathcal{A}$ \cmark, $\mathcal{B}^{\uparrow}$ \cmark) accounts for such zero-crossings by $P(\mathcal{B}_{i}^{\uparrow} \mid S_{i})$, whereas ($\mathcal{A}$ \cmark, $\mathcal{B}^{\uparrow}$ \xmark) has no such term and misses them.
The error is confined to large $\kappa$ as well: crossing zero and returning within a single interval requires the interval to be longer than the correlation length $1 / \kappa$; otherwise the process stays below zero across an endpoint and the zero-crossing is captured by $\mathcal{B}_{i}^{\downarrow}$.
\item[{\regfour}] The error shared by both variants for $w = 0.025$ at small $\tau$.
The slab lies between two consecutive samples, so both endpoints of the interval containing it are positive and the zero-crossing must be inferred from what happens inside the interval.
At small $\tau$, the process is nearly deterministic and follows the dip of the mean field inside the interval, which neither variant captures: the linear interpolation in \cref{prop:crossing_probability_up_integrand} cannot represent a dip within a single interval.
This is the interpolation error of \cref{prop:crossing_probability_up_integrand} shown in \cref{sec:interpolation_error_validation}, caused here by the curvature of the mean field rather than by $\kappa \Delta t_{i}$.
\item[{\regfive}] The error due to dropping $\mathcal{B}^{\uparrow}$ at the slab.
A zero-crossing inside the interval can be caused either by the dip of the mean field or by the noise.
The former is invisible to \cref{prop:crossing_probability_up_integrand}, as in \regfour, but the latter is exactly what $P(\mathcal{B}_{i}^{\uparrow} \mid S_{i})$ models; as $\tau$ increases, the noise alone becomes large enough to make the process cross zero inside the interval, so that ($\mathcal{A}$ \cmark, $\mathcal{B}^{\uparrow}$ \cmark) recovers the zero-crossing at the slab through the noise rather than through the curvature of the mean field, whereas ($\mathcal{A}$ \cmark, $\mathcal{B}^{\uparrow}$ \xmark) cannot.
\end{itemize}
For $w = 0.05$, \regfour{} and \regfive{} do not appear: a sample lies inside the slab, so the zero-crossing at the slab is covered by $\mathcal{B}_{i}^{\downarrow}$ and the crossing approximation has no effect on it.
\cref{fig:approximation_error_validation/crossing/phase_05_vs_00/analytic} confirms this from the other side: placing a sample inside the thinner slab removes \regfour{} and \regfive{} and makes the maps coincide with those for $w = 0.05$.
What matters is thus not the thickness itself but whether a sample falls inside the slab.

\begin{figure}[t]
\centering
\begin{minipage}[b]{0.49\linewidth}
\centering
\includegraphics[width=1.0\linewidth]{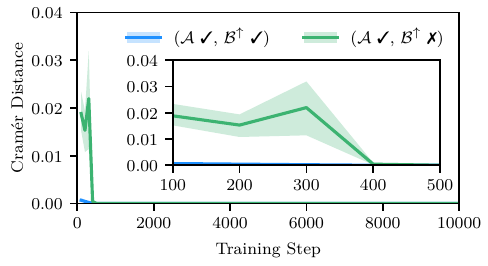}
\subcaption{$w = 0.05$}
\end{minipage}
\begin{minipage}[b]{0.49\linewidth}
\centering
\includegraphics[width=1.0\linewidth]{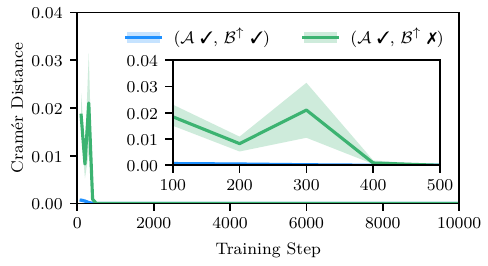}
\subcaption{$w = 0.025$}
\end{minipage}
\caption{Training dynamics of the approximation error for the crossing approximation $\mathcal{B}_{i} \equiv \{S_{i+1} \le 0\}$ on the scenes $w \in \{0.05, 0.025\}$: the Cram\'{e}r distance between the first-passage-time distribution given by each variant and the MC reference of the process learned by that variant up to each checkpoint, with the median and the interquartile range over the rays.}
\label{fig:approximation_error_validation/crossing/w0050_vs_w0025/learned}
\end{figure}

\begin{figure}[t]
\centering

\rowlabel{$w = 0.05$}
\hfill\rule[-2.5cm]{0.5pt}{5.25cm}\hfill
\begin{minipage}[c]{0.46\linewidth}
\centering
\includegraphics[width=1.0\linewidth]{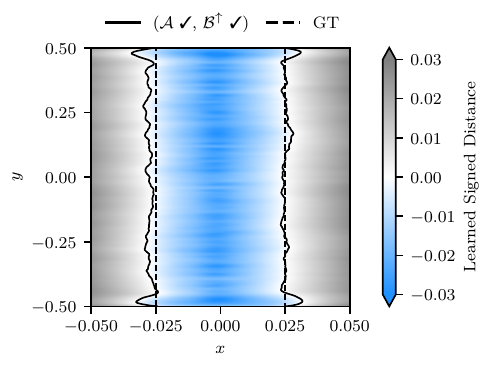}
\end{minipage}
\begin{minipage}[c]{0.46\linewidth}
\centering
\includegraphics[width=1.0\linewidth]{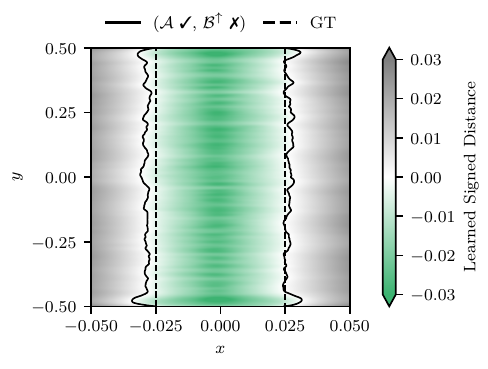}
\end{minipage}

\rowlabel{$w = 0.025$}
\hfill\rule[-2.25cm]{0.5pt}{5.25cm}\hfill
\begin{minipage}[c]{0.46\linewidth}
\centering
\includegraphics[width=1.0\linewidth]{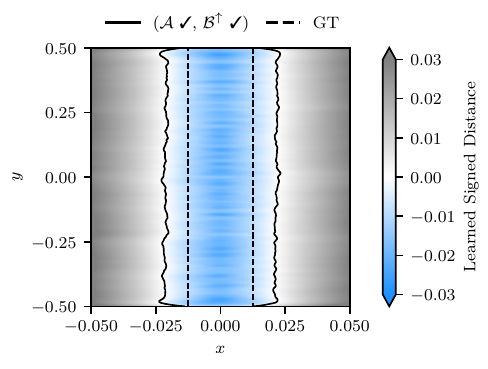}
\subcaption{$\mathcal{A}$ \cmark, $\mathcal{B}^{\uparrow}$ \cmark}
\end{minipage}
\begin{minipage}[c]{0.46\linewidth}
\centering
\includegraphics[width=1.0\linewidth]{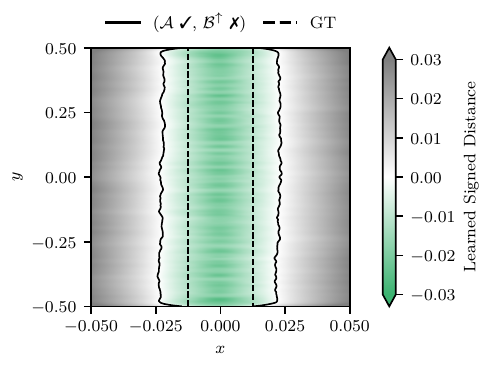}
\subcaption{$\mathcal{A}$ \cmark, $\mathcal{B}^{\uparrow}$ \xmark}
\end{minipage}

\caption{Mean field learned by each variant on the scenes $w \in \{0.05, 0.025\}$, sliced at $z = 0$. The solid and dashed lines show the learned and ground-truth zero-level sets, respectively.}
\label{fig:approximation_error_validation/crossing/sdf}
\end{figure}

The training trajectories start where $\tau$ is large, so \regthree{} appears at the beginning of training and vanishes within the first few hundred iterations, as shown in \cref{fig:approximation_error_validation/crossing/w0050_vs_w0025/learned}.
They end, however, inside \regfive{}, where the error due to dropping $\mathcal{B}^{\uparrow}$ is large when the mean field is the analytic SDF.
The error nevertheless vanishes during training for two reasons: the samples are jittered, so the slab is not fixed between the samples, and the learned surface is thickened until the samples fall inside it, as shown next.
\cref{fig:approximation_error_validation/crossing/sdf} shows the mean field learned by each variant on each scene.
For $w = 0.05$, both variants nearly recover the thickness, whereas for $w = 0.025$, both reconstruct the slab about as thick as the sampling interval: the median thickness along the rays parallel to the $x$-axis is $0.052$ and $0.057$ for $w = 0.05$, and $0.043$ and $0.046$ for $w = 0.025$, for ($\mathcal{A}$ \cmark, $\mathcal{B}^{\uparrow}$ \cmark) and ($\mathcal{A}$ \cmark, $\mathcal{B}^{\uparrow}$ \xmark), respectively.
That is, the error at a slab thinner than the sampling interval does not persist because the learned surface is thickened until the samples fall inside it, with or without $\mathcal{B}^{\uparrow}$; the approximation error is converted into a geometric error shared by both variants.
Note that ($\mathcal{A}$ \cmark, $\mathcal{B}^{\uparrow}$ \cmark) is free of error in \regfive{} and thus has the capacity to represent a slab thinner than the sampling interval through \cref{prop:crossing_probability_up_integrand}; nevertheless, training does not exploit this capacity but thickens the slab in the same way.

\clearpage

\subsection{Failure Cases of the Negative-Absorbing Approximation}
\label{sec:approximation_error_validation/negative_absorbing}

\begin{figure}[t]
\centering

\rowlabel{$(K, w) = (1, 0.05)$}
\hfill\rule[-2.75cm]{0.5pt}{5.75cm}\hfill
\begin{minipage}[c]{0.31\linewidth}
\centering
\begin{labeledmap}[width=1.0\linewidth]{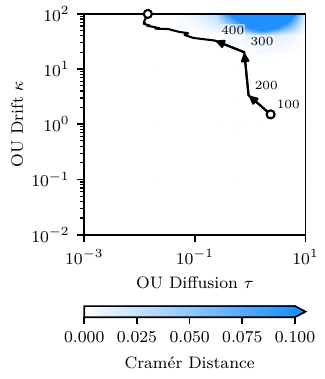}
\regrect{1}
\end{labeledmap}
\end{minipage}
\begin{minipage}[c]{0.31\linewidth}
\centering
\begin{labeledmap}[width=1.0\linewidth]{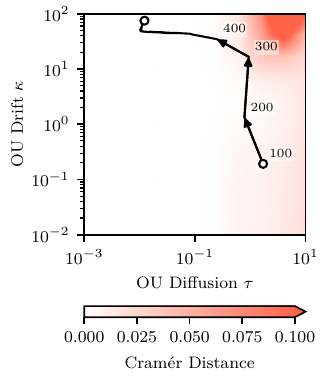}
\regrect{3}
\end{labeledmap}
\end{minipage}
\begin{minipage}[c]{0.31\linewidth}
\centering
\begin{labeledmap}[width=1.0\linewidth]{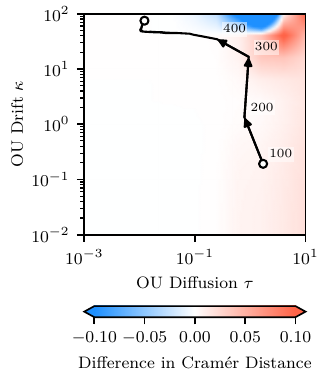}
\regrect{1}
\regrect{3}
\end{labeledmap}
\end{minipage}

\rowlabel{$(K, w) = (3, 0.025)$}
\hfill\rule[-2.5cm]{0.5pt}{5.75cm}\hfill
\begin{minipage}[c]{0.31\linewidth}
\centering
\begin{labeledmap}[width=1.0\linewidth]{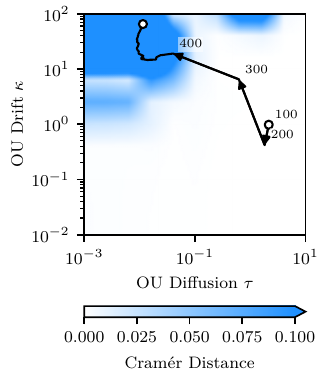}
\regrect{1}
\regrect{4}
\end{labeledmap}
\subcaption{$\mathcal{A}$ \cmark, $\mathcal{B}^{\uparrow}$ \cmark}
\end{minipage}
\begin{minipage}[c]{0.31\linewidth}
\centering
\begin{labeledmap}[width=1.0\linewidth]{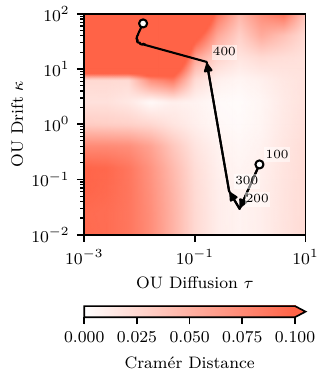}
\regrect{2}
\regrect{3}
\regrect{4}
\regrect{5}
\end{labeledmap}
\subcaption{$\mathcal{A}$ \xmark, $\mathcal{B}^{\uparrow}$ \xmark}
\end{minipage}
\begin{minipage}[c]{0.31\linewidth}
\centering
\begin{labeledmap}[width=1.0\linewidth]{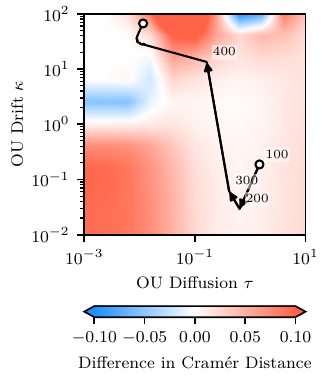}
\regrect{2}
\regrect{3}
\regrect{5}
\end{labeledmap}
\subcaption{Difference (b) $-$ (a)}
\end{minipage}
\caption{Error maps for the negative-absorbing approximation on the scenes $(K, w) \in \{(1, 0.05), (3, 0.025)\}$, with the thinner slabs between two consecutive samples: the Cram\'{e}r distance between the first-passage-time distribution given by each variant and the MC reference over the OU parameters $(\kappa, \tau)$, and the difference between the two distances. The line is the training trajectory of the OU parameters learned by the corresponding variant. The numbered regions are those of \cref{sec:approximation_error_validation/survival,sec:approximation_error_validation/crossing}.}
\label{fig:approximation_error_validation/negative_absorbing/k3w0025/analytic}
\end{figure}

The negative-absorbing approximation drops both the conditioning on $\mathcal{A}$ and $\mathcal{B}^{\uparrow}$.
We compare ($\mathcal{A}$ \cmark, $\mathcal{B}^{\uparrow}$ \cmark) and ($\mathcal{A}$ \xmark, $\mathcal{B}^{\uparrow}$ \xmark) on the scene $(K, w) = (1, 0.05)$, where neither simplification affects the slab, and on the scene $(K, w) = (3, 0.025)$, where both fail: the ray intersects multiple surfaces, and each of them is thinner than the sampling interval and placed between two consecutive samples as in \cref{sec:approximation_error_validation/crossing}.
\cref{fig:approximation_error_validation/negative_absorbing/k3w0025/analytic} shows the error maps, where the regions are marked at the same positions as in \cref{sec:approximation_error_validation/survival,sec:approximation_error_validation/crossing}.
For $(K, w) = (1, 0.05)$, only \regone{} and \regthree{} appear, which are unrelated to the slab, as explained in \cref{sec:approximation_error_validation/crossing}.
For $(K, w) = (3, 0.025)$, all of them appear: \regtwo{} of the survival approximation, \regfour{} and \regfive{} of the crossing approximation, and \regone{} and \regthree{}.
Whereas \regtwo{} has the same extent as in \cref{sec:approximation_error_validation/survival}, \regfour{} and \regfive{} extend toward larger $\tau$, up to $\tau \approx 0.1$ for ($\mathcal{A}$ \cmark, $\mathcal{B}^{\uparrow}$ \cmark) and $\tau \approx 0.5$ for ($\mathcal{A}$ \xmark, $\mathcal{B}^{\uparrow}$ \xmark), so that the error of the negative-absorbing approximation, together with \regtwo, covers most of the map.
This extension can be interpreted as follows.
In \regfour{} and \regfive{}, each variant assigns too little zero-crossing probability to the thin slab, and the rest is carried beyond the slab as survival probability.
For a single slab, this affects only the zero-crossing at that slab.
For multiple slabs, the survival probability carried beyond the first slab is spent at the next slab, i.e., first-passage probability is assigned to the wrong surface.
The conditioning on $\mathcal{A}$ cannot correct this, since it conditions on the survival computed by the variant itself, which treats the process as having survived the first slab although it has not.

\begin{figure}[t]
\centering
\begin{minipage}[b]{0.49\linewidth}
\centering
\includegraphics[width=1.0\linewidth]{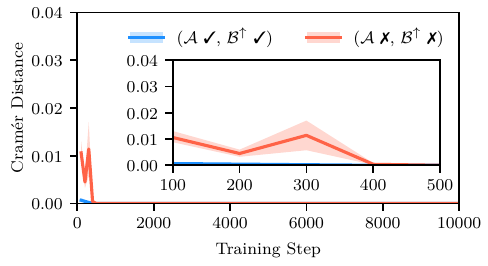}
\subcaption{$(K, w) = (1, 0.05)$}
\end{minipage}
\begin{minipage}[b]{0.49\linewidth}
\centering
\includegraphics[width=1.0\linewidth]{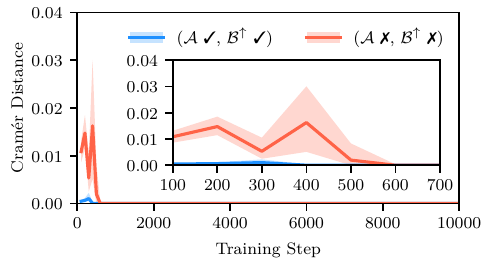}
\subcaption{$(K, w) = (3, 0.025)$}
\end{minipage}
\caption{Training dynamics of the approximation error for the negative-absorbing approximation on the scenes $(K, w) \in \{(1, 0.05), (3, 0.025)\}$: the Cram\'{e}r distance between the first-passage-time distribution given by each variant and the MC reference of the process learned by that variant up to each checkpoint, with the median and the interquartile range over the rays.}
\label{fig:approximation_error_validation/negative_absorbing/k3w0025/learned}
\end{figure}

For $(K, w) = (3, 0.025)$, the training trajectories start where $\tau$ is large, pass through the extended error region, and end inside \regfive{}.
Nevertheless, \cref{fig:approximation_error_validation/negative_absorbing/k3w0025/learned} shows that the approximation error of each variant is confined to the first few hundred iterations and remains negligible for the rest of training, for the same reasons as in \cref{sec:approximation_error_validation/survival,sec:approximation_error_validation/crossing}.
Finally, \cref{tab:approximation_error_validation/negative_absorbing/geometry} compares the full Bayesian formulation and the negative-absorbing approximation with the evaluation metrics in \cref{sec:experiments/evaluation_metrics} along with $|\text{SDF}|$, which is the mean absolute error of the learned mean field $\mu_{\bm{\theta}}$ evaluated at points sampled from the zero-level set of the ground-truth SDF.
The full Bayesian formulation is better in CRPS and ECE on every scene and no worse in MAE, which is the same trend as in \cref{tab:dtu_ablation_negative_absorbing_approx}, whereas the two variants differ by at most $0.002$ in $|\text{SDF}|$.
In summary, although \cref{sec:approximation_error_validation/survival,sec:approximation_error_validation/crossing} show that dropping $\mathcal{A}$ and $\mathcal{B}^{\uparrow}$ can incur error in principle, the error appears only in the first few hundred iterations of training and remains negligible afterward, and it does not lead to a large difference in the learned geometry, supporting the use of the negative-absorbing approximation in practice.

\begin{table}[t]
\centering
\caption{Evaluation comparison of the full Bayesian formulation and the negative-absorbing approximation with the evaluation metrics in \cref{sec:experiments/evaluation_metrics} along with $|\text{SDF}|$, which is the mean absolute error of the learned mean field $\mu_{\bm{\theta}}$ evaluated at points sampled from the zero-level set of the ground-truth SDF.}
\label{tab:approximation_error_validation/negative_absorbing/geometry}
{
\begin{tabular}{ccccrrrrr}
\toprule
$K$ & $w$ & $\mathcal{A}$ & $\mathcal{B}^{\uparrow}$ & $|\text{SDF}|$ $\downarrow$ & CRPS $\downarrow$ & MAE $\downarrow$ & ECE $\downarrow$ & Sharp. $\downarrow$ \\
\midrule
\multirow{4}{*}{1} & \multirow{2}{*}{0.05} & \cmark & \cmark & \textbf{0.002} & \textbf{0.006} & \textbf{0.011} & \textbf{0.062} & \textbf{0.014} \\
                   &                       & \xmark & \xmark & 0.004 & 0.012 & 0.012 & 0.161 & 0.015 \\
                   \cmidrule(lr){2-9}
                   & \multirow{2}{*}{0.025} & \cmark & \cmark & \textbf{0.009} & \textbf{0.009} & \textbf{0.010} & \textbf{0.200} & \textbf{0.012} \\
                   &                       & \xmark & \xmark & 0.011 & 0.010 & 0.012 & 0.212 & \textbf{0.012} \\
\midrule
\multirow{4}{*}{3} & \multirow{2}{*}{0.05} & \cmark & \cmark & 0.004 & \textbf{0.006} & \textbf{0.011} & \textbf{0.039} & 0.016 \\
                   &                       & \xmark & \xmark & \textbf{0.003} & 0.008 & \textbf{0.011} & 0.115 & \textbf{0.013} \\
                   \cmidrule(lr){2-9}
                   & \multirow{2}{*}{0.025} & \cmark & \cmark & \textbf{0.009} & \textbf{0.020} & \textbf{0.019} & \textbf{0.343} & 0.015 \\
                   &                       & \xmark & \xmark & 0.010 & 0.022 & 0.020 & 0.356 & \textbf{0.013} \\   
\bottomrule
\end{tabular}
}
\end{table}

\clearpage

\section{Per-Scene Evaluation Results}
\label{sec:per_scene_evaluation_results}

\vfill

\begin{table}[h]
\centering
\caption{Per-scene unmasked Chamfer distance (mm) and CRPS (mm) on the DTU dataset with 41M training rays, corresponding to the averages in \cref{tab:dtu_benchmark}. \textbf{Bold} and \underline{underlined} indicate the best and second-best results per scene, respectively. The last two rows report the mean and standard deviation over scenes.}
\label{tab:dtu_per_scene_41m}
\begin{subtable}{\textwidth}
\centering
\subcaption{CD $\downarrow$}
\label{tab:dtu_per_scene_41m_cd}
{
\setlength{\tabcolsep}{7pt}
\footnotesize
\begin{tabular}{rrrrrrr>{\columncolor{DodgerBlue!10}}r}
\toprule
Scan & NeuS & NeuS-Facto & UNIS-Facto (C1) & UNIS-Facto (C2) & UNIS-Facto (C3) & OaV-Facto & SSDP-Facto \\
\midrule
24 & 1.49 & 1.47 & 1.54 & 1.75 & 1.49 & \underline{1.40} & \textbf{0.93} \\
37 & 2.45 & 1.99 & \underline{1.75} & 2.18 & \textbf{1.74} & 2.06 & 1.89 \\
40 & \underline{0.66} & 2.09 & 2.30 & 2.42 & 2.37 & 1.88 & \textbf{0.50} \\
55 & \underline{0.54} & 0.63 & 0.66 & \textbf{0.50} & 0.63 & 0.81 & 0.59 \\
63 & 1.87 & \underline{1.35} & 1.37 & 1.71 & \textbf{1.34} & 2.18 & 2.02 \\
65 & \underline{0.86} & 1.75 & 1.68 & 1.30 & 1.69 & 1.72 & \textbf{0.85} \\
69 & \textbf{0.86} & 1.05 & 1.03 & 1.05 & 1.01 & 1.19 & \underline{0.89} \\
83 & \underline{1.16} & 1.28 & 1.25 & \textbf{1.12} & 1.25 & 1.37 & 1.25 \\
97 & 1.58 & 1.35 & 1.30 & \textbf{1.25} & \underline{1.28} & 1.65 & 1.30 \\
105 & 1.62 & 1.27 & \underline{1.23} & 1.34 & 1.28 & 1.34 & \textbf{0.98} \\
106 & 0.69 & \textbf{0.49} & \underline{0.56} & 0.61 & 0.57 & 0.69 & 0.67 \\
110 & \underline{2.26} & 2.63 & 2.93 & 3.22 & 2.83 & 3.11 & \textbf{1.56} \\
114 & \textbf{0.50} & 0.70 & 0.78 & 0.73 & 0.75 & 1.26 & \underline{0.62} \\
118 & 0.78 & 0.76 & 0.75 & 0.83 & \underline{0.72} & 0.80 & \textbf{0.64} \\
122 & \underline{0.68} & 0.87 & 0.91 & 0.92 & 0.91 & 0.95 & \textbf{0.61} \\
\midrule
Mean & \underline{1.20} & 1.31 & 1.34 & 1.40 & 1.32 & 1.49 & \textbf{1.02} \\
SD & 0.64 & 0.60 & 0.64 & 0.75 & 0.64 & 0.64 & 0.48 \\
\bottomrule
\end{tabular}
}
\end{subtable}

\vspace{10pt}

\begin{subtable}{\textwidth}
\centering
\subcaption{CRPS $\downarrow$}
\label{tab:dtu_per_scene_41m_crps}
{
\setlength{\tabcolsep}{7pt}
\footnotesize
\begin{tabular}{rrrrrrr>{\columncolor{DodgerBlue!10}}r}
\toprule
Scan & NeuS & NeuS-Facto & UNIS-Facto (C1) & UNIS-Facto (C2) & UNIS-Facto (C3) & OaV-Facto & SSDP-Facto \\
\midrule
24 & \underline{3.82} & 5.08 & 4.79 & 6.73 & 4.99 & 5.50 & \textbf{2.99} \\
37 & 5.61 & 5.76 & \underline{5.29} & 6.00 & \underline{5.29} & 7.25 & \textbf{4.96} \\
40 & \underline{2.04} & 6.44 & 6.61 & 9.03 & 7.35 & 9.43 & \textbf{1.03} \\
55 & 1.66 & 1.67 & 1.61 & \underline{1.40} & 1.63 & 7.76 & \textbf{0.64} \\
63 & 6.22 & \underline{5.13} & 5.23 & 5.64 & 5.19 & 5.75 & \textbf{4.77} \\
65 & \underline{1.79} & 2.89 & 2.73 & 2.87 & 2.85 & 4.15 & \textbf{1.21} \\
69 & 2.03 & 2.06 & 1.95 & \underline{1.90} & 1.94 & 2.26 & \textbf{1.57} \\
83 & 1.67 & \underline{1.63} & \underline{1.63} & 1.67 & \textbf{1.62} & 1.82 & \textbf{1.62} \\
97 & 2.41 & 2.18 & \underline{1.99} & \textbf{1.88} & 2.03 & 2.63 & 2.29 \\
105 & 1.43 & 1.29 & \underline{1.28} & \underline{1.28} & 1.29 & 1.59 & \textbf{1.21} \\
106 & 2.21 & 2.16 & 2.15 & \underline{1.99} & 2.17 & 4.38 & \textbf{1.61} \\
110 & \underline{4.27} & 5.25 & 5.25 & 5.78 & 5.47 & 4.58 & \textbf{2.65} \\
114 & 1.23 & 1.17 & 1.17 & \underline{1.10} & 1.23 & 1.44 & \textbf{0.61} \\
118 & 1.96 & 1.83 & 1.83 & \underline{1.54} & 1.73 & 2.85 & \textbf{1.30} \\
122 & 1.89 & 1.95 & 1.87 & \underline{1.73} & 1.89 & 3.95 & \textbf{1.22} \\
\midrule
Mean & \underline{2.68} & 3.10 & 3.03 & 3.37 & 3.11 & 4.35 & \textbf{1.98} \\
SD & 1.55 & 1.85 & 1.83 & 2.54 & 1.97 & 2.41 & 1.35 \\
\bottomrule
\end{tabular}
}
\end{subtable}
\end{table}

\vfill

\begin{table}[h]
\centering
\caption{Per-scene unmasked Chamfer distance (mm) and CRPS (mm) on the DTU dataset with 154M training rays, corresponding to the averages in \cref{tab:dtu_benchmark}. \textbf{Bold} and \underline{underlined} indicate the best and second-best results per scene, respectively. The last two rows report the mean and standard deviation over scenes. $^{*}$Reproduced with the official implementation.}
\label{tab:dtu_per_scene_154m}
\begin{subtable}{\textwidth}
\centering
\subcaption{CD $\downarrow$}
\label{tab:dtu_per_scene_154m_cd}
{
\setlength{\tabcolsep}{5pt}
\footnotesize
\begin{tabular}{rrrrrrrr>{\columncolor{DodgerBlue!10}}r}
\toprule
Scan & NeuS & OaV$^{*}$ & NeuS-Facto & UNIS-Facto (C1) & UNIS-Facto (C2) & UNIS-Facto (C3) & OaV-Facto & SSDP-Facto \\
\midrule
24 & 1.44 & 2.14 & 1.23 & 1.29 & 1.57 & 1.16 & \underline{1.07} & \textbf{0.86} \\
37 & 1.87 & 2.50 & 1.61 & \underline{1.50} & 1.87 & \textbf{1.47} & 1.70 & 1.73 \\
40 & \underline{0.91} & 1.69 & 2.09 & 2.10 & 2.24 & 2.89 & 2.03 & \textbf{0.46} \\
55 & \textbf{0.49} & 0.99 & 0.58 & 0.60 & \underline{0.53} & 0.58 & 0.73 & 0.65 \\
63 & 1.33 & 1.55 & \underline{1.18} & \textbf{1.14} & 1.33 & 1.20 & 1.72 & 1.73 \\
65 & \underline{0.74} & 2.10 & 1.57 & 1.43 & 1.13 & 1.45 & 1.56 & \textbf{0.67} \\
69 & \textbf{0.70} & 1.16 & 0.97 & 0.93 & 0.95 & 0.95 & 1.05 & \underline{0.80} \\
83 & \underline{0.95} & 1.57 & 1.13 & 1.03 & 1.09 & 1.07 & 1.12 & \textbf{0.81} \\
97 & 1.17 & 1.97 & \underline{1.14} & \textbf{1.10} & 1.15 & \textbf{1.10} & 1.38 & 1.19 \\
105 & 1.47 & 1.33 & 1.12 & \underline{1.06} & 1.18 & 1.15 & 1.13 & \textbf{0.85} \\
106 & 0.59 & 0.73 & \textbf{0.41} & 0.51 & \underline{0.47} & \underline{0.47} & 0.52 & 0.62 \\
110 & \underline{1.42} & 1.57 & 1.91 & 2.09 & 2.41 & 1.95 & 2.17 & \textbf{1.21} \\
114 & \textbf{0.42} & 0.81 & 0.54 & 0.62 & \underline{0.47} & 0.60 & 1.12 & 0.56 \\
118 & 0.62 & 1.20 & \underline{0.61} & \textbf{0.56} & 0.71 & 0.64 & 0.71 & \textbf{0.56} \\
122 & \textbf{0.58} & 1.67 & 0.85 & 0.89 & \underline{0.78} & 0.81 & 0.81 & \textbf{0.58} \\
\midrule
Mean & \underline{0.98} & 1.53 & 1.13 & 1.12 & 1.19 & 1.16 & 1.25 & \textbf{0.88} \\
SD & 0.44 & 0.51 & 0.50 & 0.50 & 0.61 & 0.62 & 0.49 & 0.40 \\
\bottomrule
\end{tabular}
}
\end{subtable}

\vspace{10pt}

\begin{subtable}{\textwidth}
\centering
\subcaption{CRPS $\downarrow$}
\label{tab:dtu_per_scene_154m_crps}
{
\setlength{\tabcolsep}{5pt}
\footnotesize
\begin{tabular}{rrrrrrrr>{\columncolor{DodgerBlue!10}}r}
\toprule
Scan & NeuS & OaV$^{*}$ & NeuS-Facto & UNIS-Facto (C1) & UNIS-Facto (C2) & UNIS-Facto (C3) & OaV-Facto & SSDP-Facto \\
\midrule
24 & \underline{3.30} & 4.18 & 4.38 & 4.22 & 4.82 & 4.18 & 5.10 & \textbf{2.72} \\
37 & 6.65 & \textbf{4.12} & 5.28 & 5.07 & 5.23 & 5.14 & 6.14 & \underline{4.77} \\
40 & \underline{1.99} & 2.24 & 6.24 & 6.13 & 8.52 & 6.92 & 10.36 & \textbf{1.11} \\
55 & 1.76 & \textbf{0.74} & 1.81 & 1.69 & 1.44 & 1.71 & 12.91 & \underline{0.80} \\
63 & 4.84 & \textbf{3.84} & 4.65 & 4.68 & 4.56 & 4.63 & 4.93 & \underline{4.19} \\
65 & 1.84 & \underline{1.66} & 2.78 & 2.45 & 2.82 & 2.64 & 3.99 & \textbf{1.14} \\
69 & 1.90 & \underline{1.58} & 1.94 & 1.88 & 1.75 & 1.88 & 2.22 & \textbf{1.46} \\
83 & 1.61 & \underline{1.47} & 1.58 & 1.56 & 1.60 & 1.56 & 1.79 & \textbf{1.45} \\
97 & 2.11 & 2.36 & \underline{1.92} & 1.93 & \textbf{1.85} & 1.97 & 2.48 & 2.37 \\
105 & 1.33 & 1.38 & 1.22 & \underline{1.21} & \underline{1.21} & 1.22 & 1.65 & \textbf{1.01} \\
106 & 2.20 & \textbf{1.40} & 2.03 & 2.19 & 1.89 & 2.19 & 4.39 & \underline{1.61} \\
110 & 2.96 & \textbf{1.84} & 4.02 & 3.92 & 4.33 & 3.78 & 3.53 & \underline{2.27} \\
114 & 1.14 & \textbf{0.52} & 1.12 & 1.11 & 0.93 & 1.15 & 1.51 & \underline{0.53} \\
118 & 1.87 & \textbf{1.22} & 1.84 & 1.82 & 1.63 & 1.84 & 3.43 & \underline{1.24} \\
122 & 2.00 & \underline{1.31} & 1.94 & 1.94 & 1.75 & 1.95 & 4.88 & \textbf{1.25} \\
\midrule
Mean & 2.50 & \underline{1.99} & 2.85 & 2.79 & 2.96 & 2.85 & 4.62 & \textbf{1.86} \\
SD & 1.47 & 1.17 & 1.62 & 1.58 & 2.11 & 1.69 & 3.21 & 1.22 \\
\bottomrule
\end{tabular}
}
\end{subtable}
\end{table}

\begin{table}[h]
\centering
\caption{Per-scene unmasked Chamfer distance (mm) and CRPS (mm) on the MobileBrick dataset, corresponding to the averages in \cref{tab:lego_benchmark}. \textbf{Bold} and \underline{underlined} indicate the best and second-best results per scene, respectively. The last two rows report the mean and standard deviation over scenes.}
\label{tab:lego_per_scene}
\begin{subtable}{\textwidth}
\centering
\subcaption{CD $\downarrow$}
\label{tab:lego_per_scene_cd}
{
\setlength{\tabcolsep}{5pt}
\footnotesize
\begin{tabular}{lrrrrrr>{\columncolor{DodgerBlue!10}}r}
\toprule
Scene & NeuS & NeuS-Facto & UNIS-Facto (C1) & UNIS-Facto (C2) & UNIS-Facto (C3) & OaV-Facto & SSDP-Facto \\
\midrule
Aston & 8.31 & 6.04 & 5.68 & \underline{5.60} & \textbf{5.55} & 8.06 & 8.16 \\
Audi & 3.86 & 4.16 & 3.97 & \underline{3.73} & 4.13 & 4.38 & \textbf{3.41} \\
Beetles & 5.40 & 5.12 & 5.02 & \underline{4.96} & \textbf{4.90} & 5.36 & 5.38 \\
Big Ben & 6.94 & 6.86 & 6.98 & 8.36 & \underline{6.47} & 9.15 & \textbf{4.17} \\
Boat & 3.60 & 3.35 & 3.38 & 3.37 & \underline{3.25} & 3.52 & \textbf{3.24} \\
Bridge & \underline{2.57} & 2.81 & 3.04 & 3.35 & 2.89 & 3.58 & \textbf{2.42} \\
Cabin & 4.03 & \underline{3.69} & 3.80 & 4.10 & 3.77 & 4.20 & \textbf{3.28} \\
Camera & \textbf{1.60} & 2.91 & 2.91 & 3.06 & 2.60 & 3.08 & \underline{2.20} \\
Castle & 9.60 & \textbf{7.50} & 8.41 & 9.46 & \underline{7.83} & 8.17 & 9.13 \\
Colosseum & 11.28 & 16.92 & 21.69 & \textbf{9.54} & 22.21 & 22.57 & \underline{10.07} \\
Convertible & 7.42 & 5.71 & \underline{5.48} & 5.67 & 5.55 & \textbf{5.24} & 5.59 \\
Ferrari & 5.49 & 4.33 & \textbf{4.18} & 5.55 & \underline{4.27} & 4.32 & 4.79 \\
Jeep & \underline{2.90} & 3.25 & 3.28 & 3.08 & 3.24 & 5.59 & \textbf{2.62} \\
London Bus & \textbf{4.95} & \underline{5.29} & 6.09 & 6.50 & 5.77 & 5.38 & 5.40 \\
Motorcycle & 2.40 & \textbf{2.12} & 2.21 & 2.30 & \underline{2.19} & 2.29 & 2.20 \\
Porsche & 9.08 & 5.79 & \textbf{5.28} & 5.75 & \underline{5.34} & 7.20 & 7.19 \\
Satellite & 16.60 & 8.96 & 16.17 & \underline{4.42} & 15.42 & 18.01 & \textbf{4.31} \\
Space Shuttle & 7.09 & 6.68 & 6.60 & \underline{6.06} & 6.83 & 7.54 & \textbf{5.82} \\
\midrule
Mean & 6.28 & 5.64 & 6.34 & \underline{5.27} & 6.23 & 7.09 & \textbf{4.97} \\
SD & 3.76 & 3.35 & 4.95 & 2.15 & 4.97 & 5.23 & 2.38 \\
\bottomrule
\end{tabular}
}
\end{subtable}

\vspace{10pt}

\begin{subtable}{\textwidth}
\centering
\subcaption{CRPS $\downarrow$}
\label{tab:lego_per_scene_crps}
{
\setlength{\tabcolsep}{5pt}
\footnotesize
\begin{tabular}{lrrrrrr>{\columncolor{DodgerBlue!10}}r}
\toprule
Scene & NeuS & NeuS-Facto & UNIS-Facto (C1) & UNIS-Facto (C2) & UNIS-Facto (C3) & OaV-Facto & SSDP-Facto \\
\midrule
Aston & \textbf{4.92} & 5.91 & 5.90 & 5.61 & 5.88 & \underline{5.05} & 5.48 \\
Audi & \textbf{1.78} & 2.13 & 2.11 & \textbf{1.78} & \underline{2.02} & 2.55 & 2.88 \\
Beetles & 3.83 & 3.77 & \underline{3.59} & \textbf{3.48} & \underline{3.59} & 4.24 & 3.76 \\
Big Ben & 5.82 & 5.37 & 5.42 & 9.25 & \underline{4.91} & 7.95 & \textbf{3.29} \\
Boat & 2.97 & \textbf{2.28} & 2.42 & \underline{2.32} & 2.35 & 2.94 & 2.50 \\
Bridge & 5.45 & 4.72 & \underline{4.71} & 4.90 & \underline{4.71} & \textbf{4.69} & 5.08 \\
Cabin & \textbf{2.65} & 2.93 & 2.88 & 3.15 & 3.03 & 3.22 & \underline{2.70} \\
Camera & 1.67 & 1.47 & \underline{1.39} & 1.54 & \textbf{1.34} & 1.46 & 1.50 \\
Castle & 6.91 & 5.00 & \textbf{4.93} & 5.23 & 4.96 & \underline{4.94} & 5.28 \\
Colosseum & \underline{9.98} & 12.69 & 19.37 & \textbf{7.57} & 19.55 & 19.02 & 10.18 \\
Convertible & 5.95 & 5.02 & 4.89 & \textbf{4.79} & \underline{4.85} & 4.97 & 5.63 \\
Ferrari & 6.29 & 6.80 & 5.97 & \textbf{5.31} & 6.70 & 7.11 & \underline{5.46} \\
Jeep & 2.12 & 1.78 & 1.68 & 2.17 & \underline{1.62} & 1.73 & \textbf{1.39} \\
London Bus & 3.69 & 3.62 & \textbf{3.54} & 3.87 & \underline{3.58} & 3.59 & 4.05 \\
Motorcycle & 2.32 & \underline{2.04} & 2.05 & \textbf{1.91} & 2.11 & 2.63 & 2.06 \\
Porsche & 7.44 & 7.07 & \textbf{6.69} & 7.14 & 6.88 & 7.20 & \underline{6.74} \\
Satellite & 5.41 & 4.93 & 5.82 & \underline{4.49} & 5.70 & 5.72 & \textbf{4.25} \\
Space Shuttle & 3.94 & \underline{2.74} & \textbf{2.73} & 2.76 & 2.81 & 3.14 & \underline{2.74} \\
\midrule
Mean & 4.62 & 4.46 & 4.78 & \underline{4.29} & 4.81 & 5.12 & \textbf{4.17} \\
SD & 2.25 & 2.68 & 4.01 & 2.17 & 4.06 & 3.94 & 2.16 \\
\bottomrule
\end{tabular}
}
\end{subtable}
\end{table}

\clearpage

\section{Additional Visualization Results}
\label{sec:additional_visualization_results}

\vfill

\begin{figure}[h]

\centering

\begin{minipage}[b]{0.245\linewidth}
\centering
\includegraphics[width=1.0\linewidth]{images/dtu/scan24/gt/000023.jpg}
\end{minipage}
\begin{minipage}[b]{0.245\linewidth}
\centering
\includegraphics[width=1.0\linewidth]{images/dtu/scan24/neus-facto-quadruple/000023.jpg}
\end{minipage}
\begin{minipage}[b]{0.245\linewidth}
\centering
\includegraphics[width=1.0\linewidth]{images/dtu/scan24/oav-facto-quadruple/000023.jpg}
\end{minipage}
\begin{minipage}[b]{0.245\linewidth}
\centering
\includegraphics[width=1.0\linewidth]{images/dtu/scan24/ssdp-facto-quadruple/000023.jpg}
\end{minipage}

\begin{minipage}[b]{0.245\linewidth}
\centering
\includegraphics[width=1.0\linewidth]{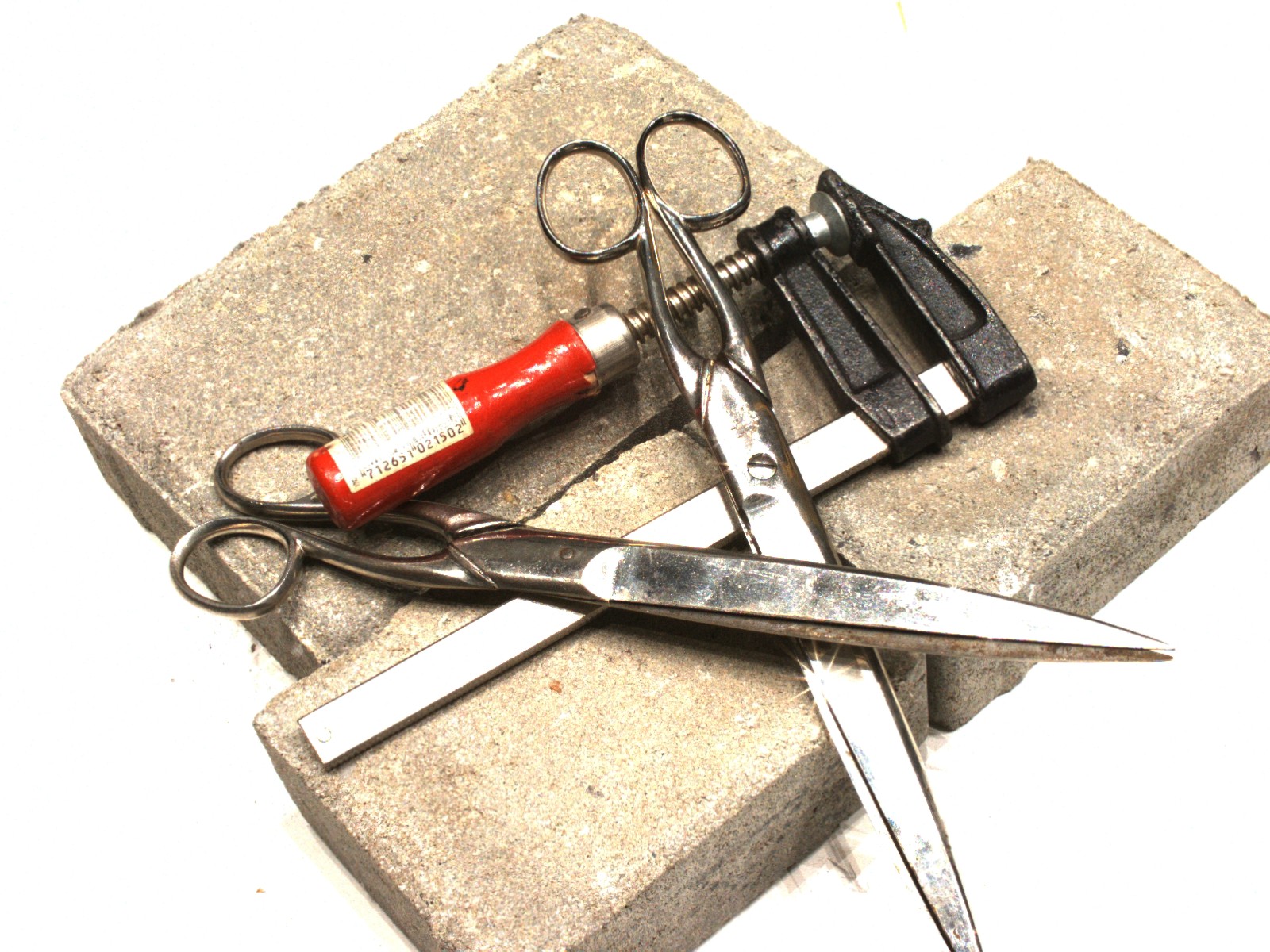}
\end{minipage}
\begin{minipage}[b]{0.245\linewidth}
\centering
\includegraphics[width=1.0\linewidth]{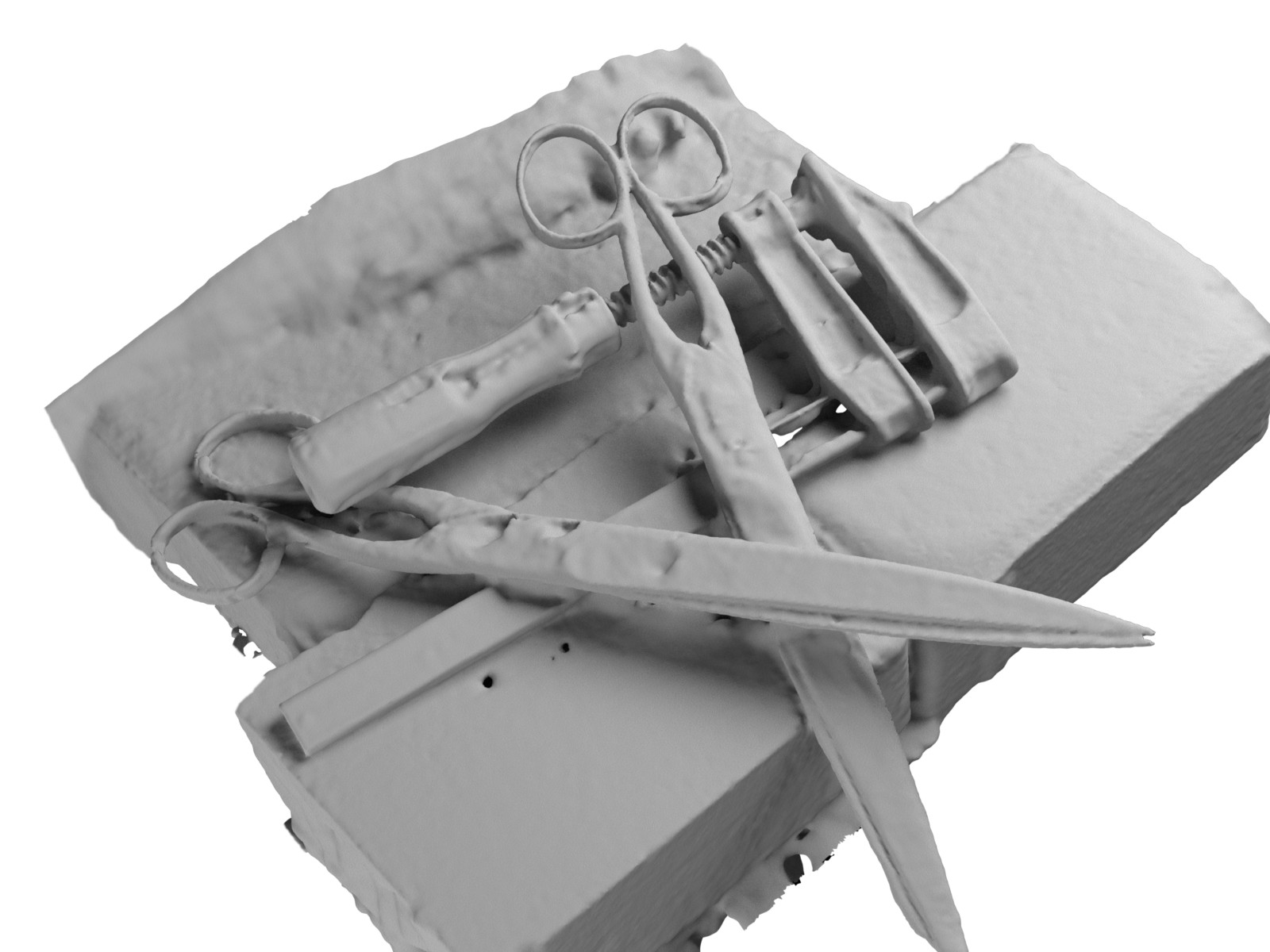}
\end{minipage}
\begin{minipage}[b]{0.245\linewidth}
\centering
\includegraphics[width=1.0\linewidth]{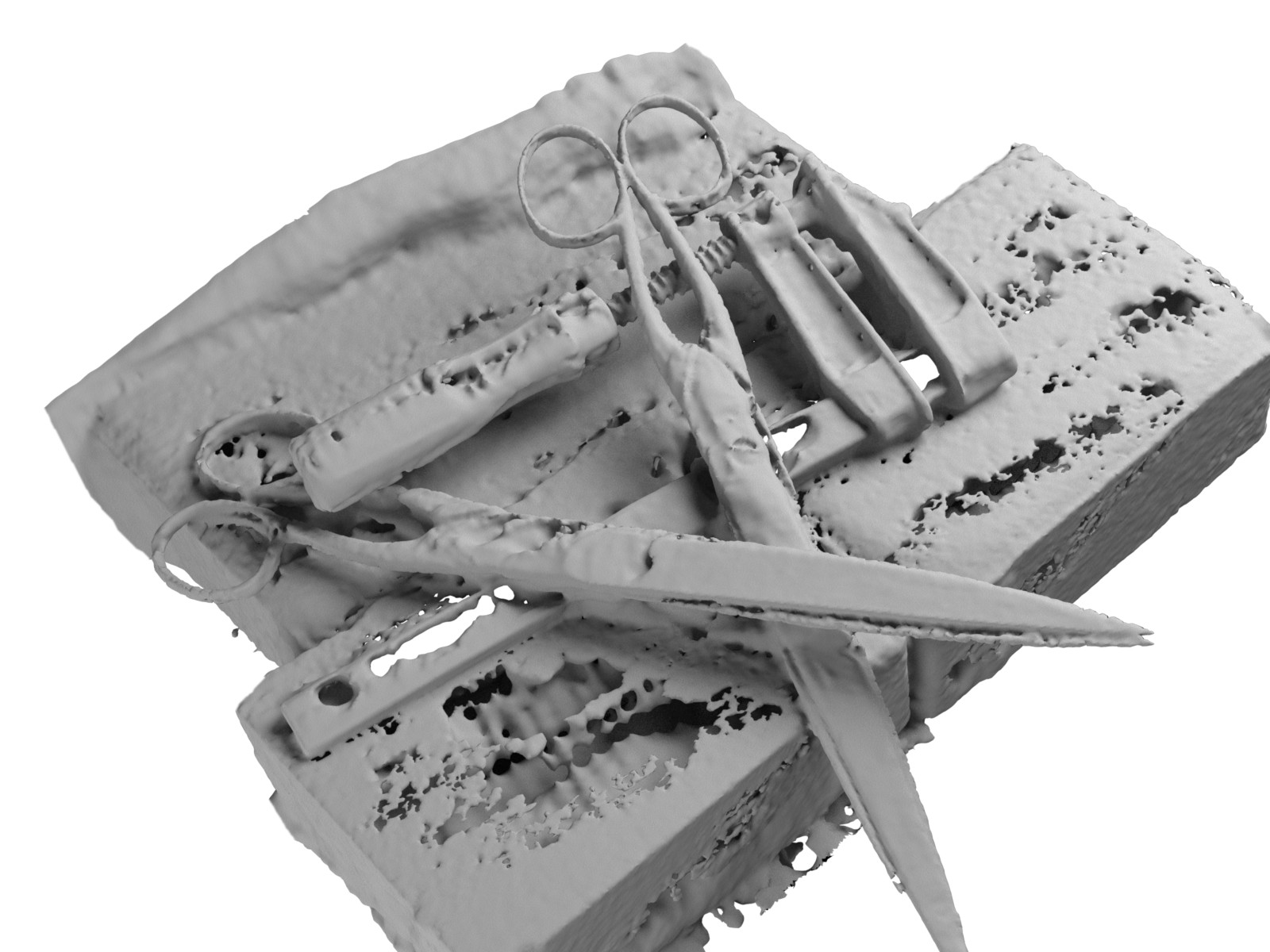}
\end{minipage}
\begin{minipage}[b]{0.245\linewidth}
\centering
\includegraphics[width=1.0\linewidth]{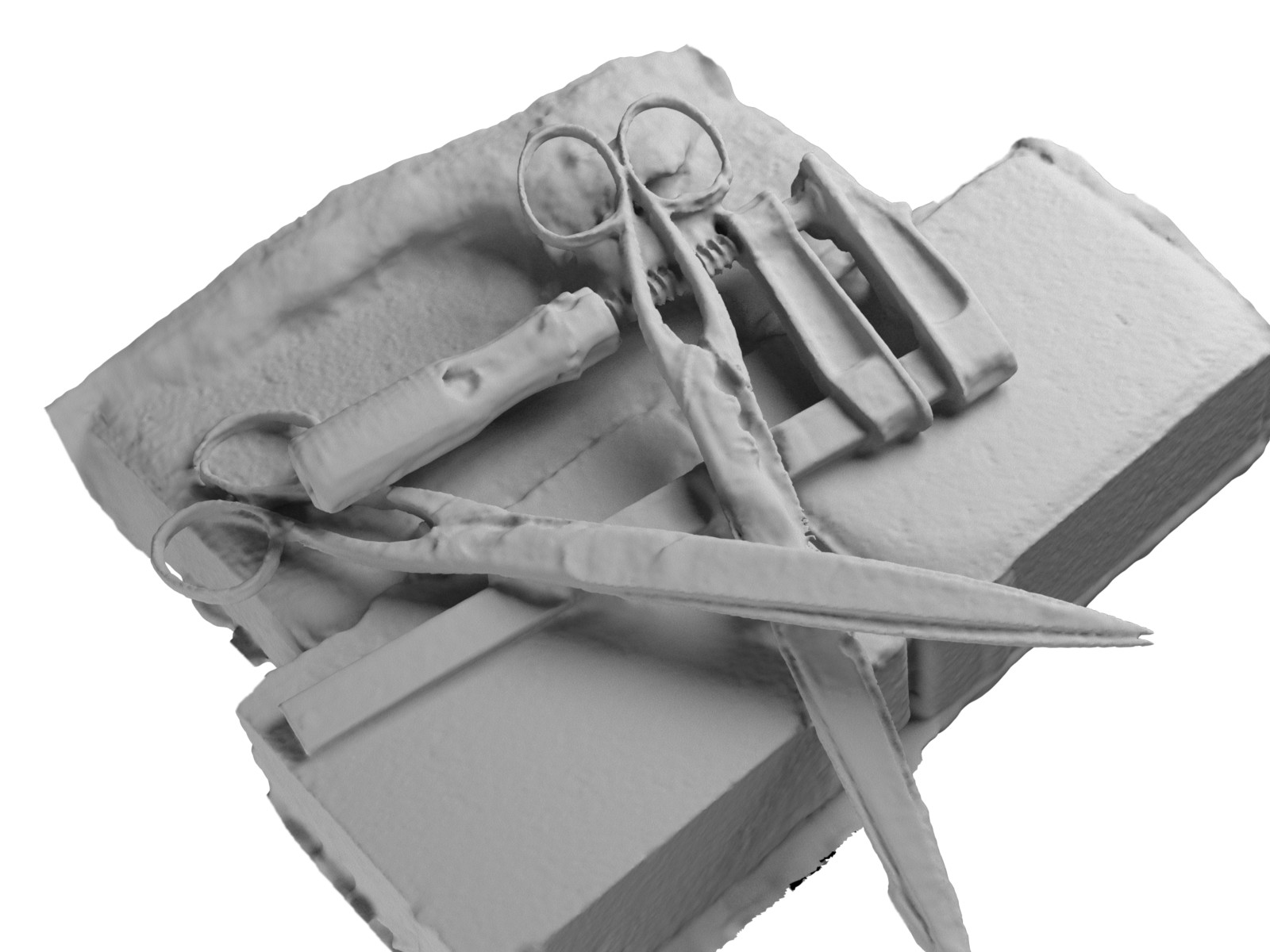}
\end{minipage}

\begin{minipage}[b]{0.245\linewidth}
\centering
\includegraphics[width=1.0\linewidth]{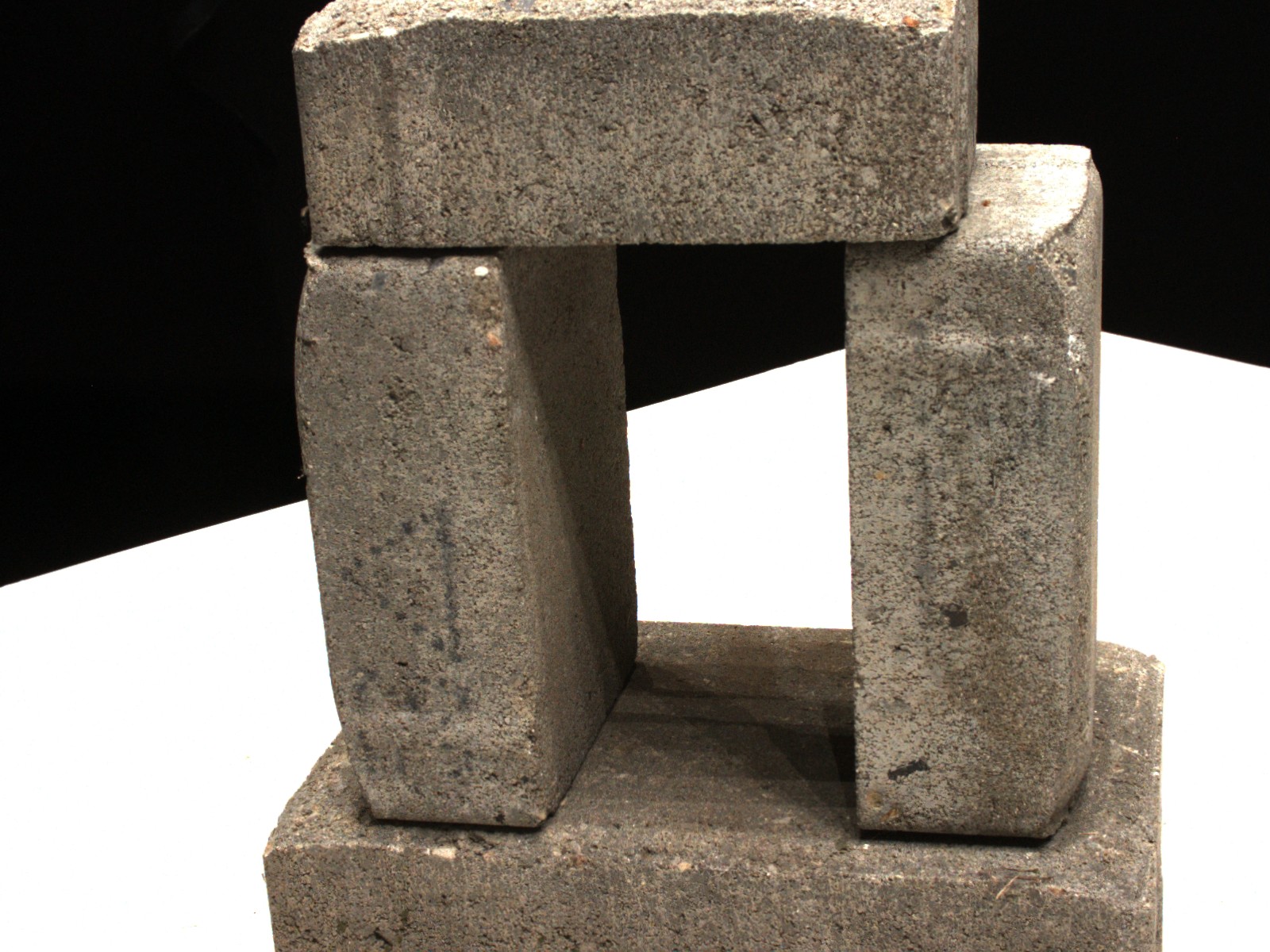}
\end{minipage}
\begin{minipage}[b]{0.245\linewidth}
\centering
\includegraphics[width=1.0\linewidth]{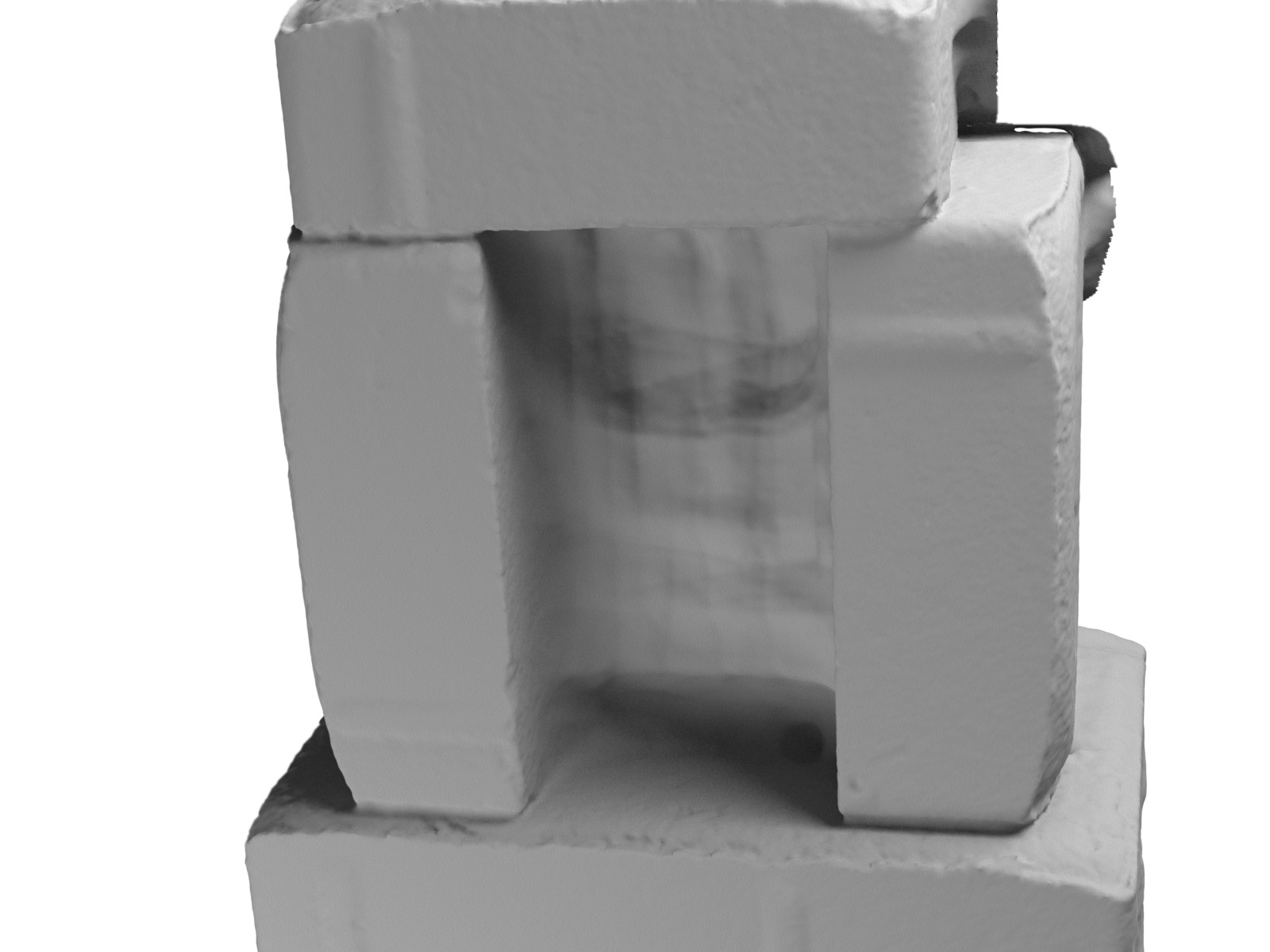}
\end{minipage}
\begin{minipage}[b]{0.245\linewidth}
\centering
\includegraphics[width=1.0\linewidth]{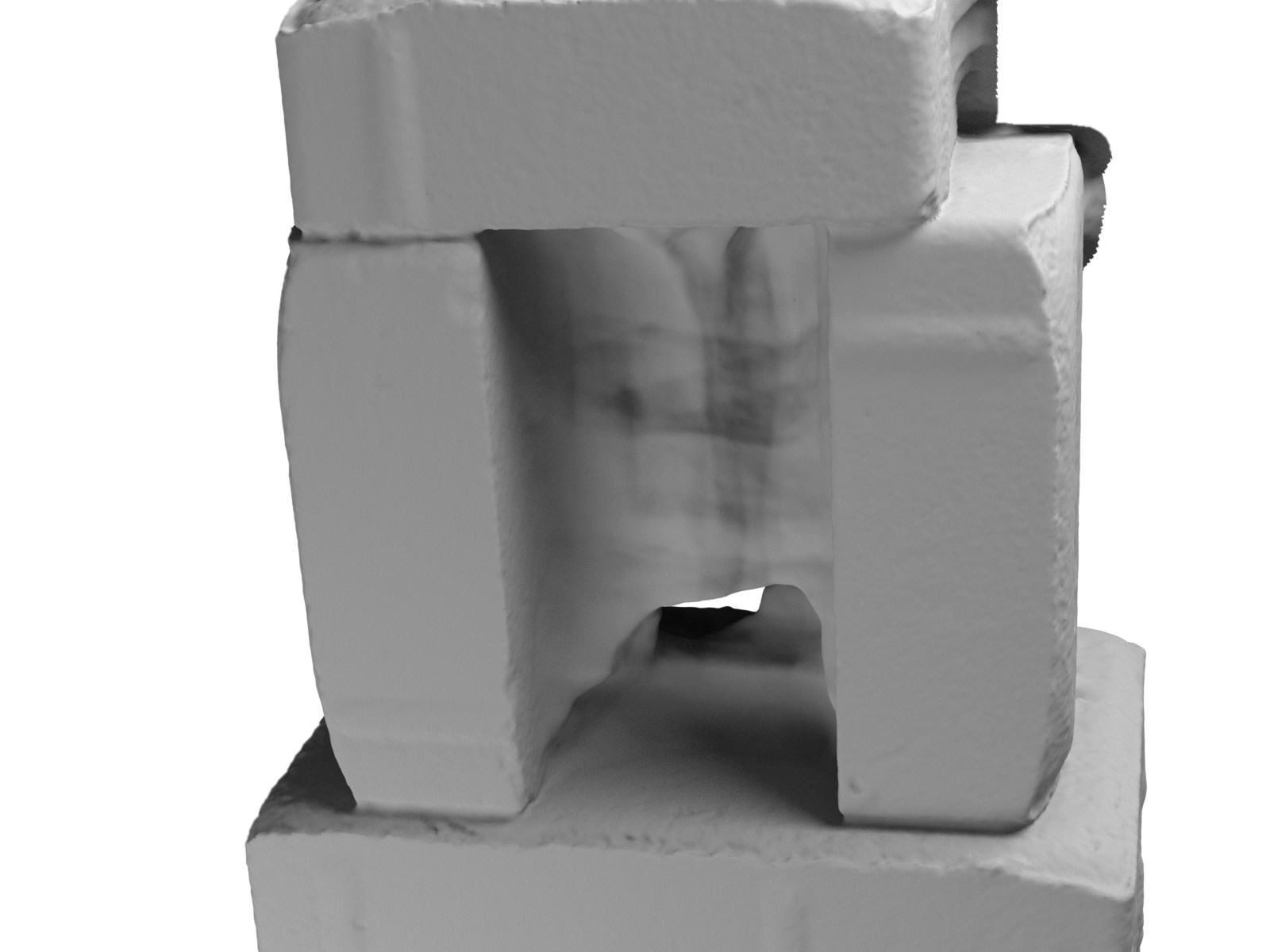}
\end{minipage}
\begin{minipage}[b]{0.245\linewidth}
\centering
\includegraphics[width=1.0\linewidth]{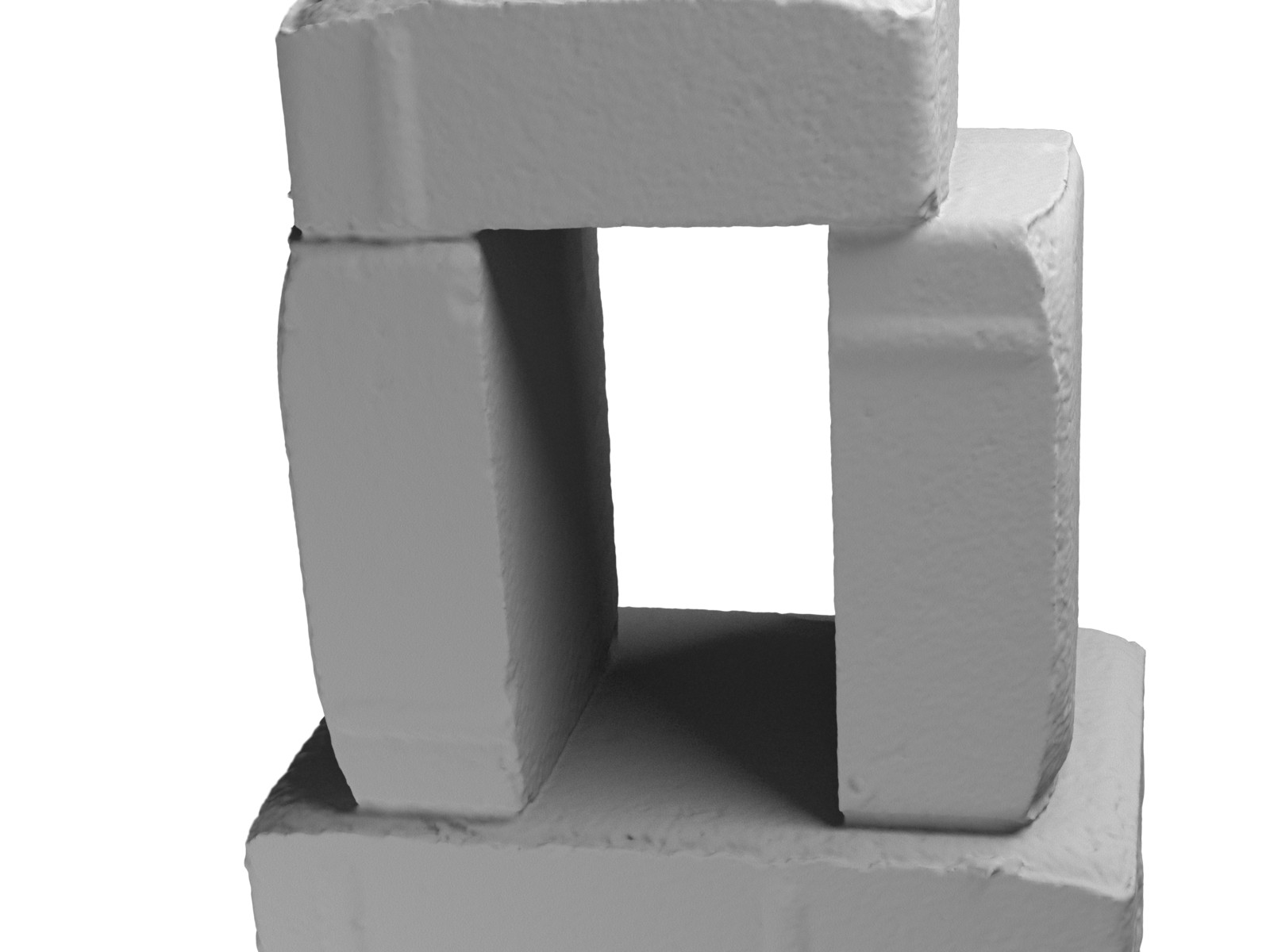}
\end{minipage}

\begin{minipage}[b]{0.245\linewidth}
\centering
\includegraphics[width=1.0\linewidth]{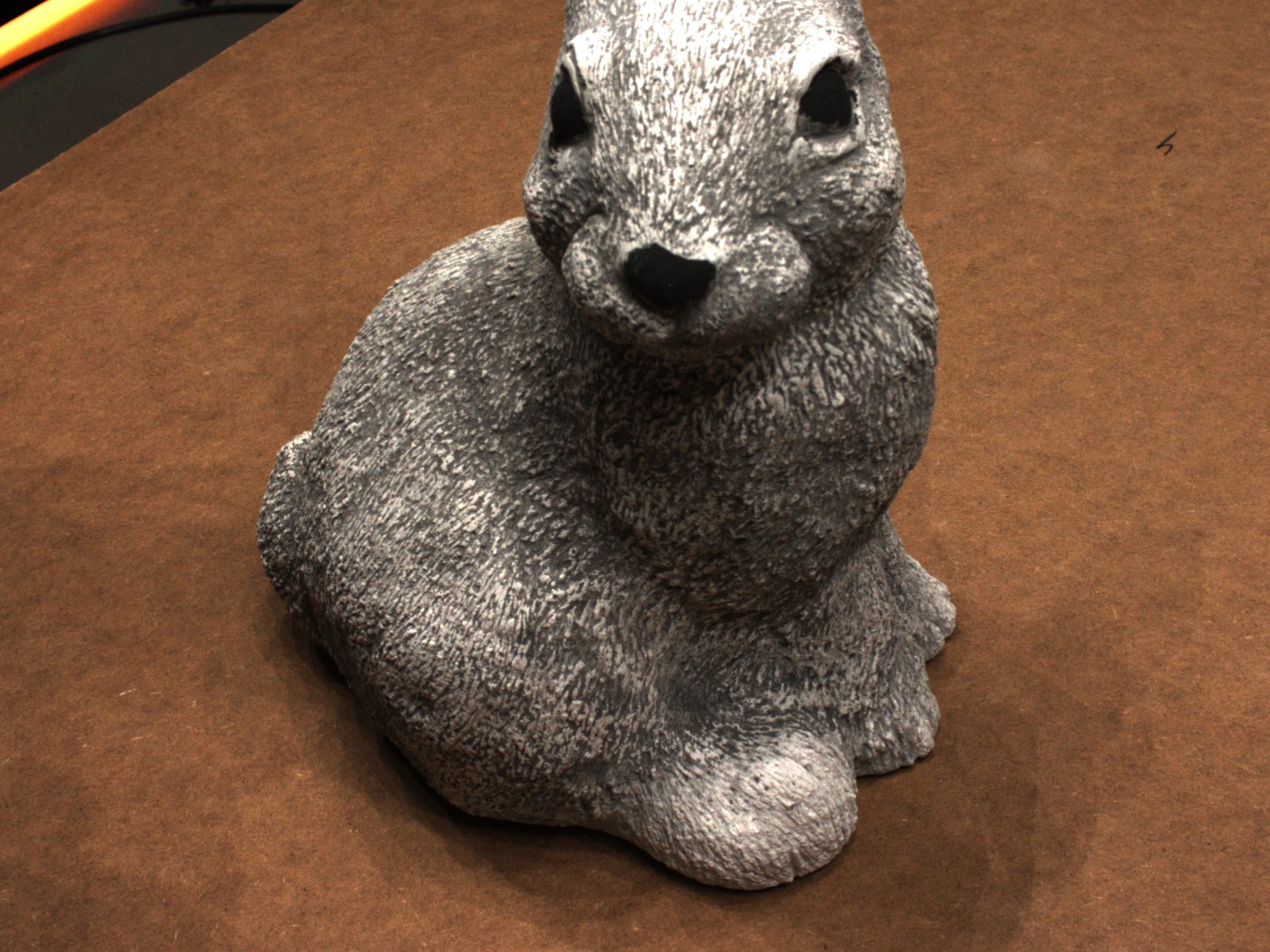}
\end{minipage}
\begin{minipage}[b]{0.245\linewidth}
\centering
\includegraphics[width=1.0\linewidth]{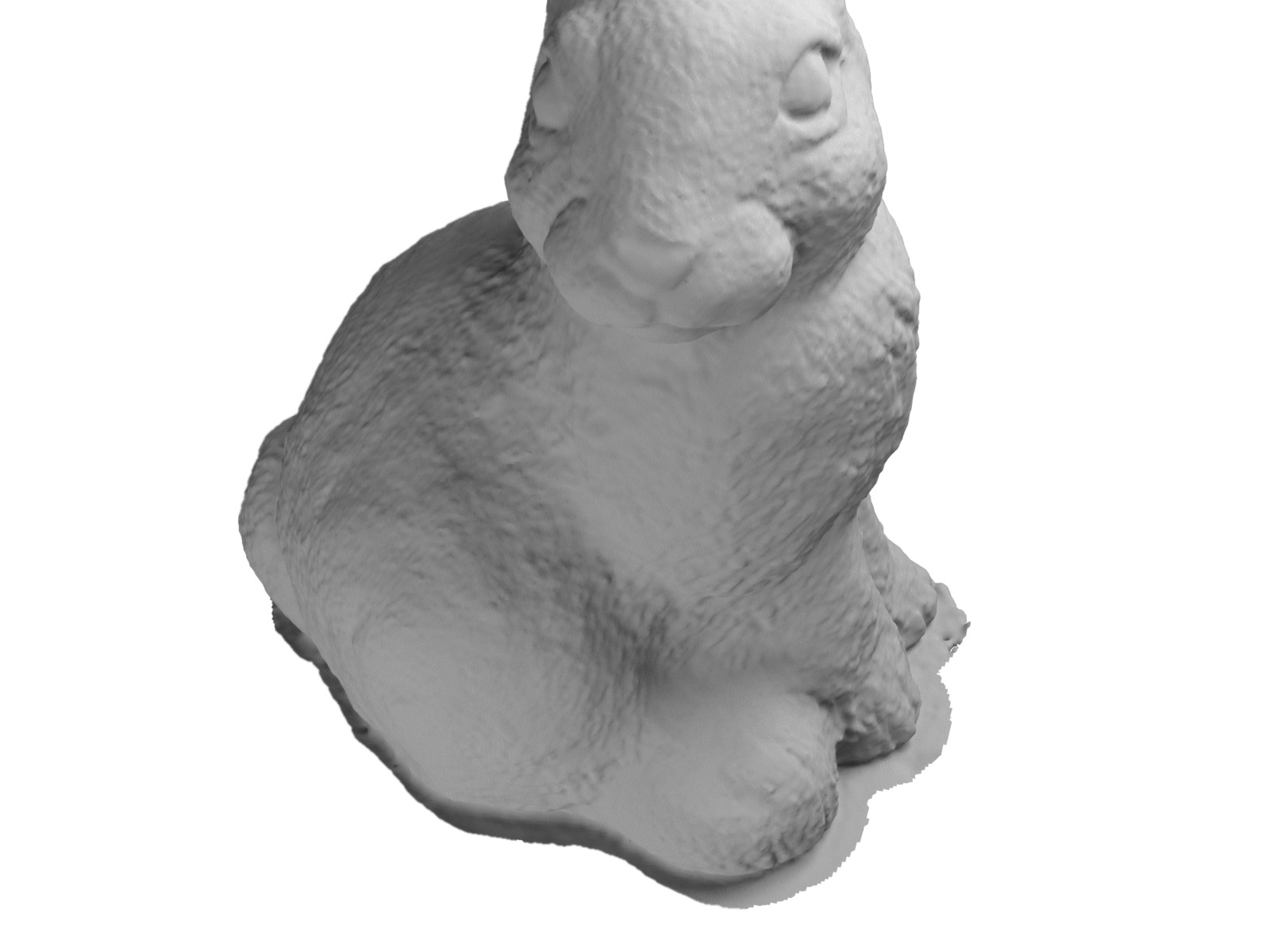}
\end{minipage}
\begin{minipage}[b]{0.245\linewidth}
\centering
\includegraphics[width=1.0\linewidth]{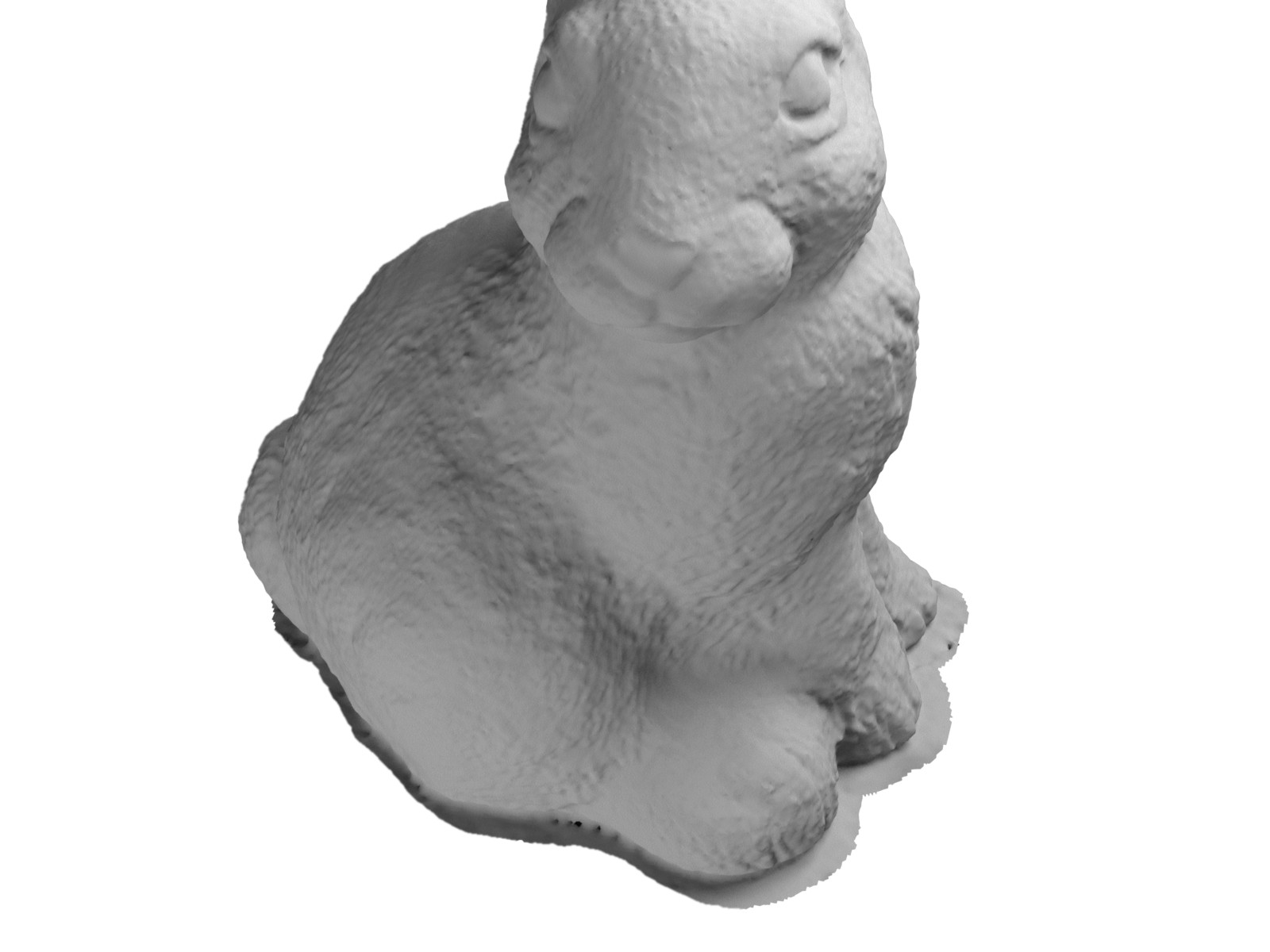}
\end{minipage}
\begin{minipage}[b]{0.245\linewidth}
\centering
\includegraphics[width=1.0\linewidth]{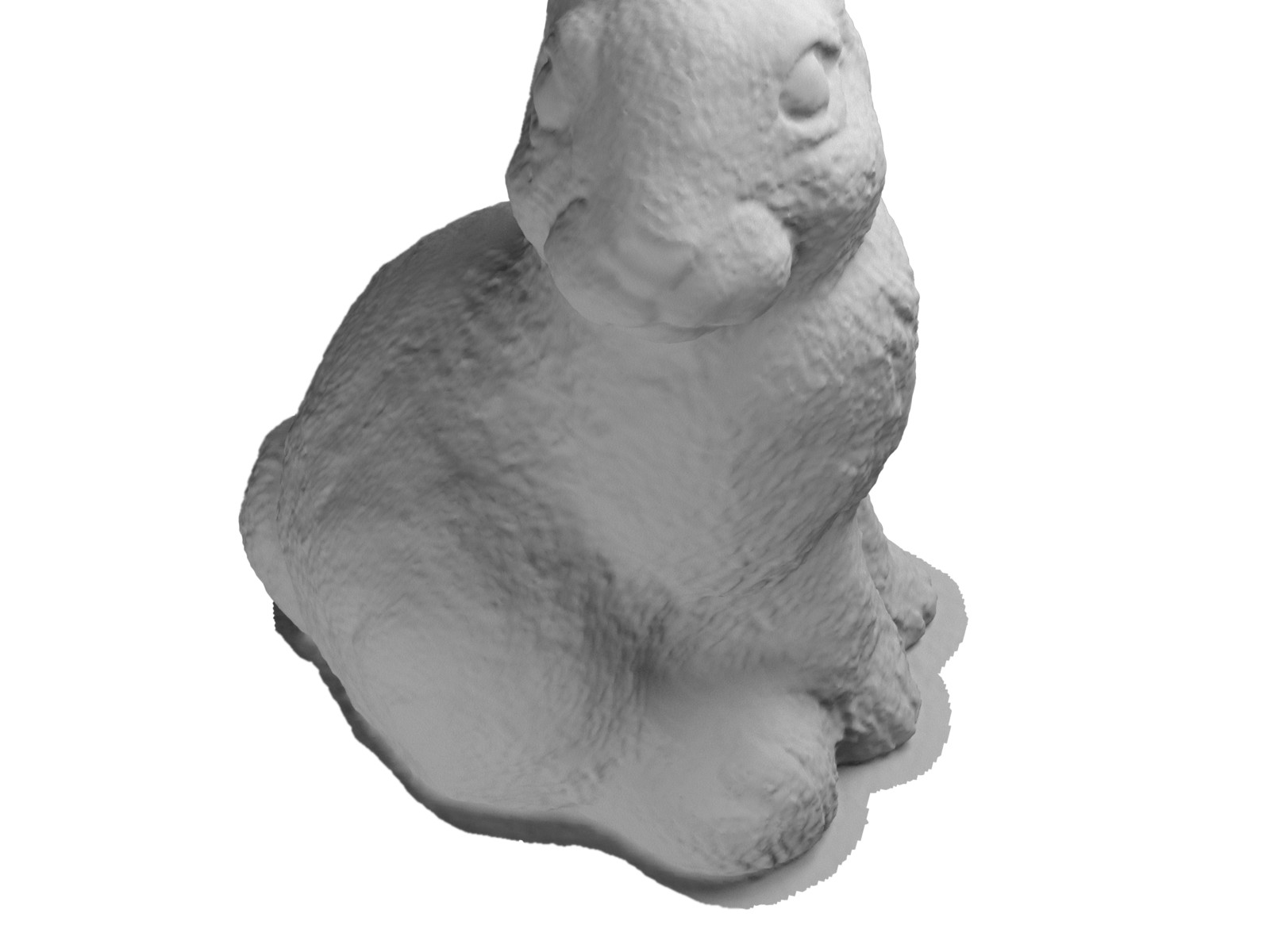}
\end{minipage}

\begin{minipage}[b]{0.245\linewidth}
\centering
\includegraphics[width=1.0\linewidth]{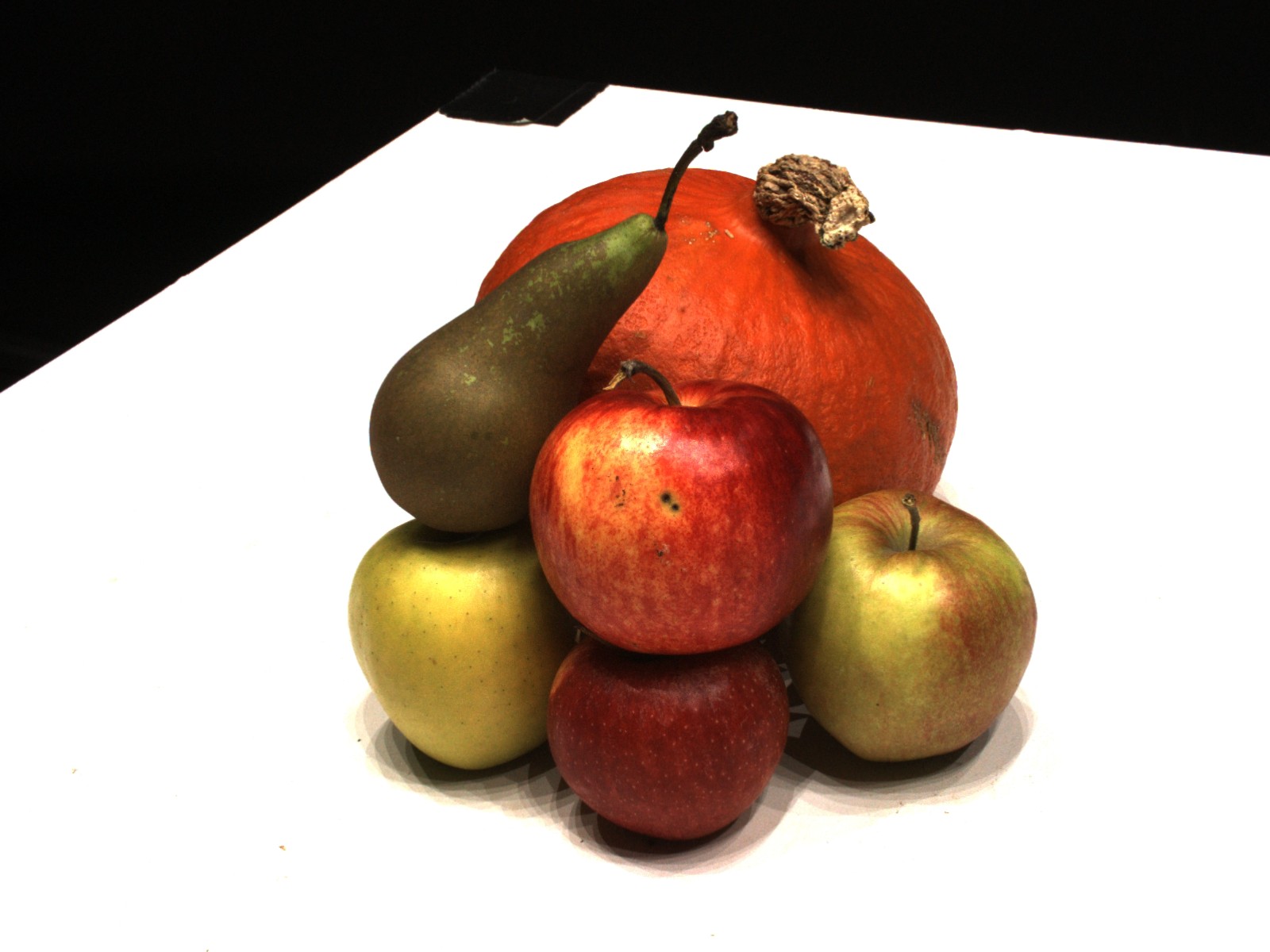}
\subcaption{Ground Truth}
\end{minipage}
\begin{minipage}[b]{0.245\linewidth}
\centering
\includegraphics[width=1.0\linewidth]{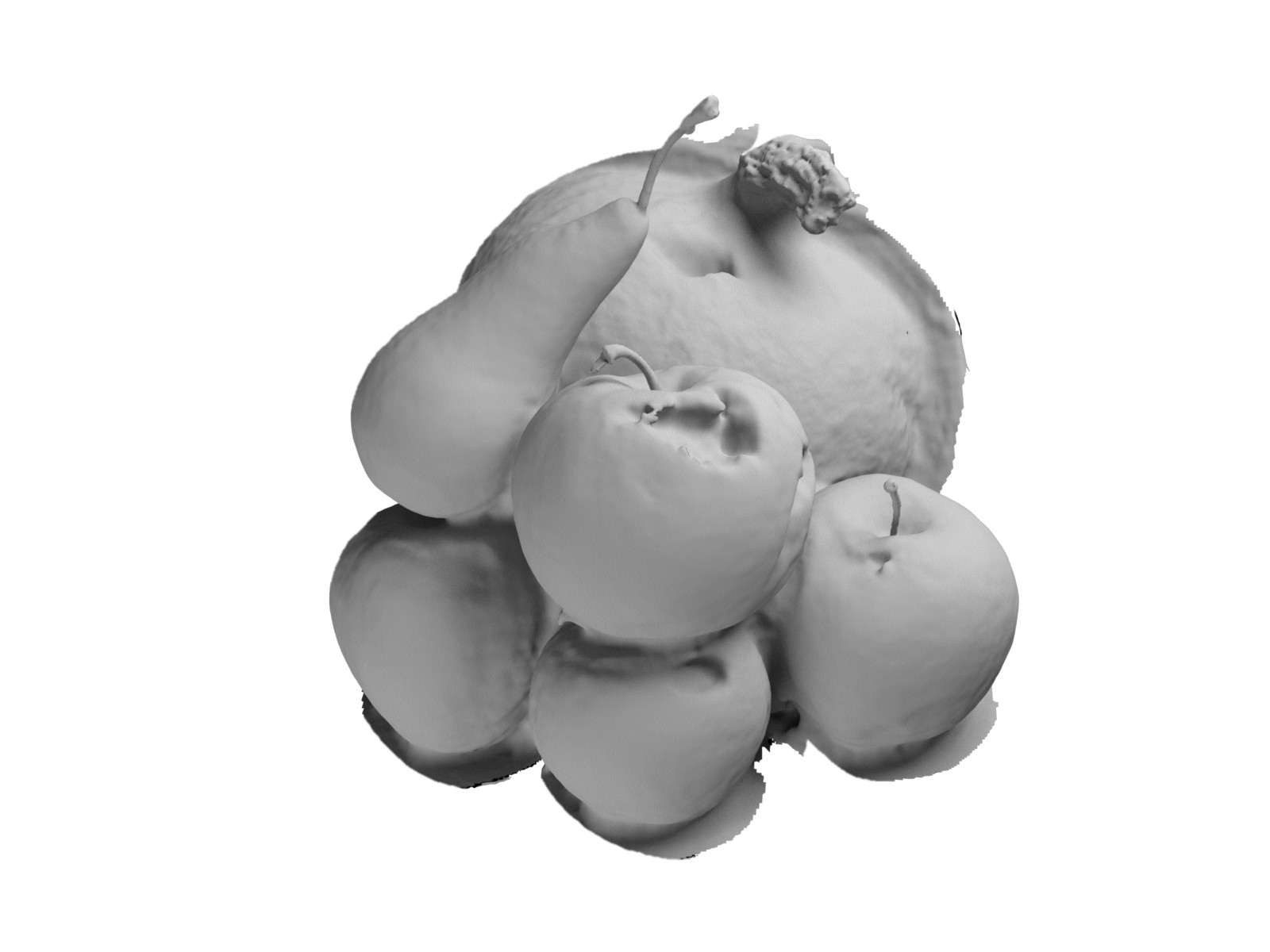}
\subcaption{NeuS-Facto}
\end{minipage}
\begin{minipage}[b]{0.245\linewidth}
\centering
\includegraphics[width=1.0\linewidth]{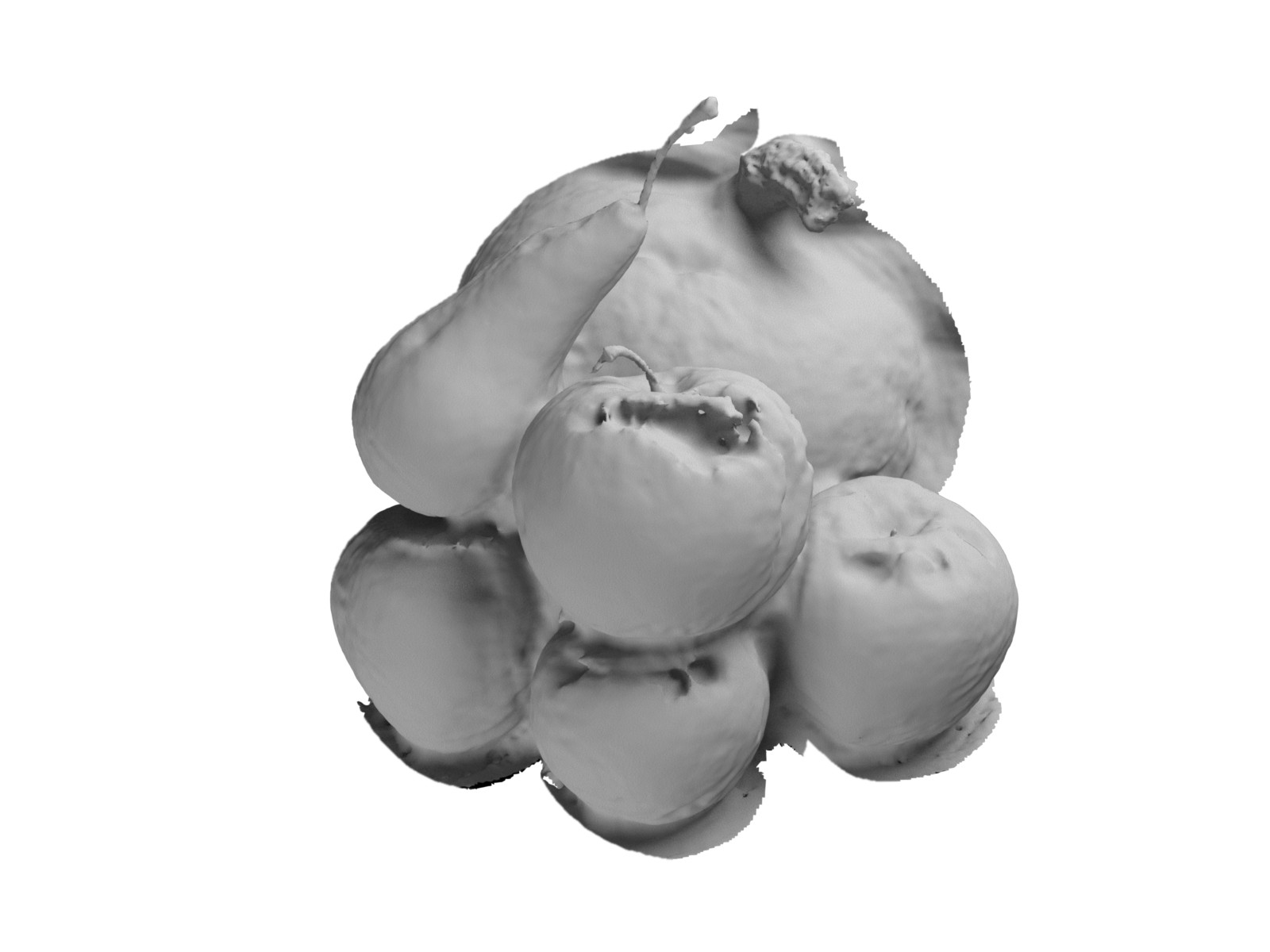}
\subcaption{OaV-Facto}
\end{minipage}
\begin{minipage}[b]{0.245\linewidth}
\centering
\includegraphics[width=1.0\linewidth]{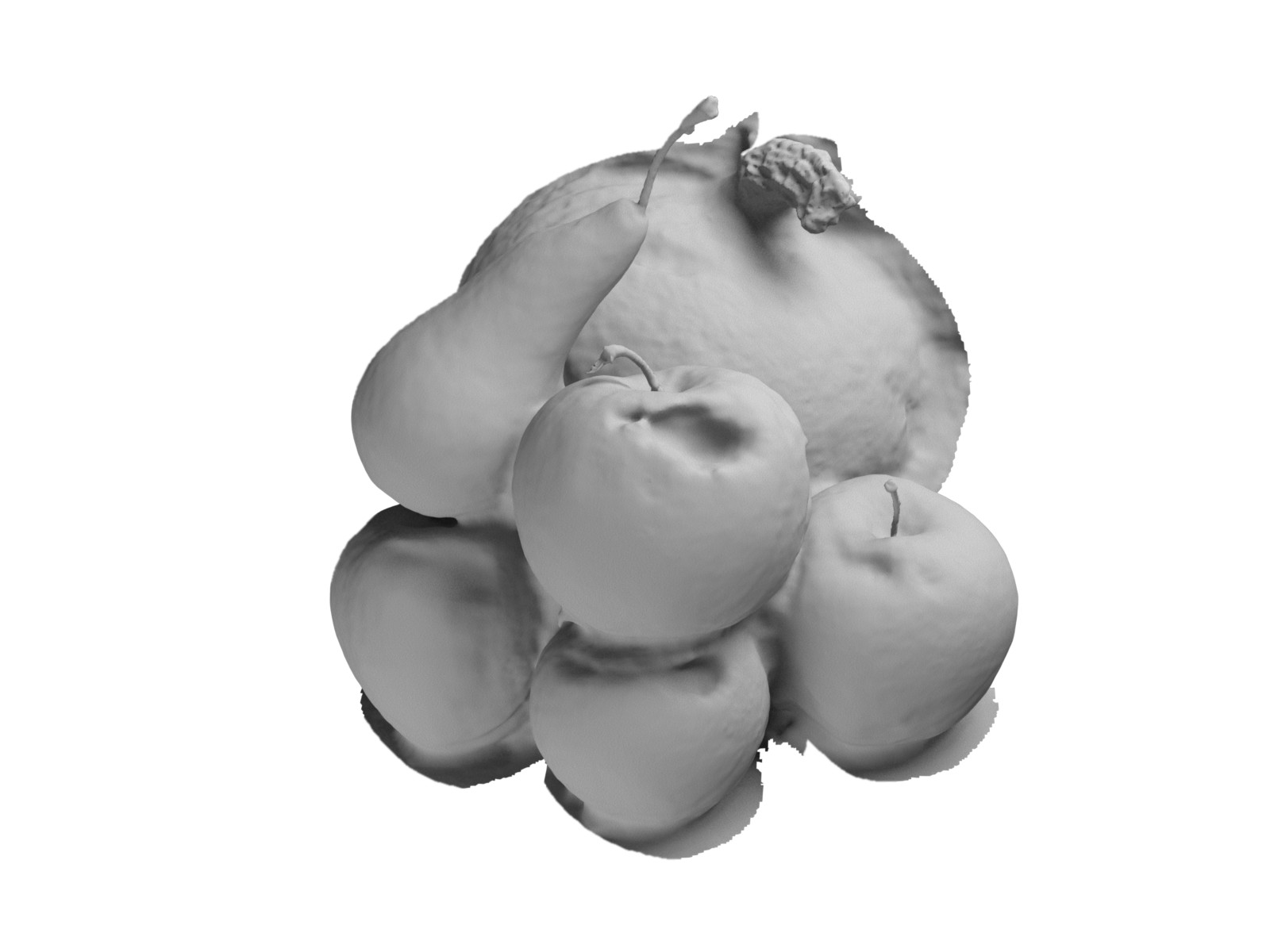}
\subcaption{SSDP-Facto}
\end{minipage}

\caption{Visualization examples on the DTU dataset.}
\label{fig:dtu_visualization_examples_all}
\end{figure}

\vfill

\begin{figure}

\ContinuedFloat
\setcounter{subfigure}{0}

\centering

\begin{minipage}[b]{0.245\linewidth}
\centering
\includegraphics[width=1.0\linewidth]{images/dtu/scan65/gt/000014.jpg}
\end{minipage}
\begin{minipage}[b]{0.245\linewidth}
\centering
\includegraphics[width=1.0\linewidth]{images/dtu/scan65/neus-facto-quadruple/000014.jpg}
\end{minipage}
\begin{minipage}[b]{0.245\linewidth}
\centering
\includegraphics[width=1.0\linewidth]{images/dtu/scan65/oav-facto-quadruple/000014.jpg}
\end{minipage}
\begin{minipage}[b]{0.245\linewidth}
\centering
\includegraphics[width=1.0\linewidth]{images/dtu/scan65/ssdp-facto-quadruple/000014.jpg}
\end{minipage}

\begin{minipage}[b]{0.245\linewidth}
\centering
\includegraphics[width=1.0\linewidth]{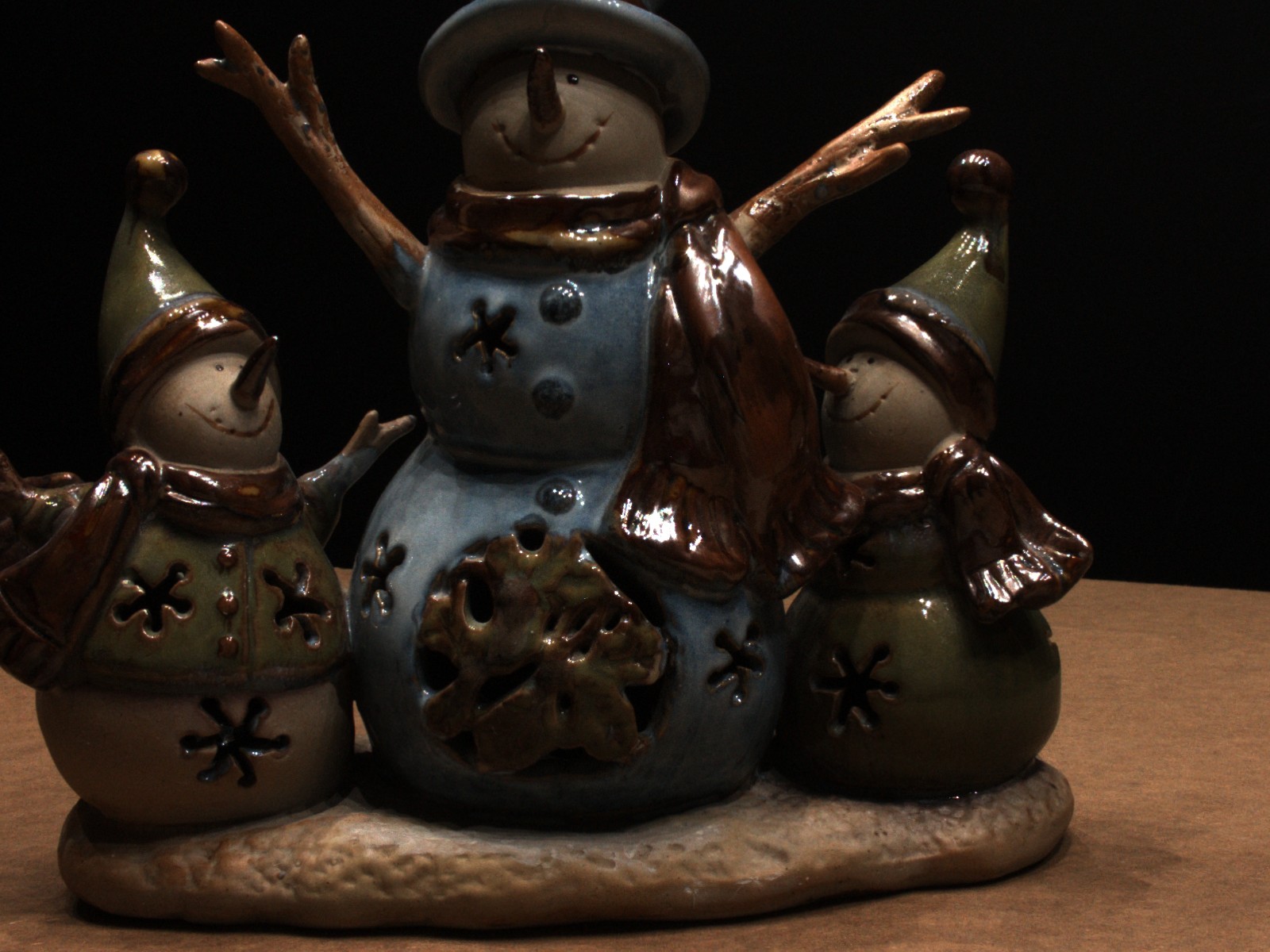}
\end{minipage}
\begin{minipage}[b]{0.245\linewidth}
\centering
\includegraphics[width=1.0\linewidth]{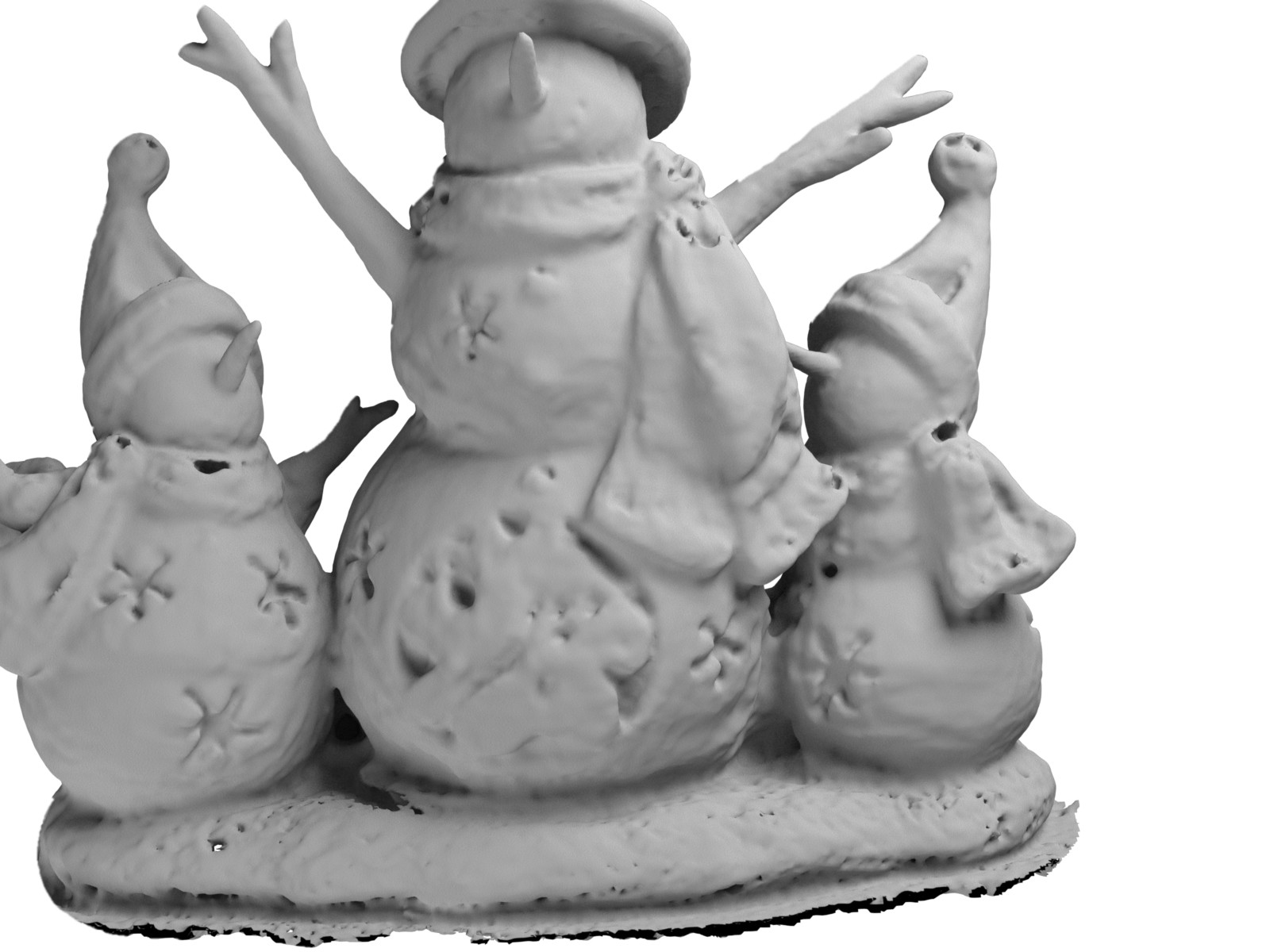}
\end{minipage}
\begin{minipage}[b]{0.245\linewidth}
\centering
\includegraphics[width=1.0\linewidth]{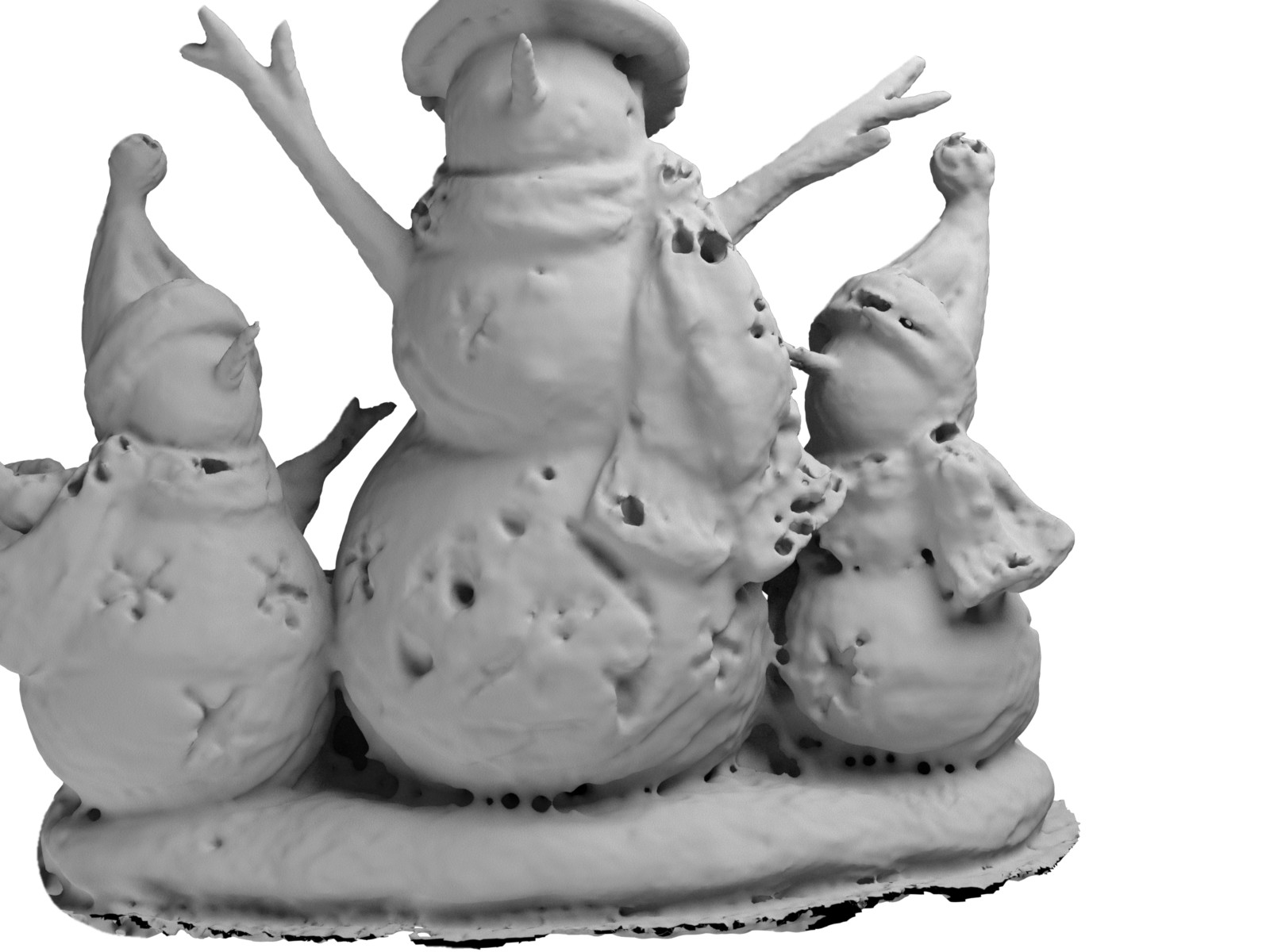}
\end{minipage}
\begin{minipage}[b]{0.245\linewidth}
\centering
\includegraphics[width=1.0\linewidth]{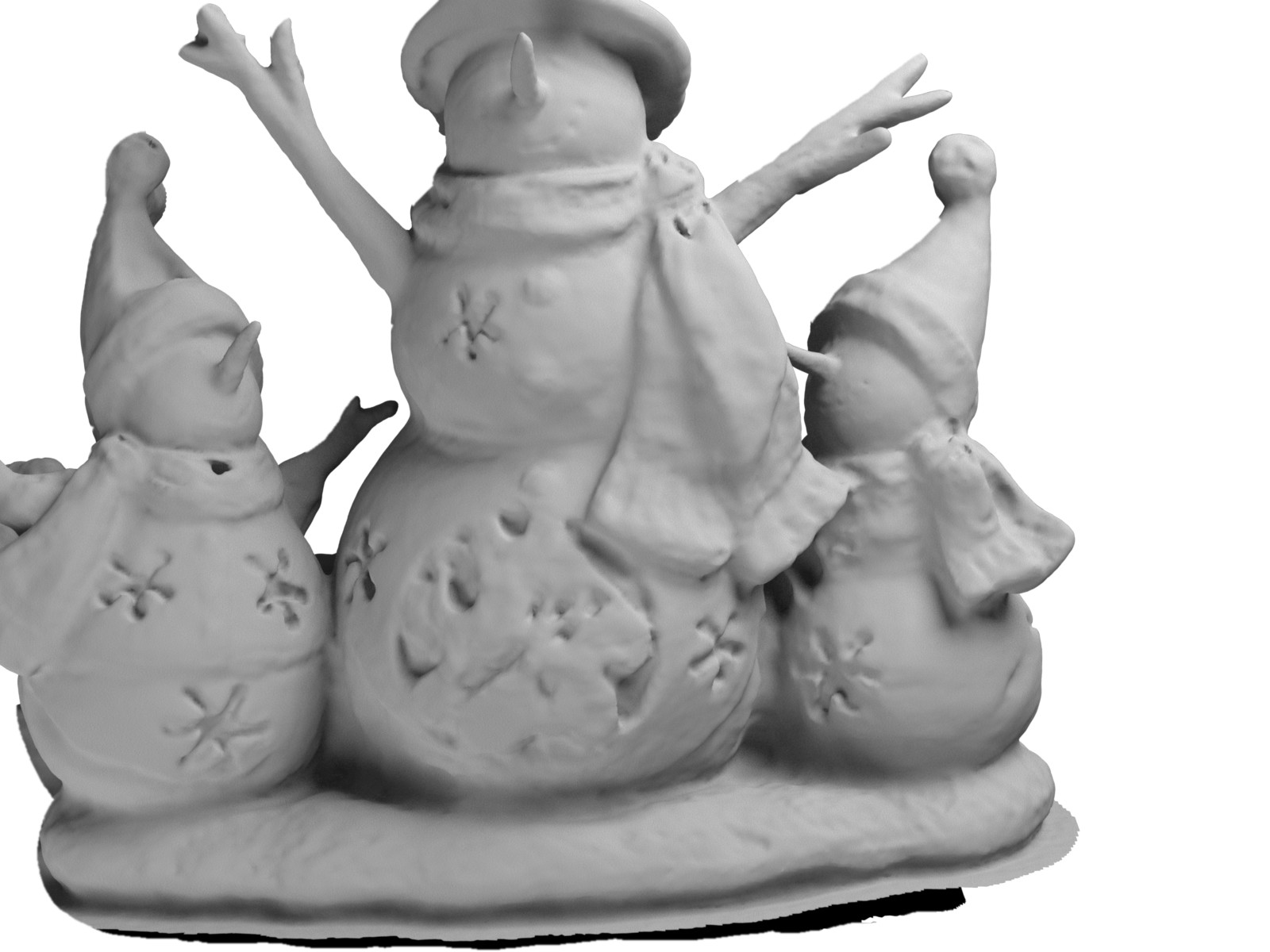}
\end{minipage}

\begin{minipage}[b]{0.245\linewidth}
\centering
\includegraphics[width=1.0\linewidth]{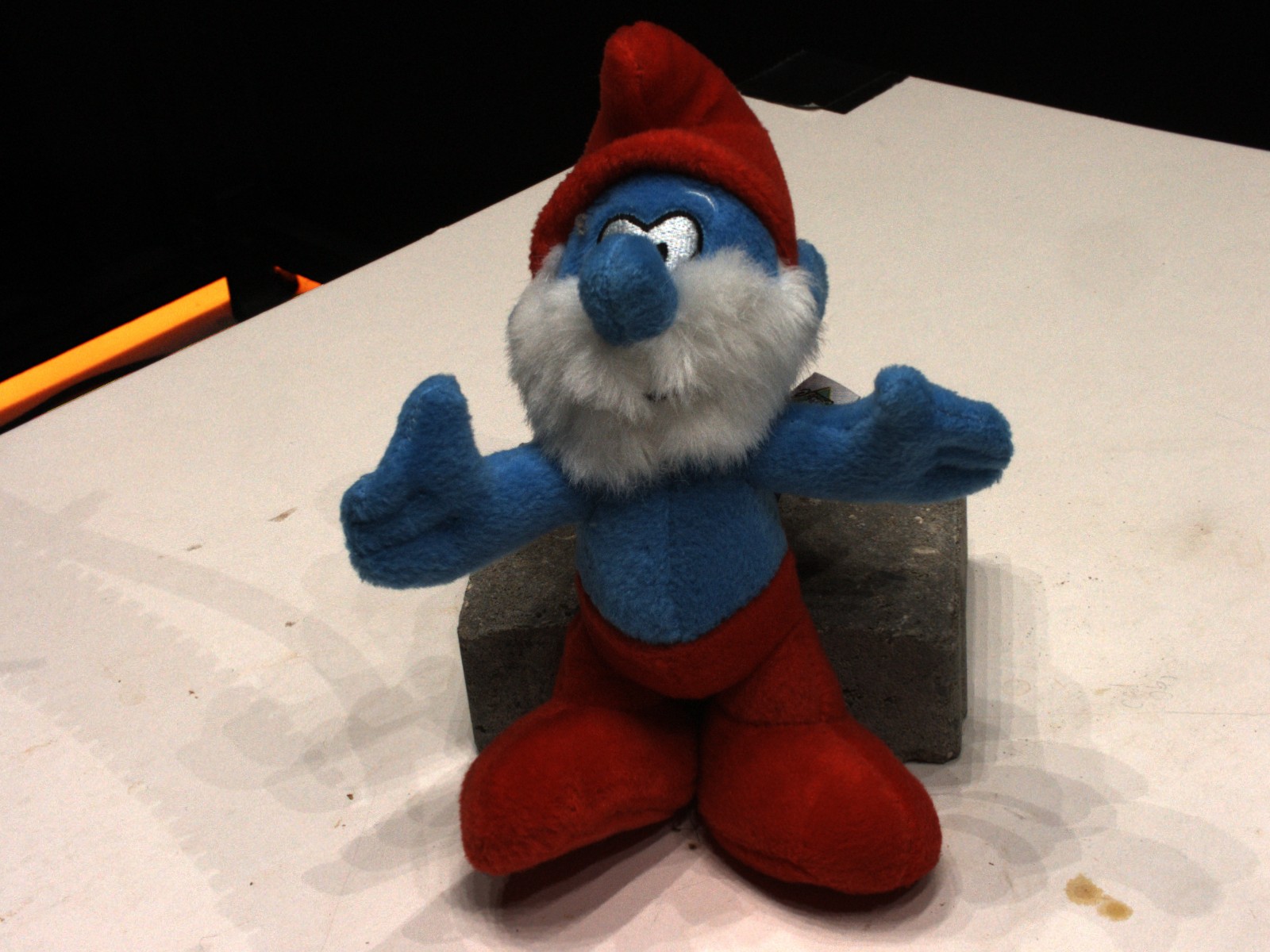}
\end{minipage}
\begin{minipage}[b]{0.245\linewidth}
\centering
\includegraphics[width=1.0\linewidth]{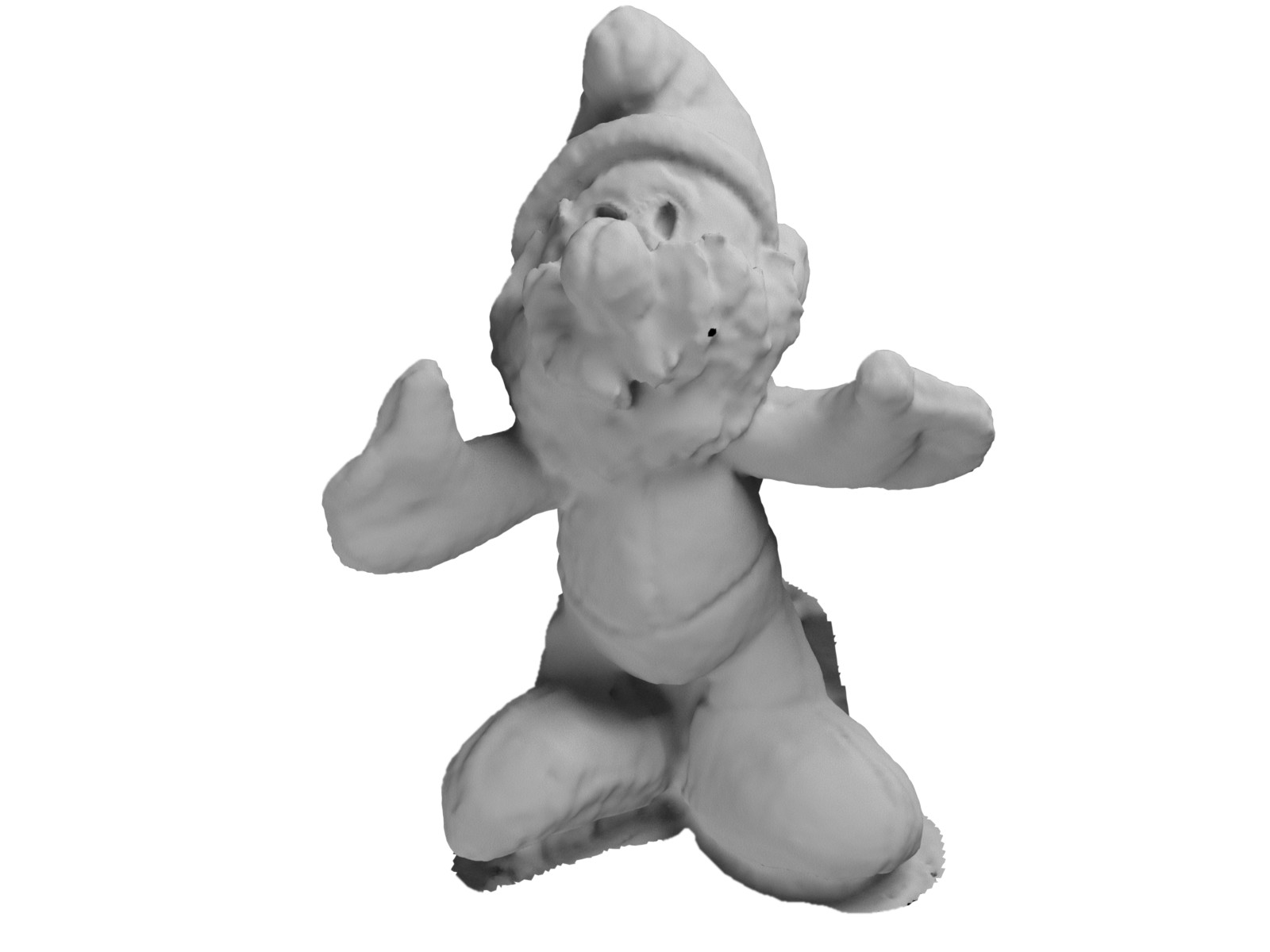}
\end{minipage}
\begin{minipage}[b]{0.245\linewidth}
\centering
\includegraphics[width=1.0\linewidth]{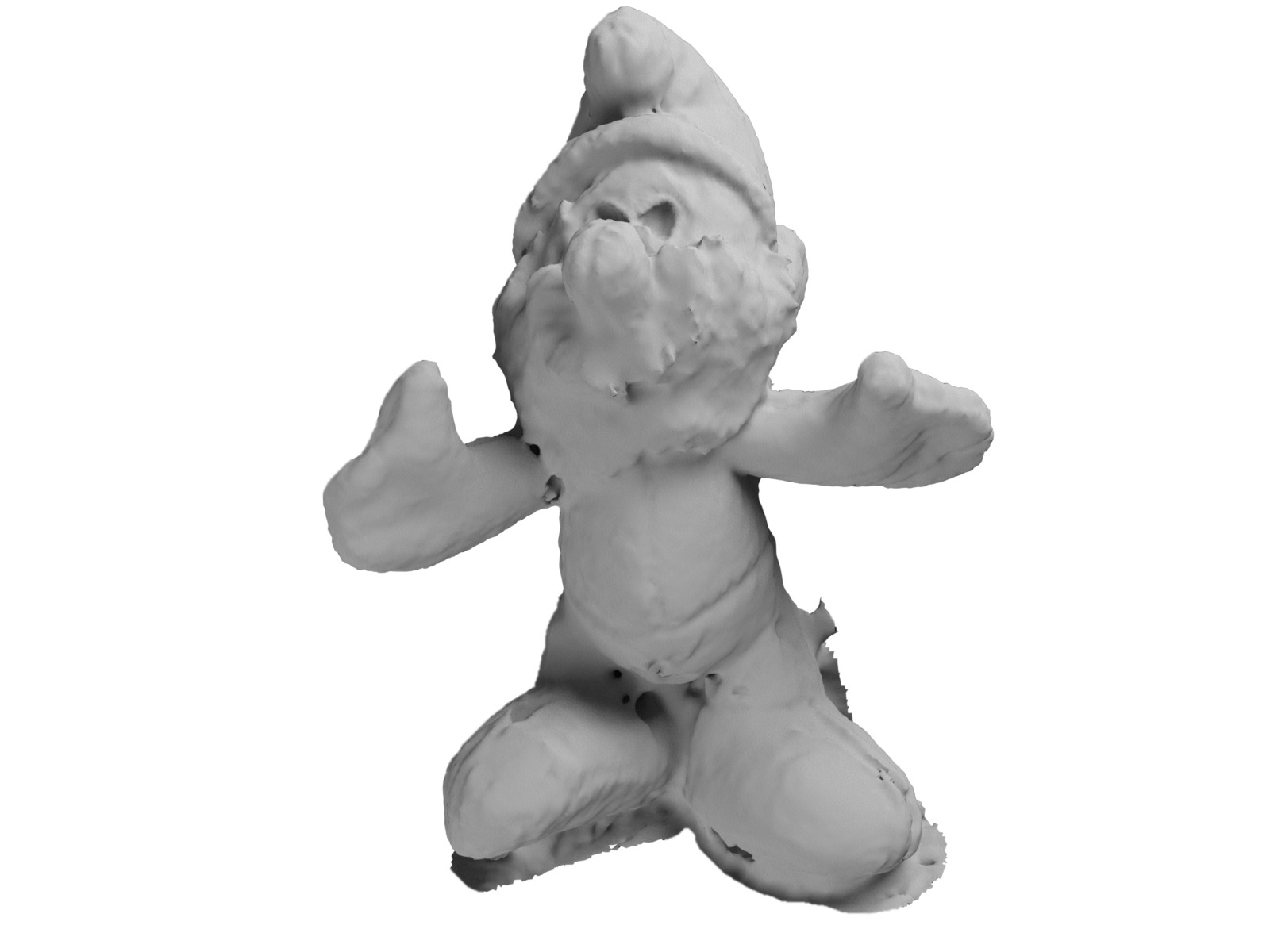}
\end{minipage}
\begin{minipage}[b]{0.245\linewidth}
\centering
\includegraphics[width=1.0\linewidth]{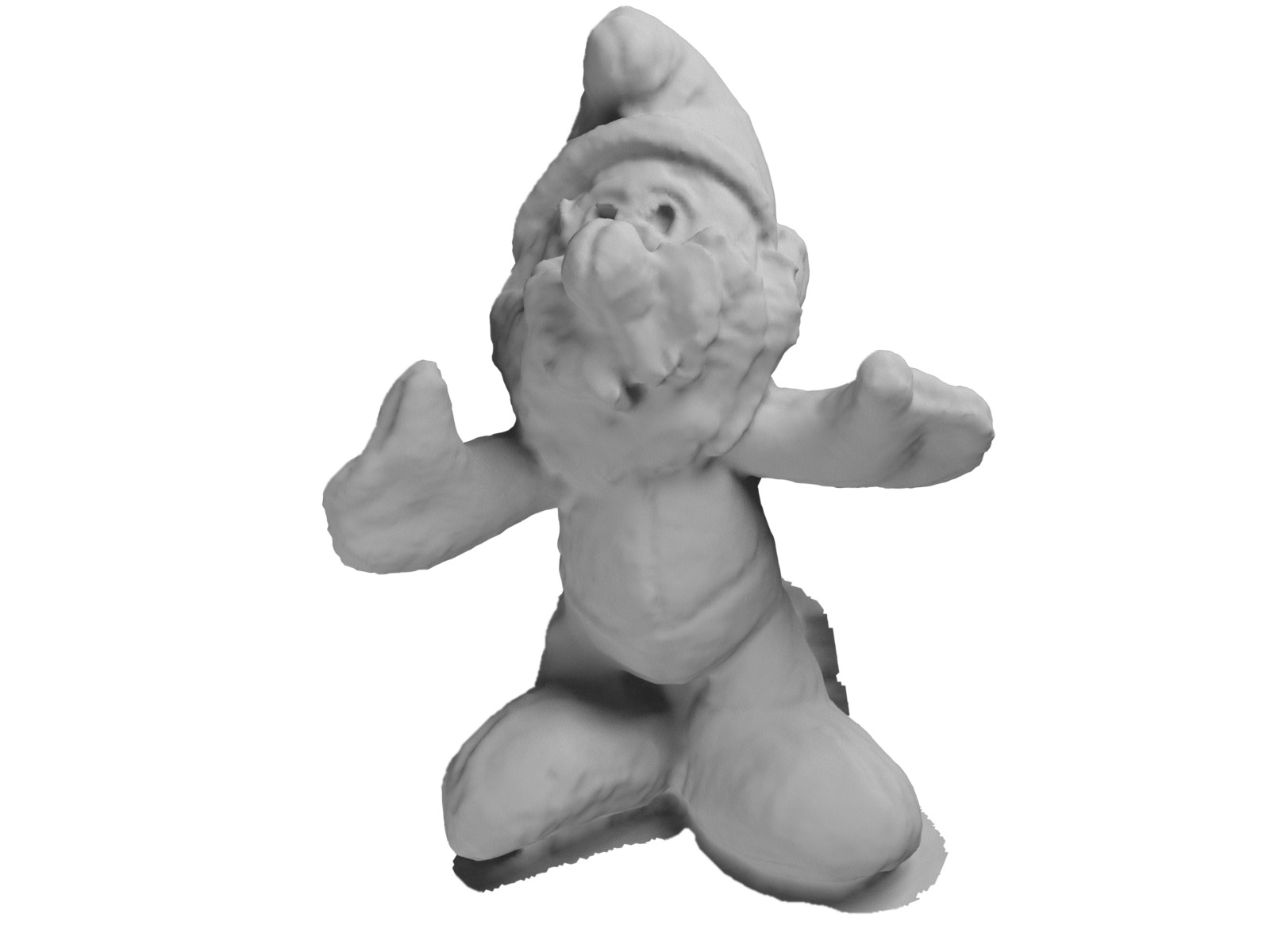}
\end{minipage}

\begin{minipage}[b]{0.245\linewidth}
\centering
\includegraphics[width=1.0\linewidth]{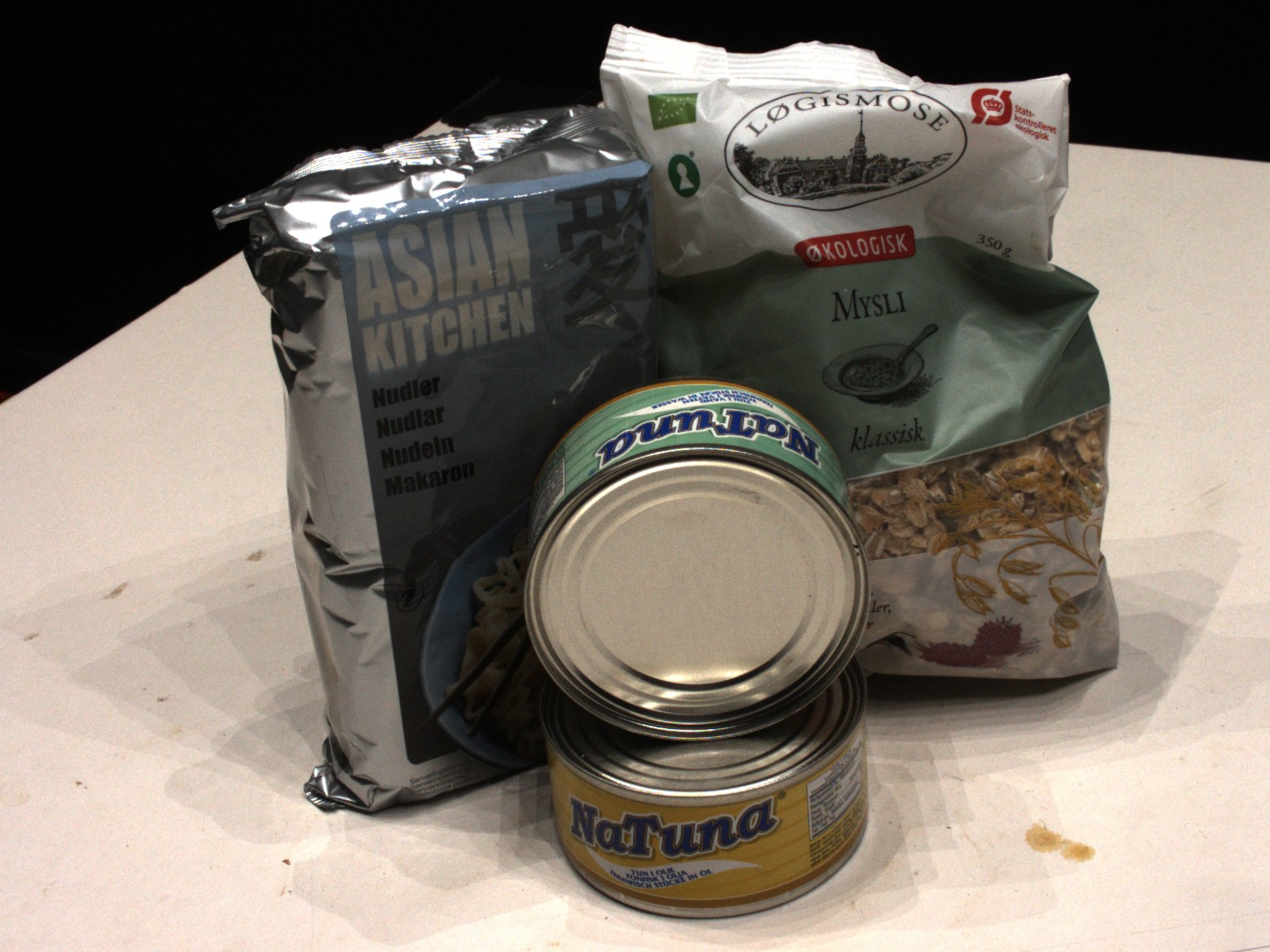}
\end{minipage}
\begin{minipage}[b]{0.245\linewidth}
\centering
\includegraphics[width=1.0\linewidth]{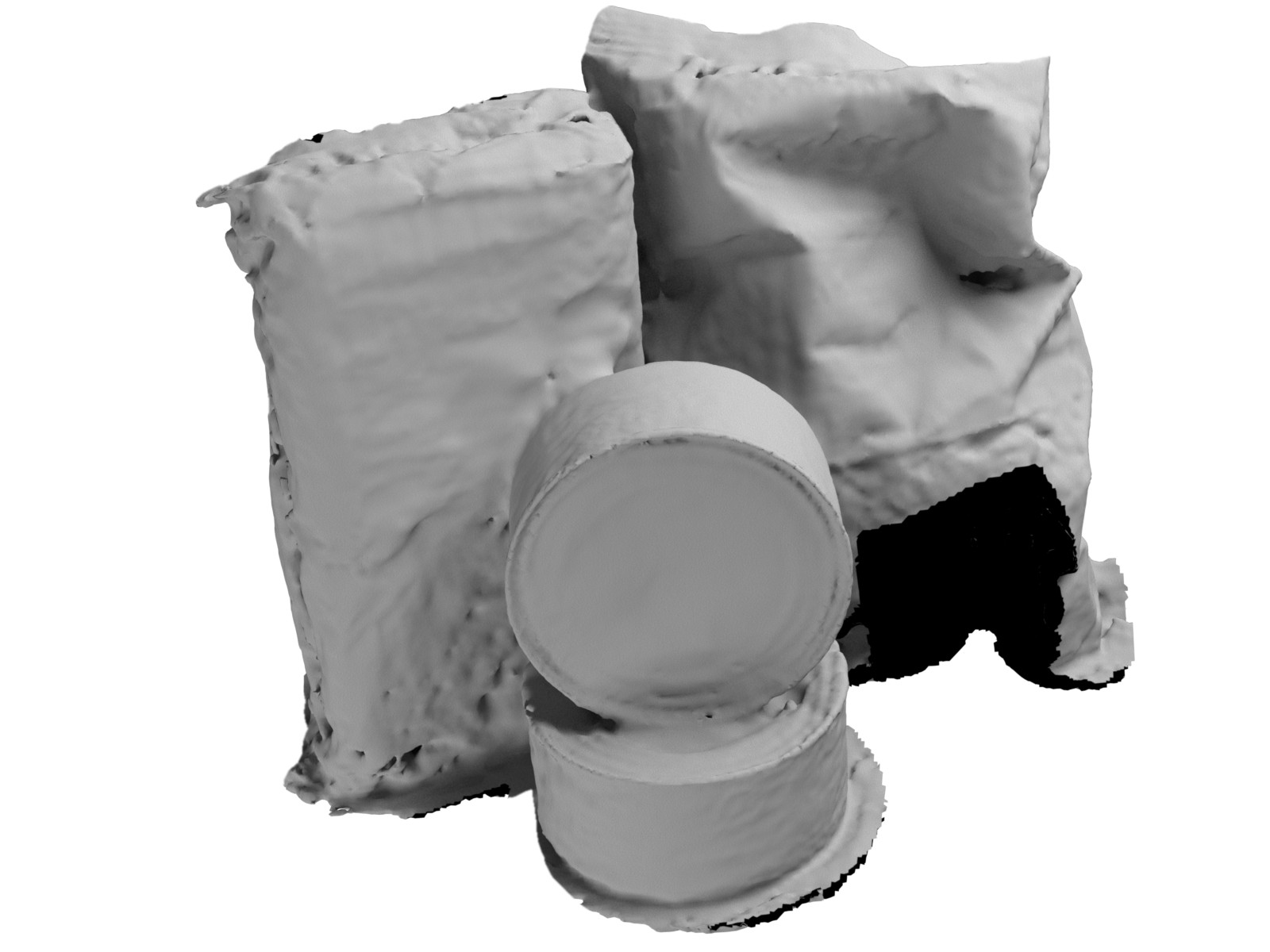}
\end{minipage}
\begin{minipage}[b]{0.245\linewidth}
\centering
\includegraphics[width=1.0\linewidth]{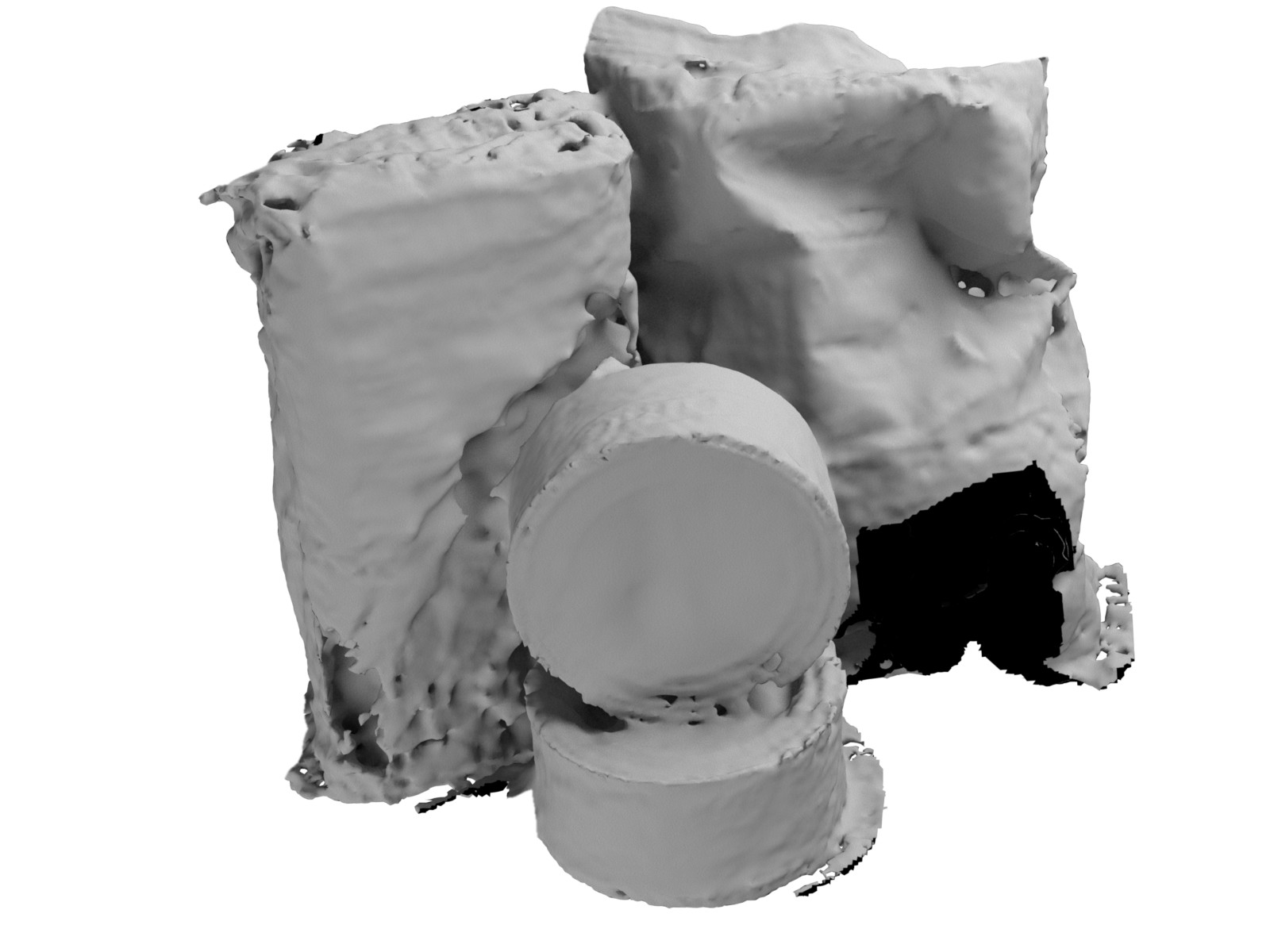}
\end{minipage}
\begin{minipage}[b]{0.245\linewidth}
\centering
\includegraphics[width=1.0\linewidth]{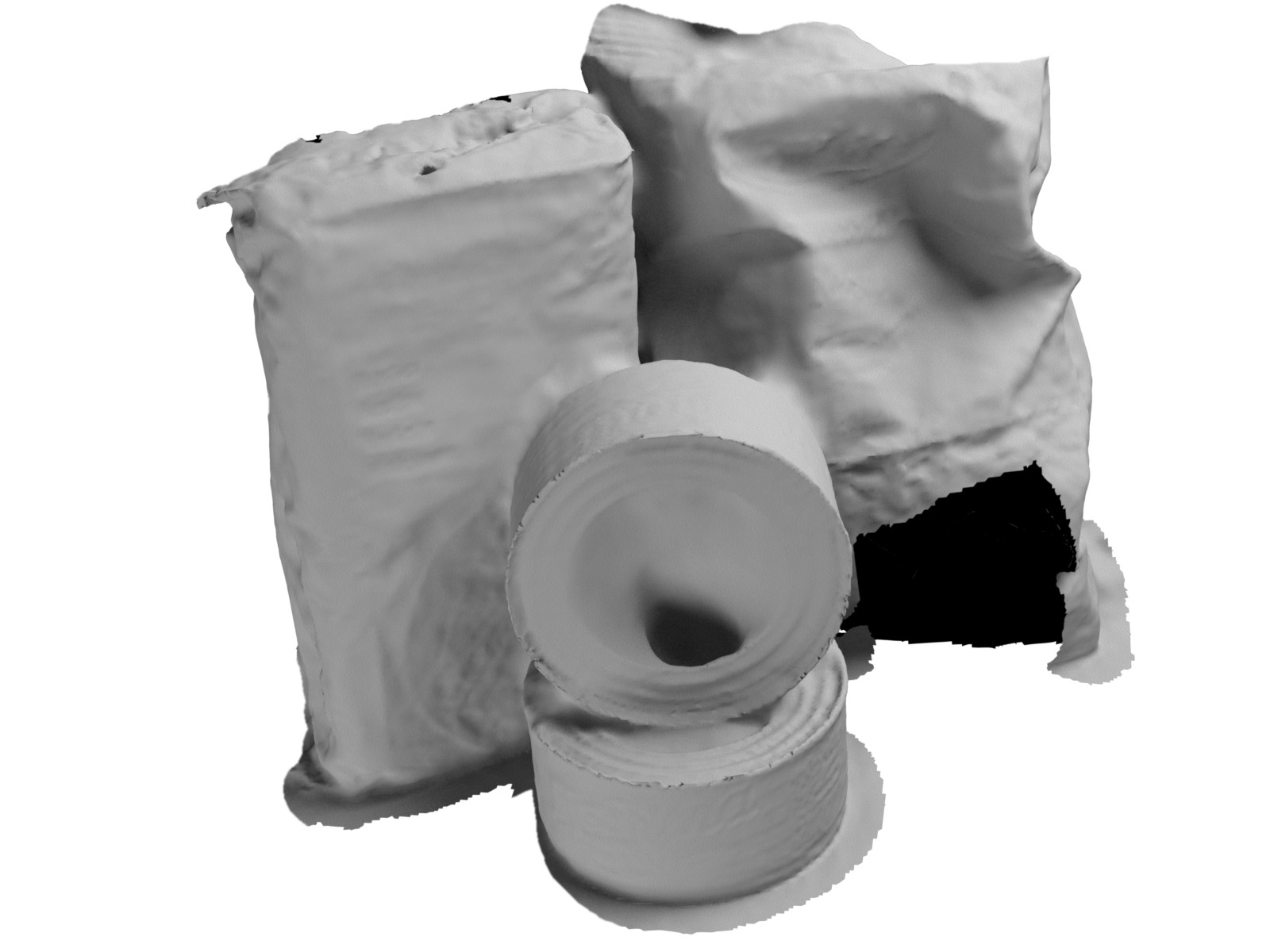}
\end{minipage}

\begin{minipage}[b]{0.245\linewidth}
\centering
\includegraphics[width=1.0\linewidth]{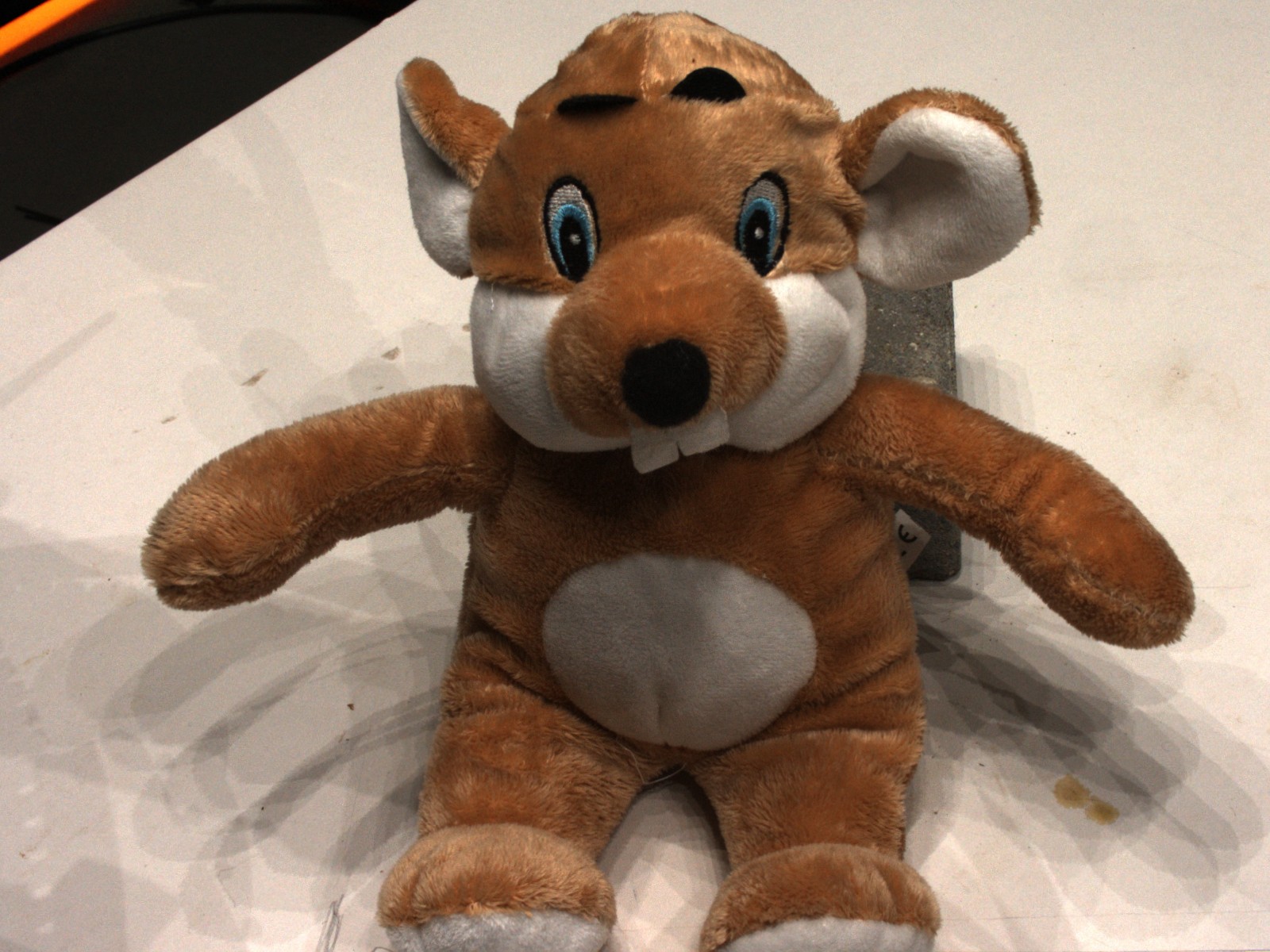}
\subcaption{Ground Truth}
\end{minipage}
\begin{minipage}[b]{0.245\linewidth}
\centering
\includegraphics[width=1.0\linewidth]{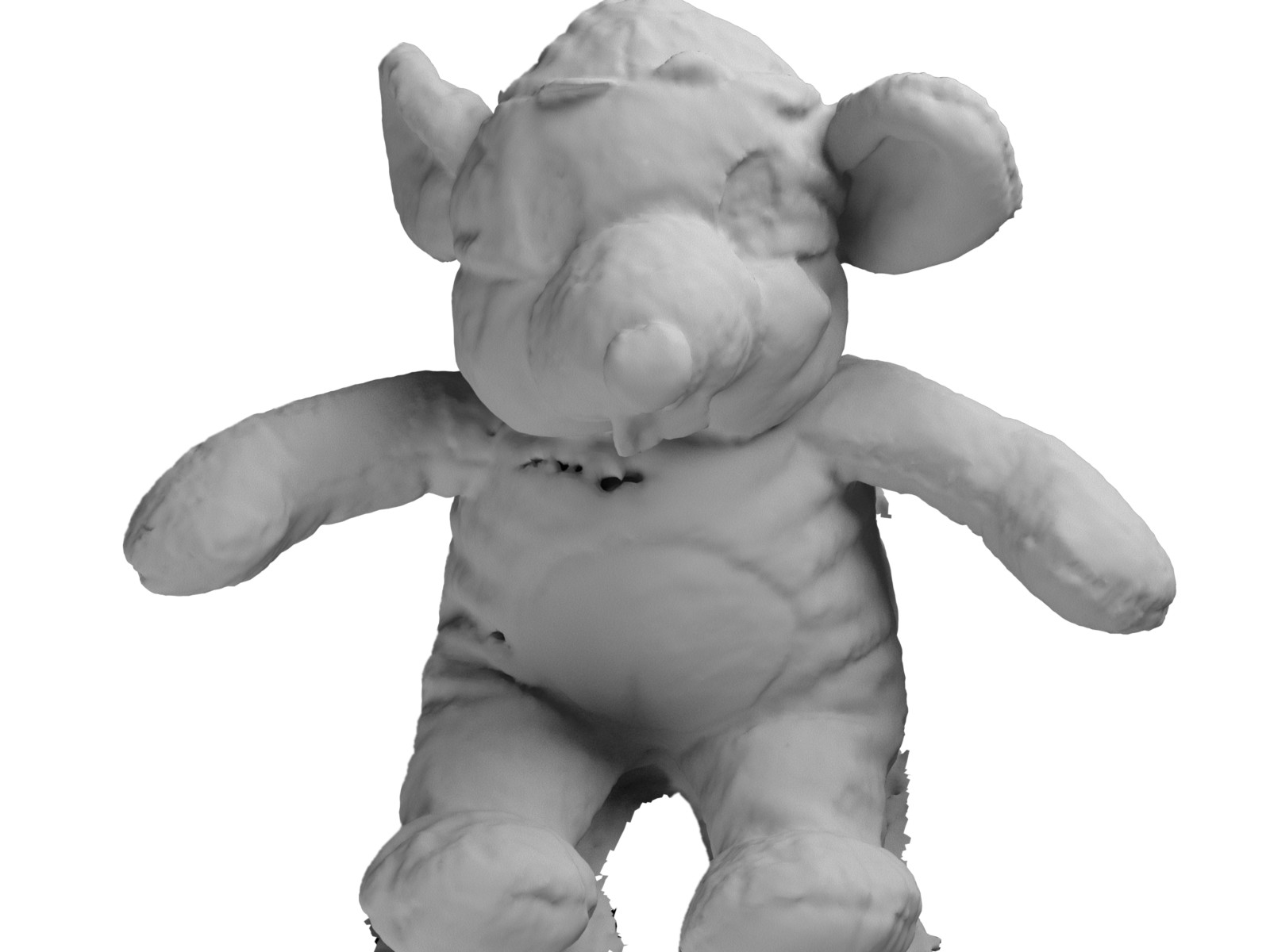}
\subcaption{NeuS-Facto}
\end{minipage}
\begin{minipage}[b]{0.245\linewidth}
\centering
\includegraphics[width=1.0\linewidth]{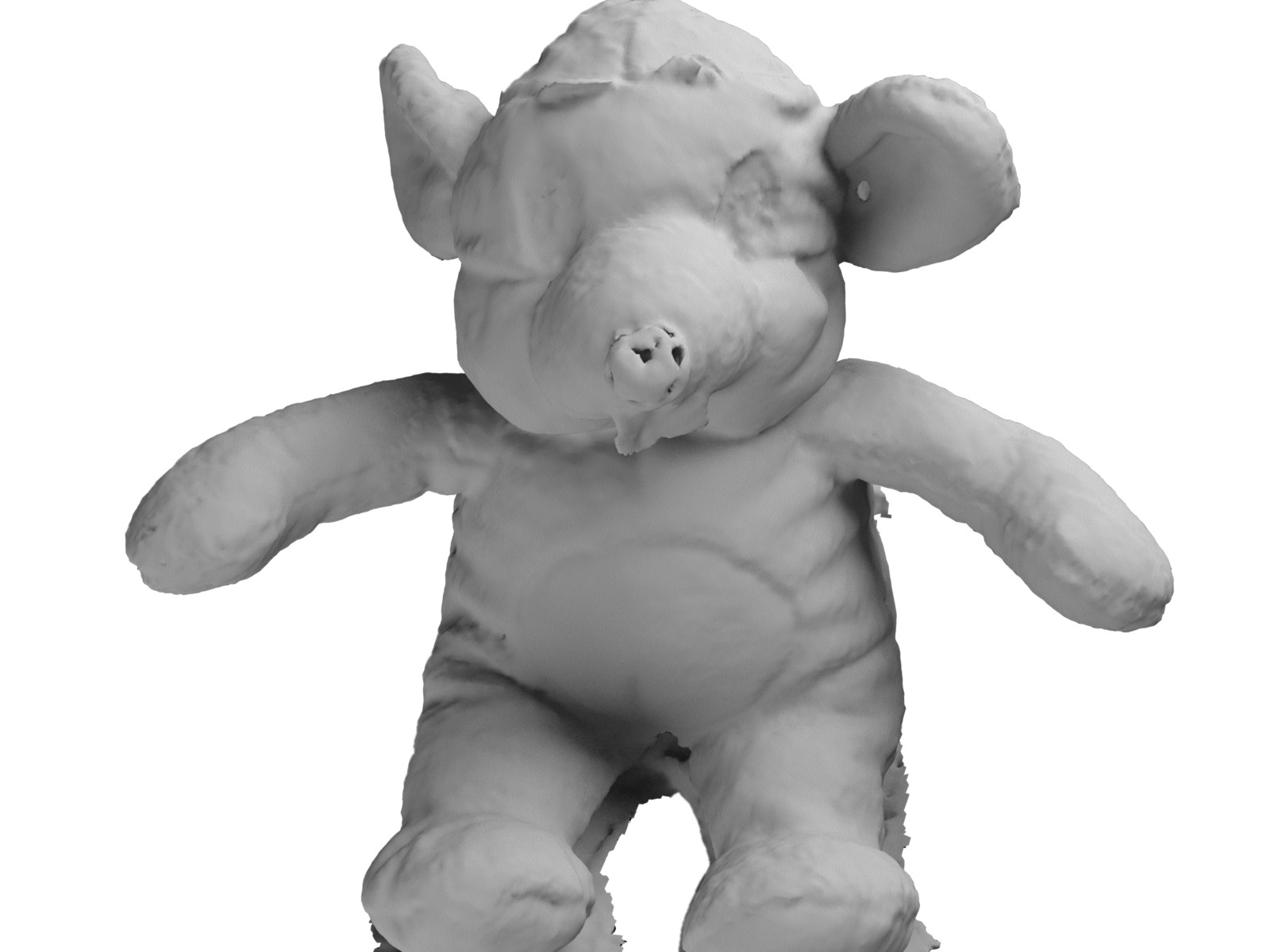}
\subcaption{OaV-Facto}
\end{minipage}
\begin{minipage}[b]{0.245\linewidth}
\centering
\includegraphics[width=1.0\linewidth]{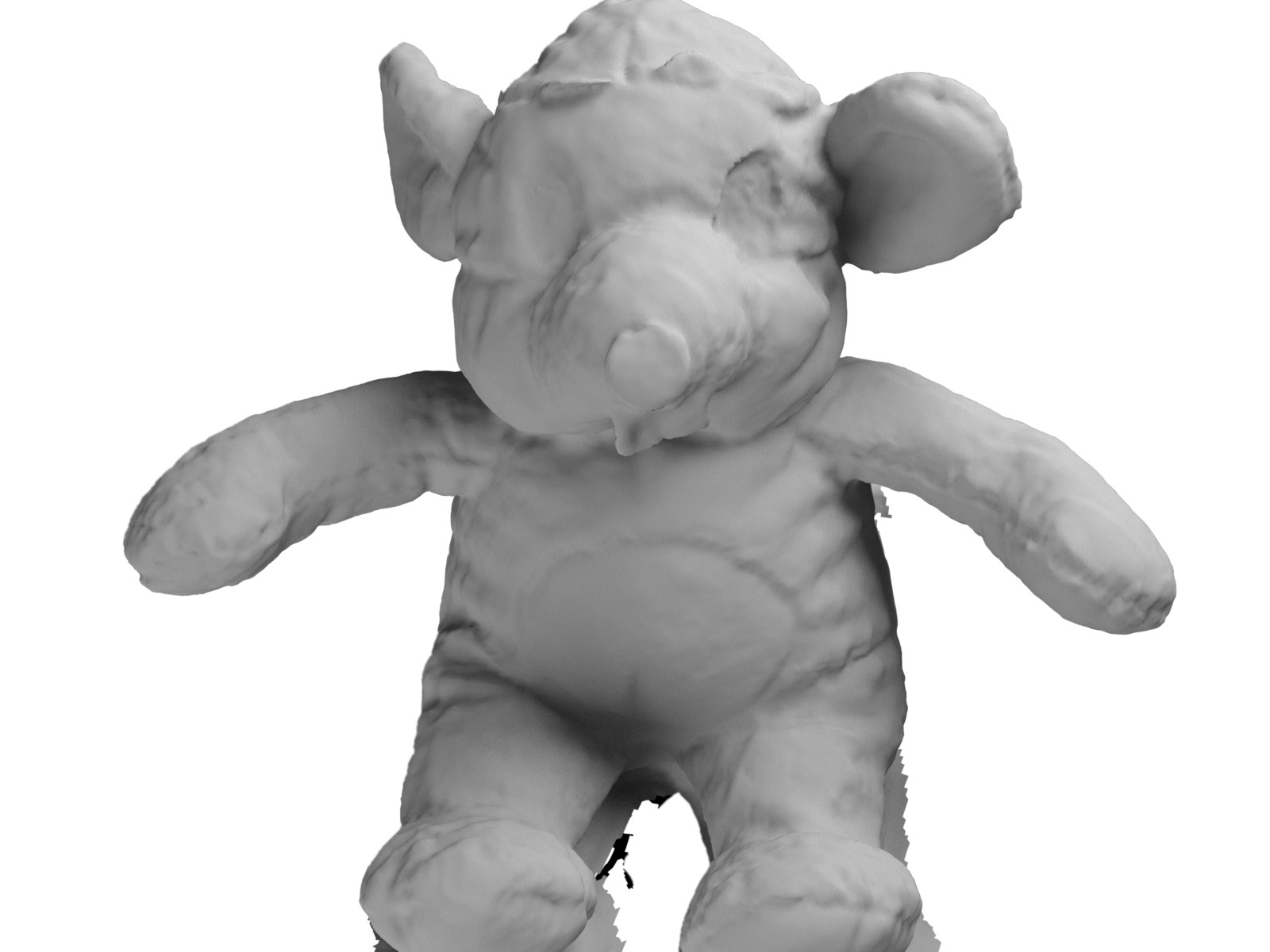}
\subcaption{SSDP-Facto}
\end{minipage}

\caption{Visualization examples on the DTU dataset.}
\end{figure}

\begin{figure}

\ContinuedFloat
\setcounter{subfigure}{0}

\centering

\begin{minipage}[b]{0.245\linewidth}
\centering
\includegraphics[width=1.0\linewidth]{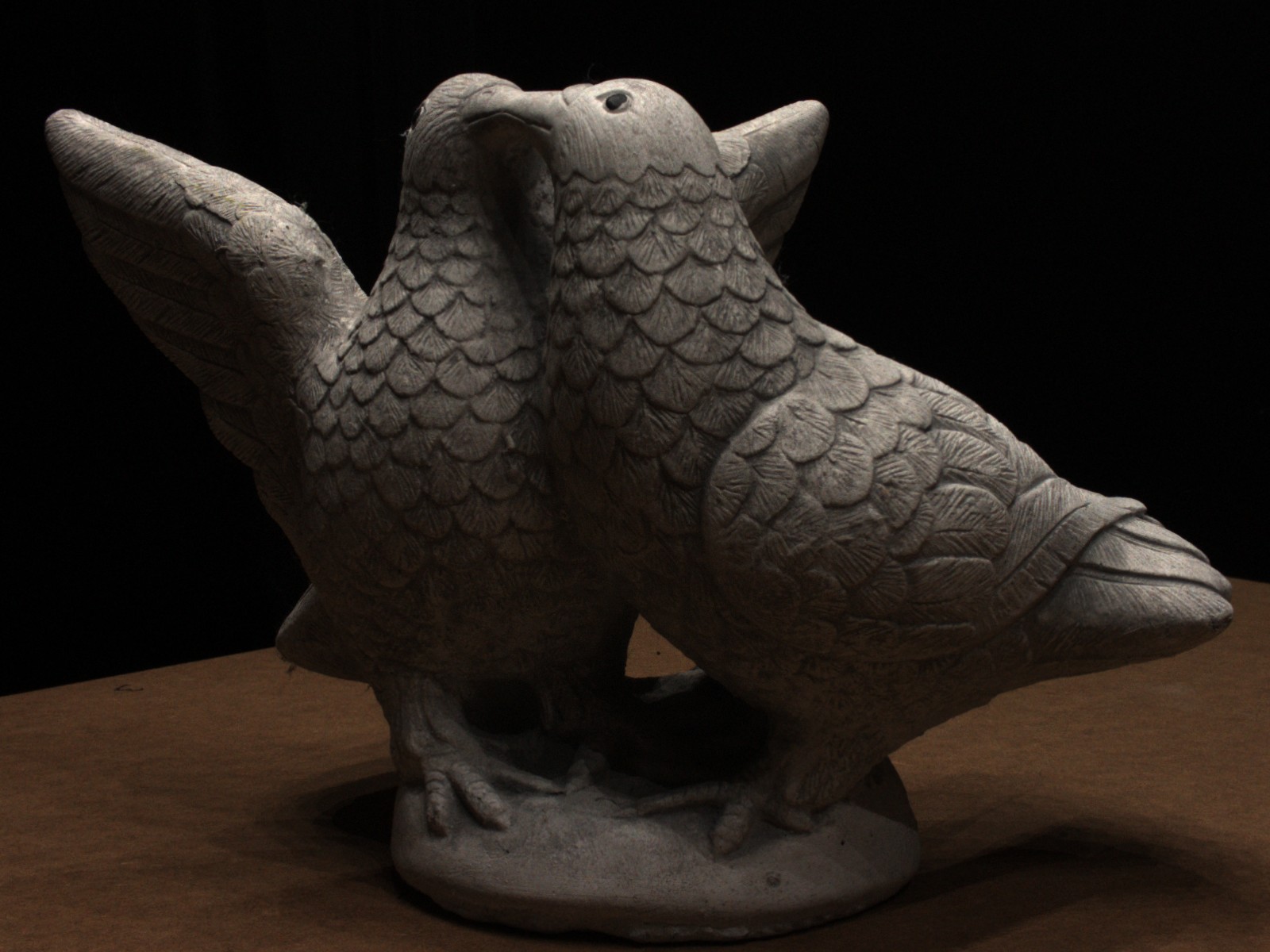}
\end{minipage}
\begin{minipage}[b]{0.245\linewidth}
\centering
\includegraphics[width=1.0\linewidth]{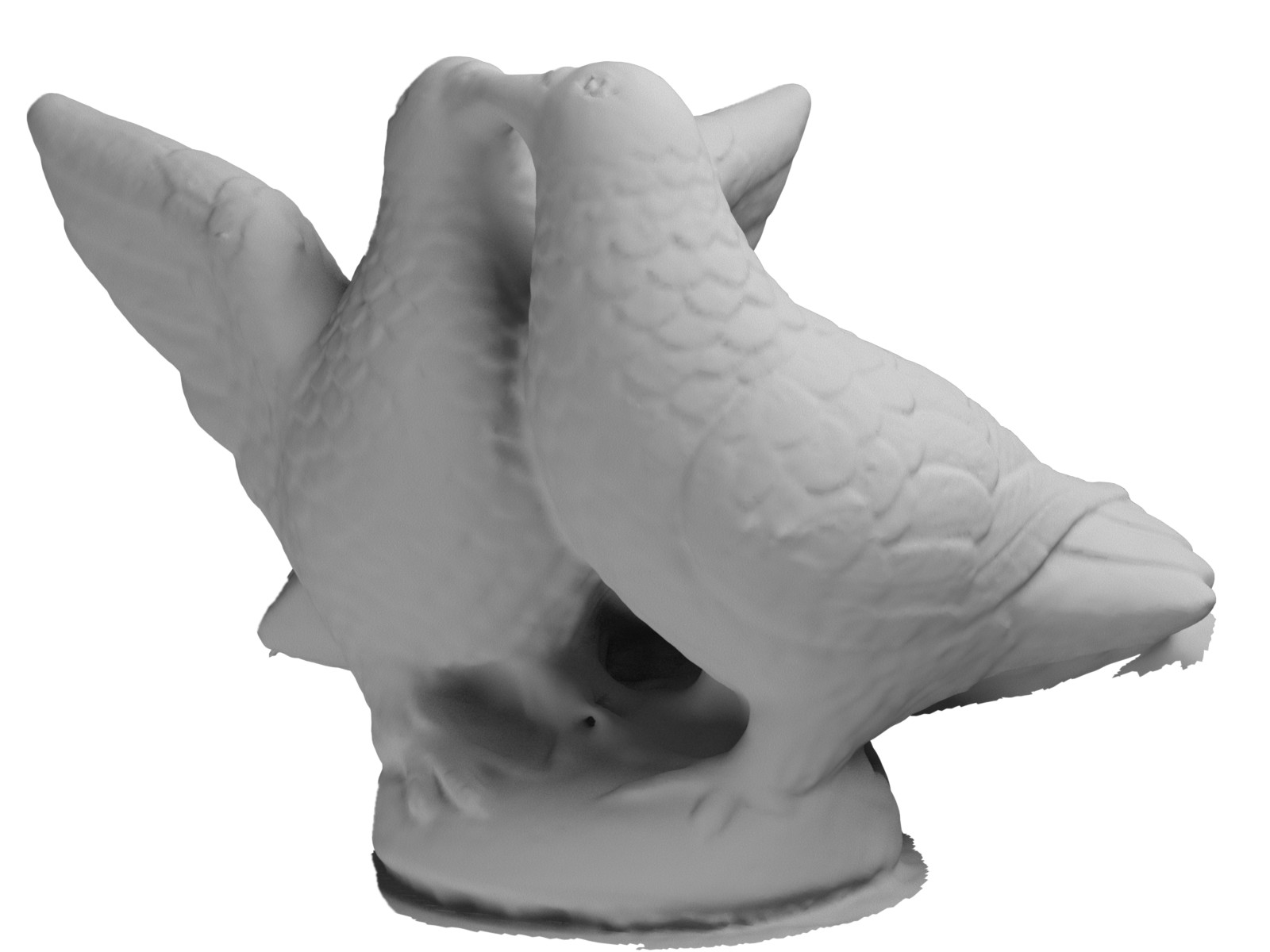}
\end{minipage}
\begin{minipage}[b]{0.245\linewidth}
\centering
\includegraphics[width=1.0\linewidth]{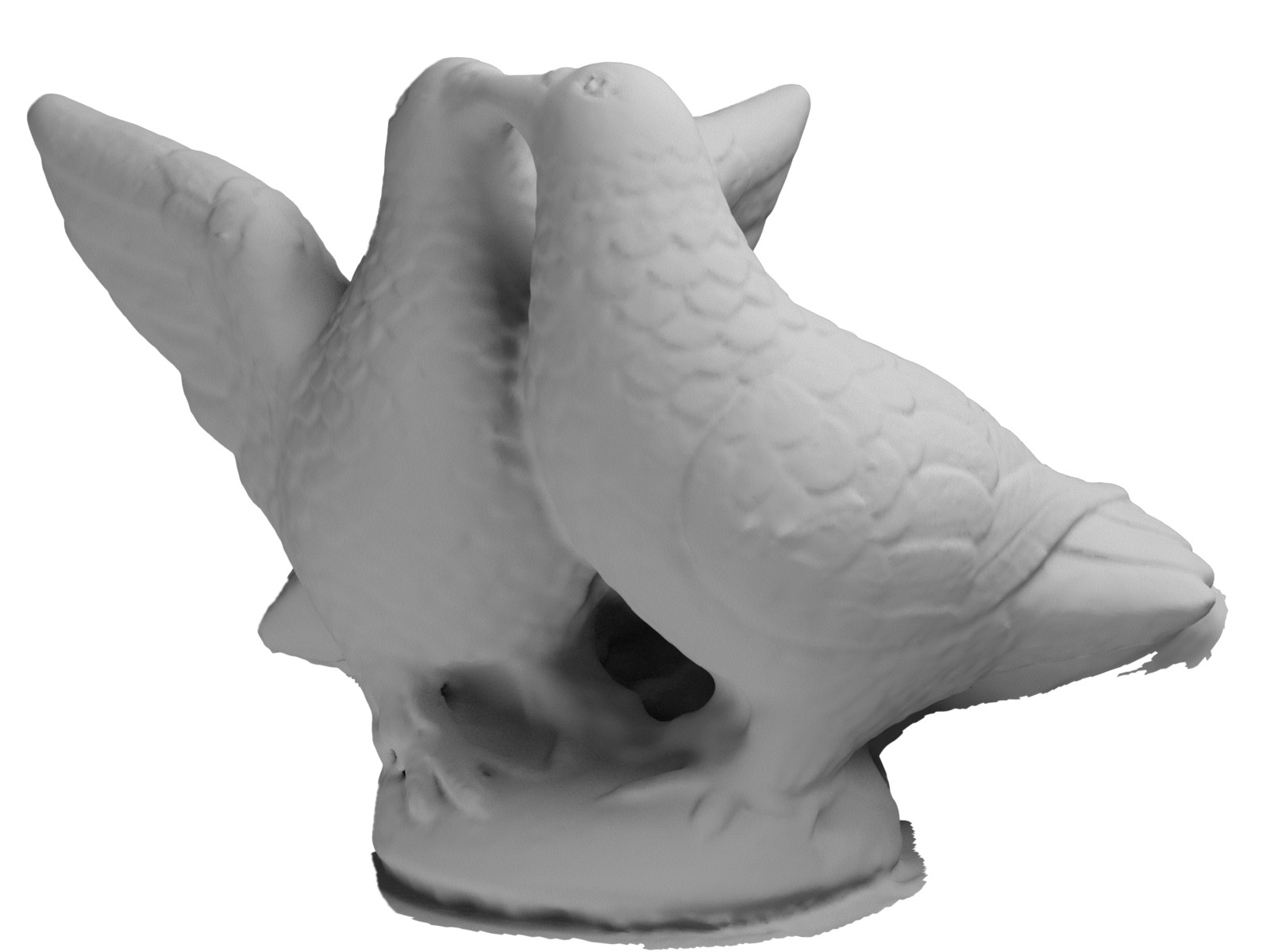}
\end{minipage}
\begin{minipage}[b]{0.245\linewidth}
\centering
\includegraphics[width=1.0\linewidth]{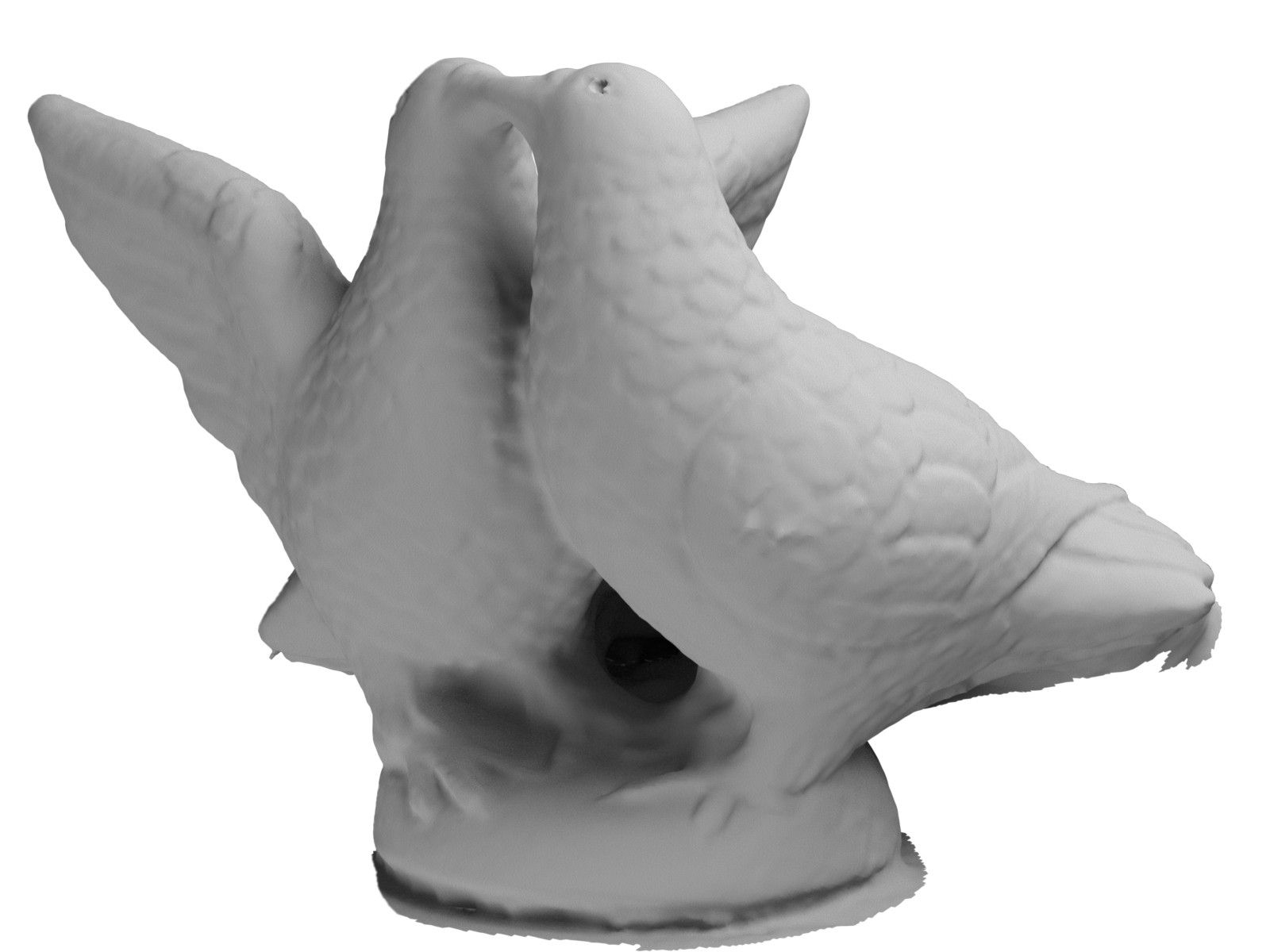}
\end{minipage}

\begin{minipage}[b]{0.245\linewidth}
\centering
\includegraphics[width=1.0\linewidth]{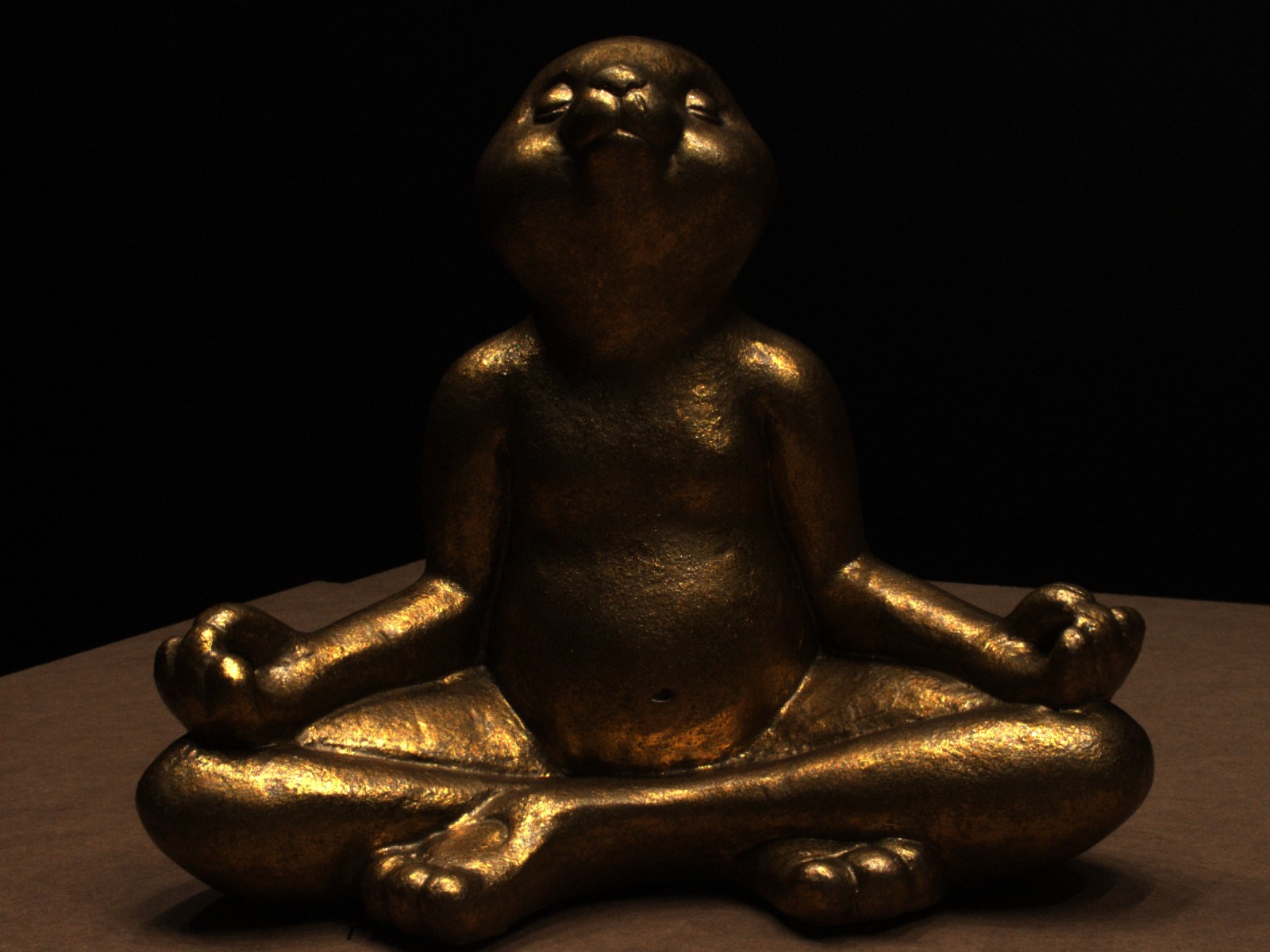}
\end{minipage}
\begin{minipage}[b]{0.245\linewidth}
\centering
\includegraphics[width=1.0\linewidth]{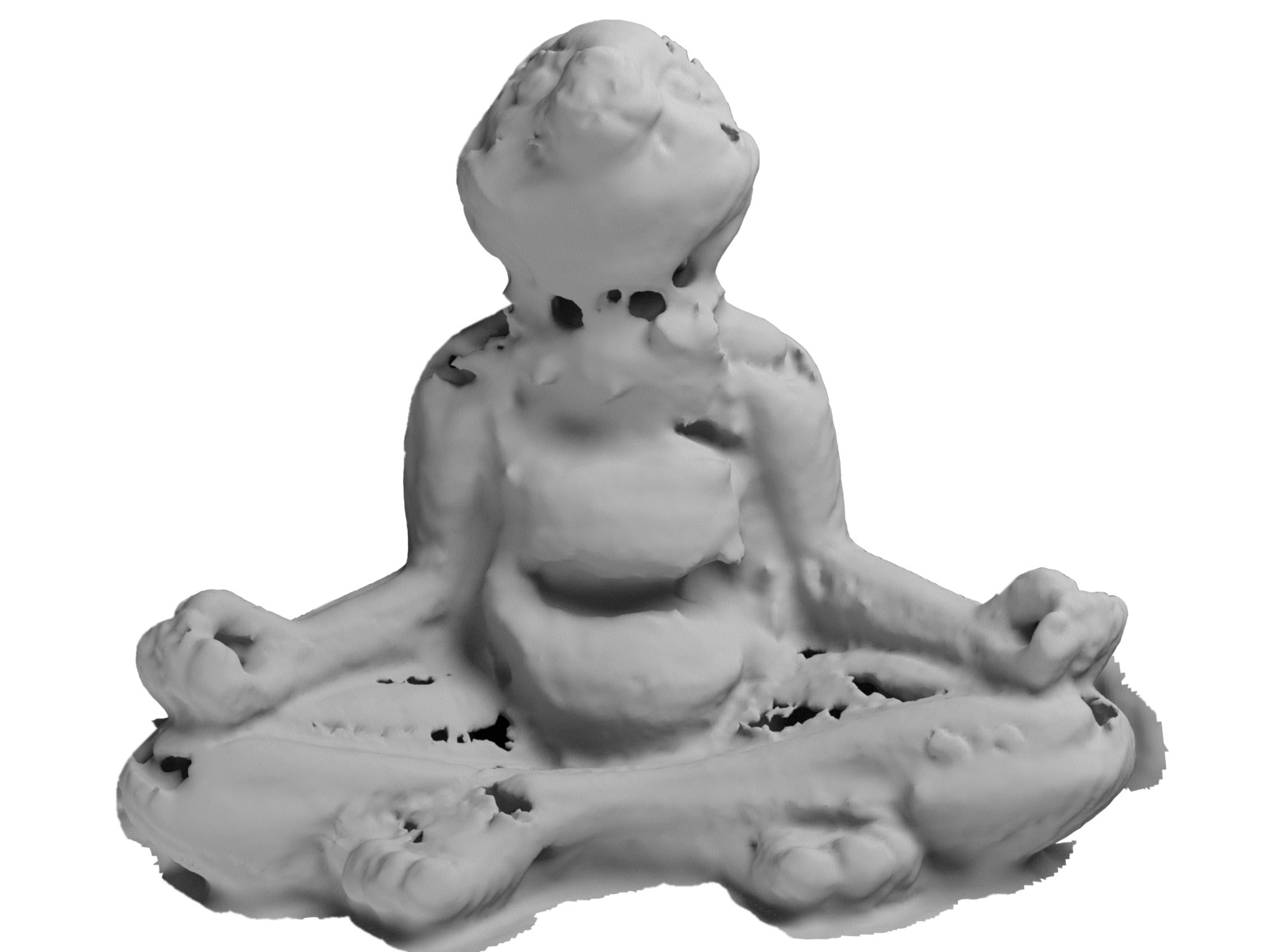}
\end{minipage}
\begin{minipage}[b]{0.245\linewidth}
\centering
\includegraphics[width=1.0\linewidth]{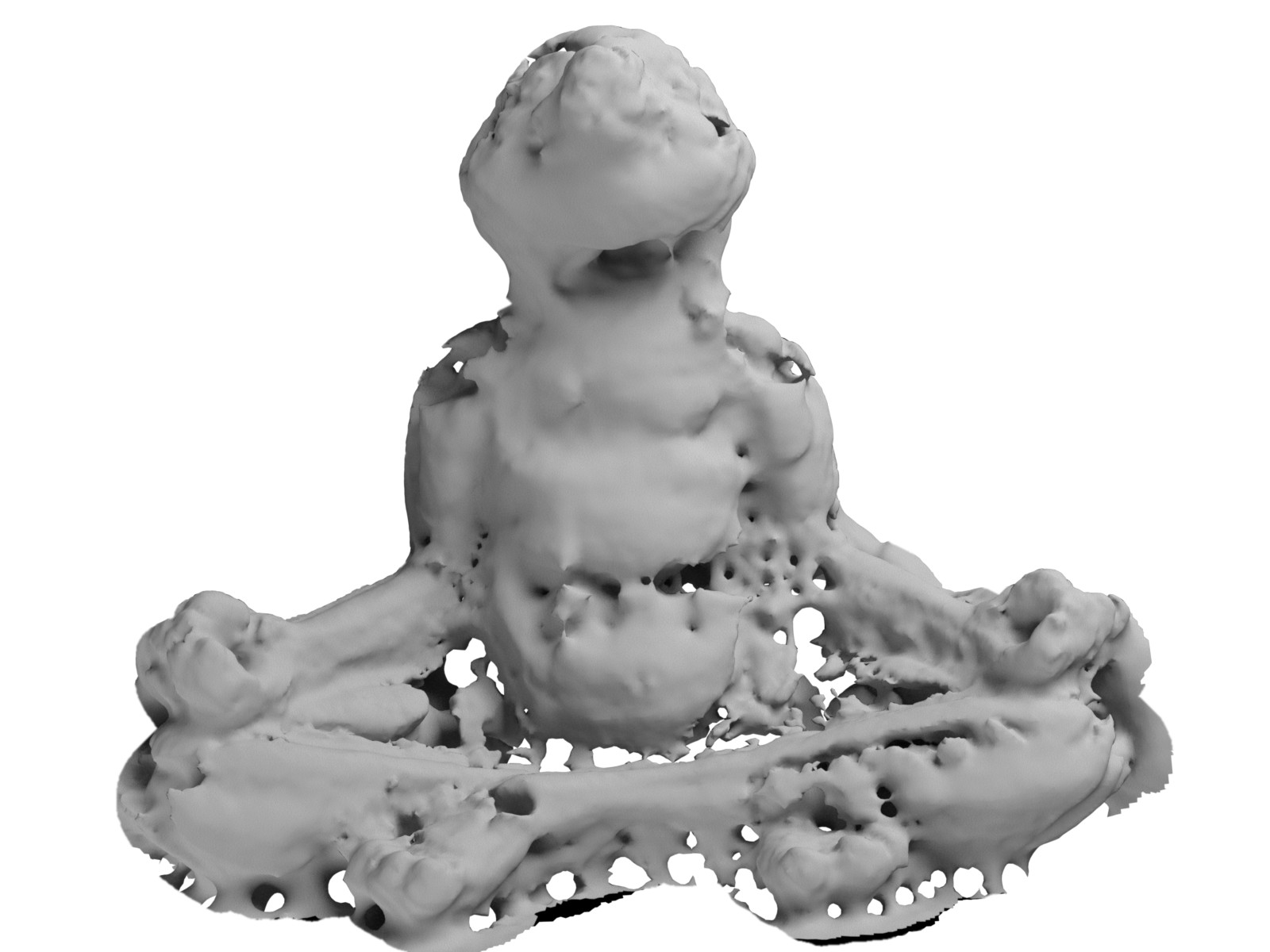}
\end{minipage}
\begin{minipage}[b]{0.245\linewidth}
\centering
\includegraphics[width=1.0\linewidth]{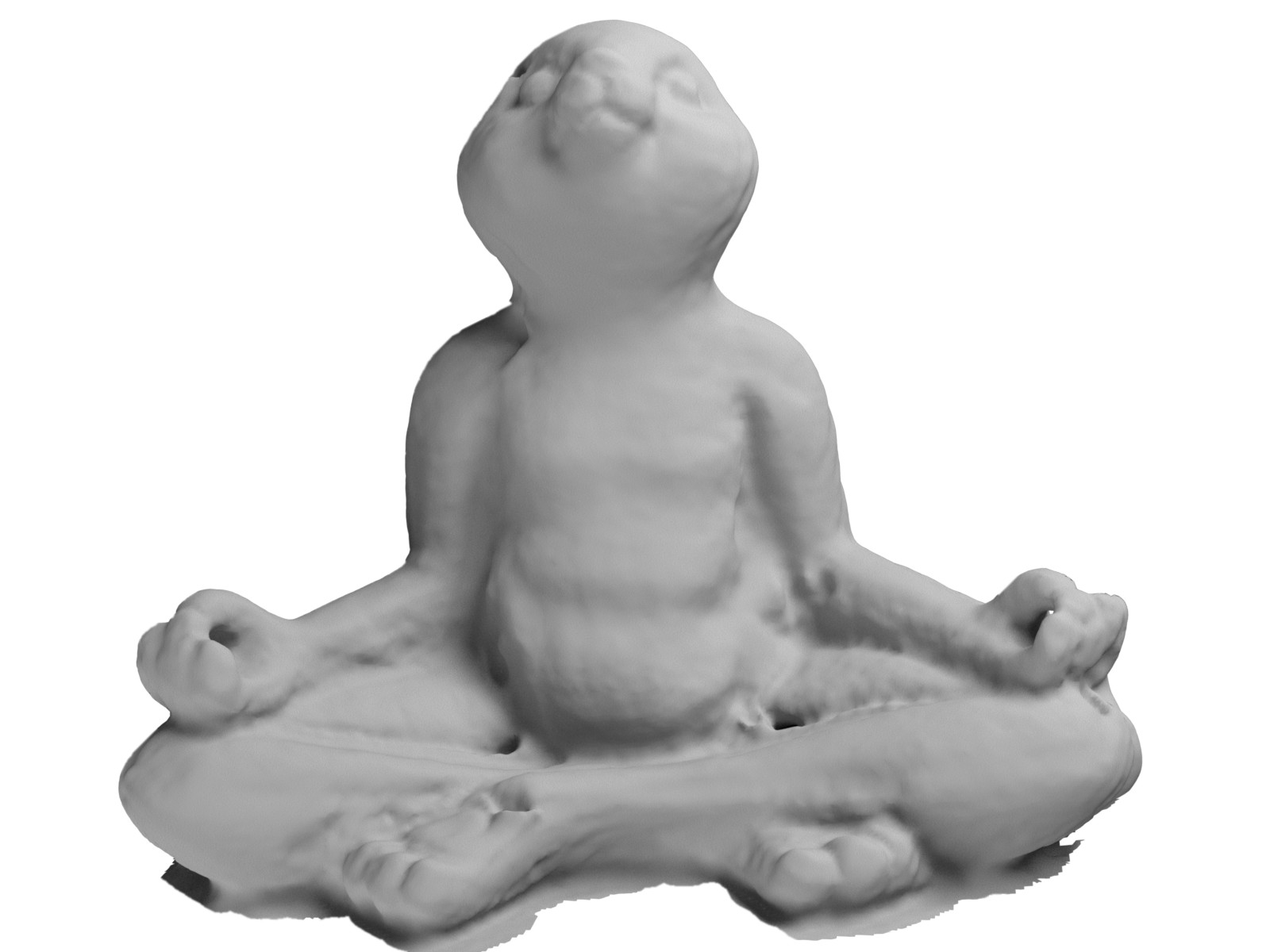}
\end{minipage}

\begin{minipage}[b]{0.245\linewidth}
\centering
\includegraphics[width=1.0\linewidth]{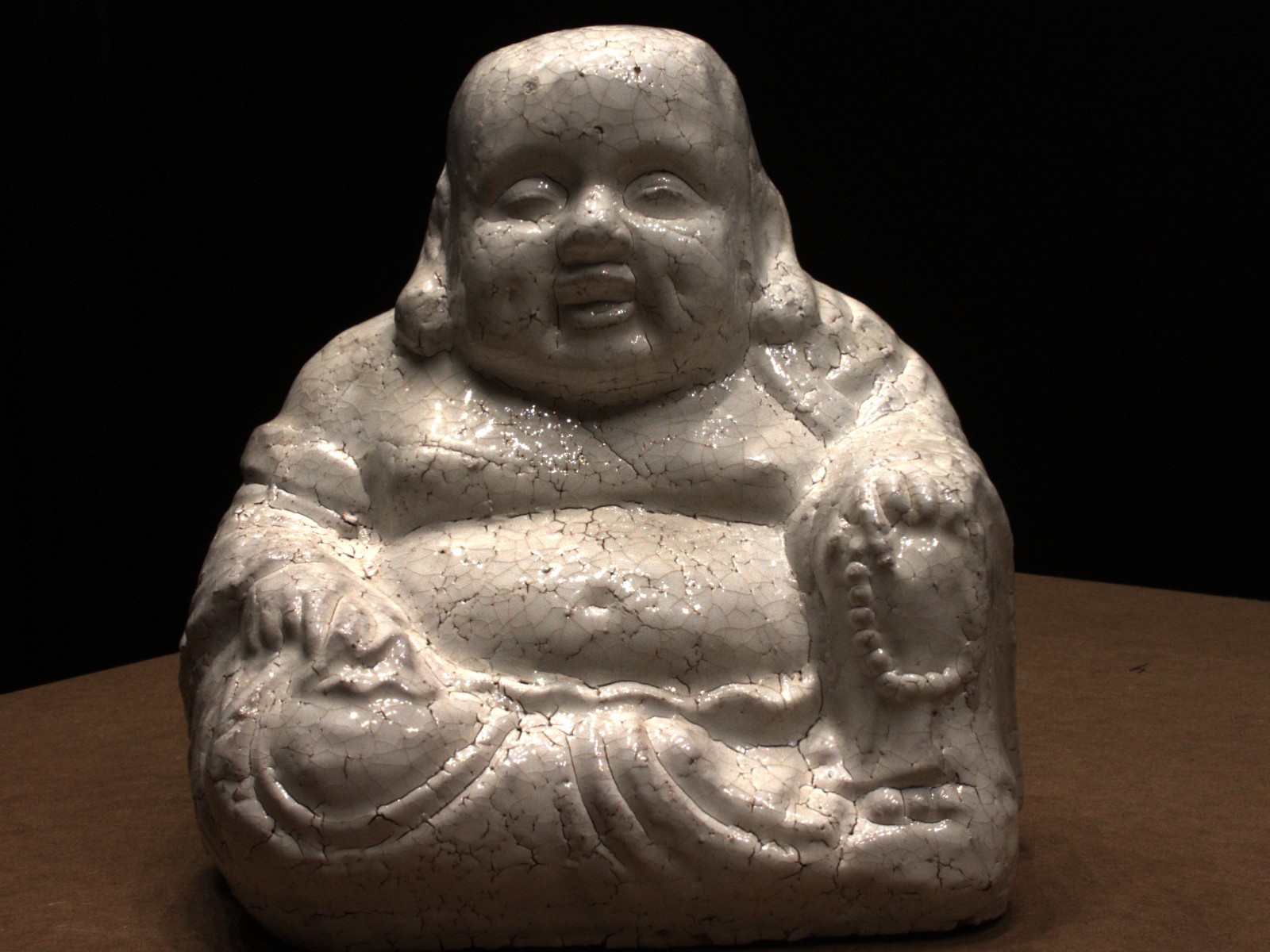}
\end{minipage}
\begin{minipage}[b]{0.245\linewidth}
\centering
\includegraphics[width=1.0\linewidth]{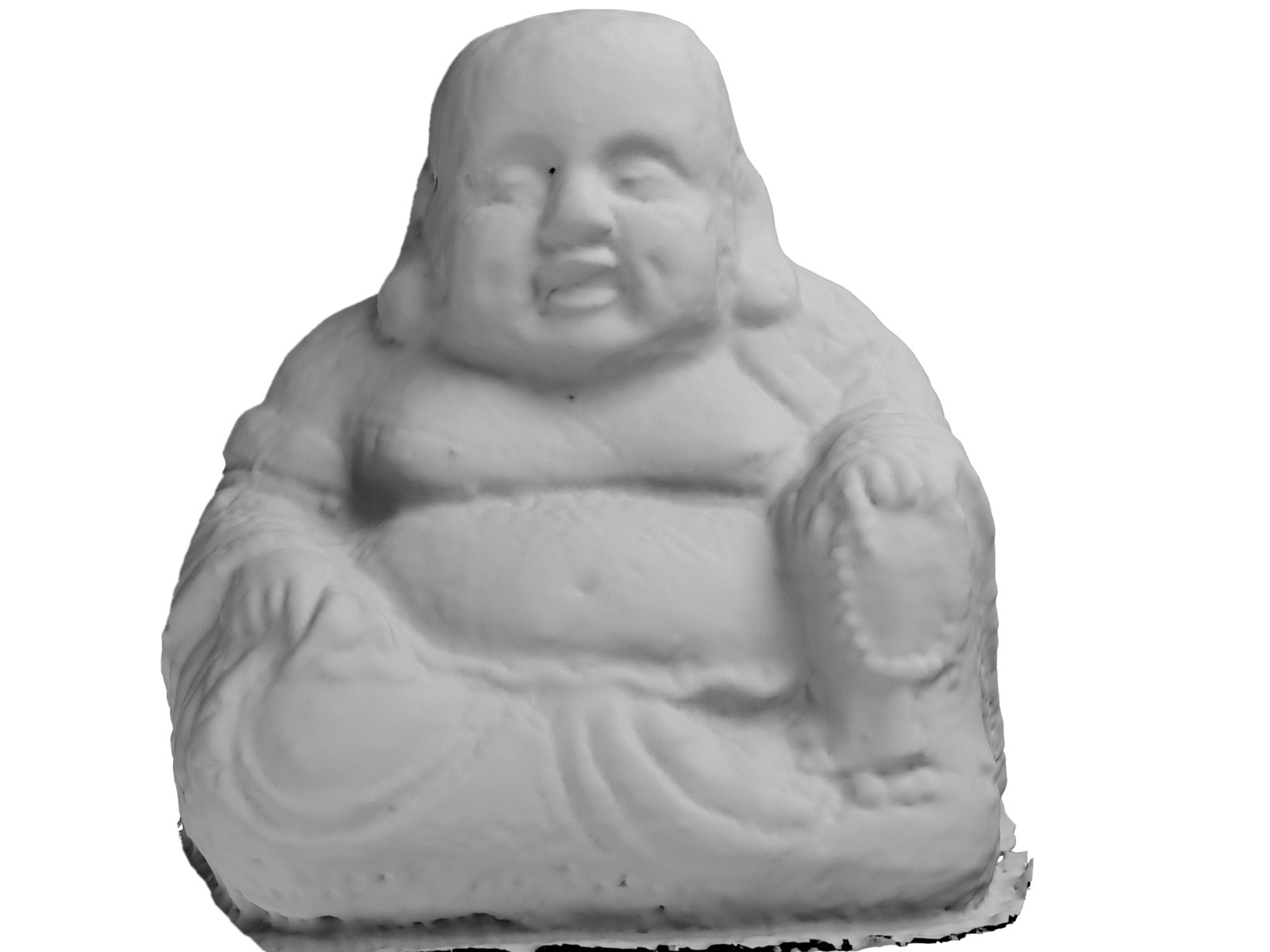}
\end{minipage}
\begin{minipage}[b]{0.245\linewidth}
\centering
\includegraphics[width=1.0\linewidth]{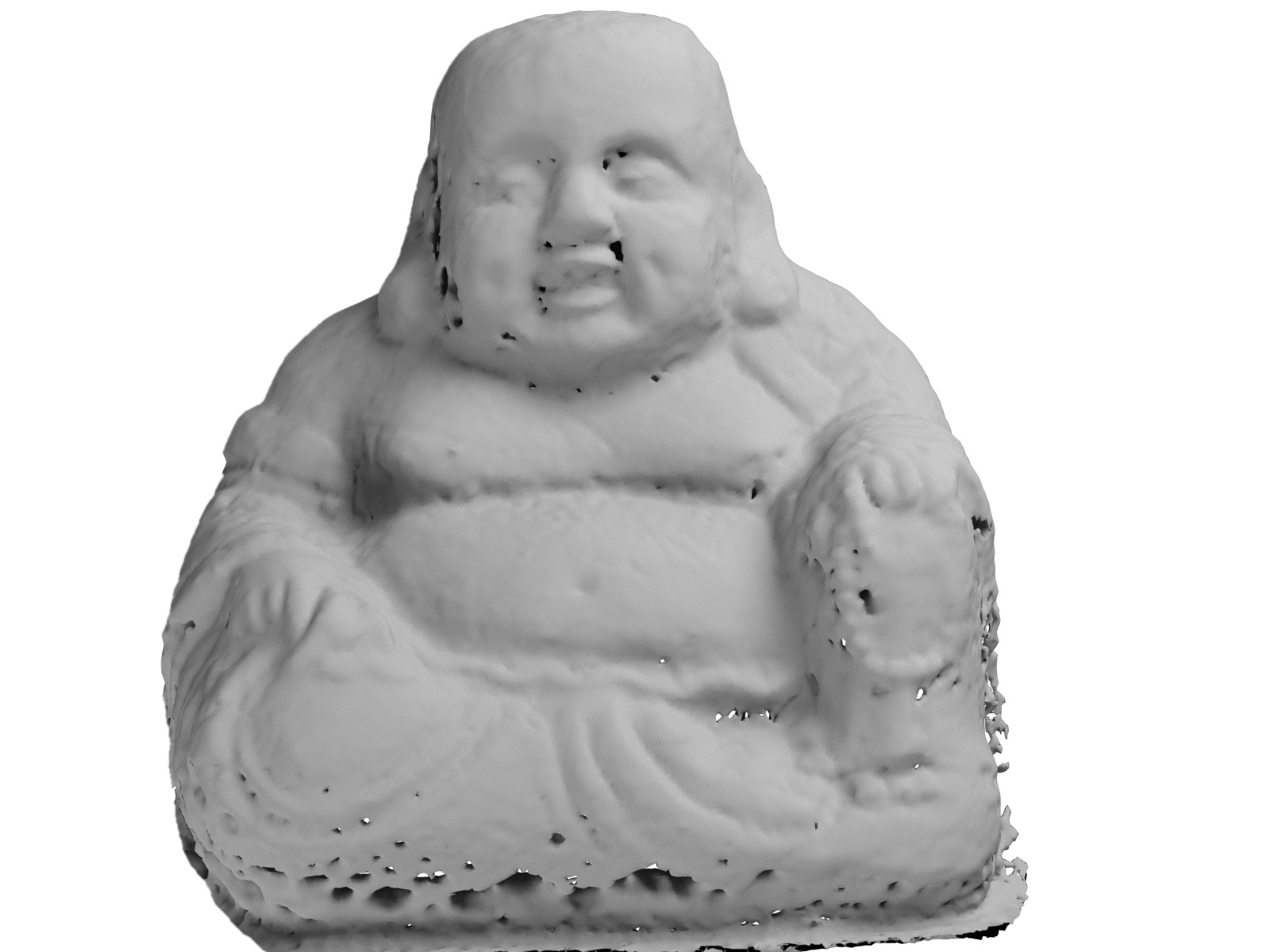}
\end{minipage}
\begin{minipage}[b]{0.245\linewidth}
\centering
\includegraphics[width=1.0\linewidth]{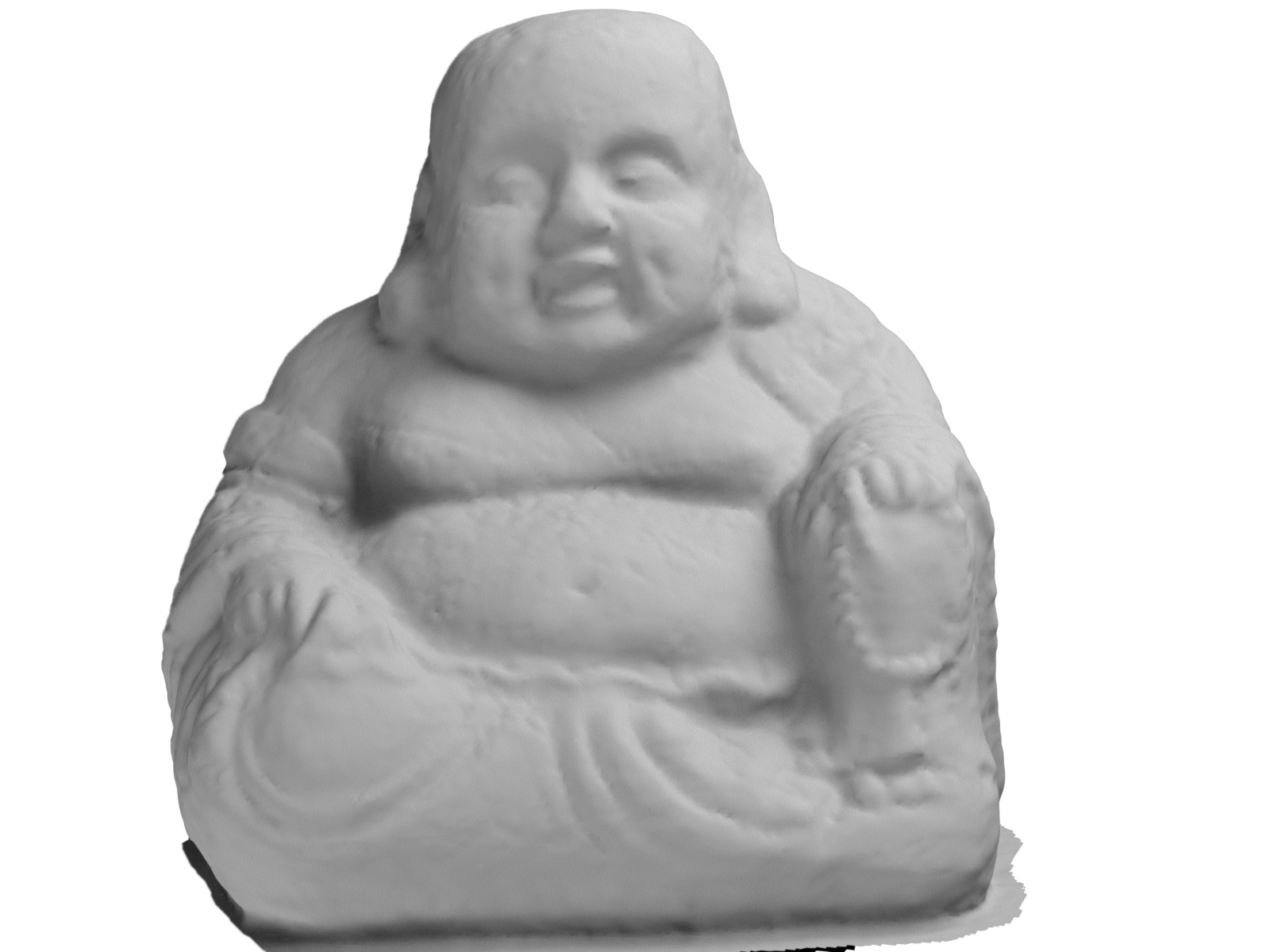}
\end{minipage}

\begin{minipage}[b]{0.245\linewidth}
\centering
\includegraphics[width=1.0\linewidth]{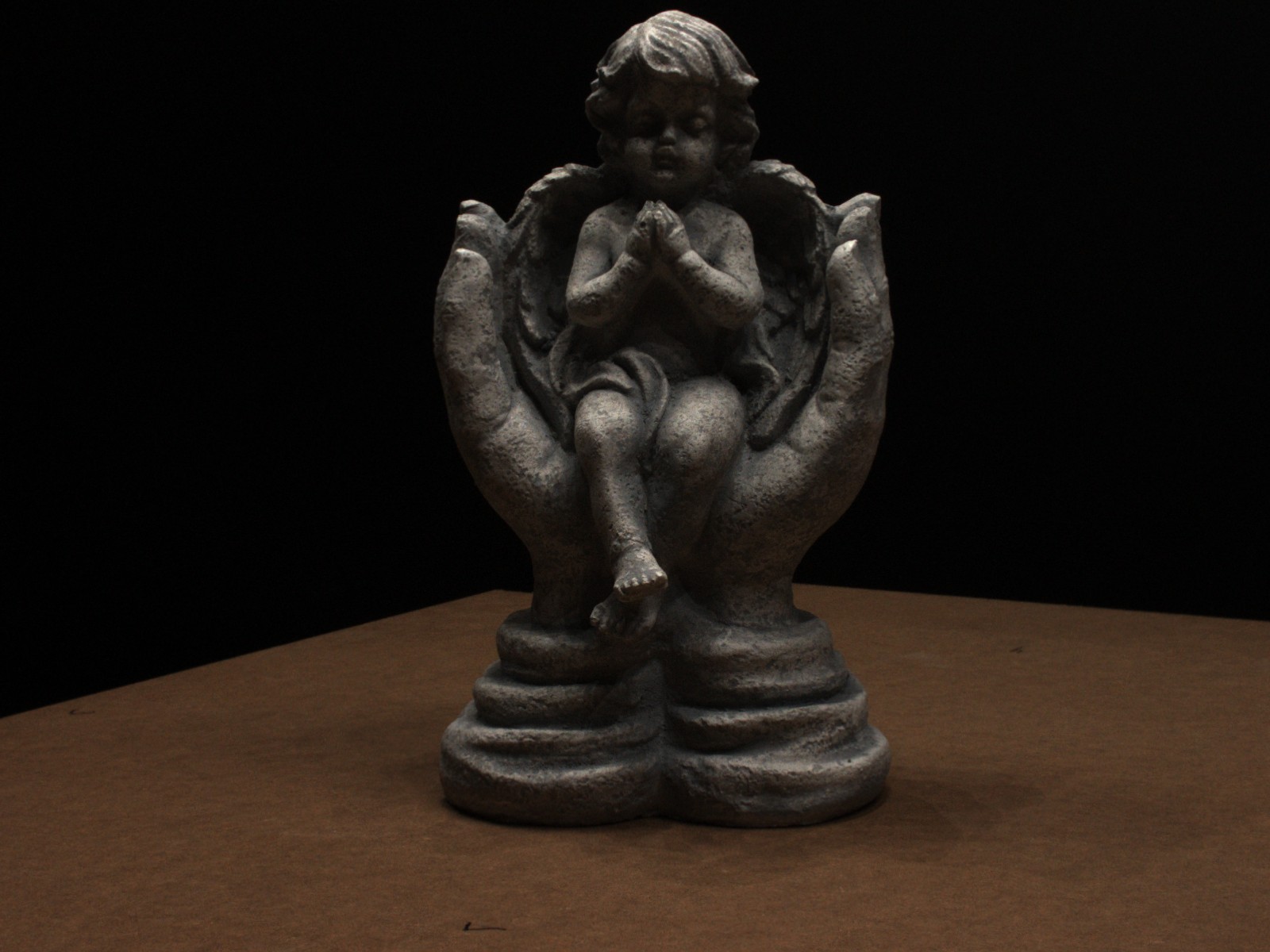}
\end{minipage}
\begin{minipage}[b]{0.245\linewidth}
\centering
\includegraphics[width=1.0\linewidth]{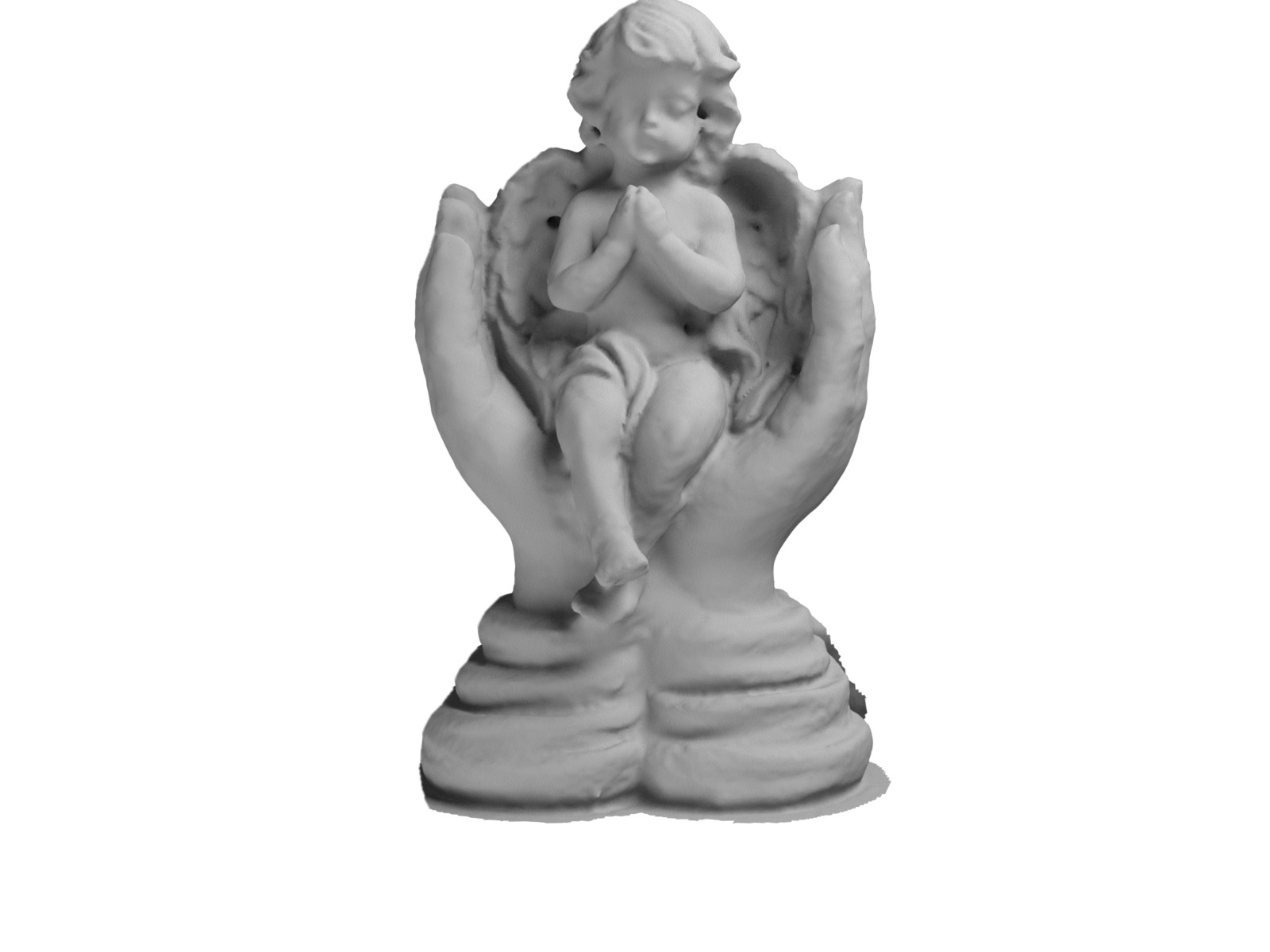}
\end{minipage}
\begin{minipage}[b]{0.245\linewidth}
\centering
\includegraphics[width=1.0\linewidth]{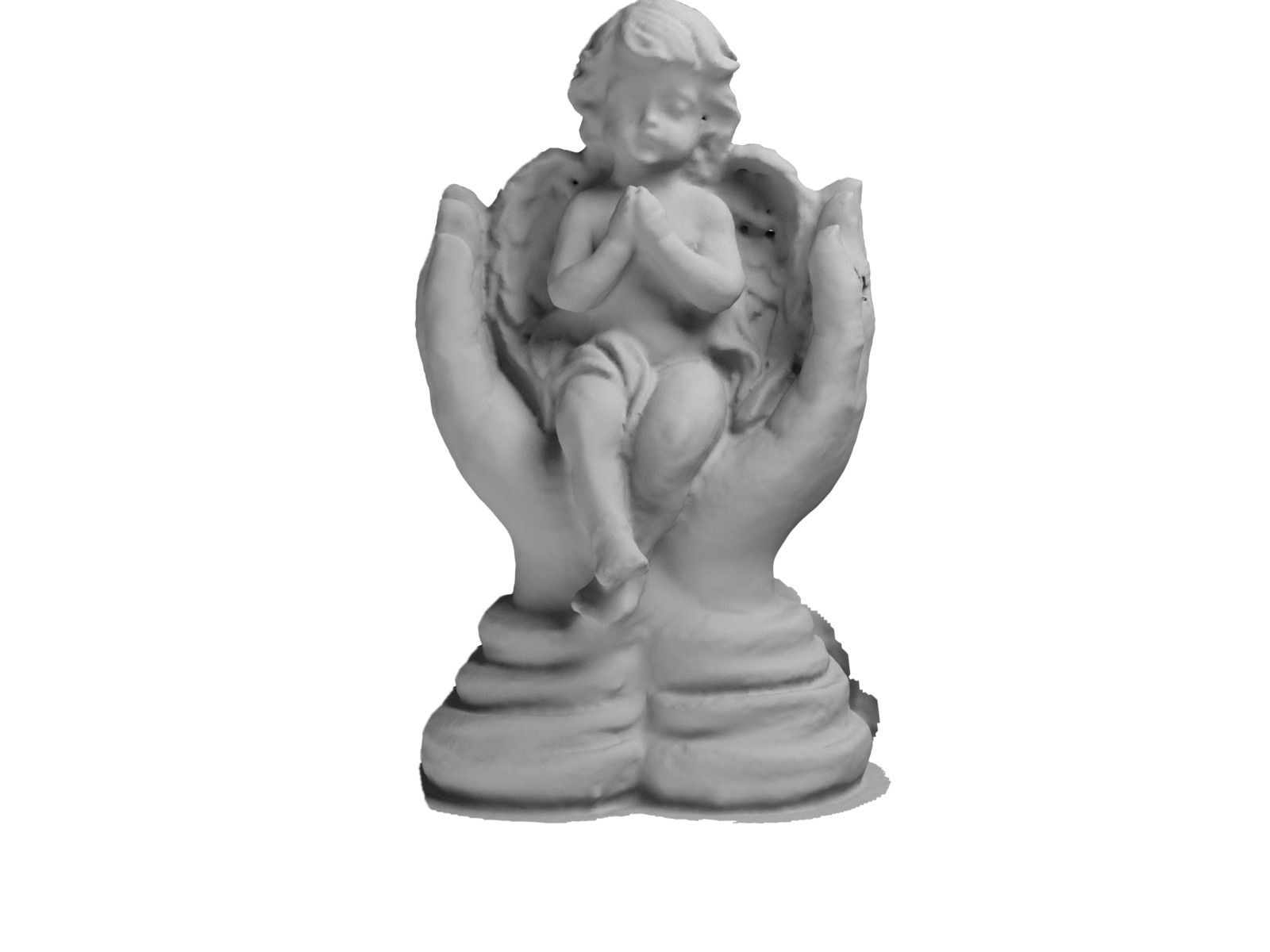}
\end{minipage}
\begin{minipage}[b]{0.245\linewidth}
\centering
\includegraphics[width=1.0\linewidth]{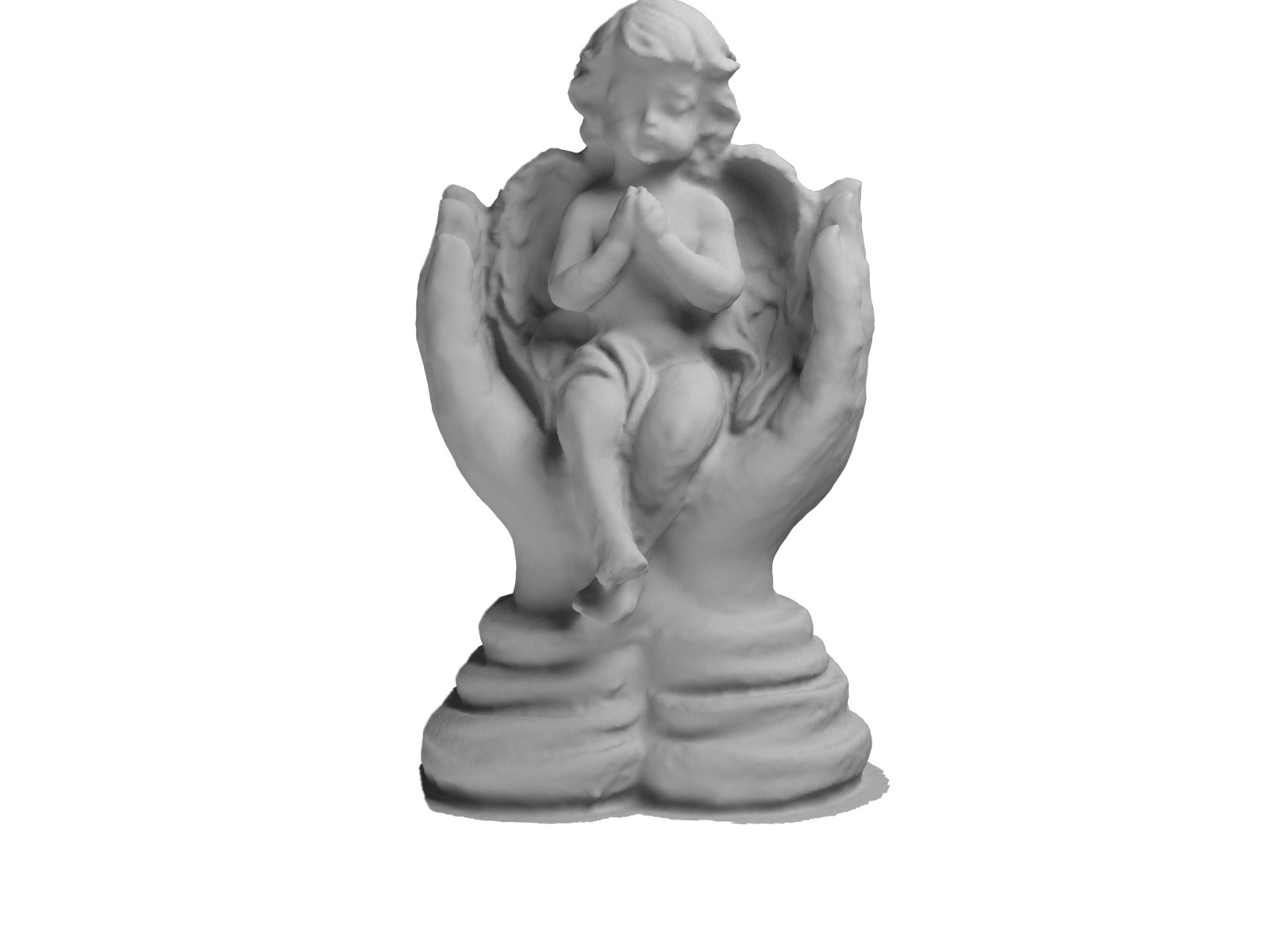}
\end{minipage}

\begin{minipage}[b]{0.245\linewidth}
\centering
\includegraphics[width=1.0\linewidth]{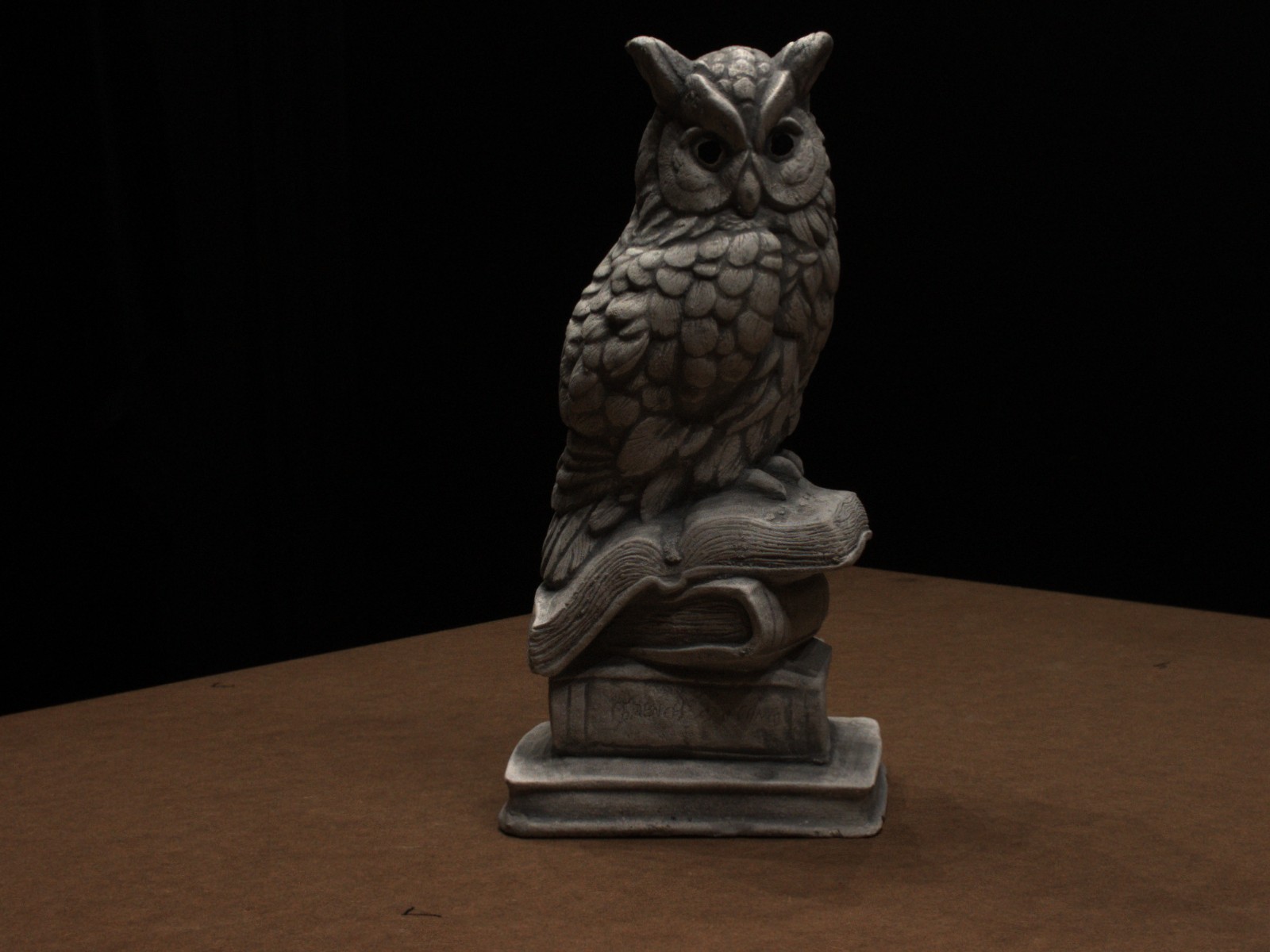}
\subcaption{Ground Truth}
\end{minipage}
\begin{minipage}[b]{0.245\linewidth}
\centering
\includegraphics[width=1.0\linewidth]{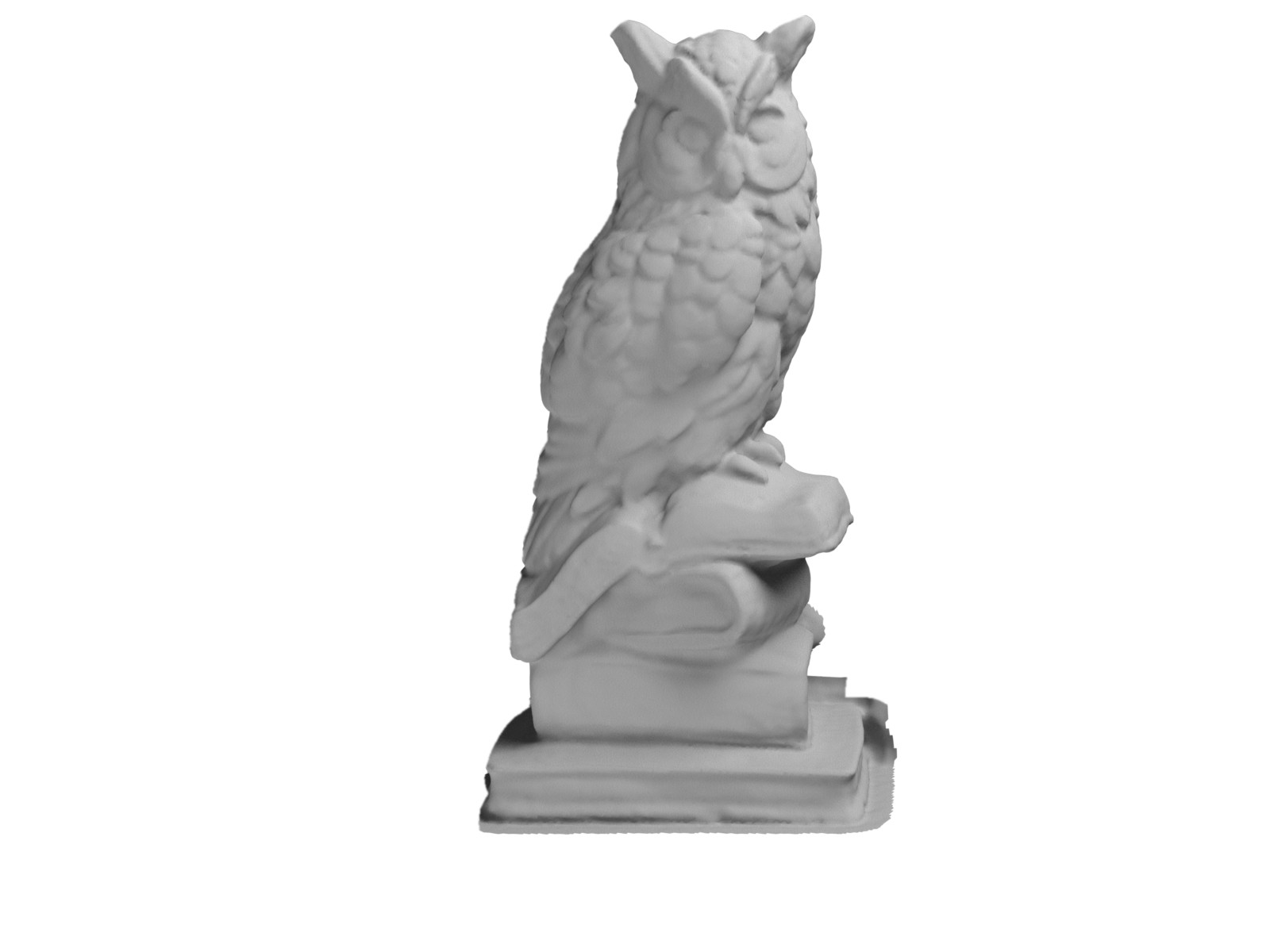}
\subcaption{NeuS-Facto}
\end{minipage}
\begin{minipage}[b]{0.245\linewidth}
\centering
\includegraphics[width=1.0\linewidth]{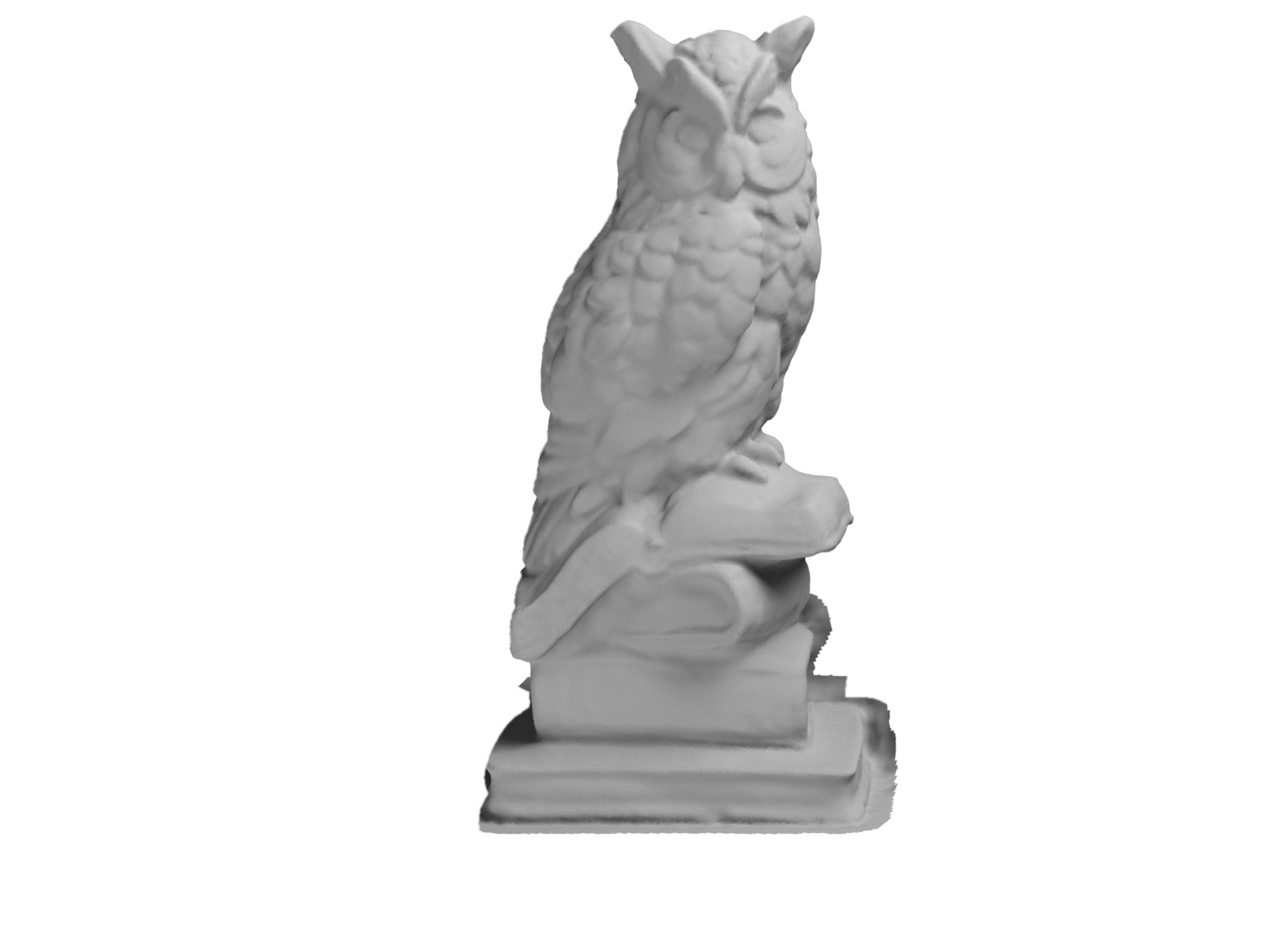}
\subcaption{OaV-Facto}
\end{minipage}
\begin{minipage}[b]{0.245\linewidth}
\centering
\includegraphics[width=1.0\linewidth]{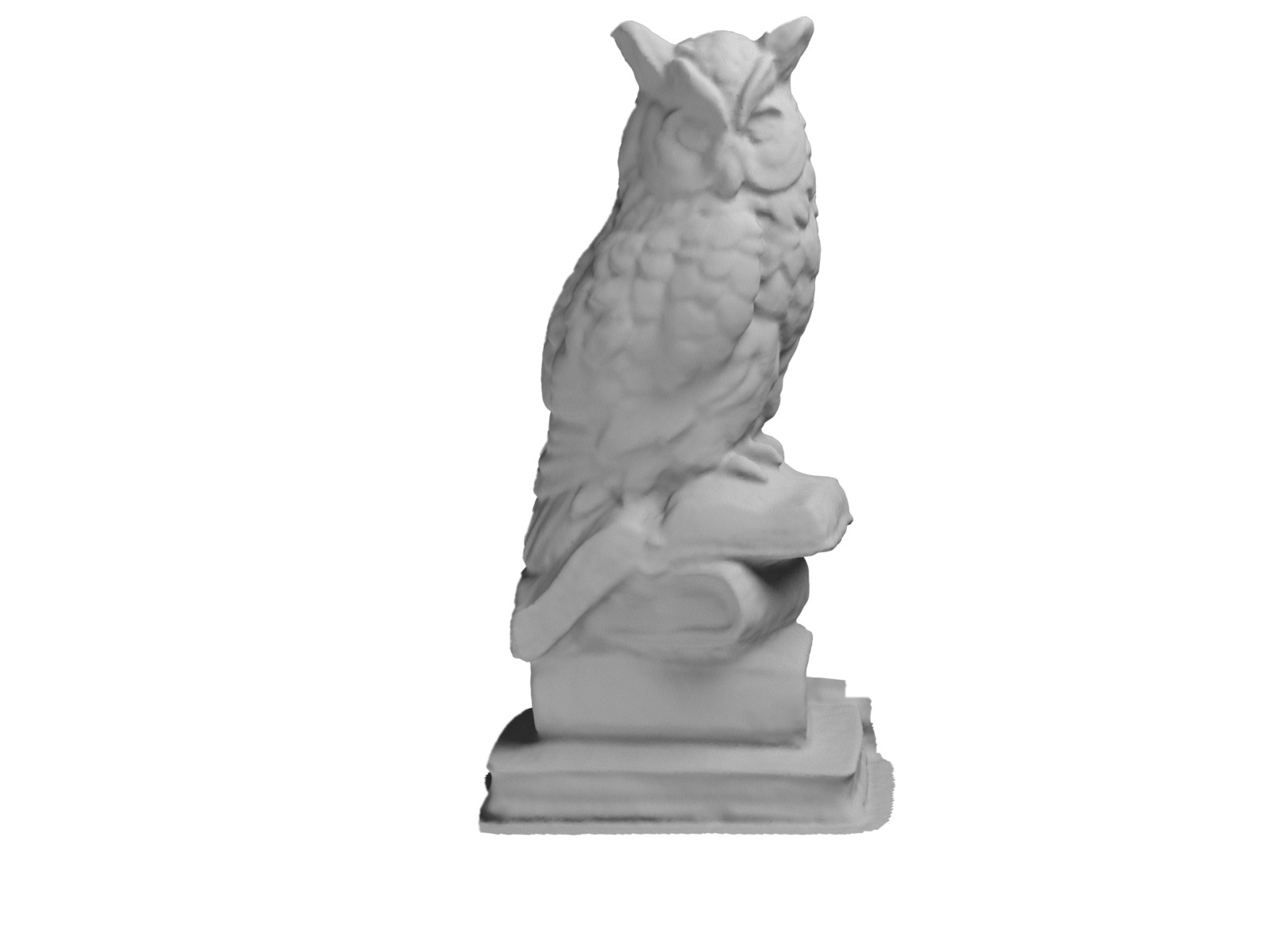}
\subcaption{SSDP-Facto}
\end{minipage}

\caption{Visualization examples on the DTU dataset.}
\end{figure}

\begin{figure}

\centering

\begin{minipage}[b]{0.245\linewidth}
\centering
\includegraphics[width=1.0\linewidth]{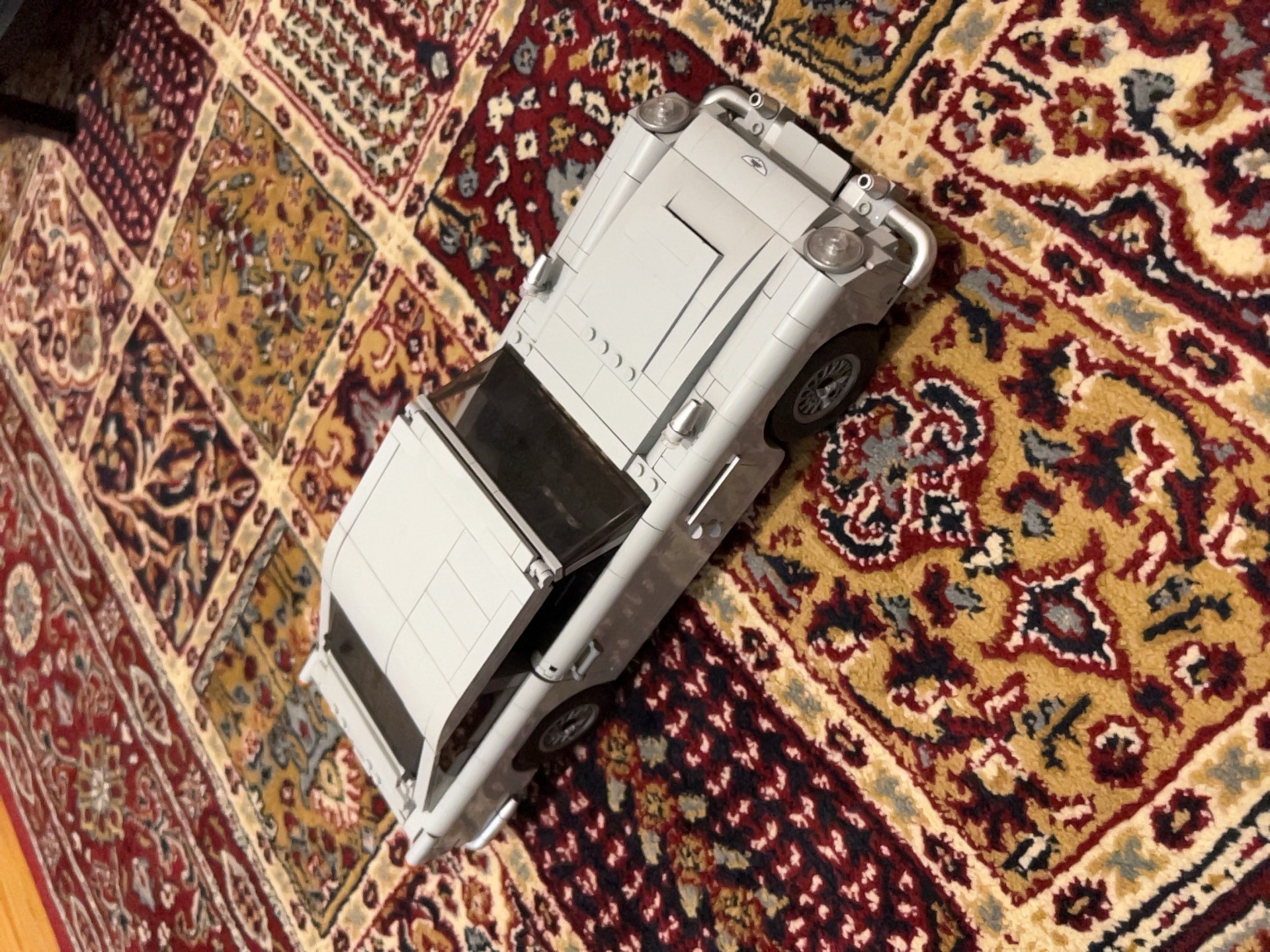}
\end{minipage}
\begin{minipage}[b]{0.245\linewidth}
\centering
\includegraphics[width=1.0\linewidth]{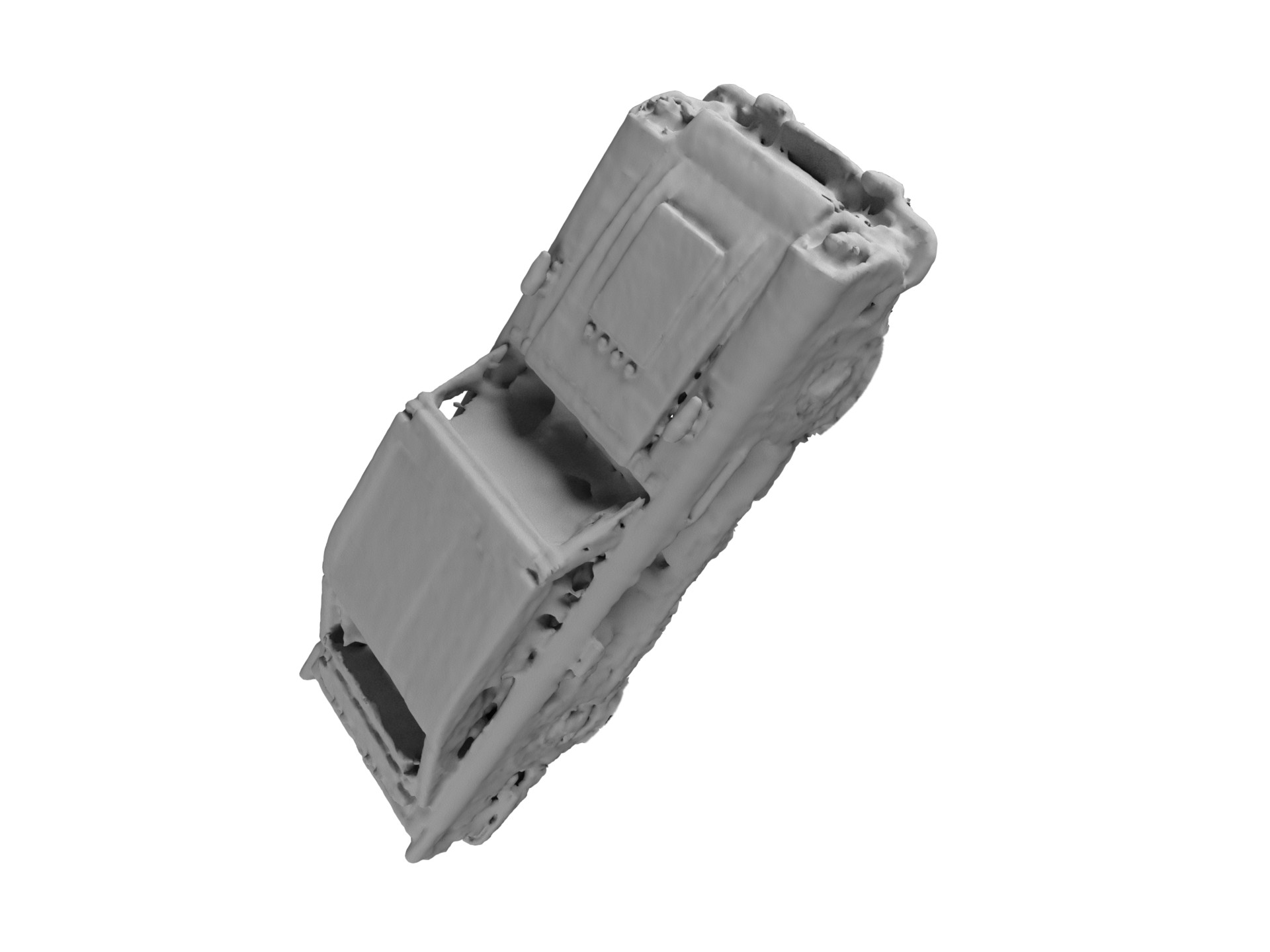}
\end{minipage}
\begin{minipage}[b]{0.245\linewidth}
\centering
\includegraphics[width=1.0\linewidth]{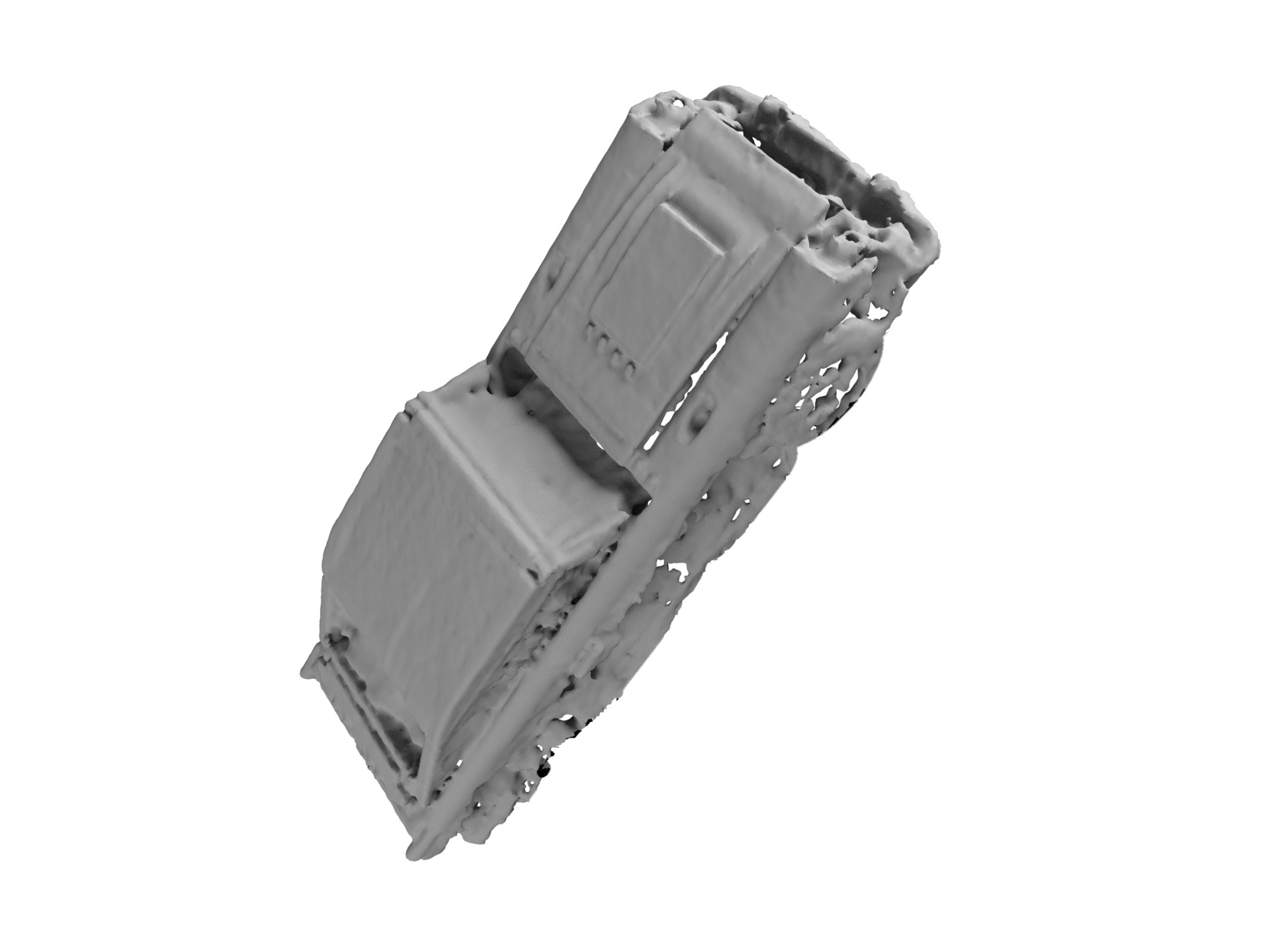}
\end{minipage}
\begin{minipage}[b]{0.245\linewidth}
\centering
\includegraphics[width=1.0\linewidth]{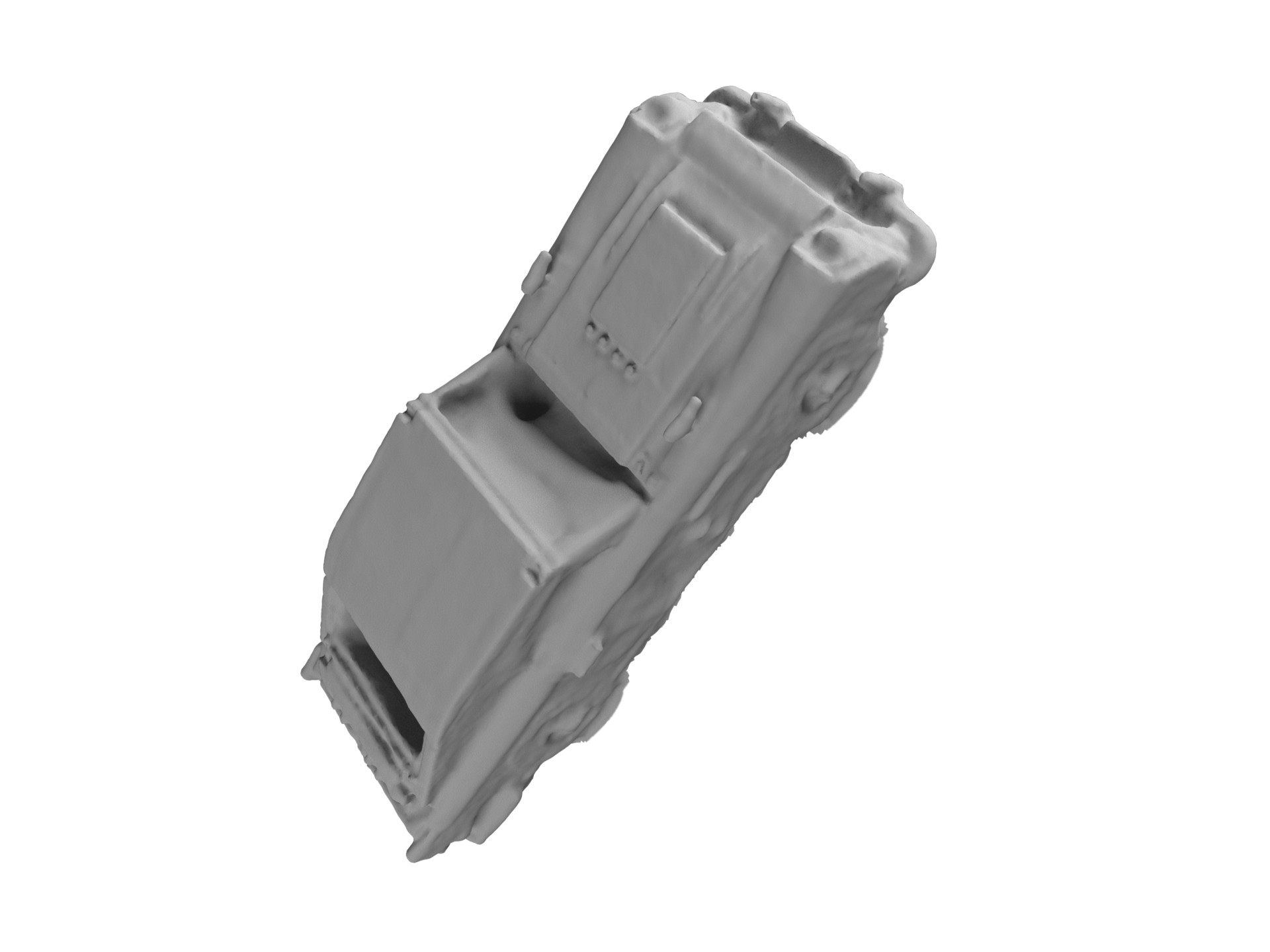}
\end{minipage}

\begin{minipage}[b]{0.245\linewidth}
\centering
\includegraphics[width=1.0\linewidth]{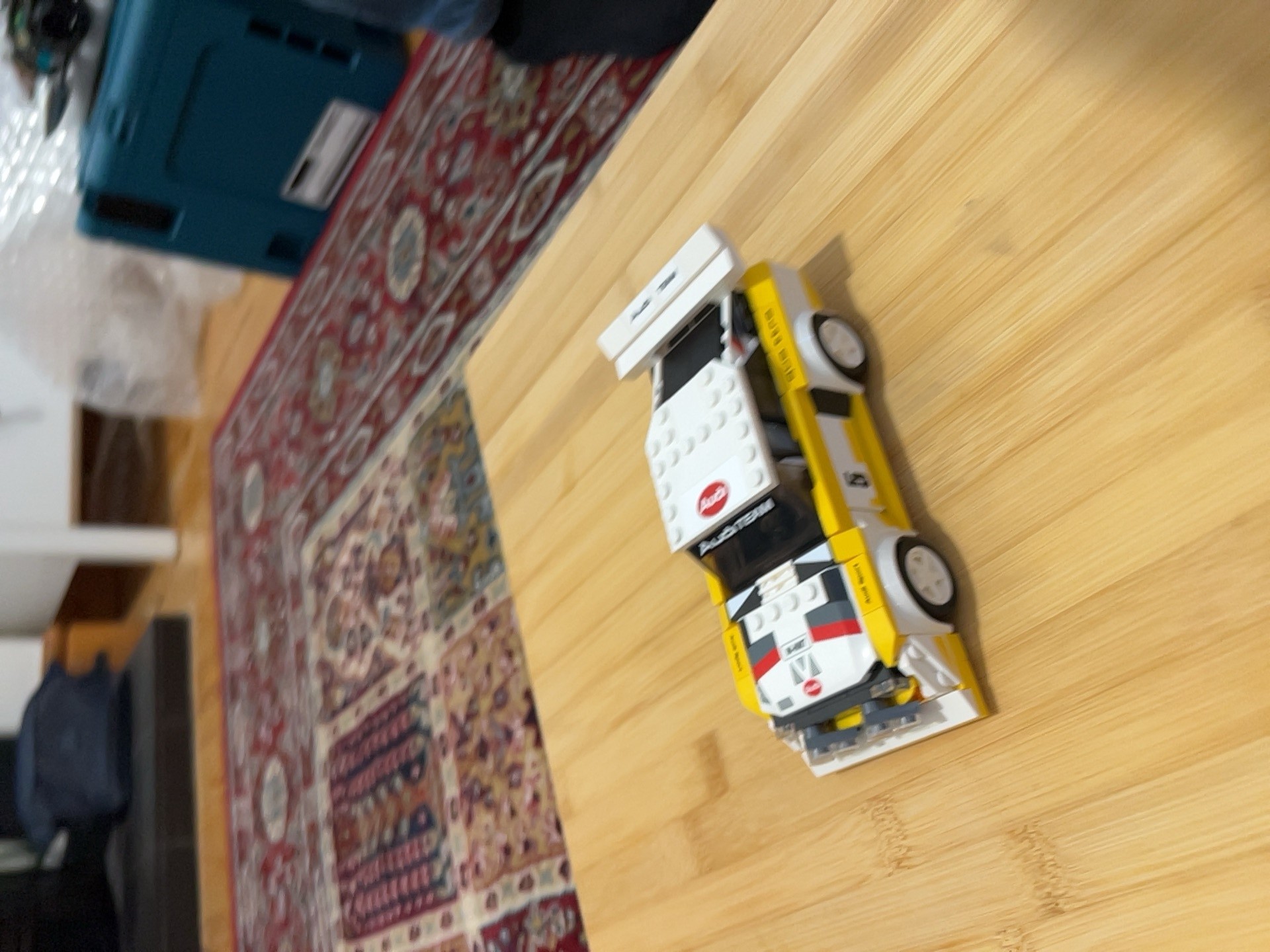}
\end{minipage}
\begin{minipage}[b]{0.245\linewidth}
\centering
\includegraphics[width=1.0\linewidth]{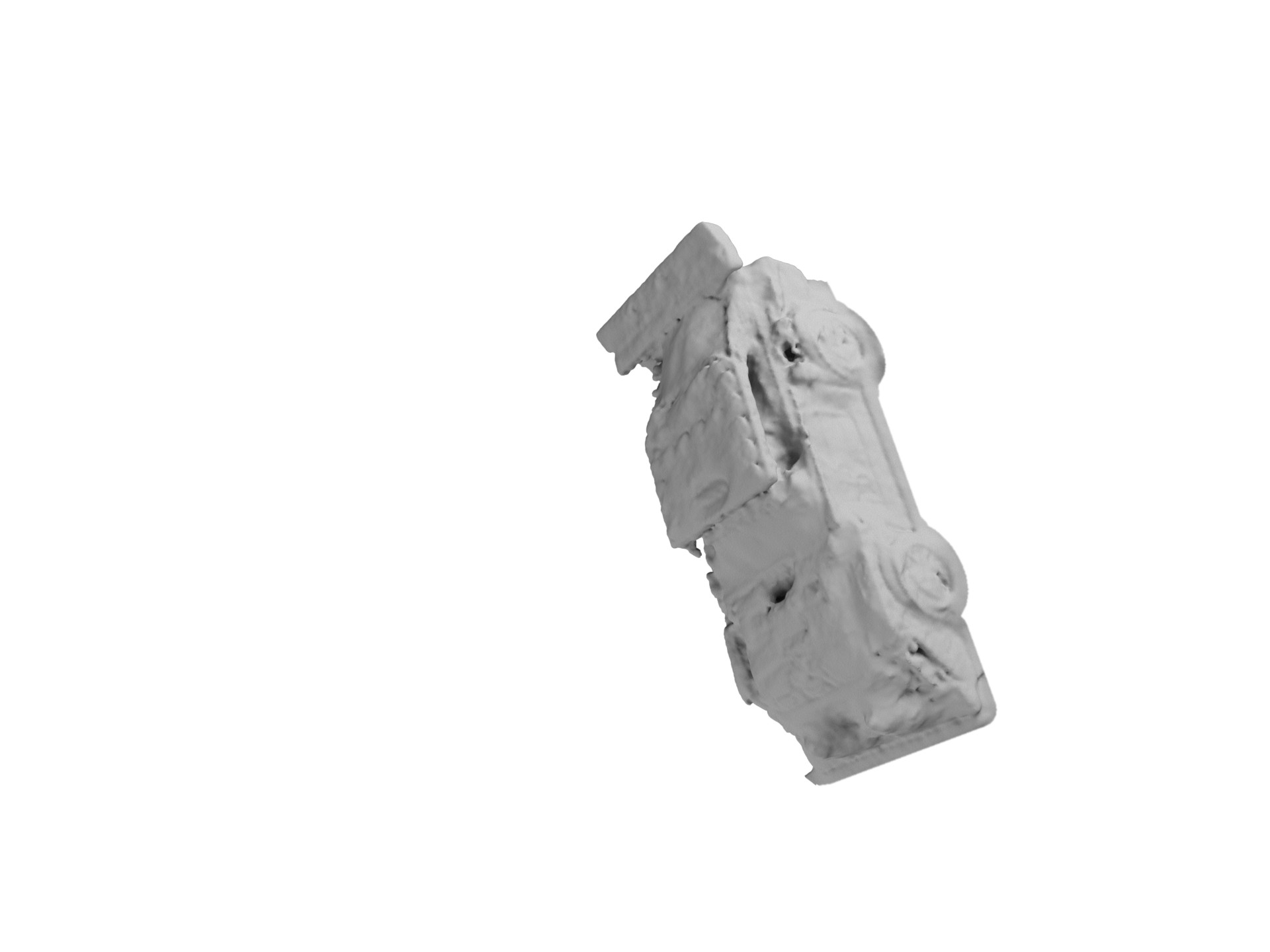}
\end{minipage}
\begin{minipage}[b]{0.245\linewidth}
\centering
\includegraphics[width=1.0\linewidth]{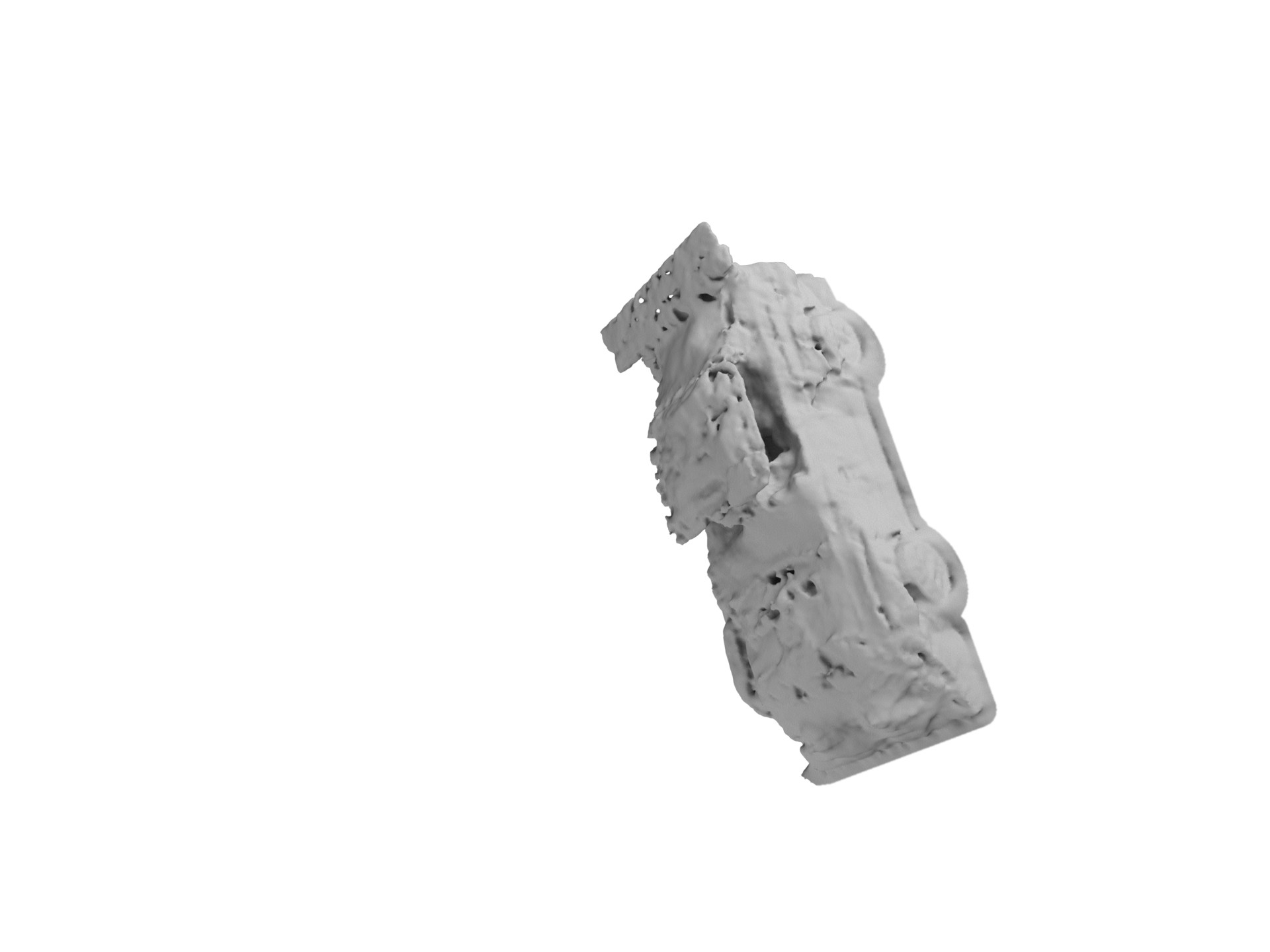}
\end{minipage}
\begin{minipage}[b]{0.245\linewidth}
\centering
\includegraphics[width=1.0\linewidth]{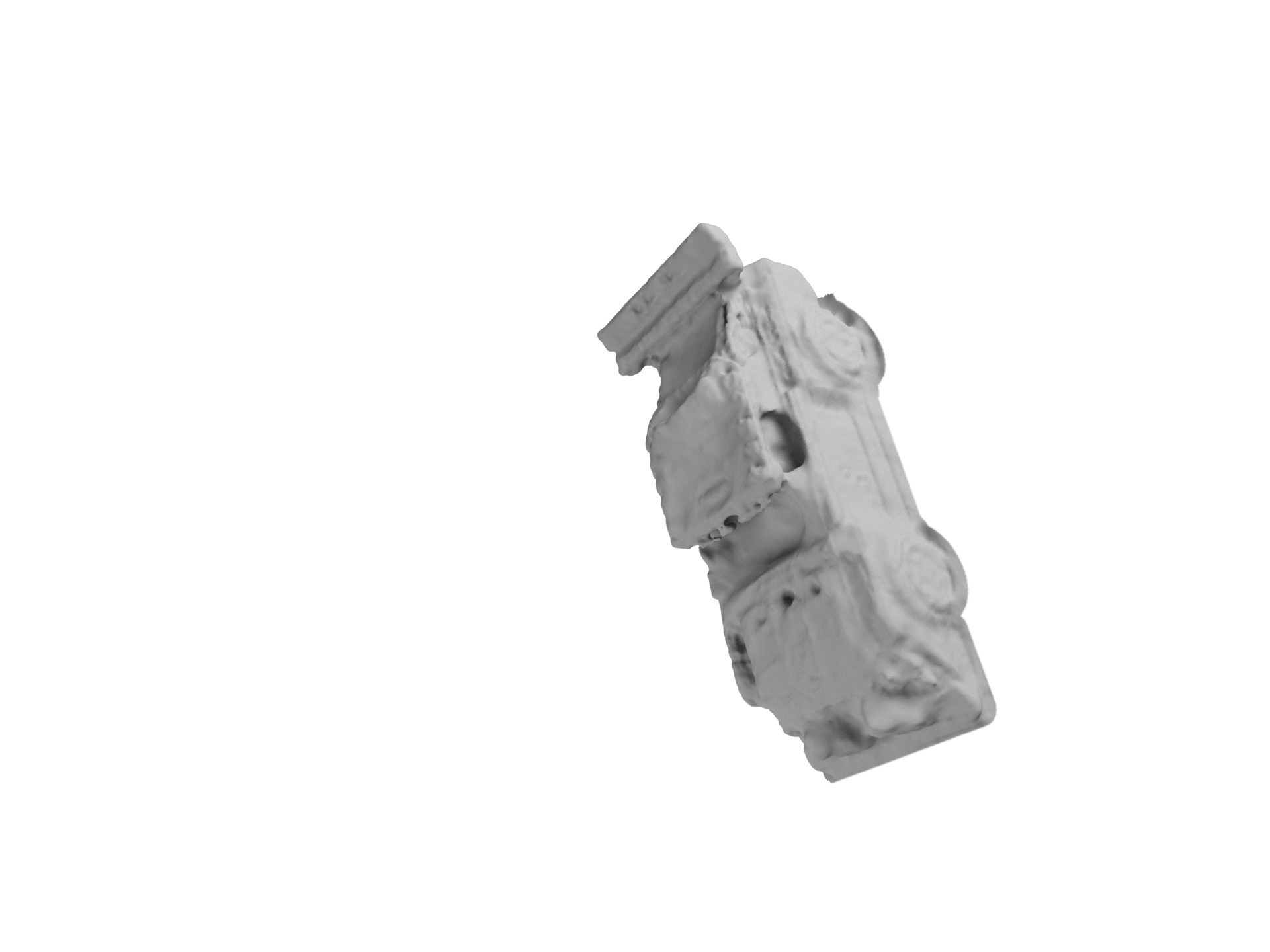}
\end{minipage}

\begin{minipage}[b]{0.245\linewidth}
\centering
\includegraphics[width=1.0\linewidth]{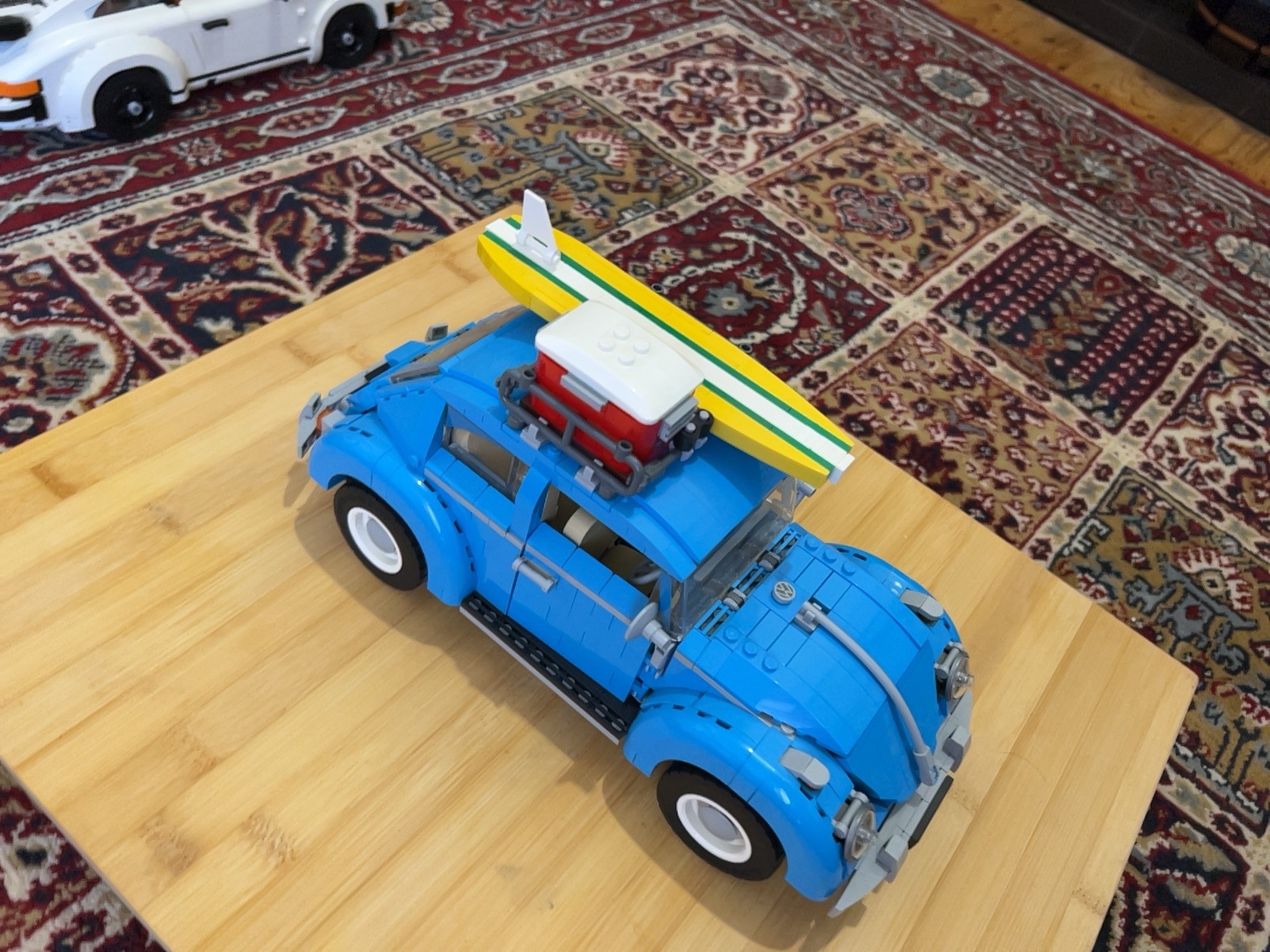}
\end{minipage}
\begin{minipage}[b]{0.245\linewidth}
\centering
\includegraphics[width=1.0\linewidth]{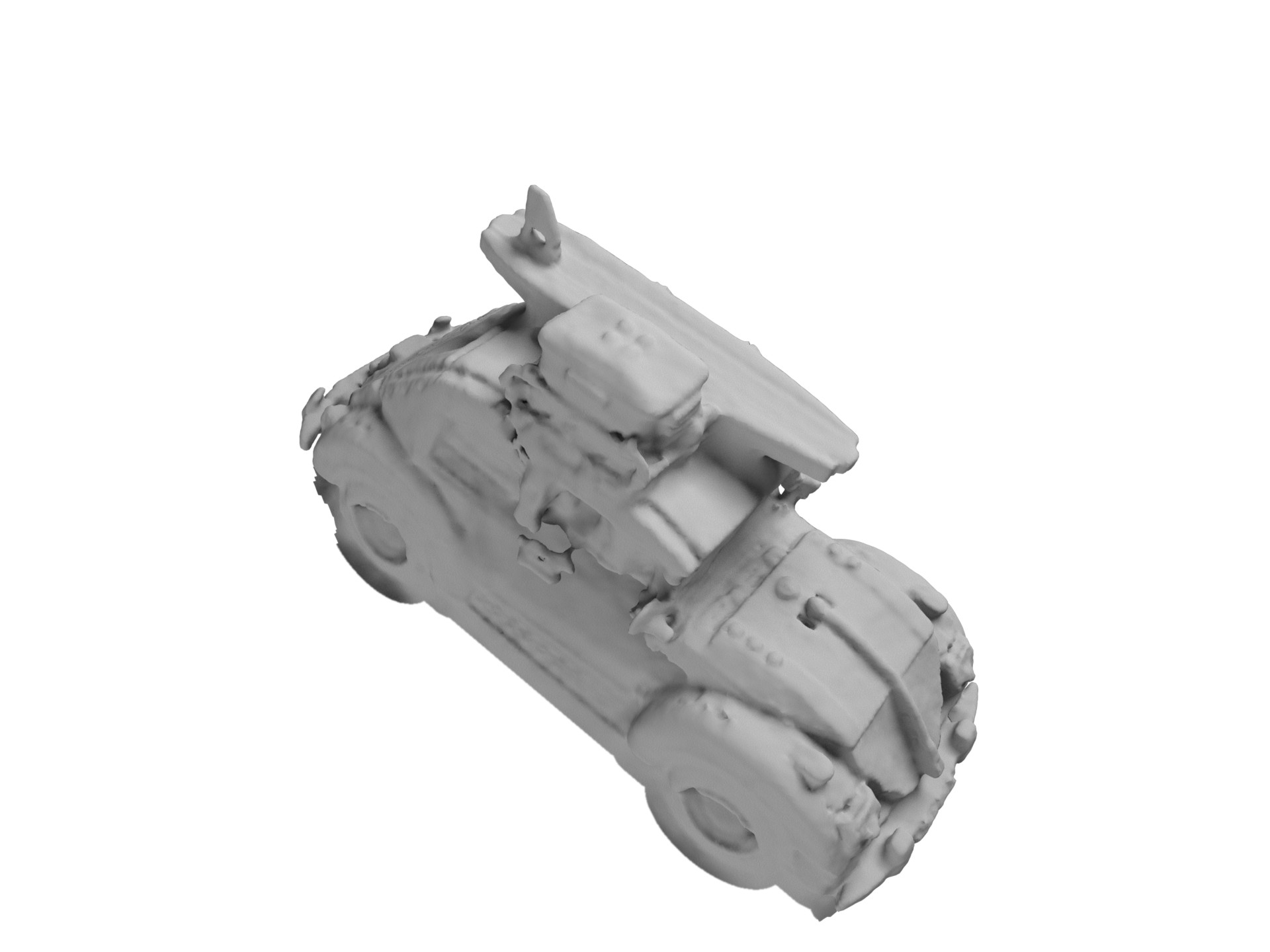}
\end{minipage}
\begin{minipage}[b]{0.245\linewidth}
\centering
\includegraphics[width=1.0\linewidth]{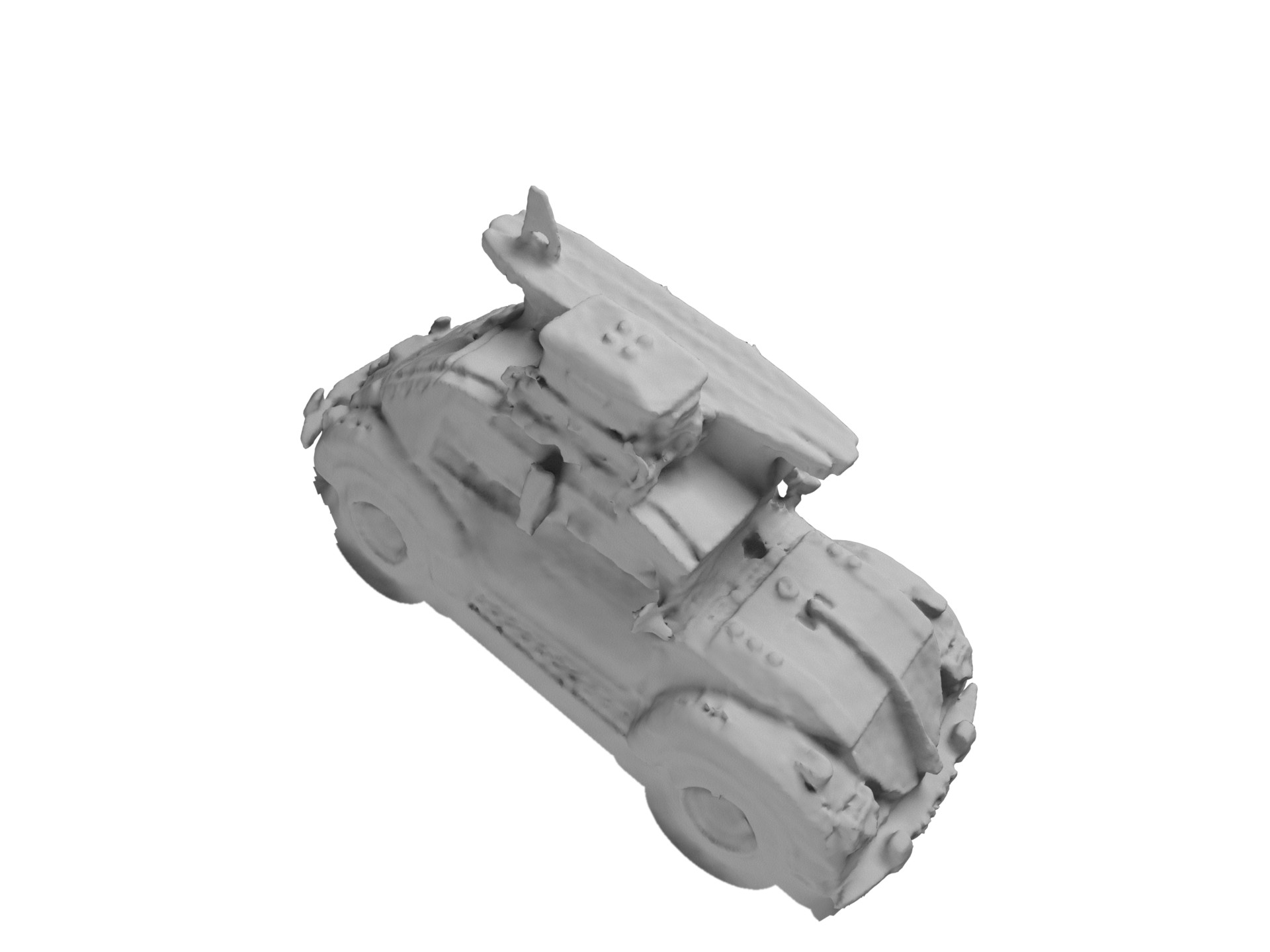}
\end{minipage}
\begin{minipage}[b]{0.245\linewidth}
\centering
\includegraphics[width=1.0\linewidth]{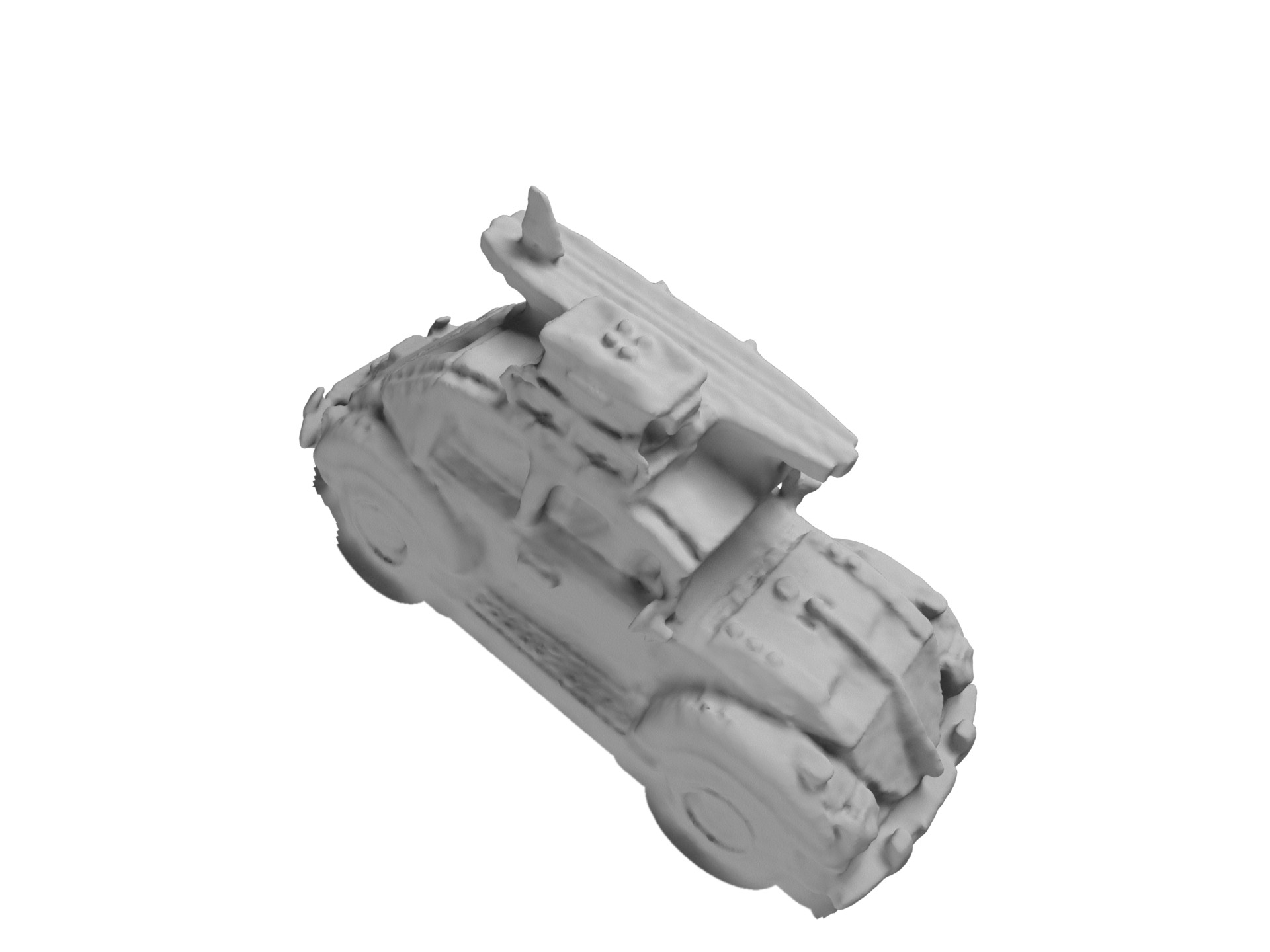}
\end{minipage}

\begin{minipage}[b]{0.245\linewidth}
\centering
\includegraphics[width=1.0\linewidth]{images/lego/big_ben/gt/000013.jpg}
\end{minipage}
\begin{minipage}[b]{0.245\linewidth}
\centering
\includegraphics[width=1.0\linewidth]{images/lego/big_ben/neus-facto-base/000013.jpg}
\end{minipage}
\begin{minipage}[b]{0.245\linewidth}
\centering
\includegraphics[width=1.0\linewidth]{images/lego/big_ben/oav-facto-base/000013.jpg}
\end{minipage}
\begin{minipage}[b]{0.245\linewidth}
\centering
\includegraphics[width=1.0\linewidth]{images/lego/big_ben/ssdp-facto-base/000013.jpg}
\end{minipage}

\begin{minipage}[b]{0.245\linewidth}
\centering
\includegraphics[width=1.0\linewidth]{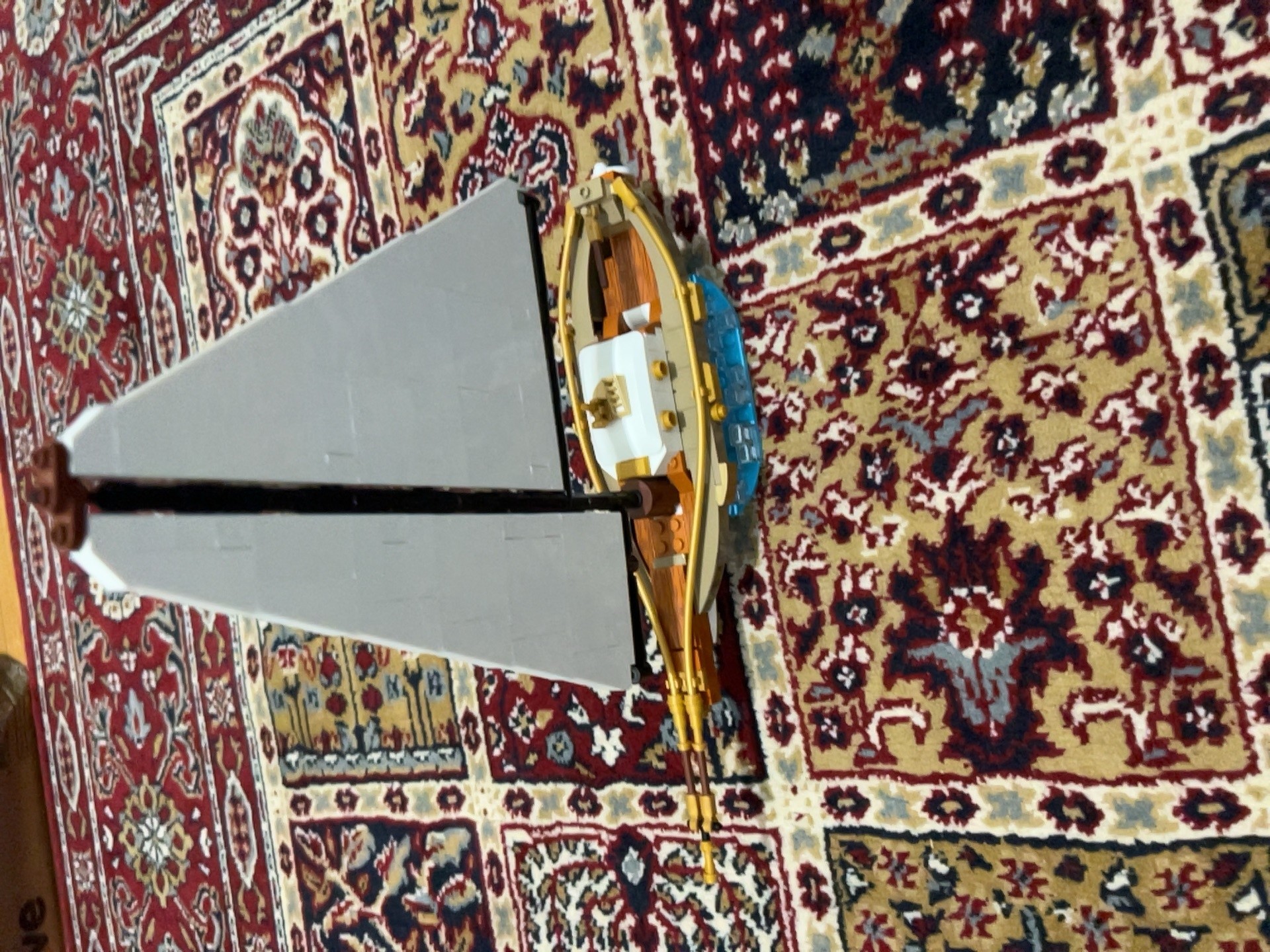}
\end{minipage}
\begin{minipage}[b]{0.245\linewidth}
\centering
\includegraphics[width=1.0\linewidth]{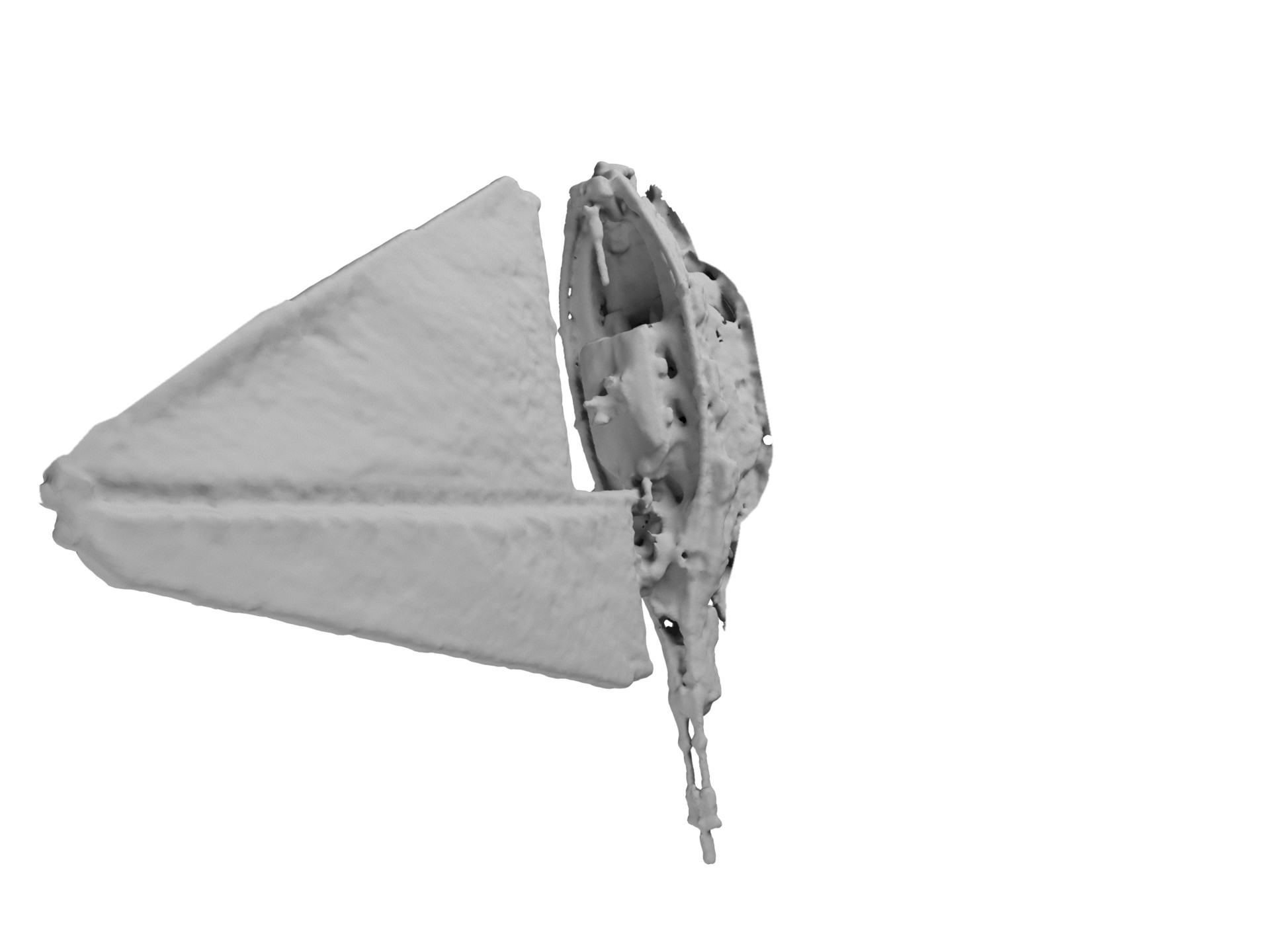}
\end{minipage}
\begin{minipage}[b]{0.245\linewidth}
\centering
\includegraphics[width=1.0\linewidth]{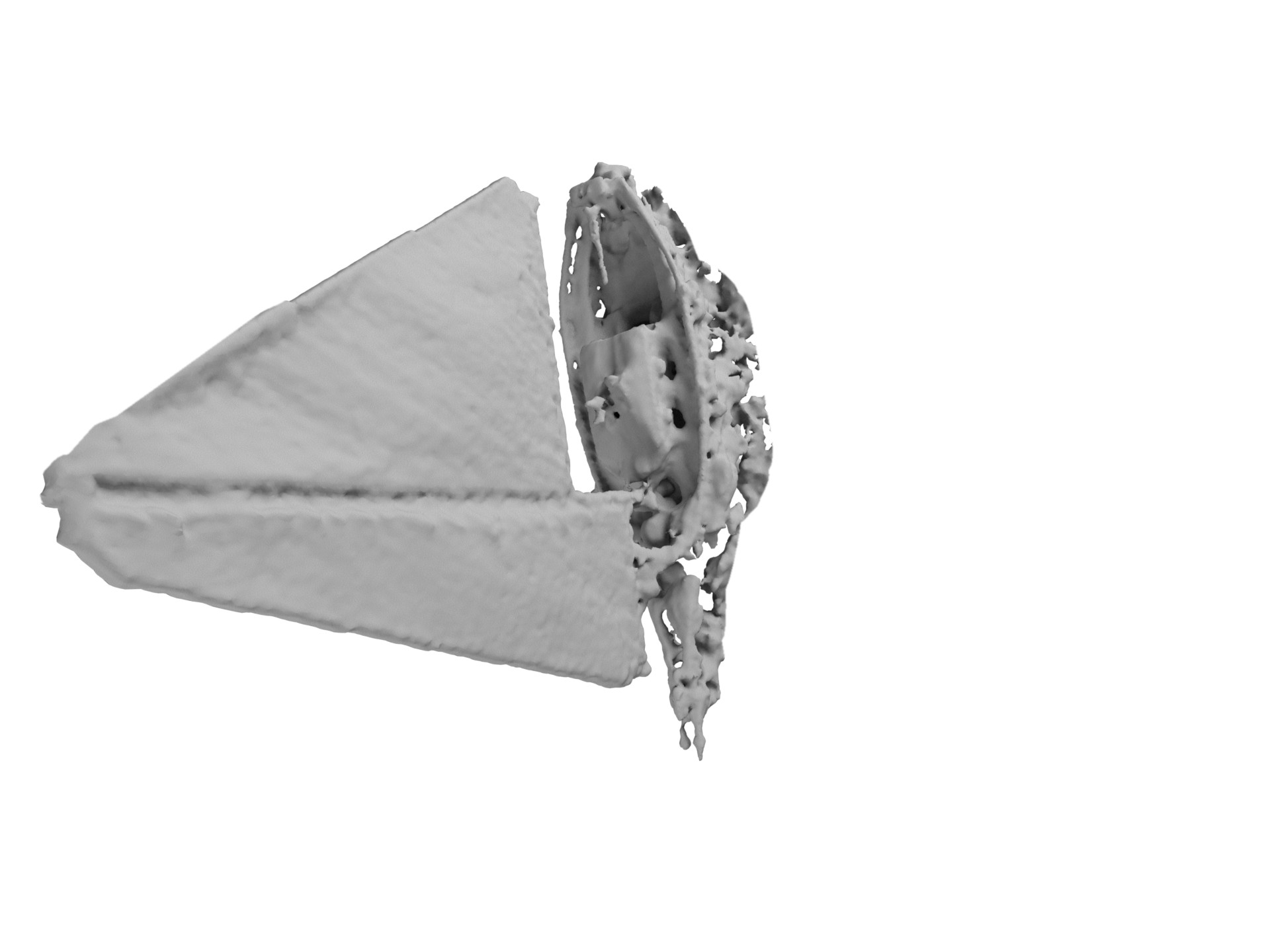}
\end{minipage}
\begin{minipage}[b]{0.245\linewidth}
\centering
\includegraphics[width=1.0\linewidth]{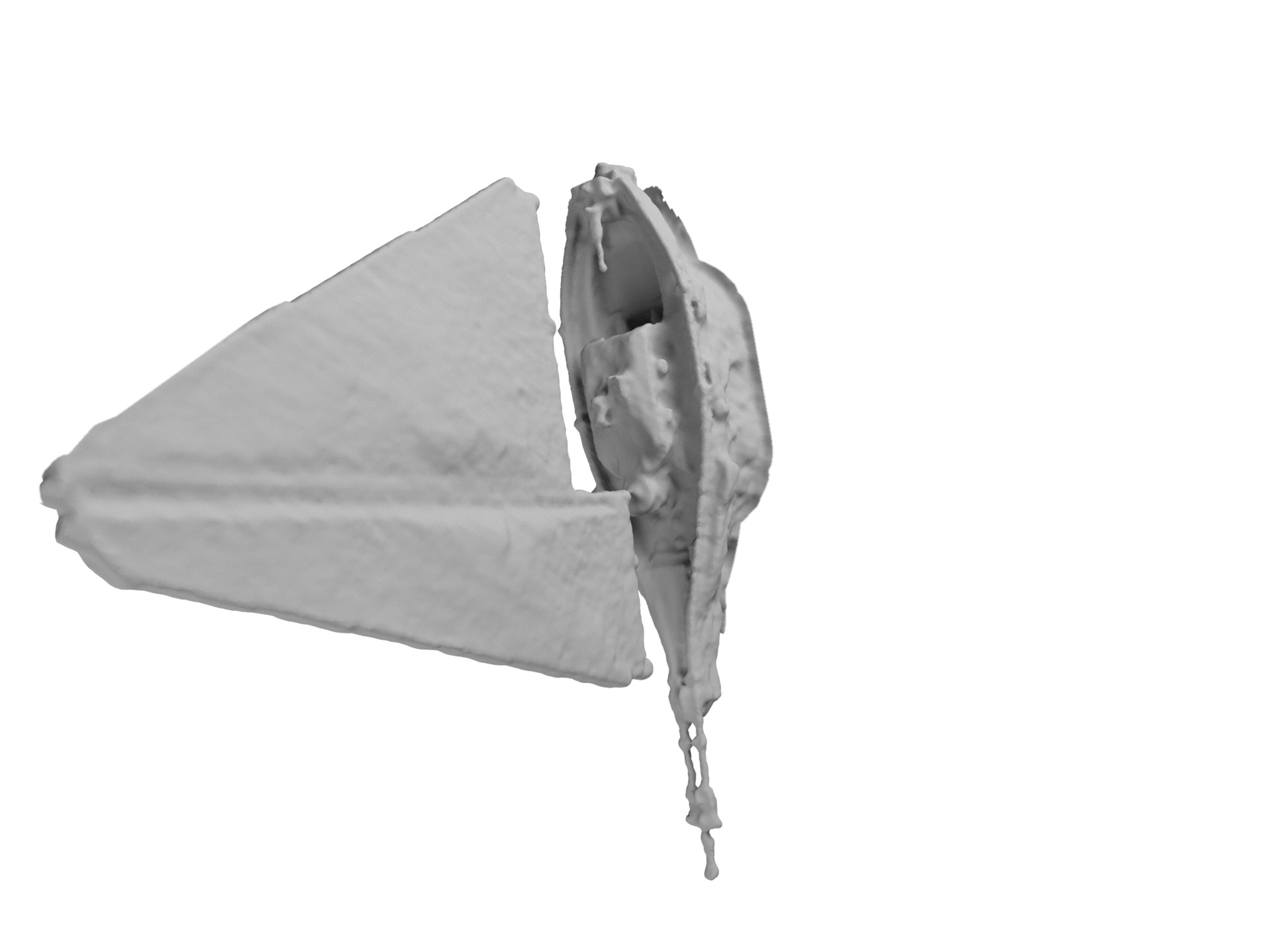}
\end{minipage}

\begin{minipage}[b]{0.245\linewidth}
\centering
\includegraphics[width=1.0\linewidth]{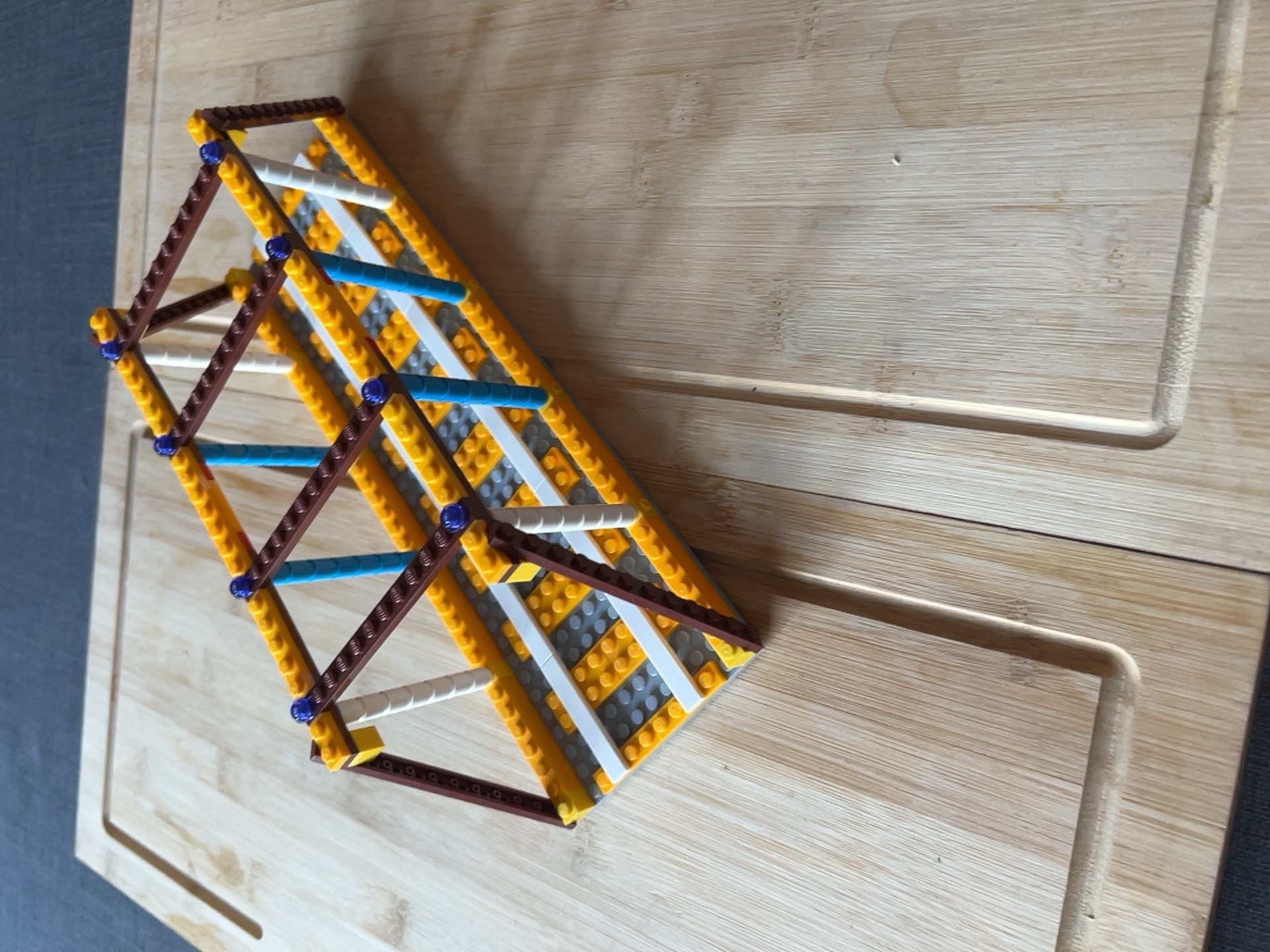}
\subcaption{Ground Truth}
\end{minipage}
\begin{minipage}[b]{0.245\linewidth}
\centering
\includegraphics[width=1.0\linewidth]{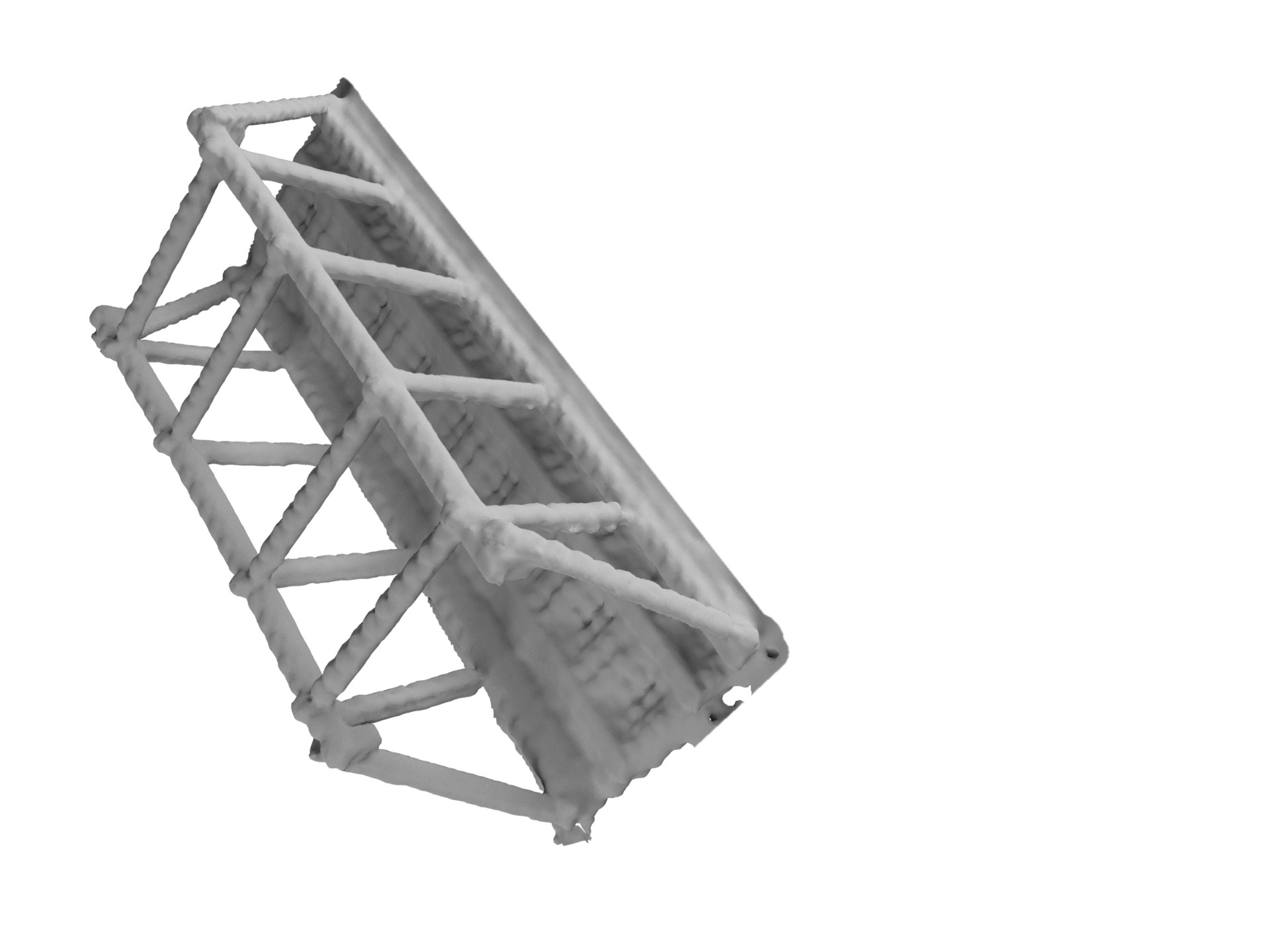}
\subcaption{NeuS-Facto}
\end{minipage}
\begin{minipage}[b]{0.245\linewidth}
\centering
\includegraphics[width=1.0\linewidth]{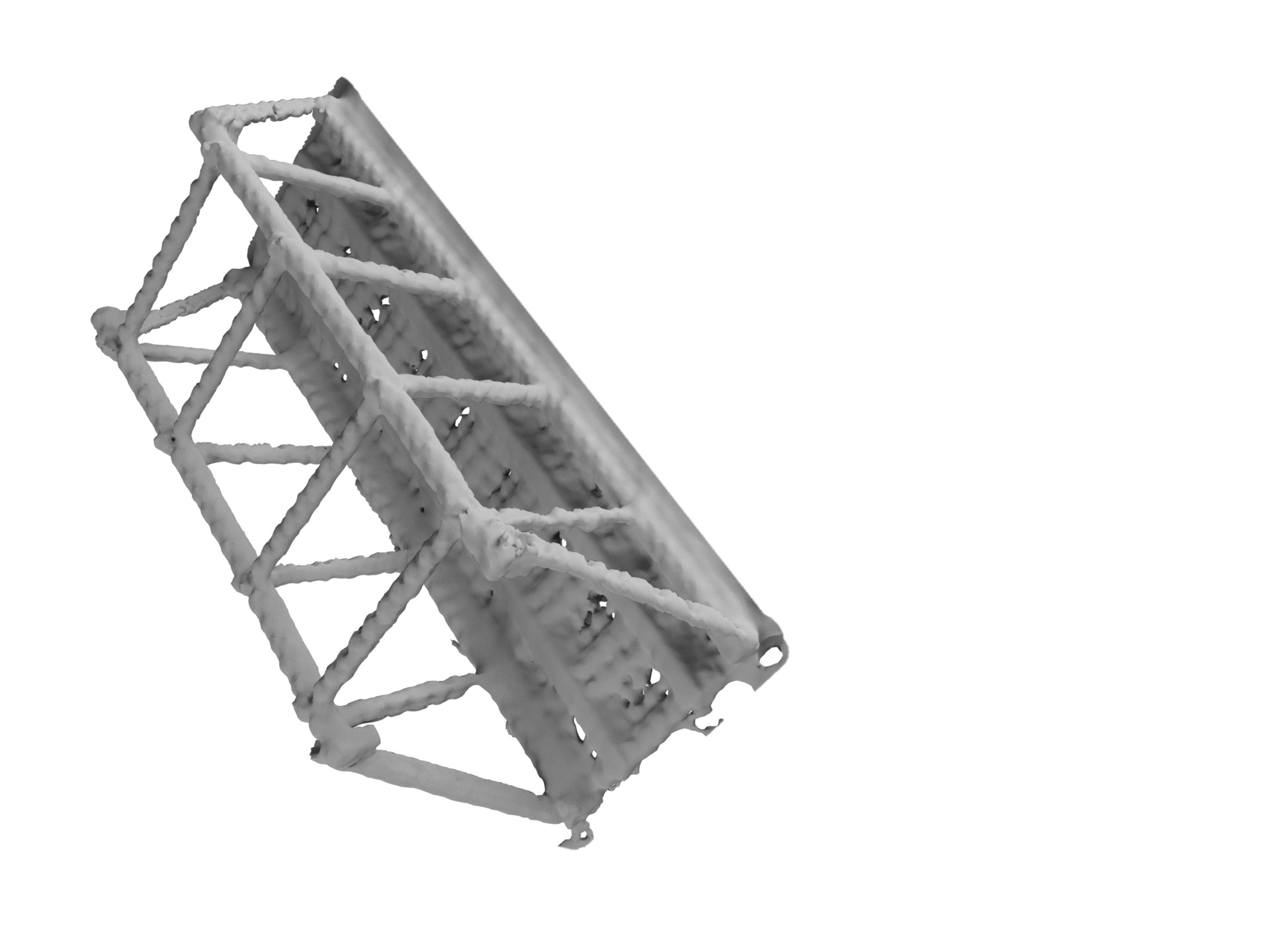}
\subcaption{OaV-Facto}
\end{minipage}
\begin{minipage}[b]{0.245\linewidth}
\centering
\includegraphics[width=1.0\linewidth]{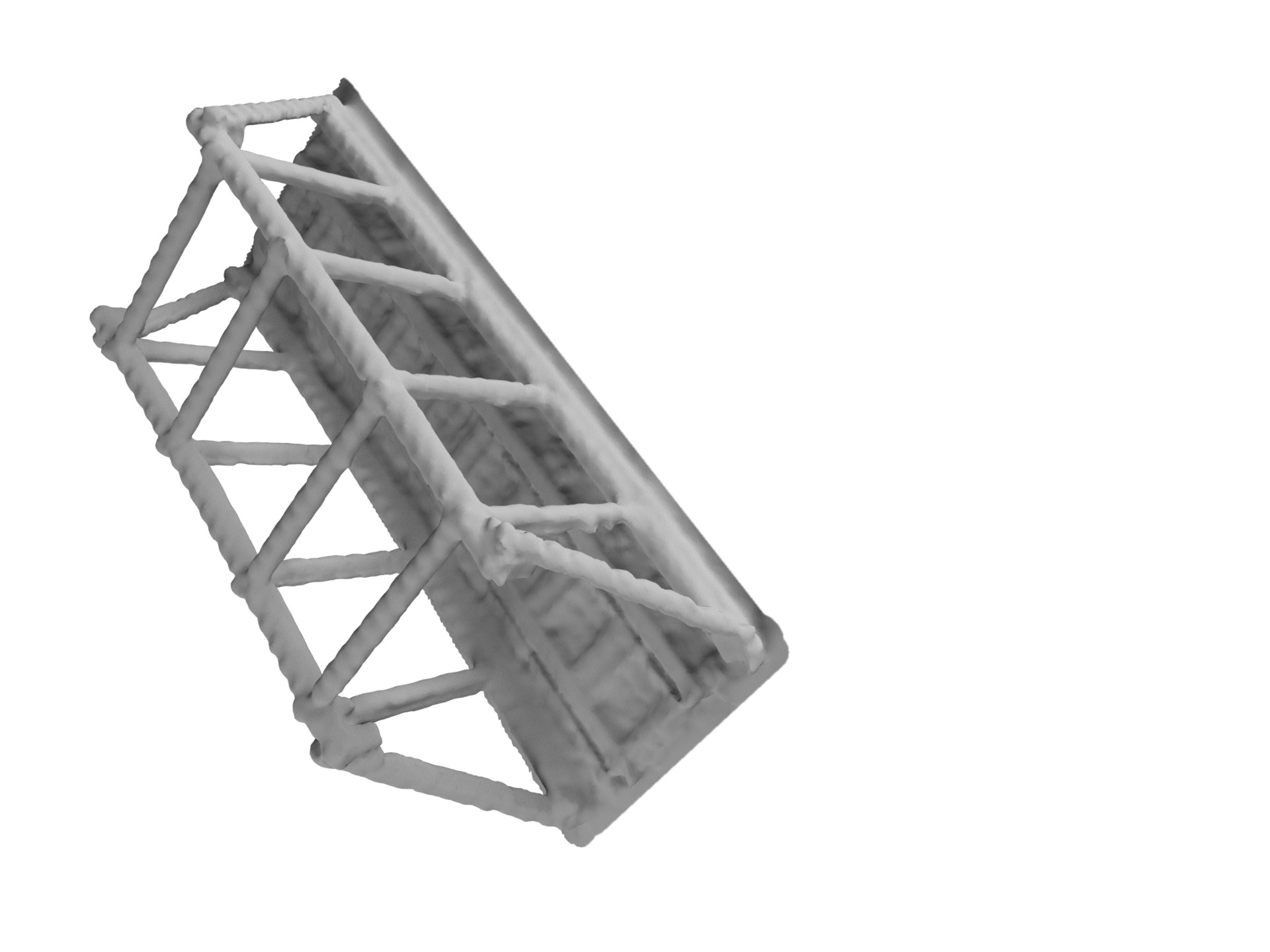}
\subcaption{SSDP-Facto}
\end{minipage}

\caption{Visualization examples on the MobileBrick dataset.}
\label{fig:lego_visualization_examples_all}
\end{figure}

\begin{figure}

\ContinuedFloat
\setcounter{subfigure}{0}

\centering

\begin{minipage}[b]{0.245\linewidth}
\centering
\includegraphics[width=1.0\linewidth]{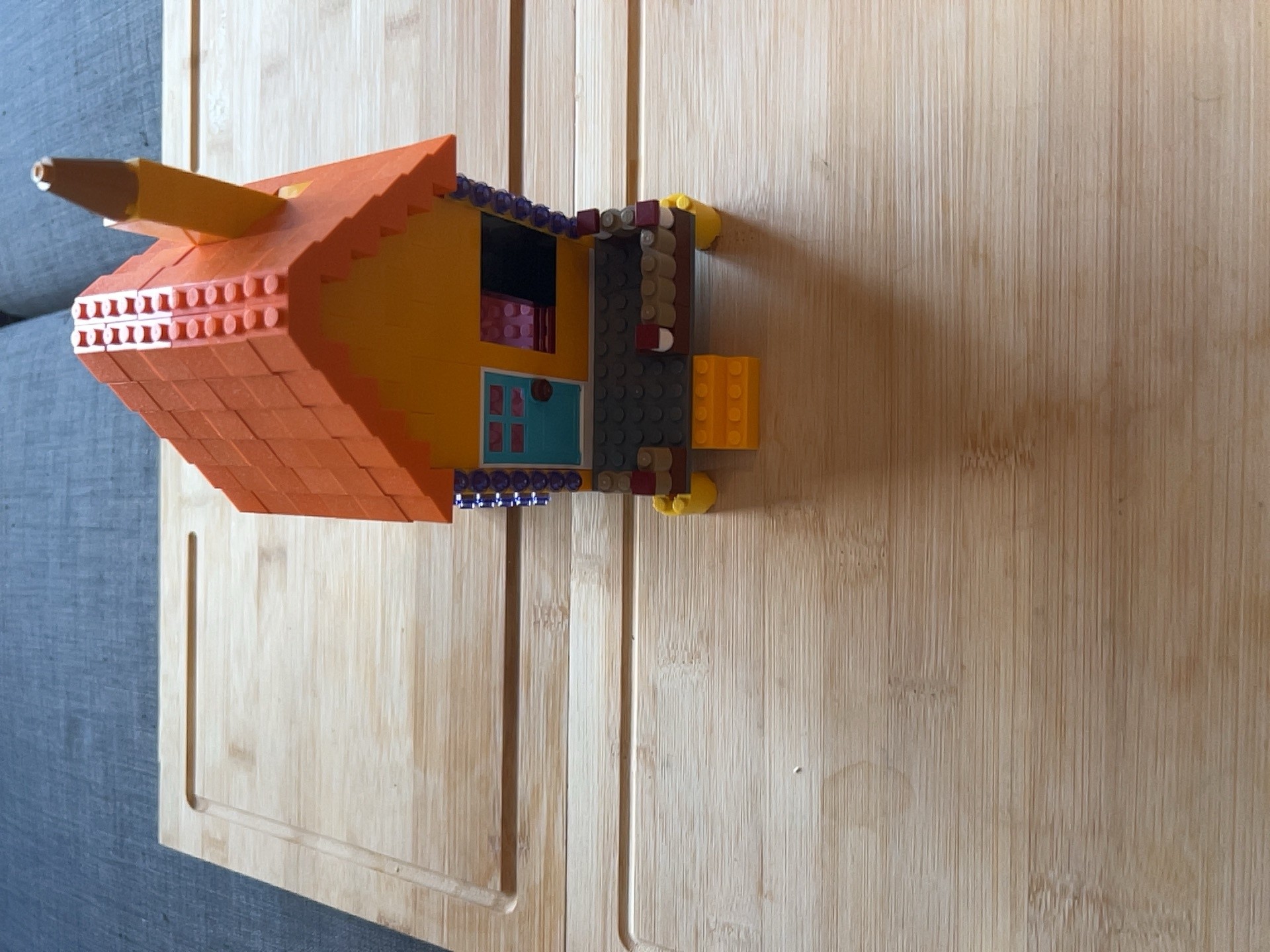}
\end{minipage}
\begin{minipage}[b]{0.245\linewidth}
\centering
\includegraphics[width=1.0\linewidth]{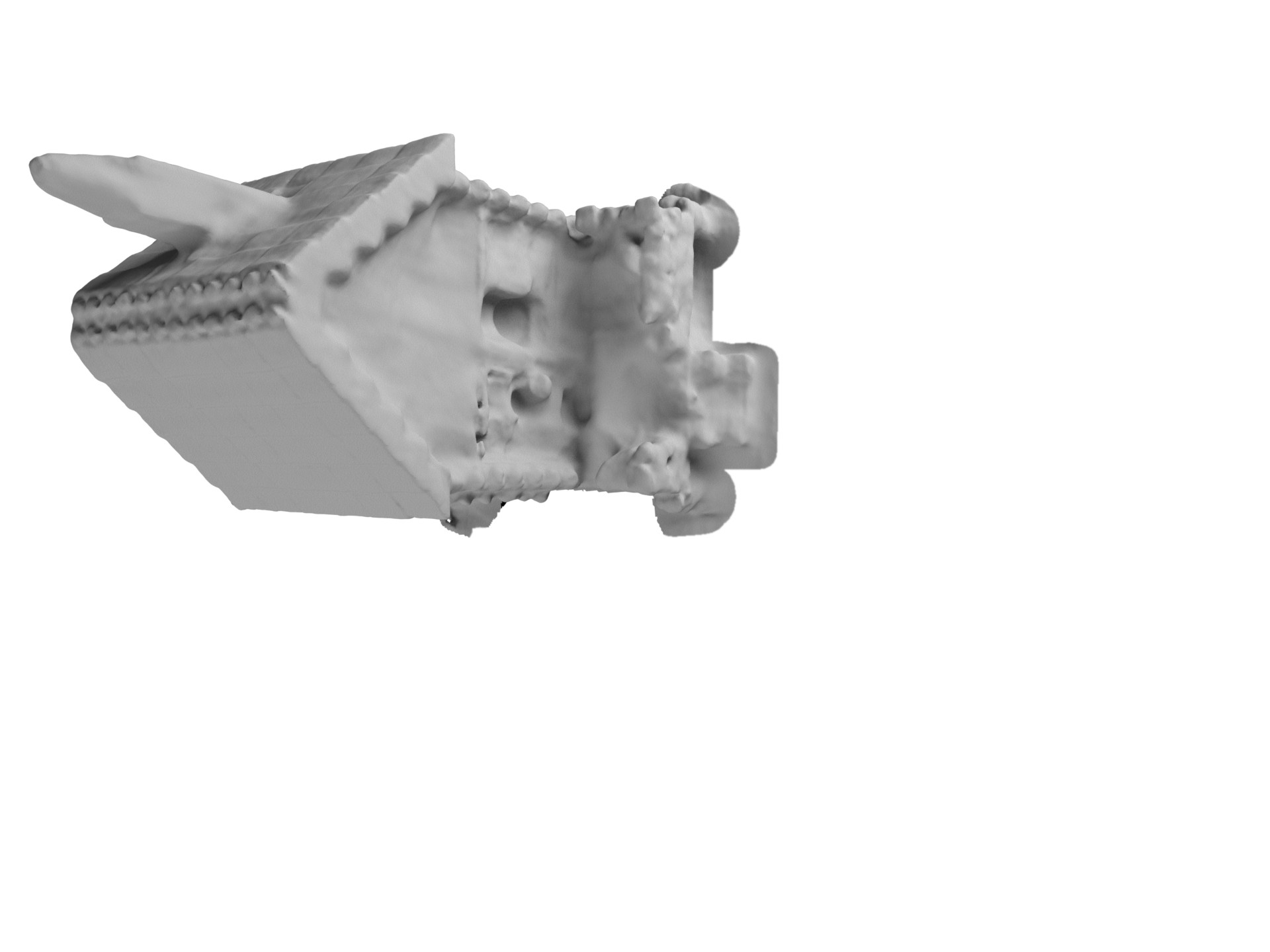}
\end{minipage}
\begin{minipage}[b]{0.245\linewidth}
\centering
\includegraphics[width=1.0\linewidth]{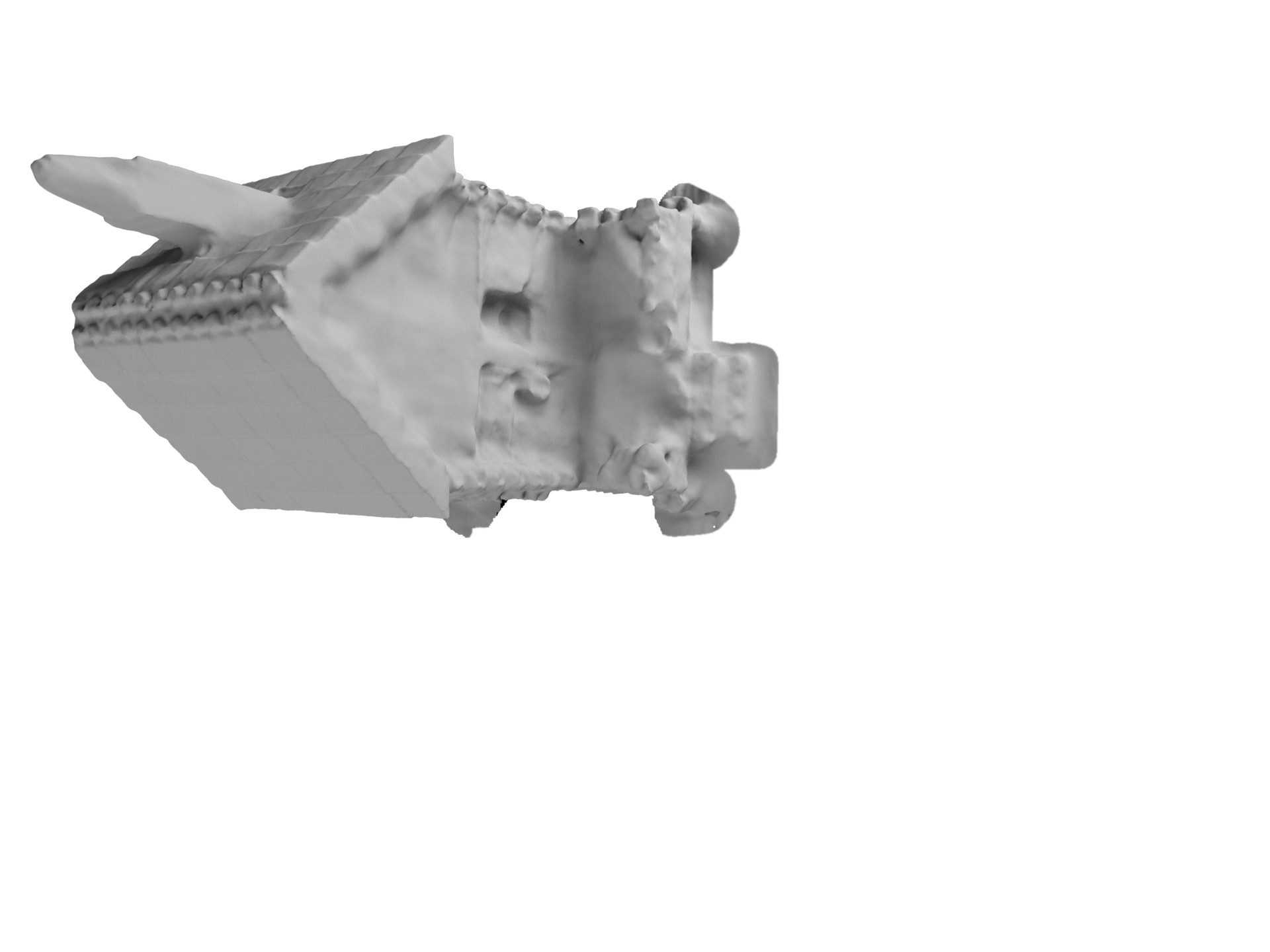}
\end{minipage}
\begin{minipage}[b]{0.245\linewidth}
\centering
\includegraphics[width=1.0\linewidth]{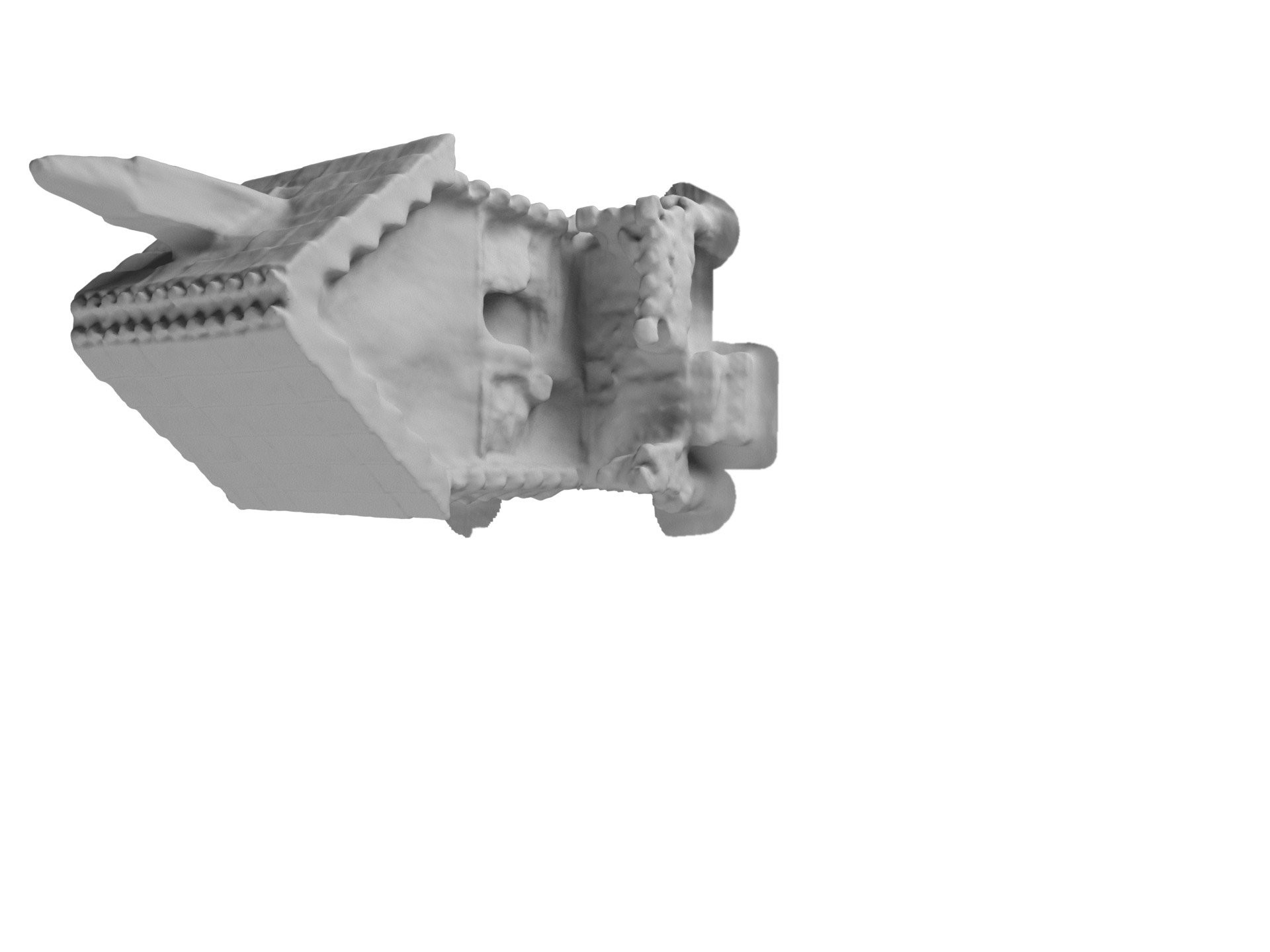}
\end{minipage}

\begin{minipage}[b]{0.245\linewidth}
\centering
\includegraphics[width=1.0\linewidth]{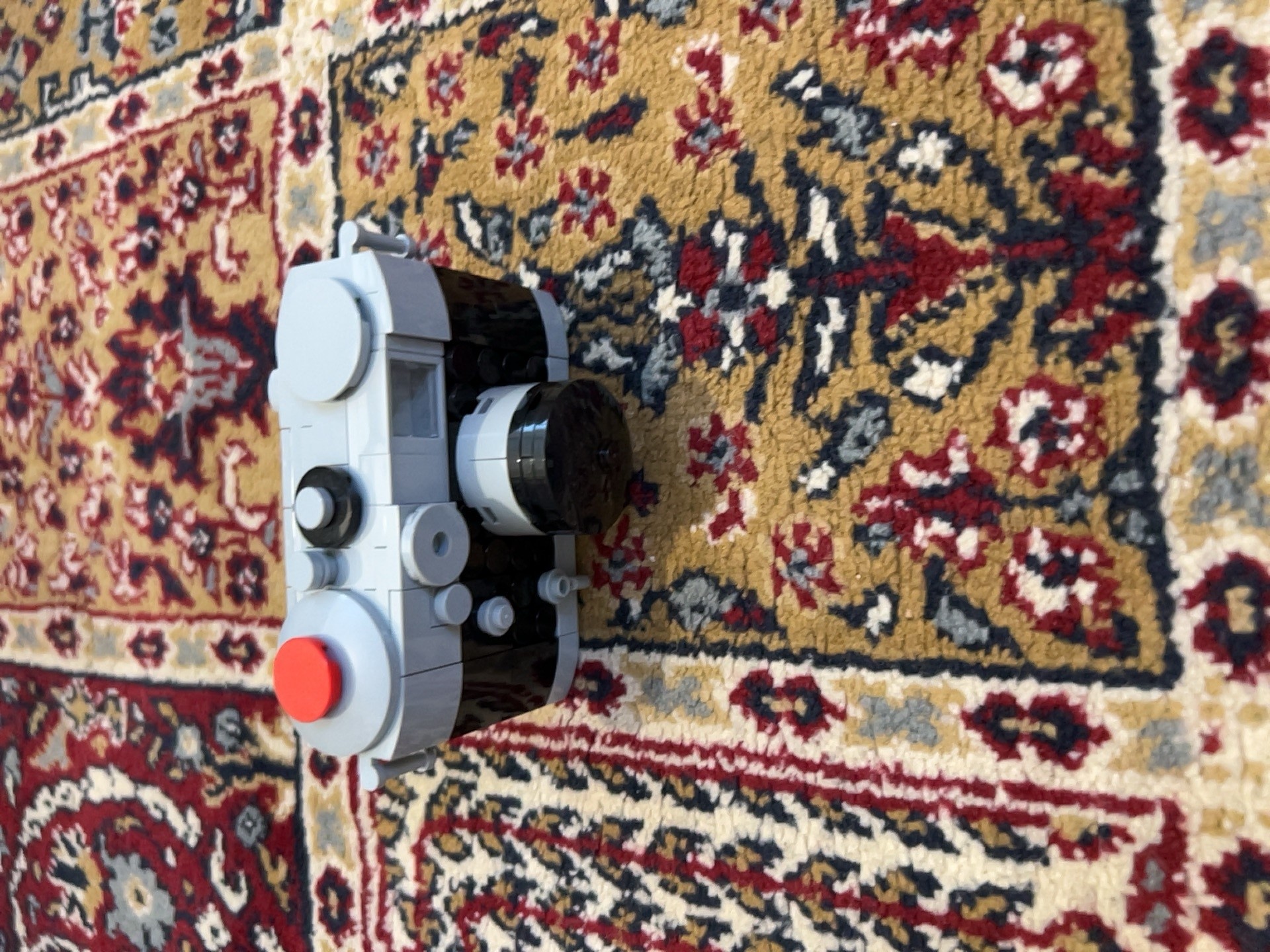}
\end{minipage}
\begin{minipage}[b]{0.245\linewidth}
\centering
\includegraphics[width=1.0\linewidth]{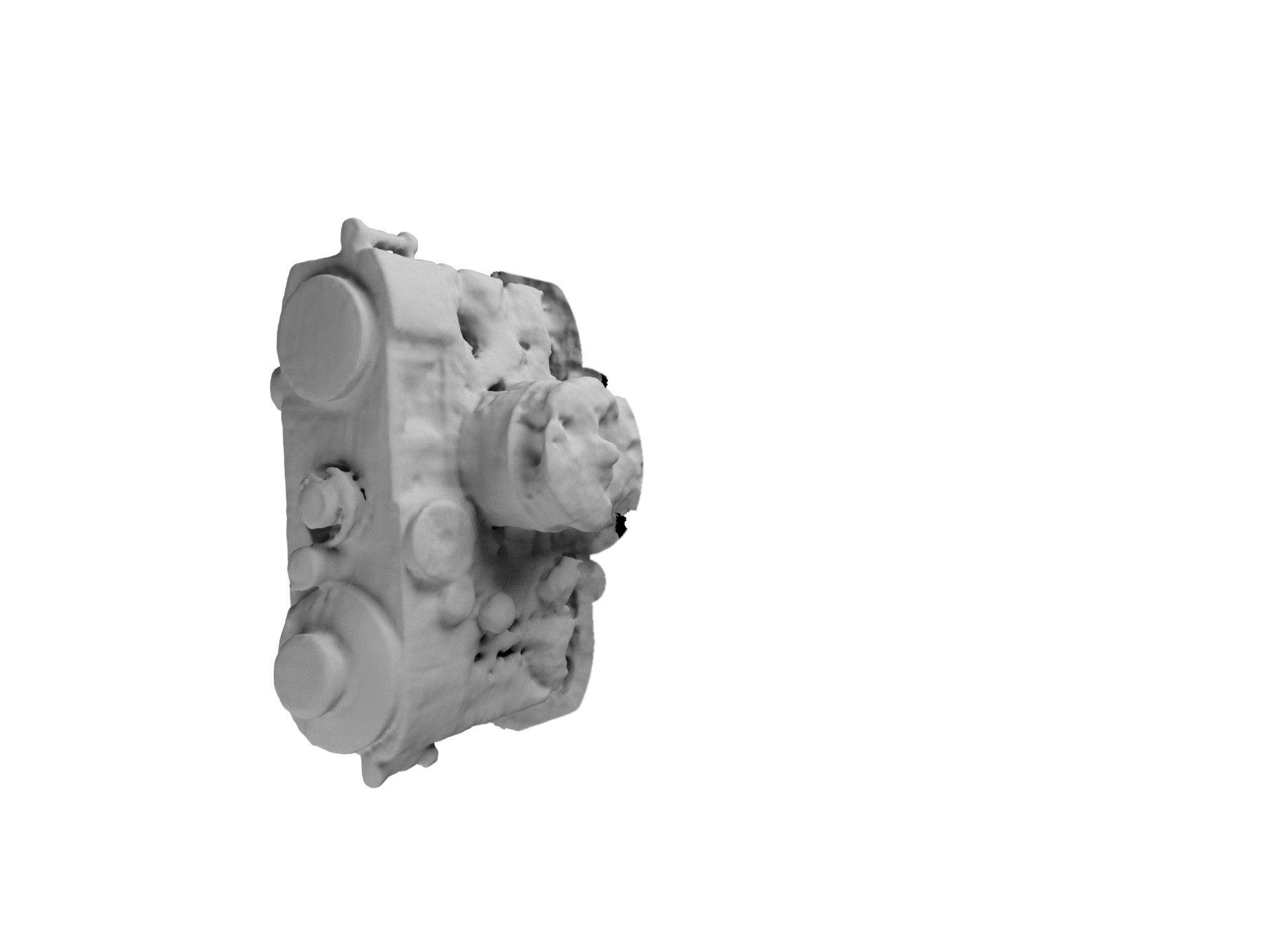}
\end{minipage}
\begin{minipage}[b]{0.245\linewidth}
\centering
\includegraphics[width=1.0\linewidth]{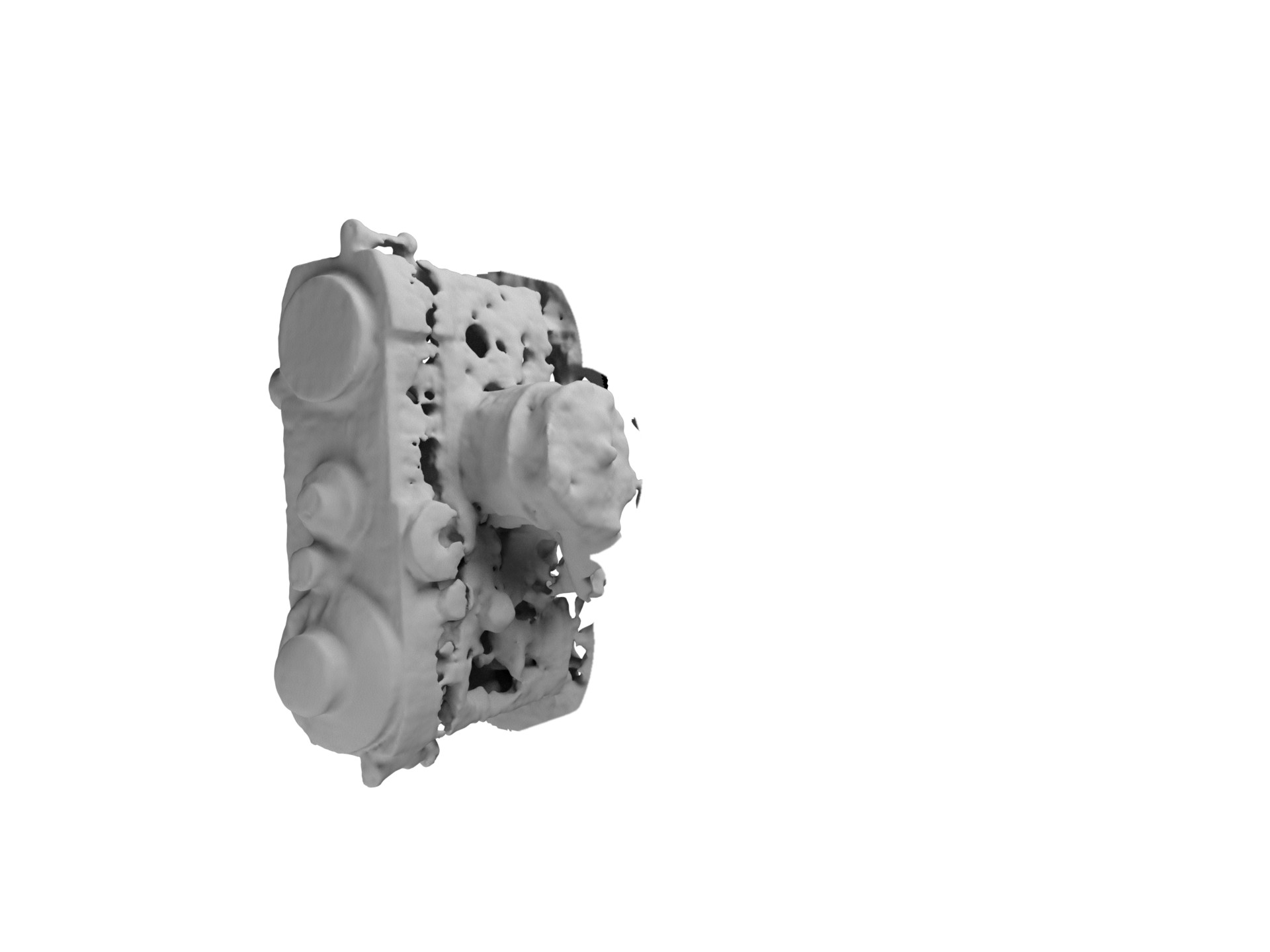}
\end{minipage}
\begin{minipage}[b]{0.245\linewidth}
\centering
\includegraphics[width=1.0\linewidth]{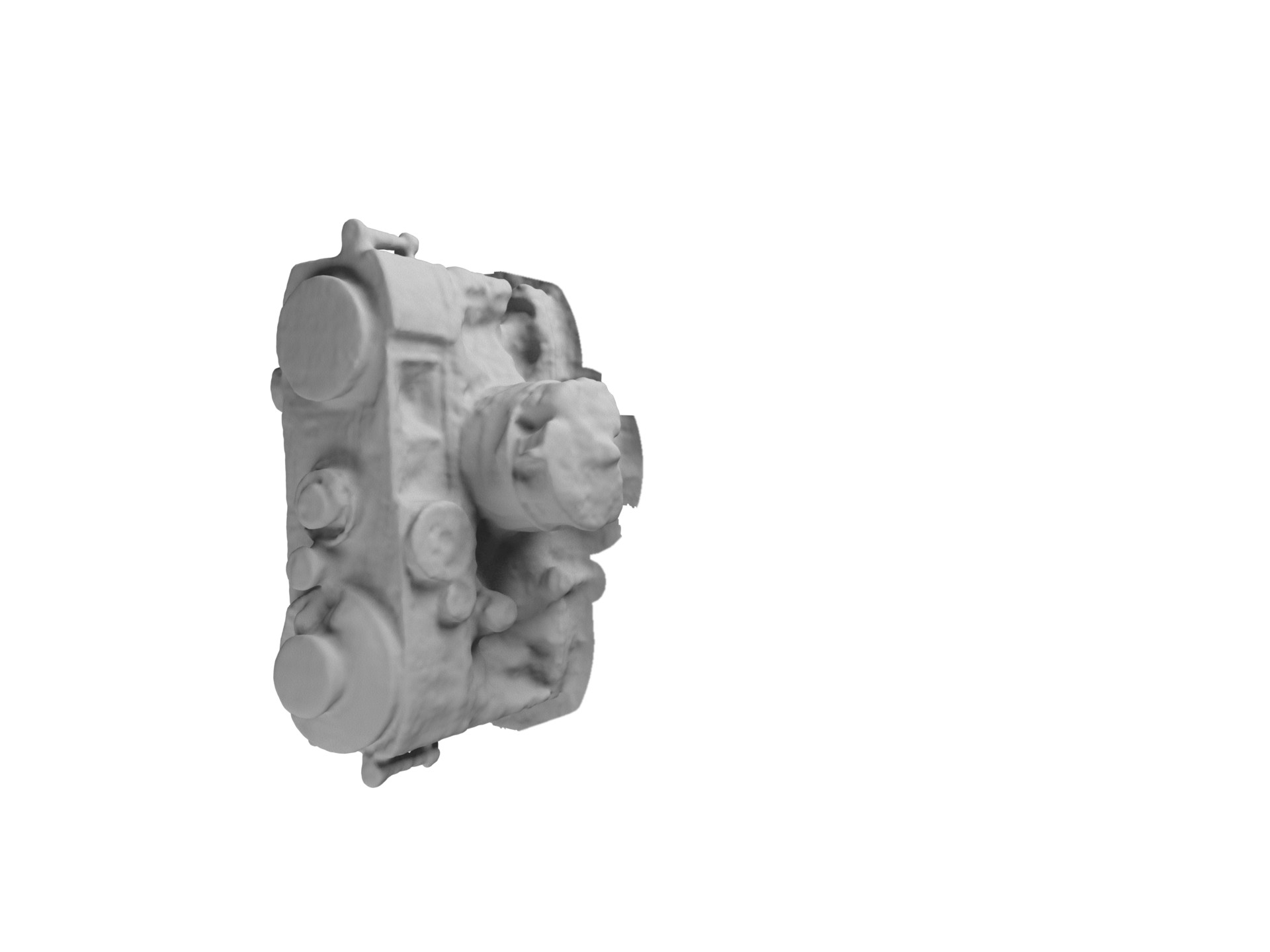}
\end{minipage}

\begin{minipage}[b]{0.245\linewidth}
\centering
\includegraphics[width=1.0\linewidth]{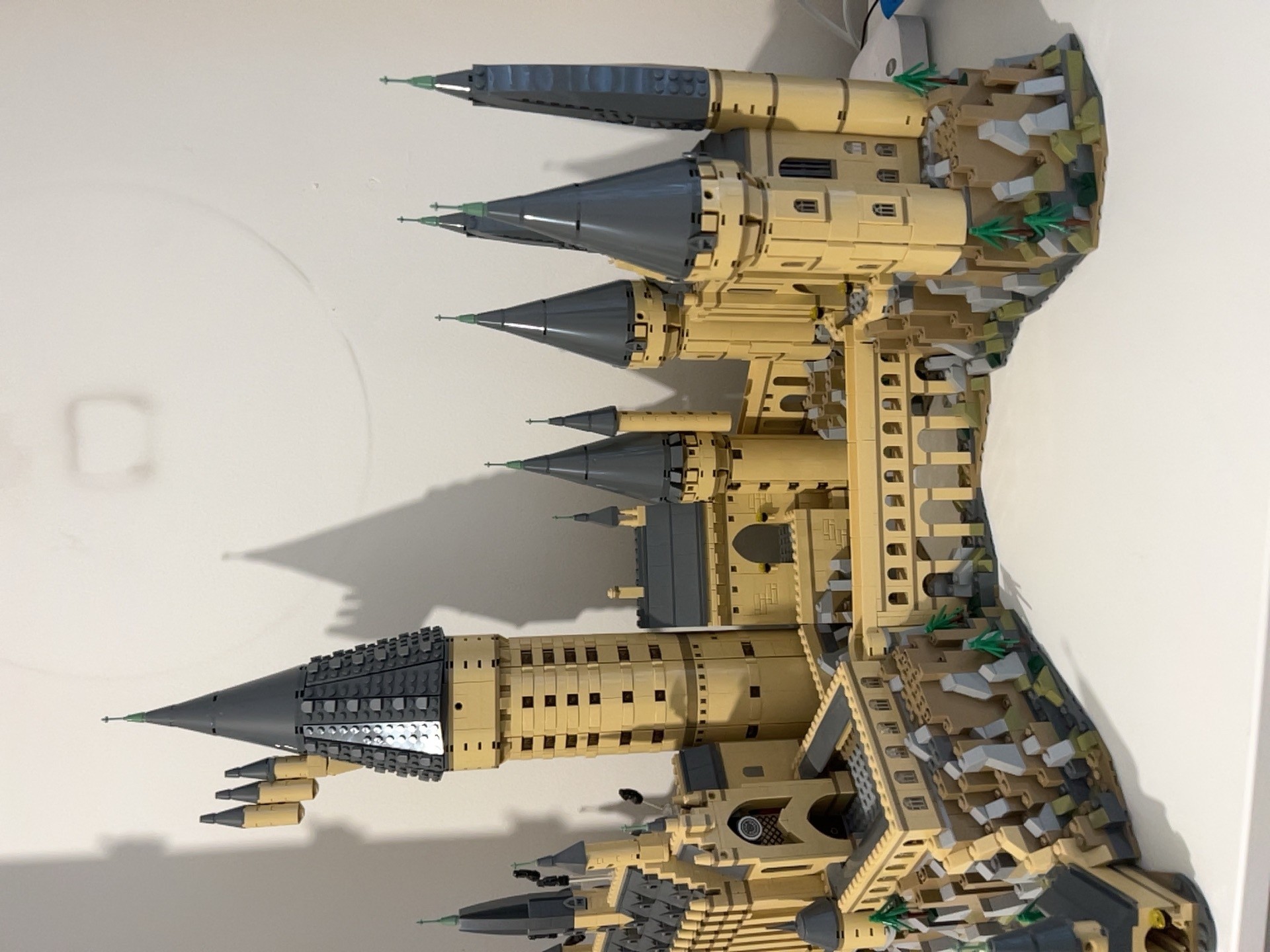}
\end{minipage}
\begin{minipage}[b]{0.245\linewidth}
\centering
\includegraphics[width=1.0\linewidth]{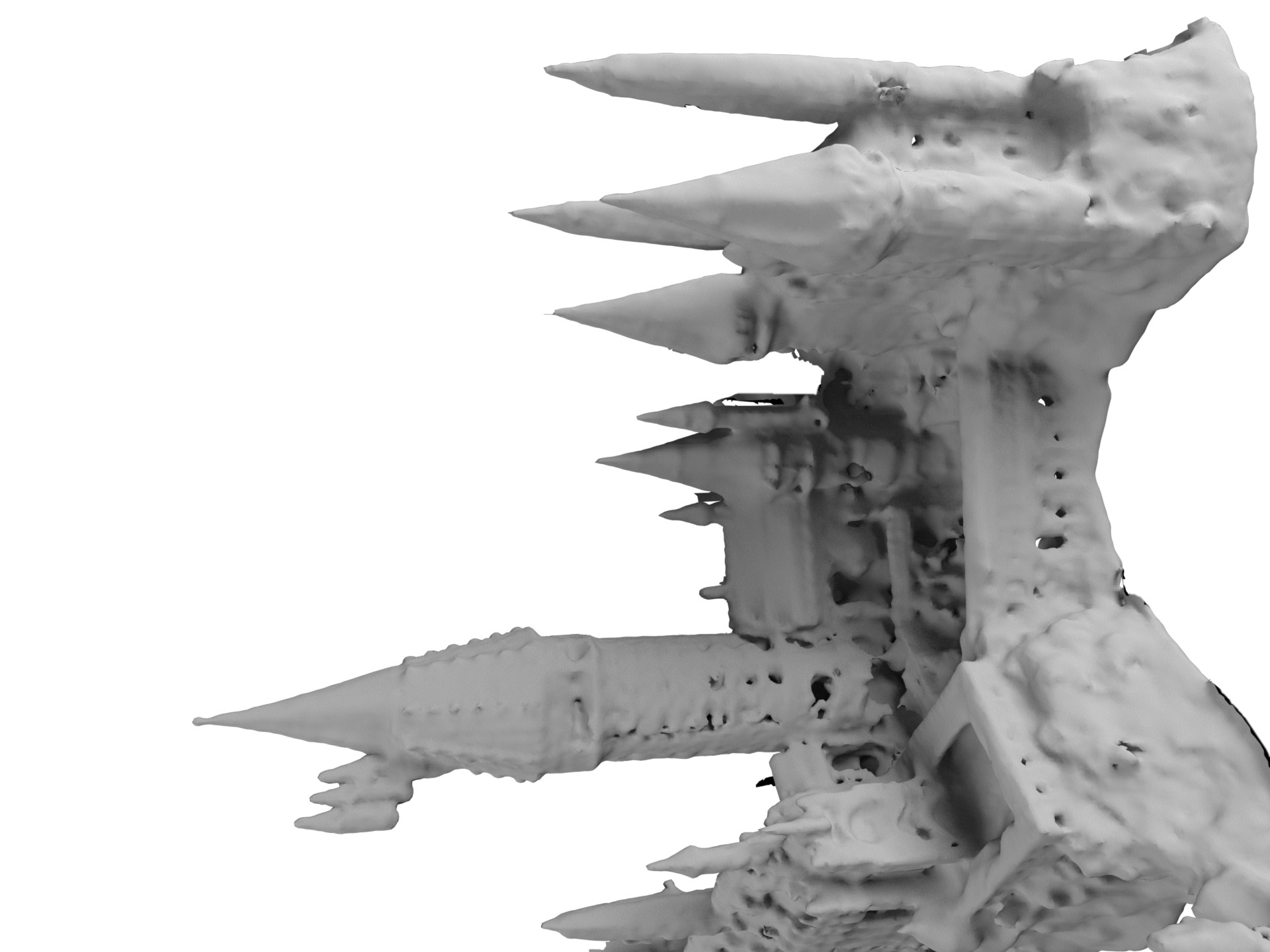}
\end{minipage}
\begin{minipage}[b]{0.245\linewidth}
\centering
\includegraphics[width=1.0\linewidth]{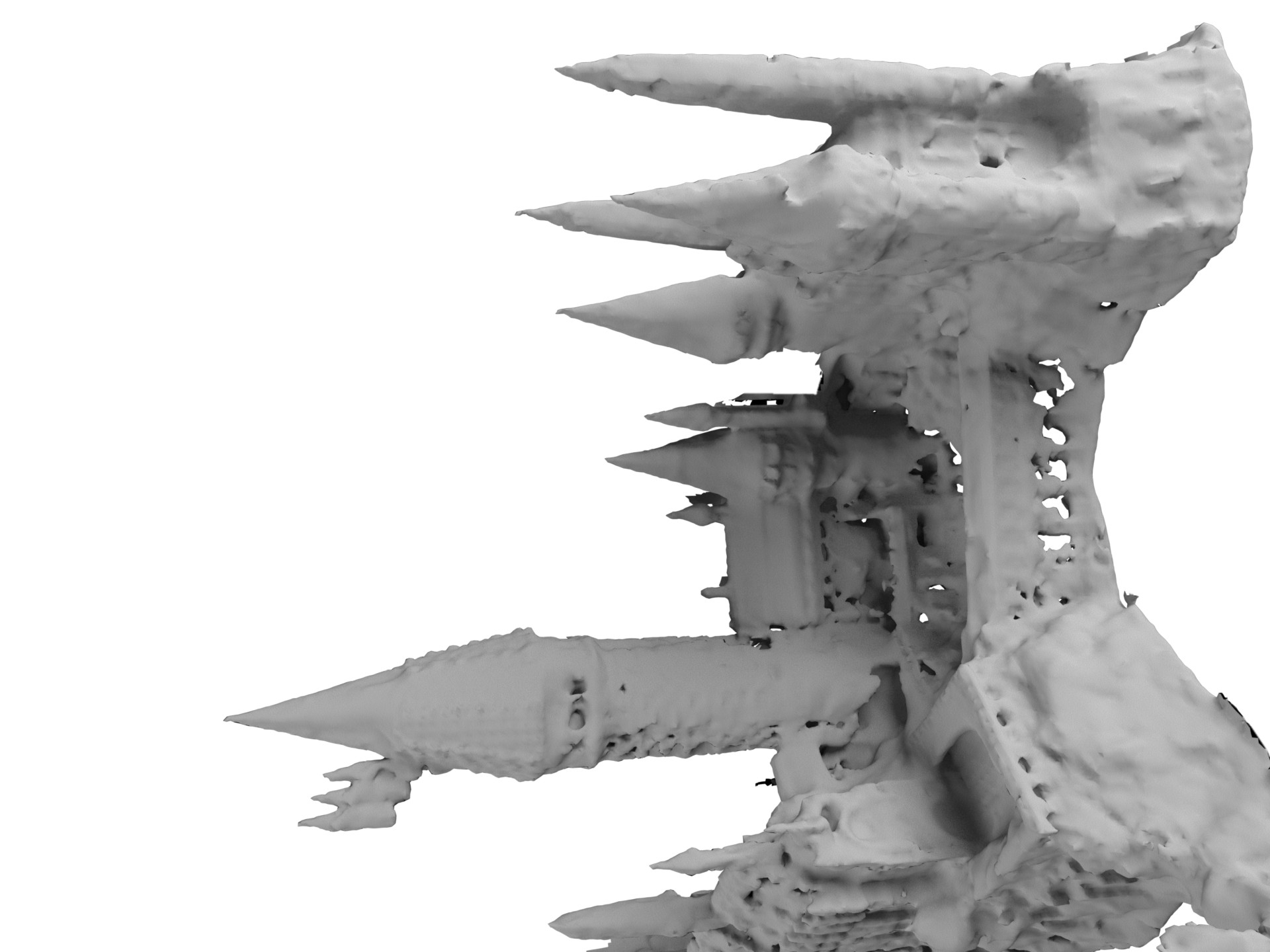}
\end{minipage}
\begin{minipage}[b]{0.245\linewidth}
\centering
\includegraphics[width=1.0\linewidth]{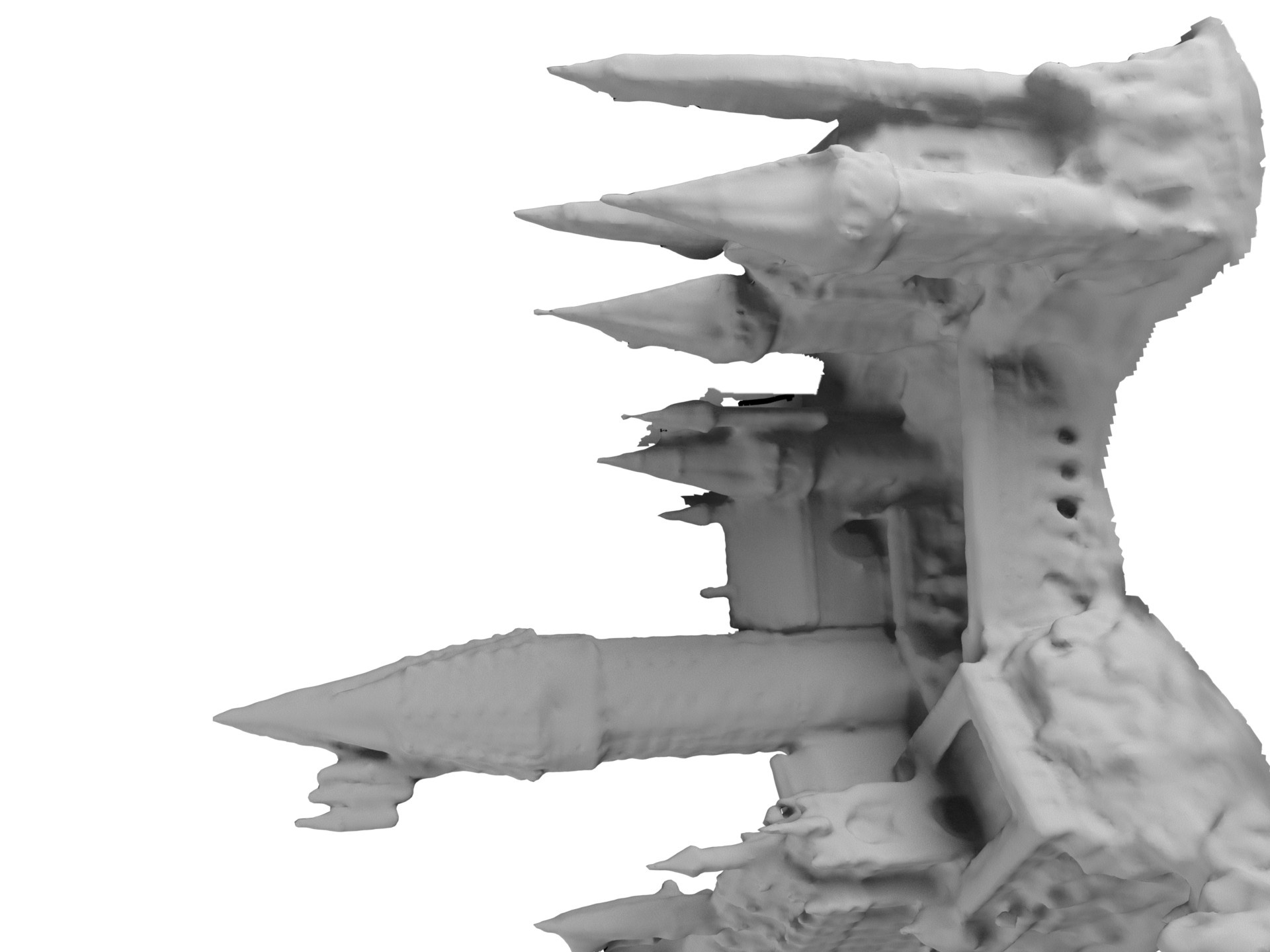}
\end{minipage}

\begin{minipage}[b]{0.245\linewidth}
\centering
\includegraphics[width=1.0\linewidth]{images/lego/colosseum/gt/000010.jpg}
\end{minipage}
\begin{minipage}[b]{0.245\linewidth}
\centering
\includegraphics[width=1.0\linewidth]{images/lego/colosseum/neus-facto-base/000010.jpg}
\end{minipage}
\begin{minipage}[b]{0.245\linewidth}
\centering
\includegraphics[width=1.0\linewidth]{images/lego/colosseum/oav-facto-base/000010.jpg}
\end{minipage}
\begin{minipage}[b]{0.245\linewidth}
\centering
\includegraphics[width=1.0\linewidth]{images/lego/colosseum/ssdp-facto-base/000010.jpg}
\end{minipage}

\begin{minipage}[b]{0.245\linewidth}
\centering
\includegraphics[width=1.0\linewidth]{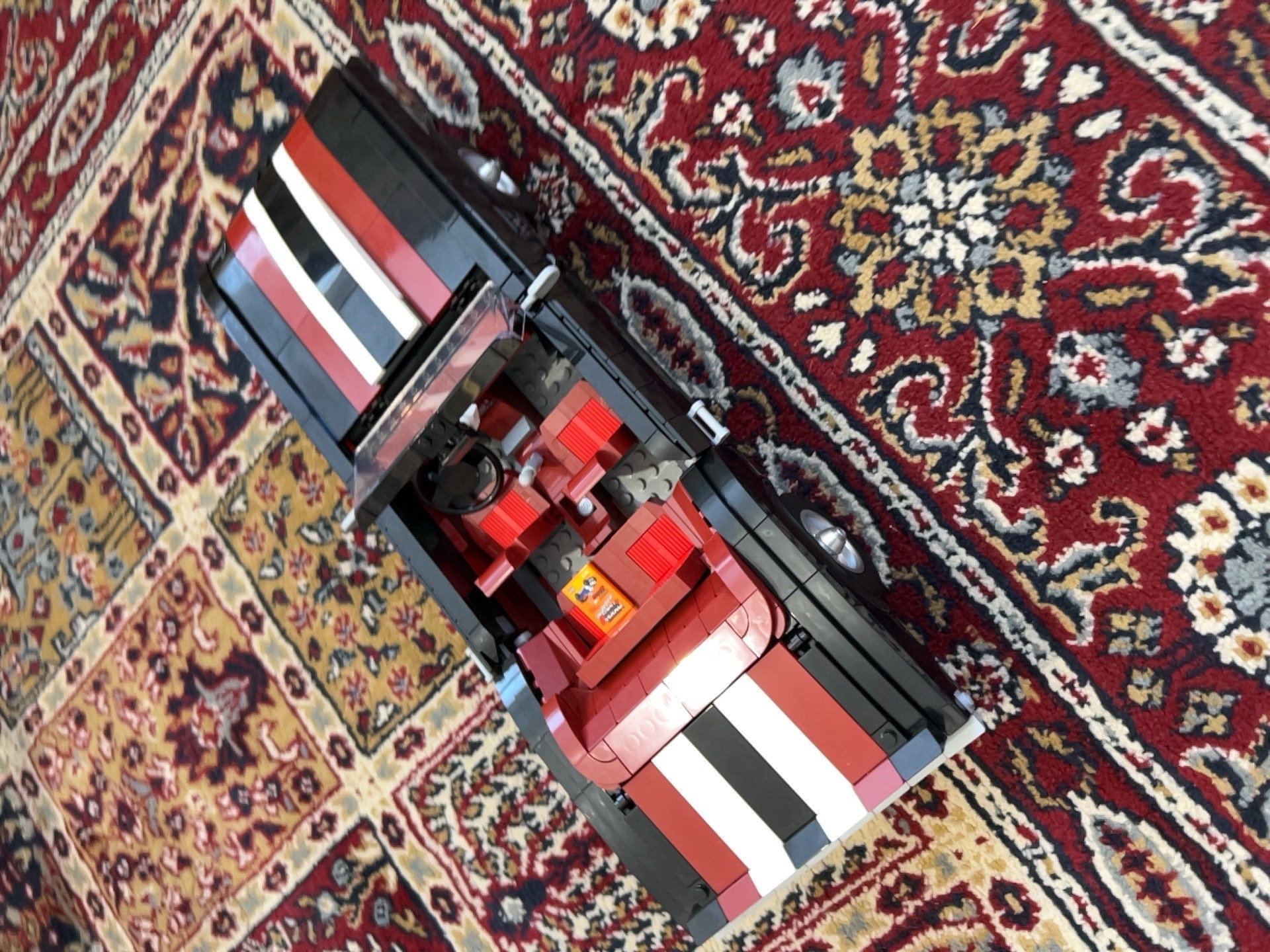}
\end{minipage}
\begin{minipage}[b]{0.245\linewidth}
\centering
\includegraphics[width=1.0\linewidth]{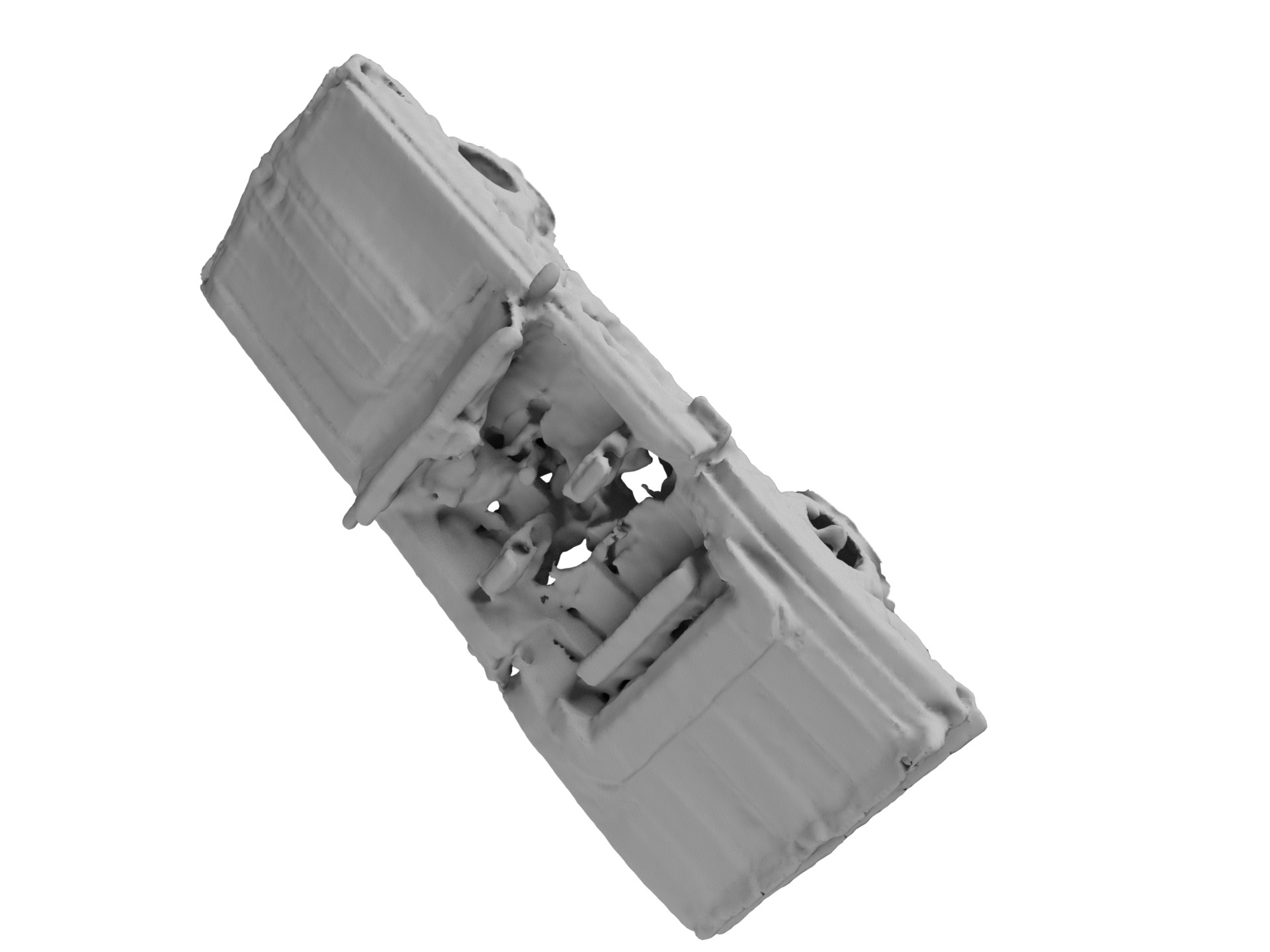}
\end{minipage}
\begin{minipage}[b]{0.245\linewidth}
\centering
\includegraphics[width=1.0\linewidth]{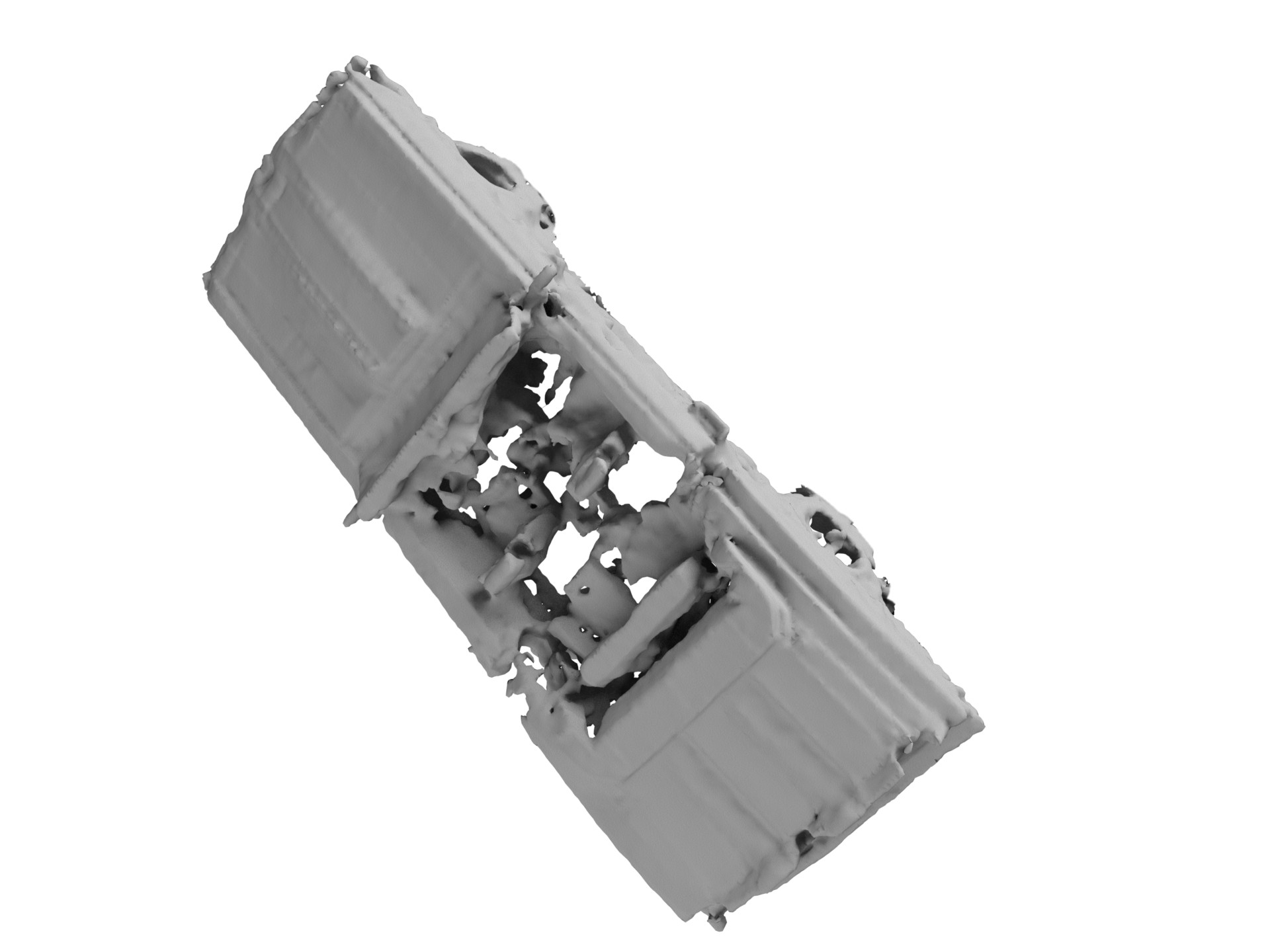}
\end{minipage}
\begin{minipage}[b]{0.245\linewidth}
\centering
\includegraphics[width=1.0\linewidth]{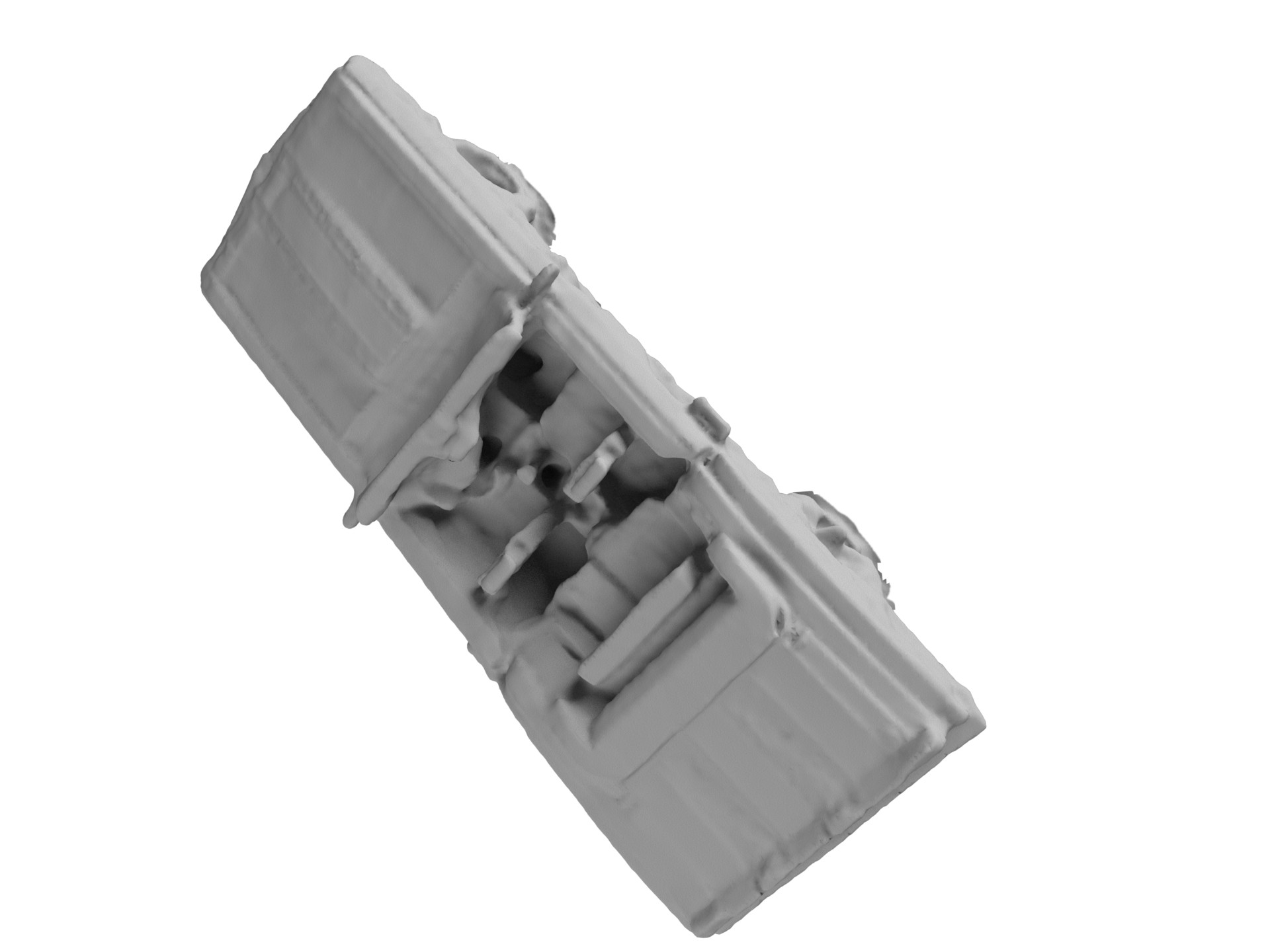}
\end{minipage}

\begin{minipage}[b]{0.245\linewidth}
\centering
\includegraphics[width=1.0\linewidth]{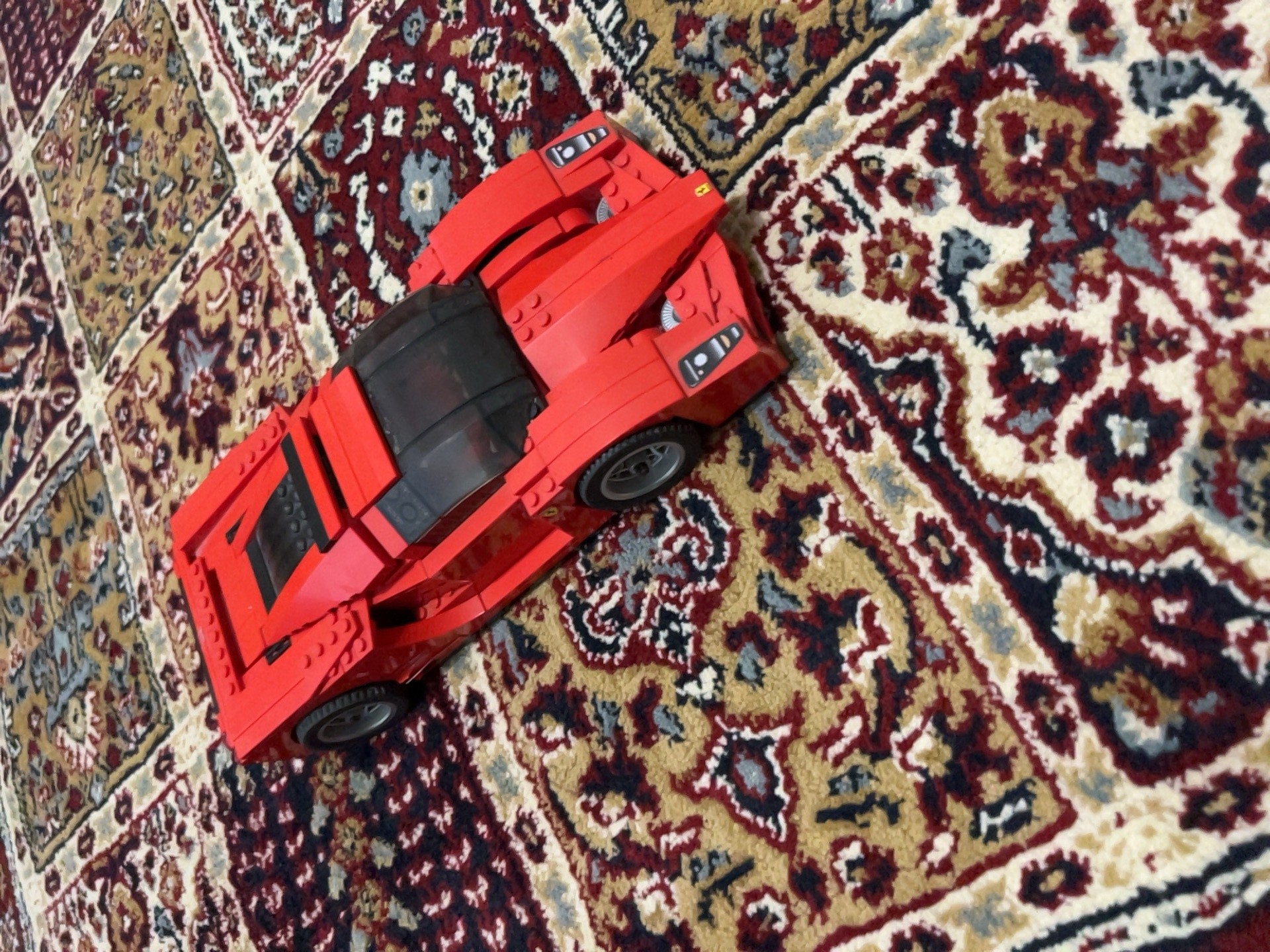}
\subcaption{Ground Truth}
\end{minipage}
\begin{minipage}[b]{0.245\linewidth}
\centering
\includegraphics[width=1.0\linewidth]{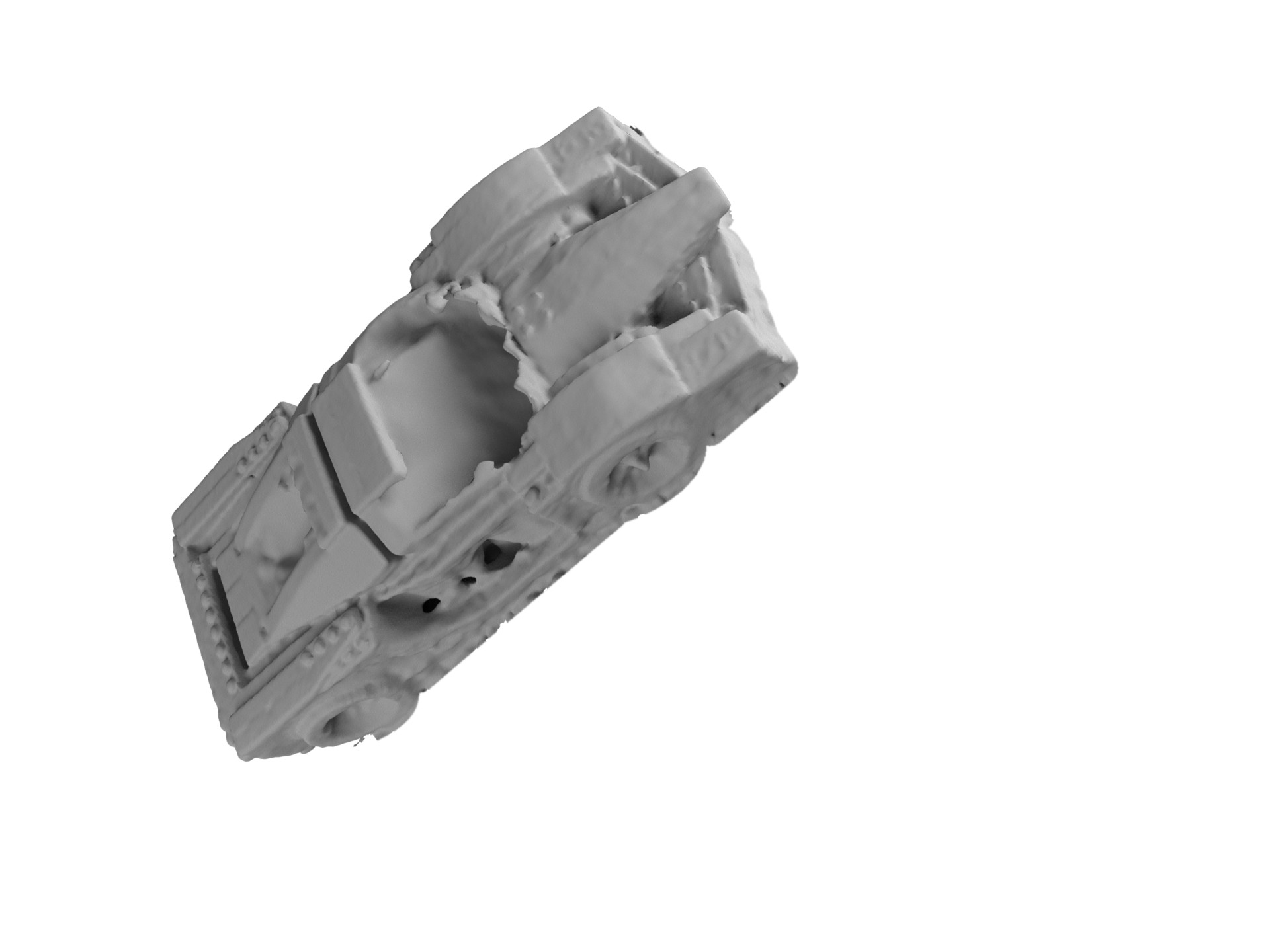}
\subcaption{NeuS-Facto}
\end{minipage}
\begin{minipage}[b]{0.245\linewidth}
\centering
\includegraphics[width=1.0\linewidth]{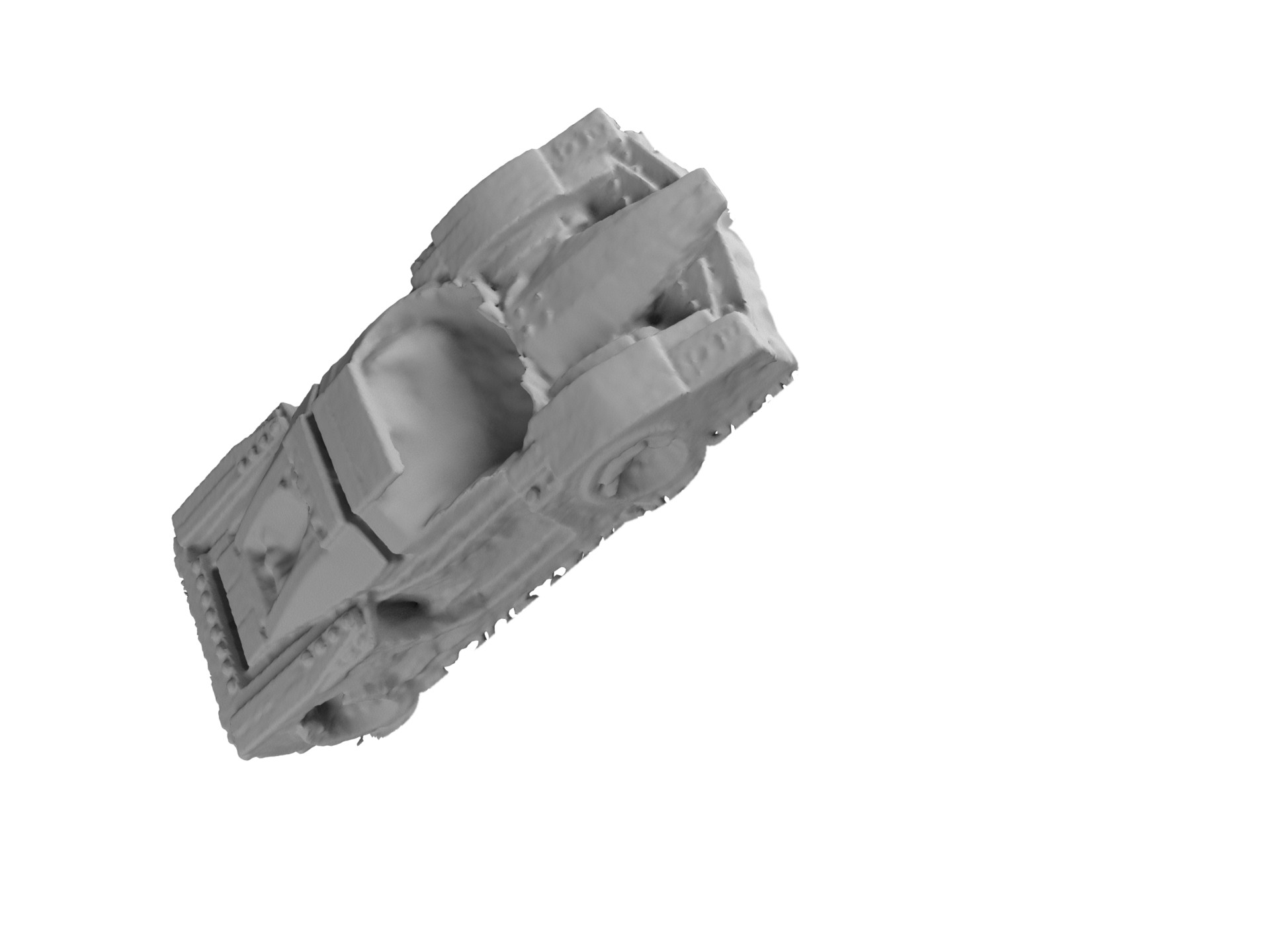}
\subcaption{OaV-Facto}
\end{minipage}
\begin{minipage}[b]{0.245\linewidth}
\centering
\includegraphics[width=1.0\linewidth]{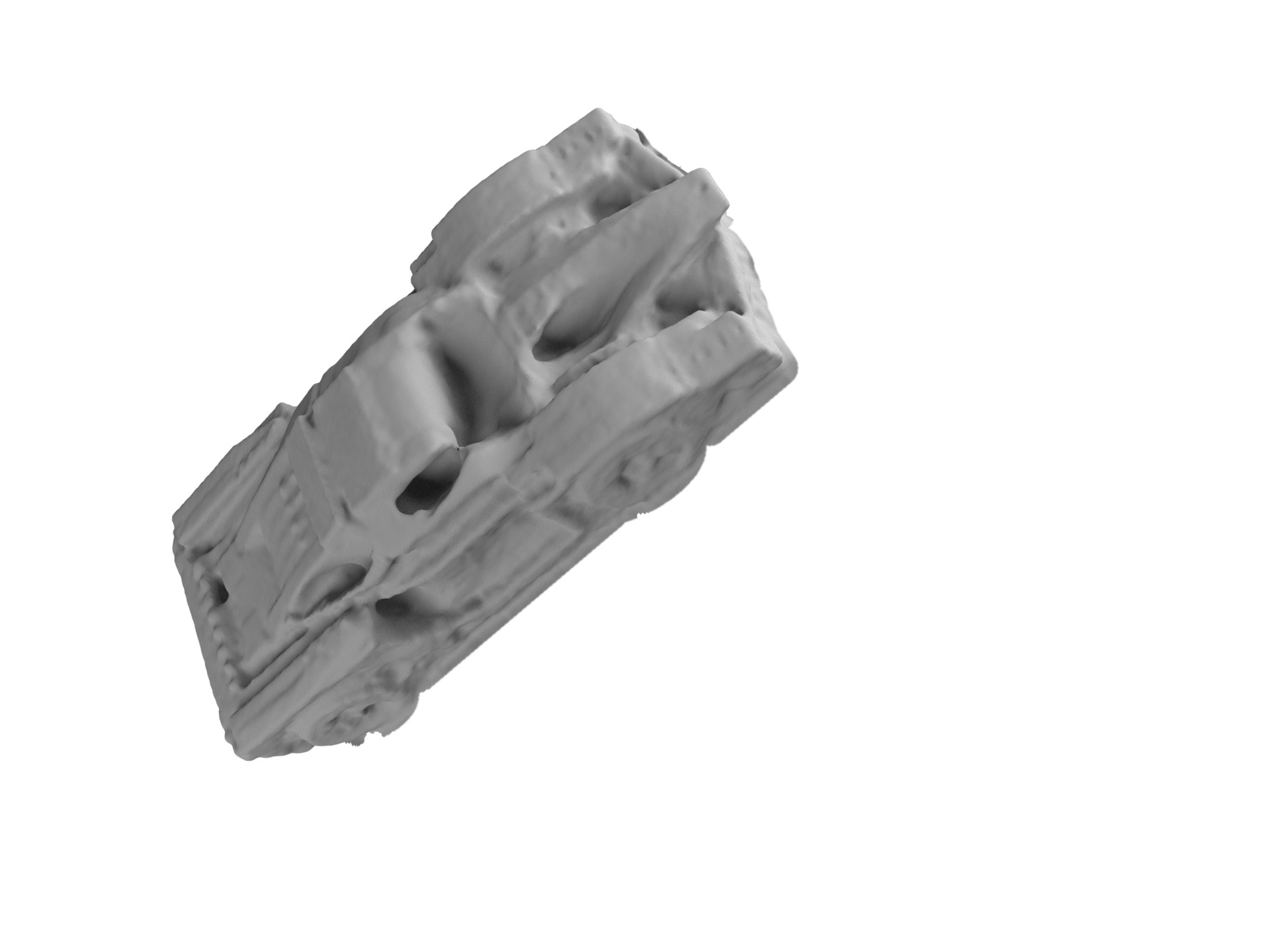}
\subcaption{SSDP-Facto}
\end{minipage}

\caption{Visualization examples on the MobileBrick dataset.}
\end{figure}

\begin{figure}

\ContinuedFloat
\setcounter{subfigure}{0}

\centering

\begin{minipage}[b]{0.245\linewidth}
\centering
\includegraphics[width=1.0\linewidth]{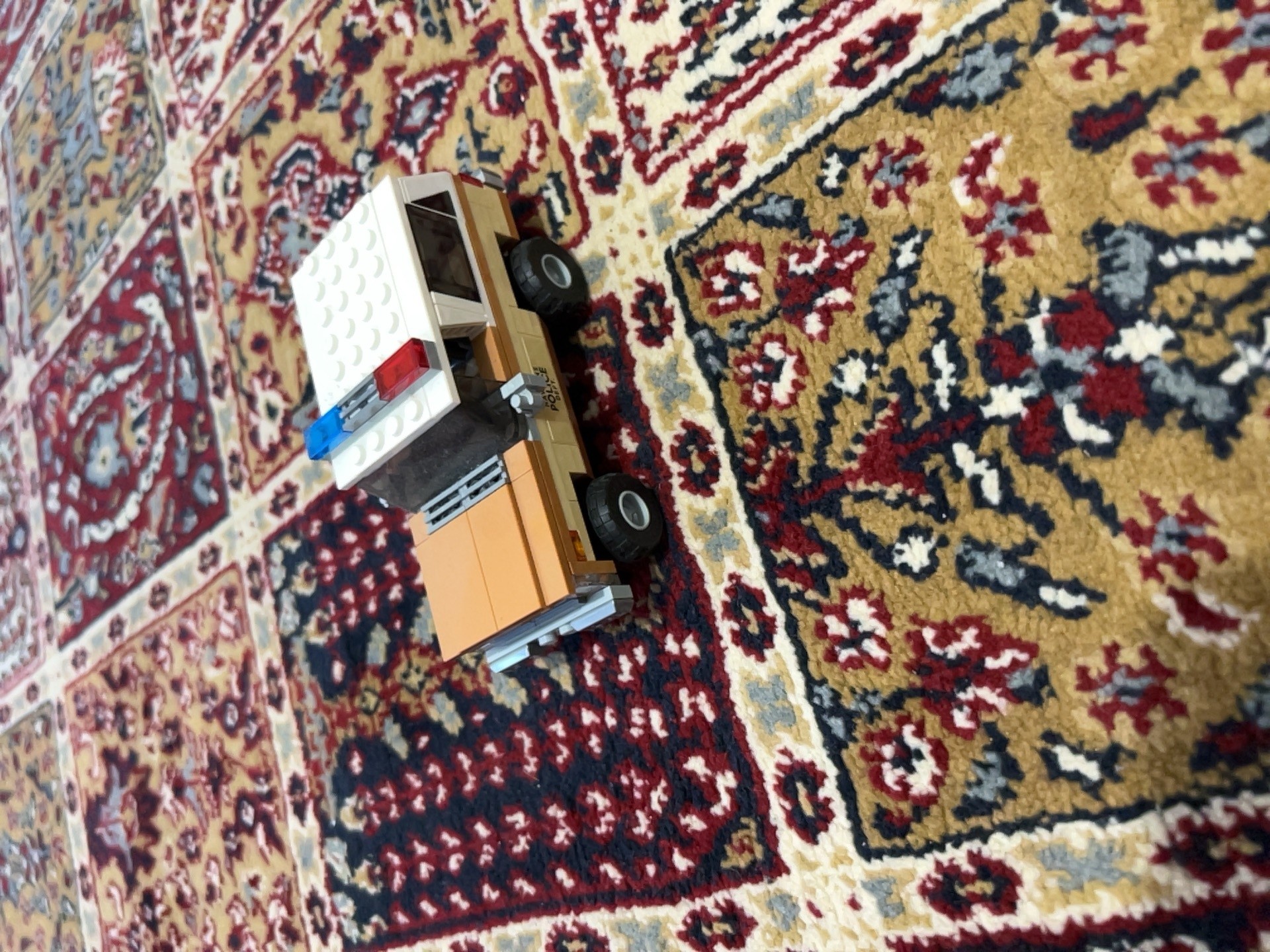}
\end{minipage}
\begin{minipage}[b]{0.245\linewidth}
\centering
\includegraphics[width=1.0\linewidth]{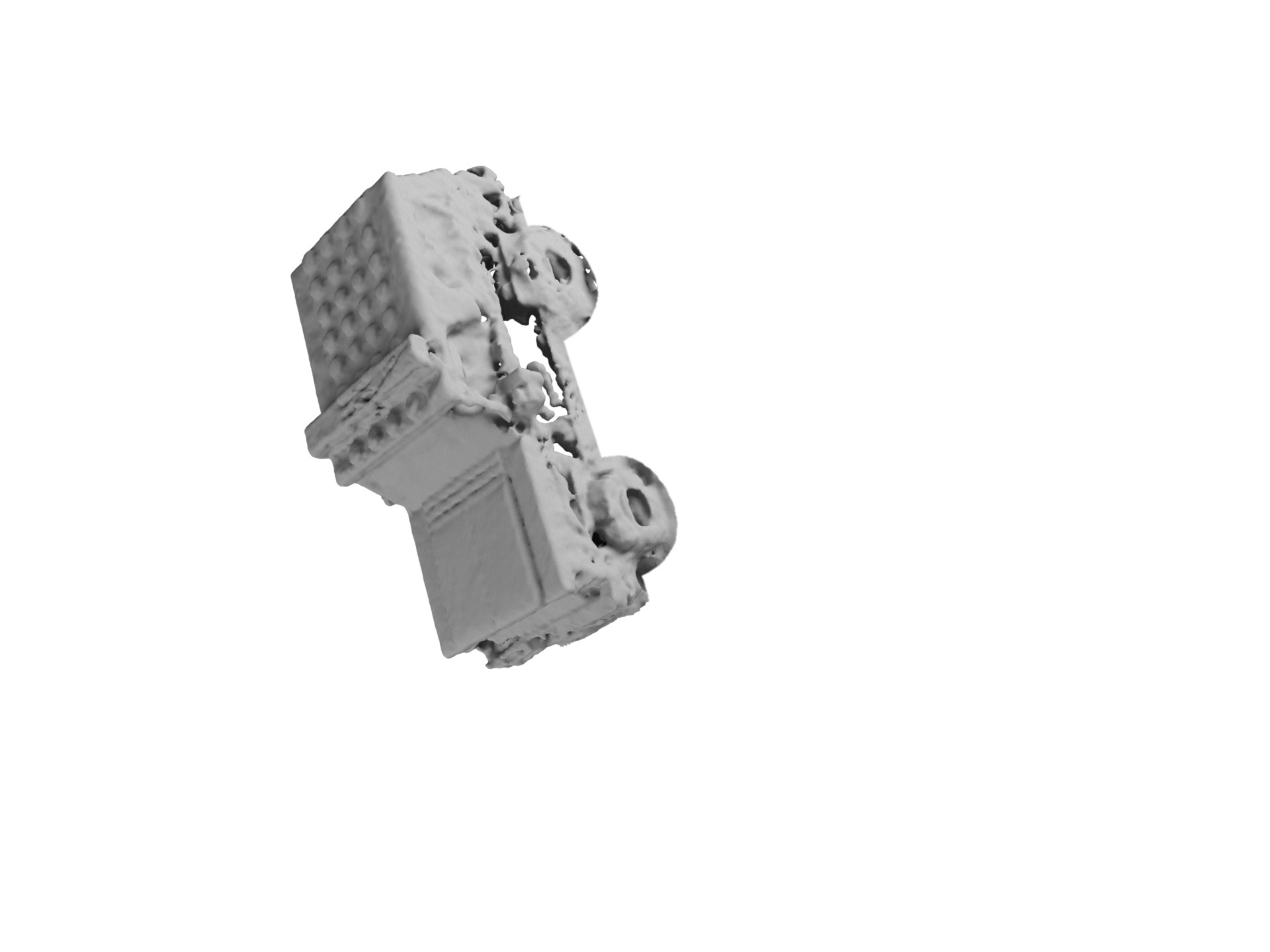}
\end{minipage}
\begin{minipage}[b]{0.245\linewidth}
\centering
\includegraphics[width=1.0\linewidth]{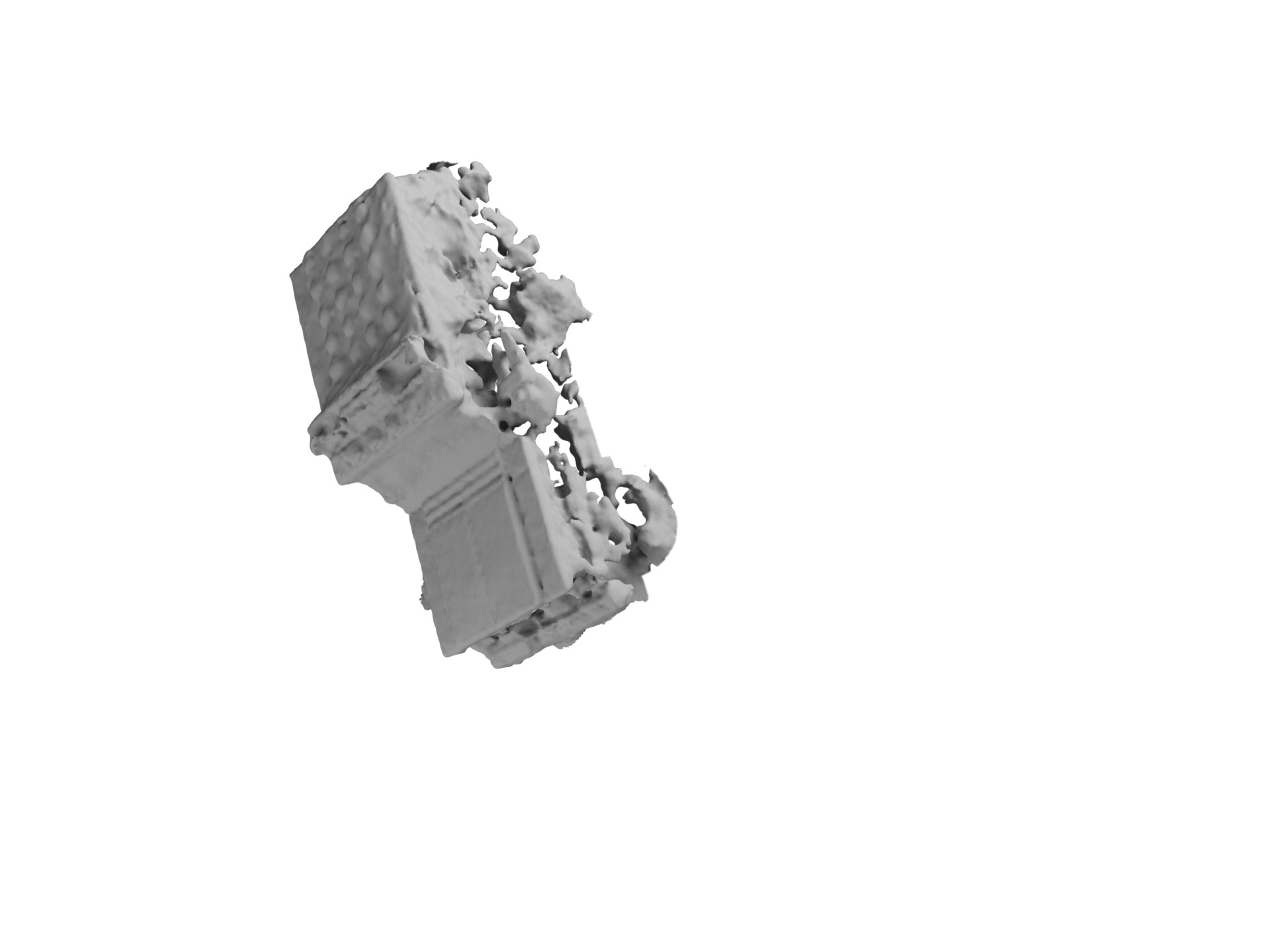}
\end{minipage}
\begin{minipage}[b]{0.245\linewidth}
\centering
\includegraphics[width=1.0\linewidth]{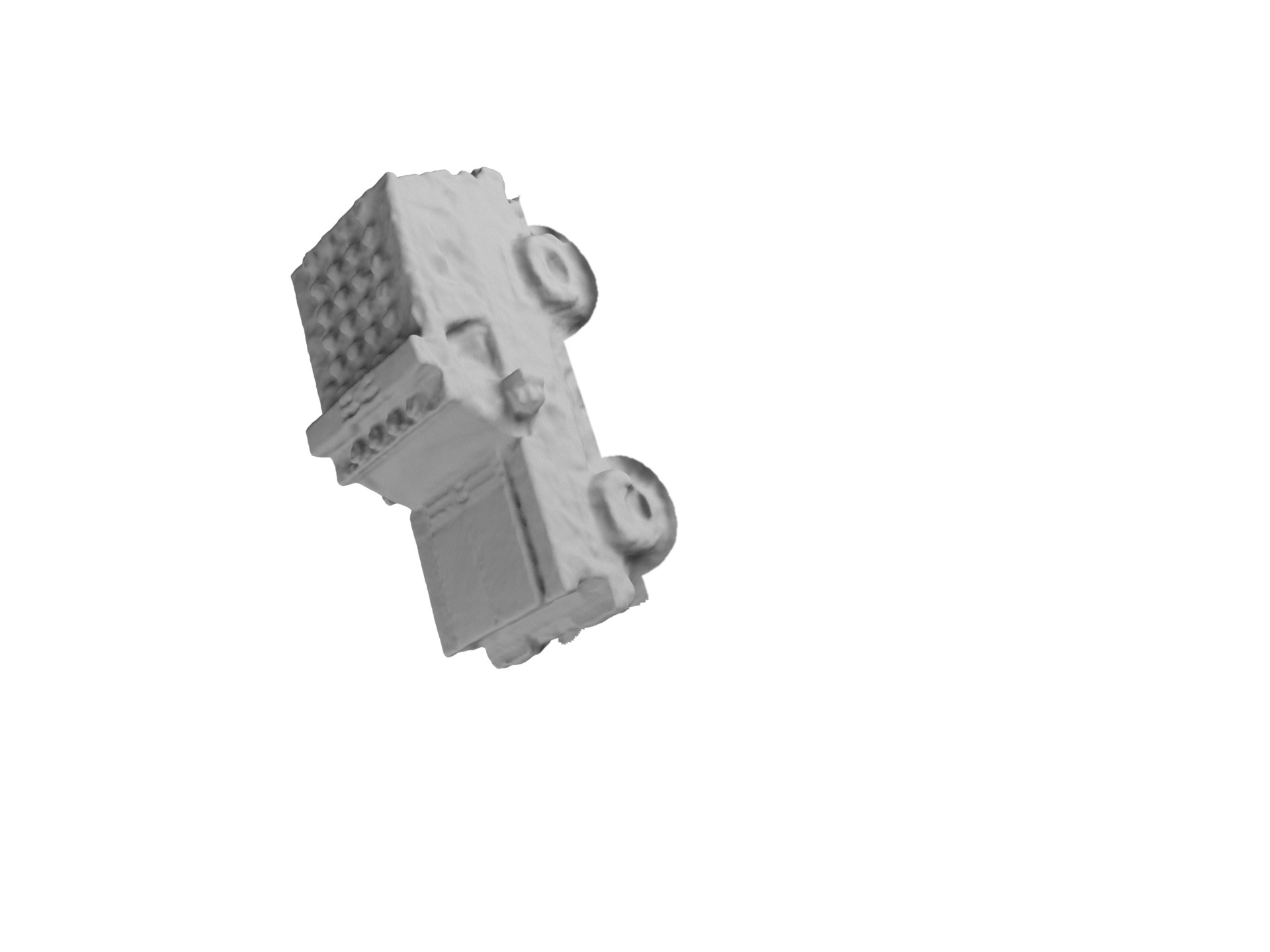}
\end{minipage}

\begin{minipage}[b]{0.245\linewidth}
\centering
\includegraphics[width=1.0\linewidth]{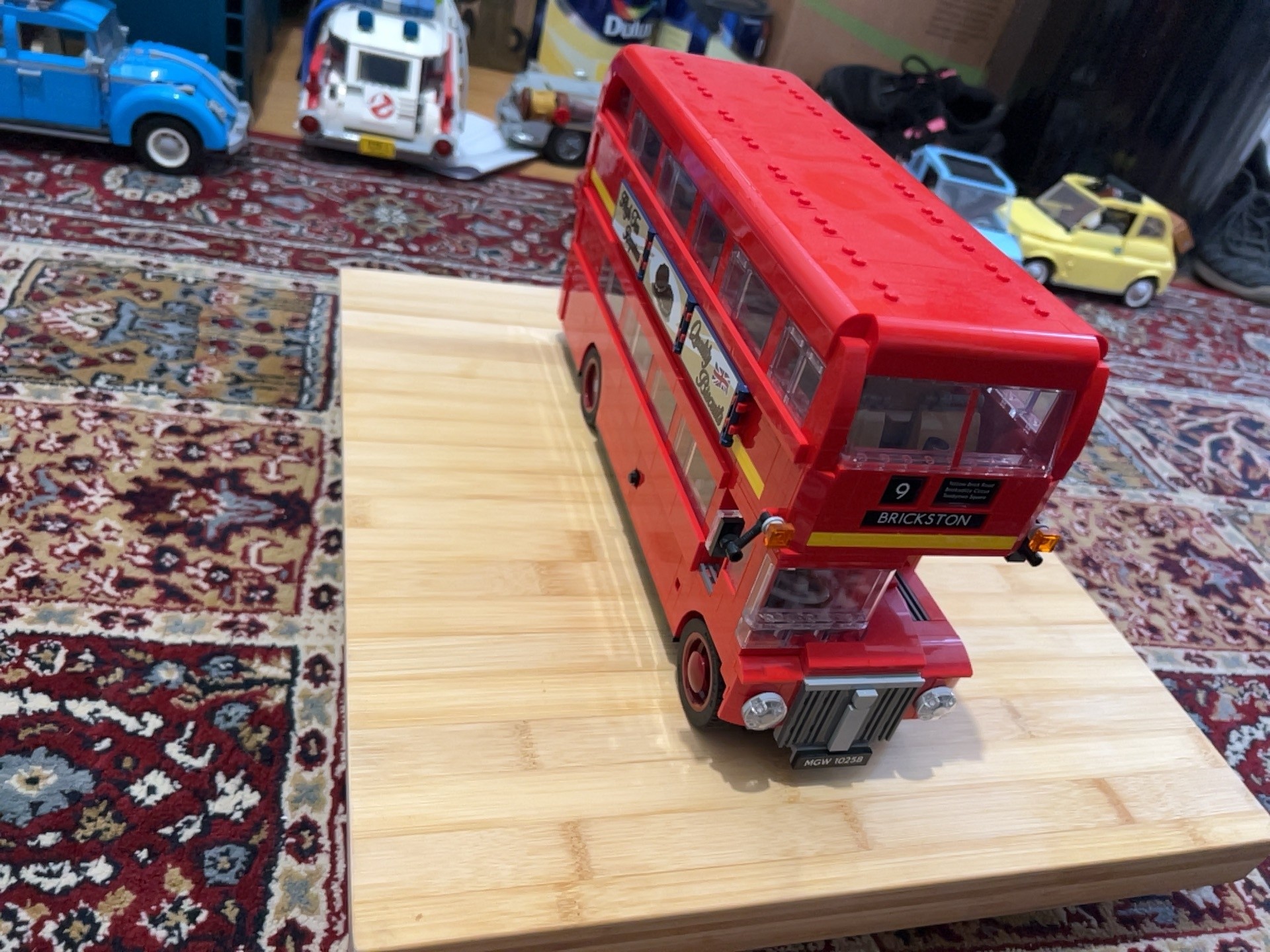}
\end{minipage}
\begin{minipage}[b]{0.245\linewidth}
\centering
\includegraphics[width=1.0\linewidth]{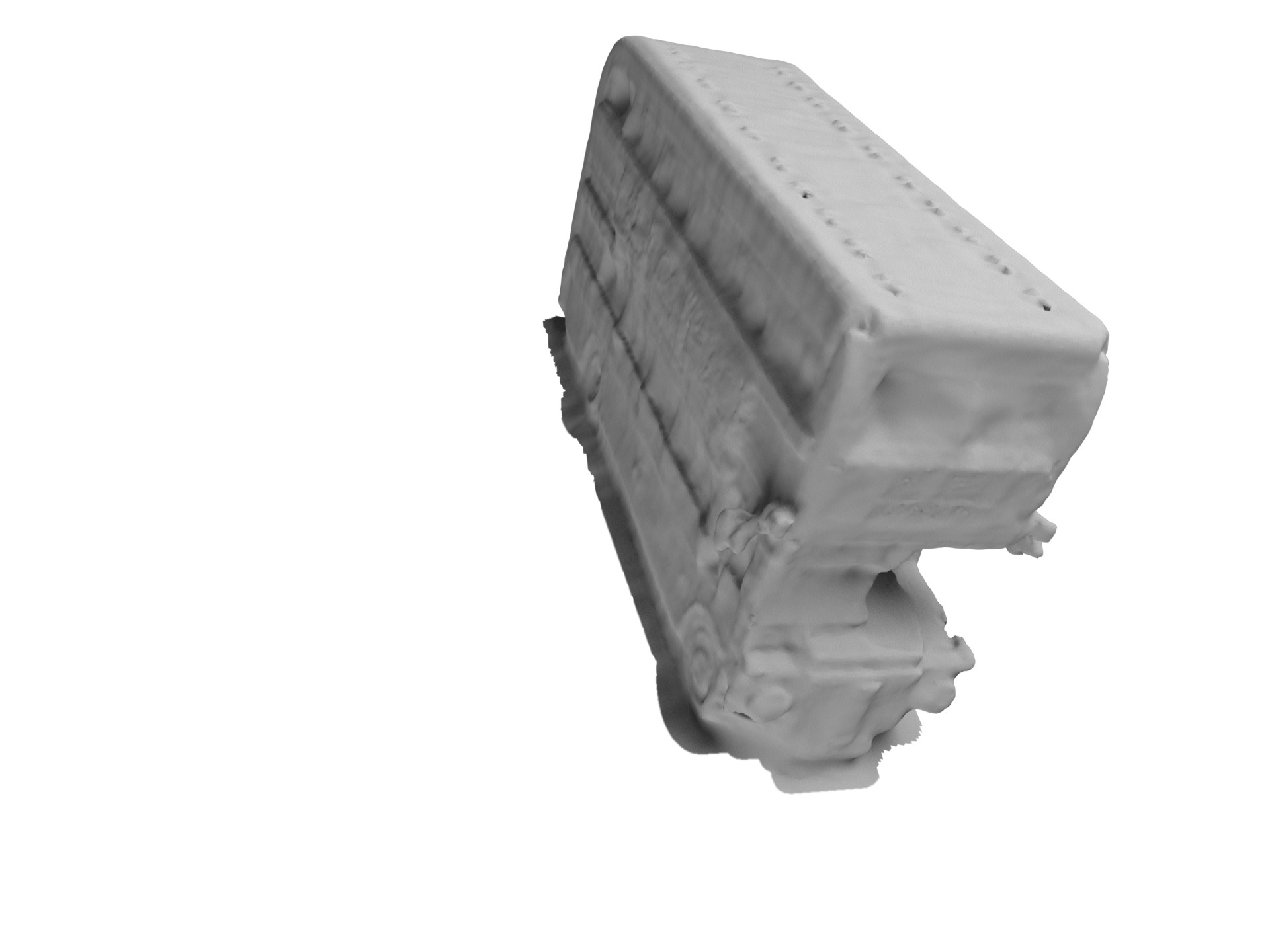}
\end{minipage}
\begin{minipage}[b]{0.245\linewidth}
\centering
\includegraphics[width=1.0\linewidth]{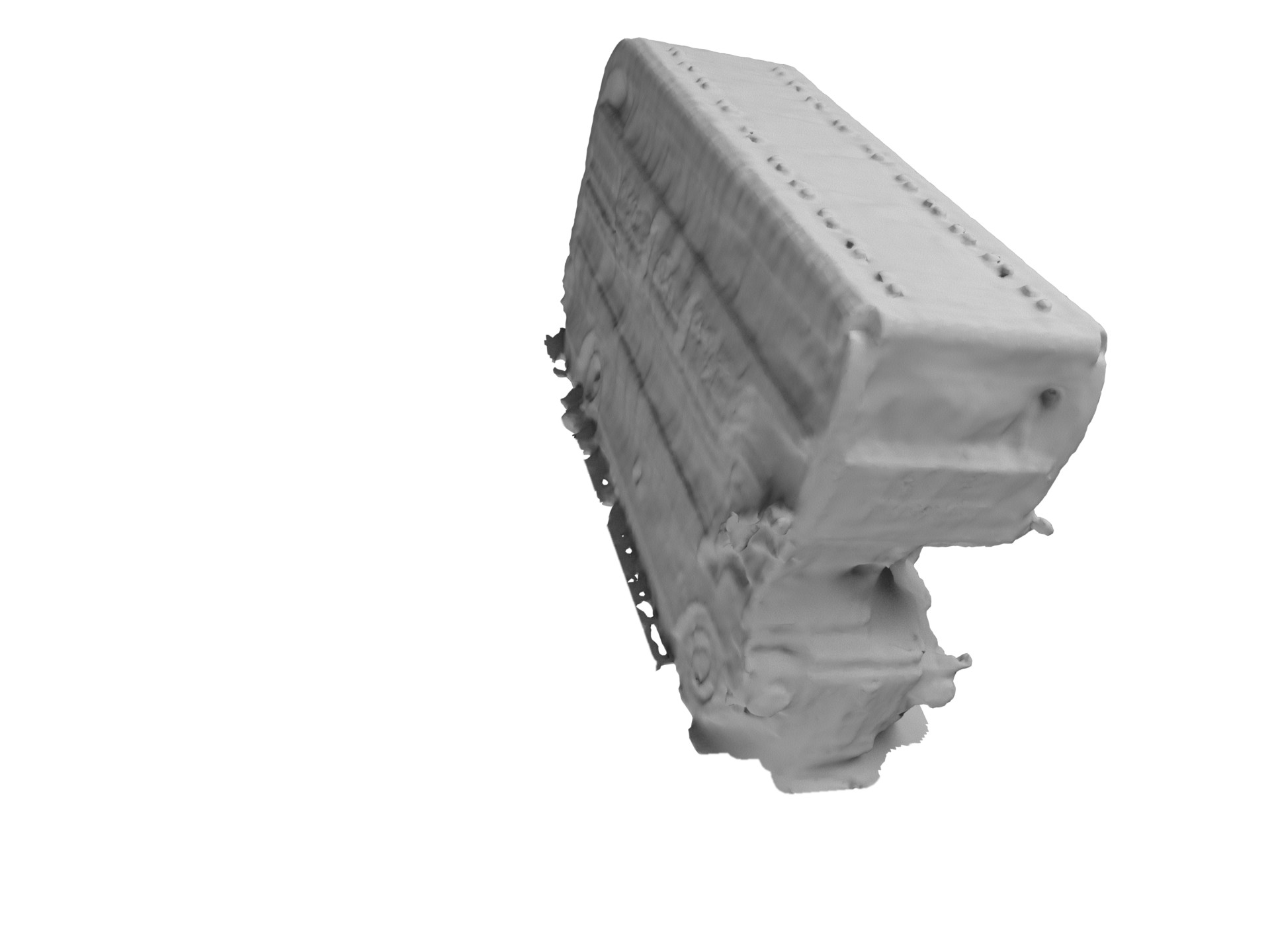}
\end{minipage}
\begin{minipage}[b]{0.245\linewidth}
\centering
\includegraphics[width=1.0\linewidth]{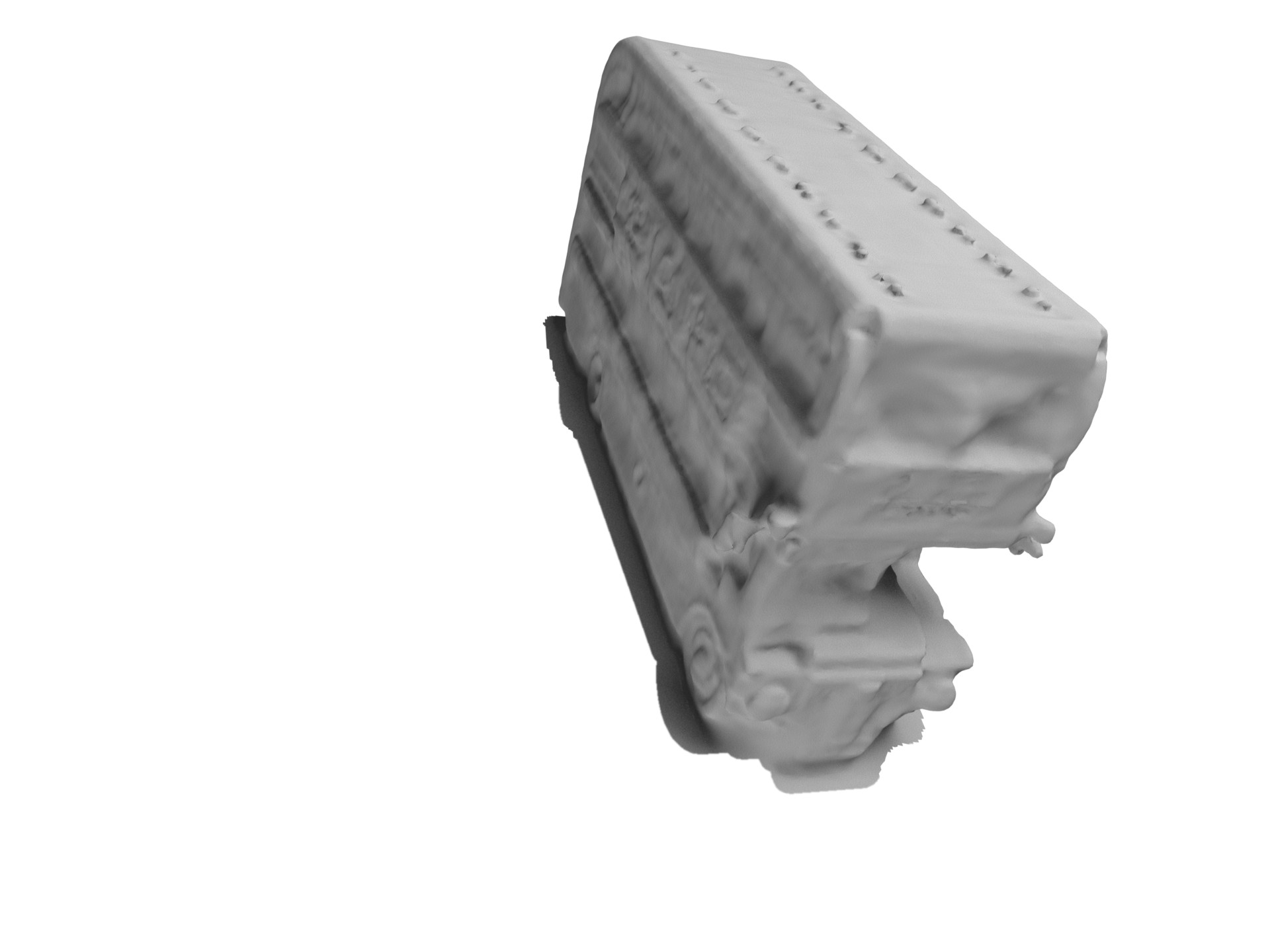}
\end{minipage}

\begin{minipage}[b]{0.245\linewidth}
\centering
\includegraphics[width=1.0\linewidth]{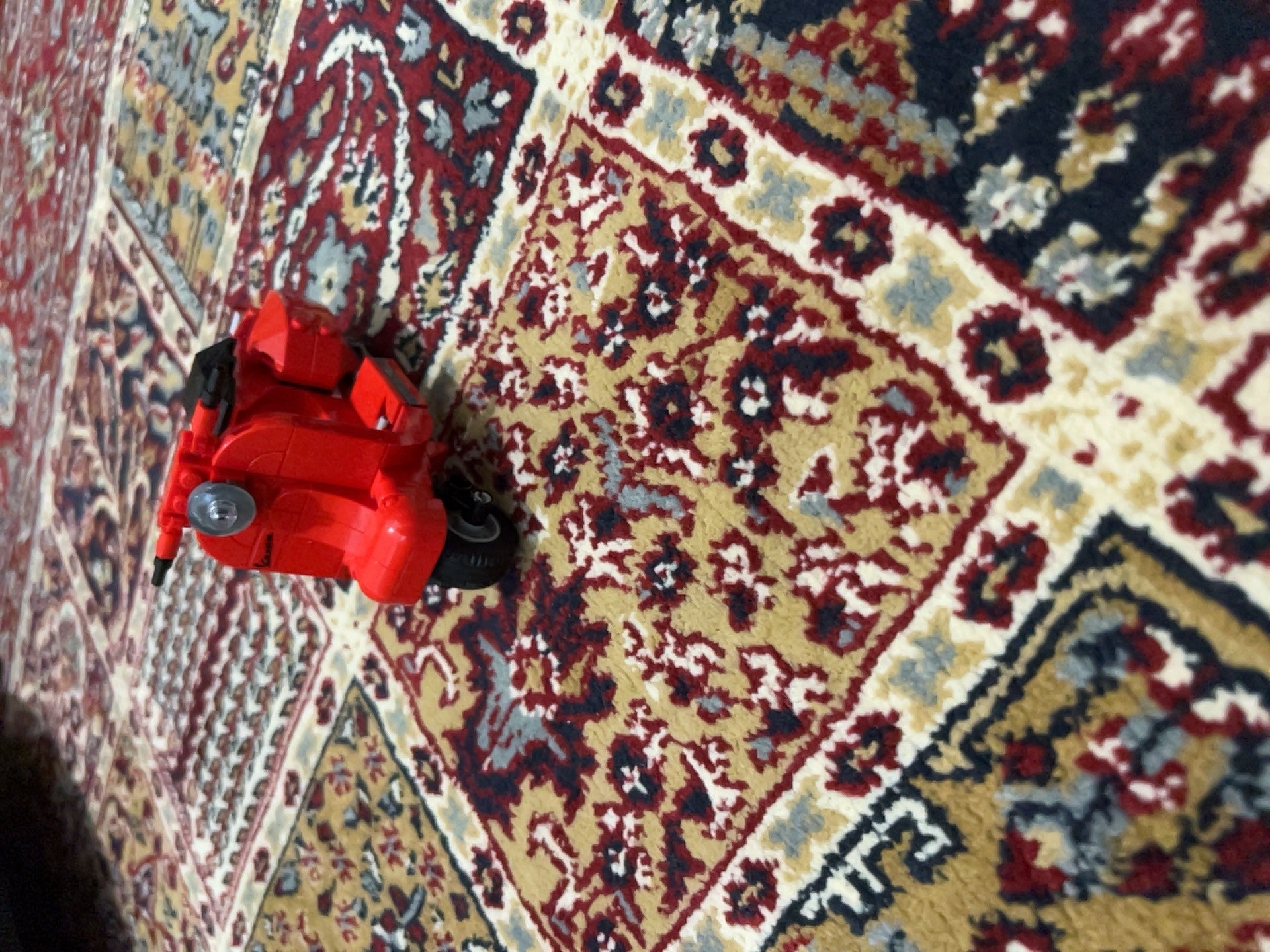}
\end{minipage}
\begin{minipage}[b]{0.245\linewidth}
\centering
\includegraphics[width=1.0\linewidth]{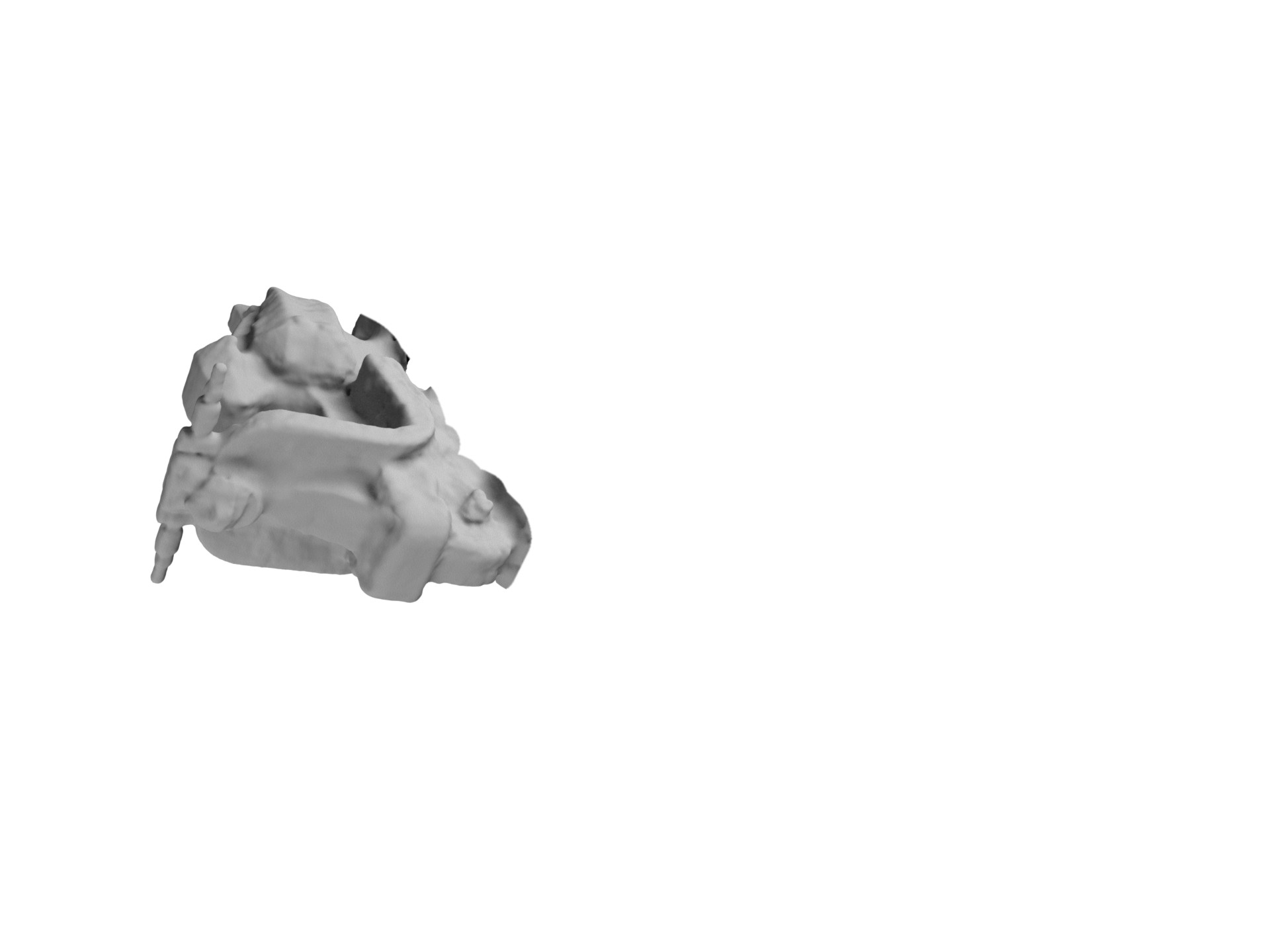}
\end{minipage}
\begin{minipage}[b]{0.245\linewidth}
\centering
\includegraphics[width=1.0\linewidth]{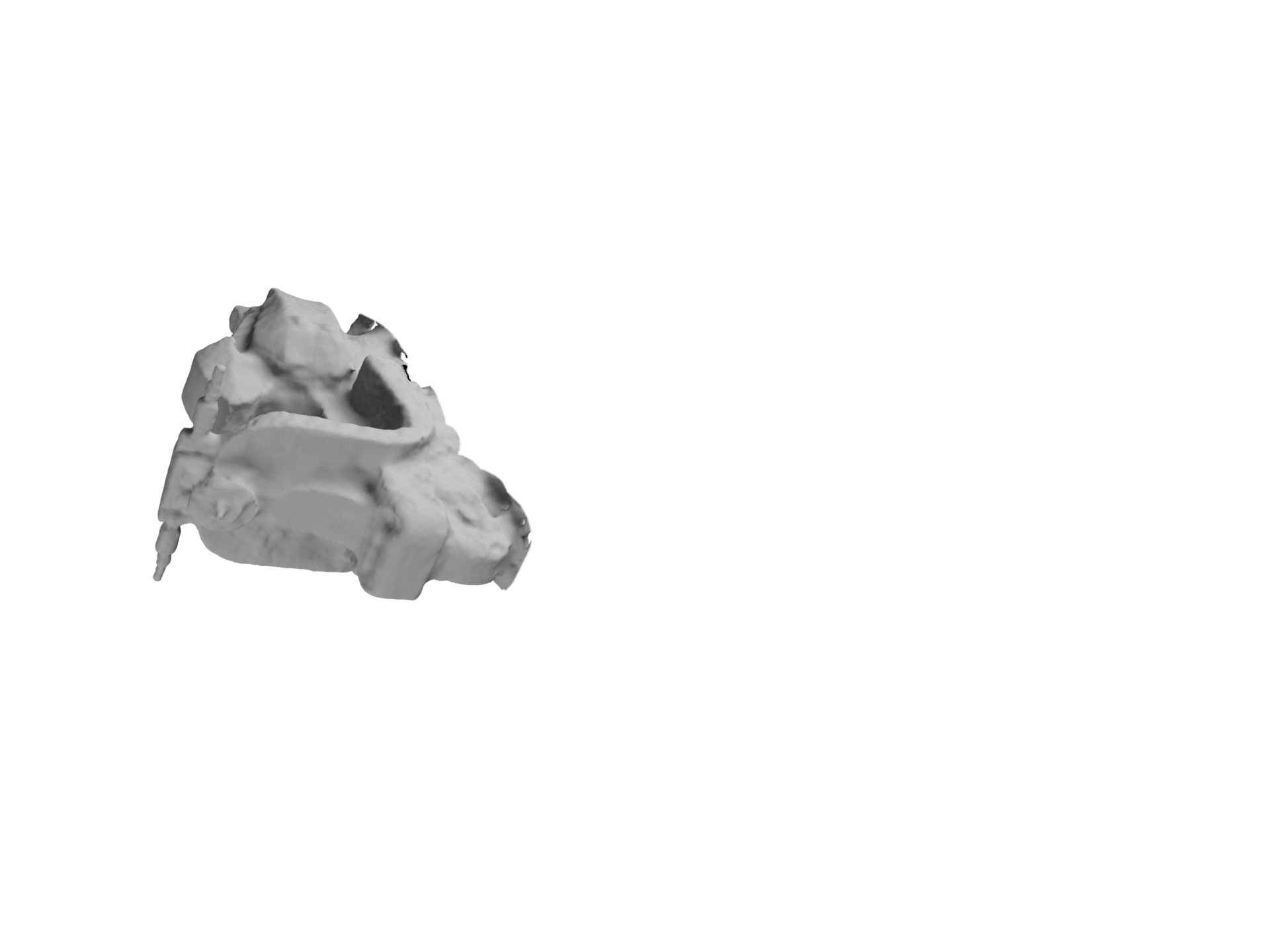}
\end{minipage}
\begin{minipage}[b]{0.245\linewidth}
\centering
\includegraphics[width=1.0\linewidth]{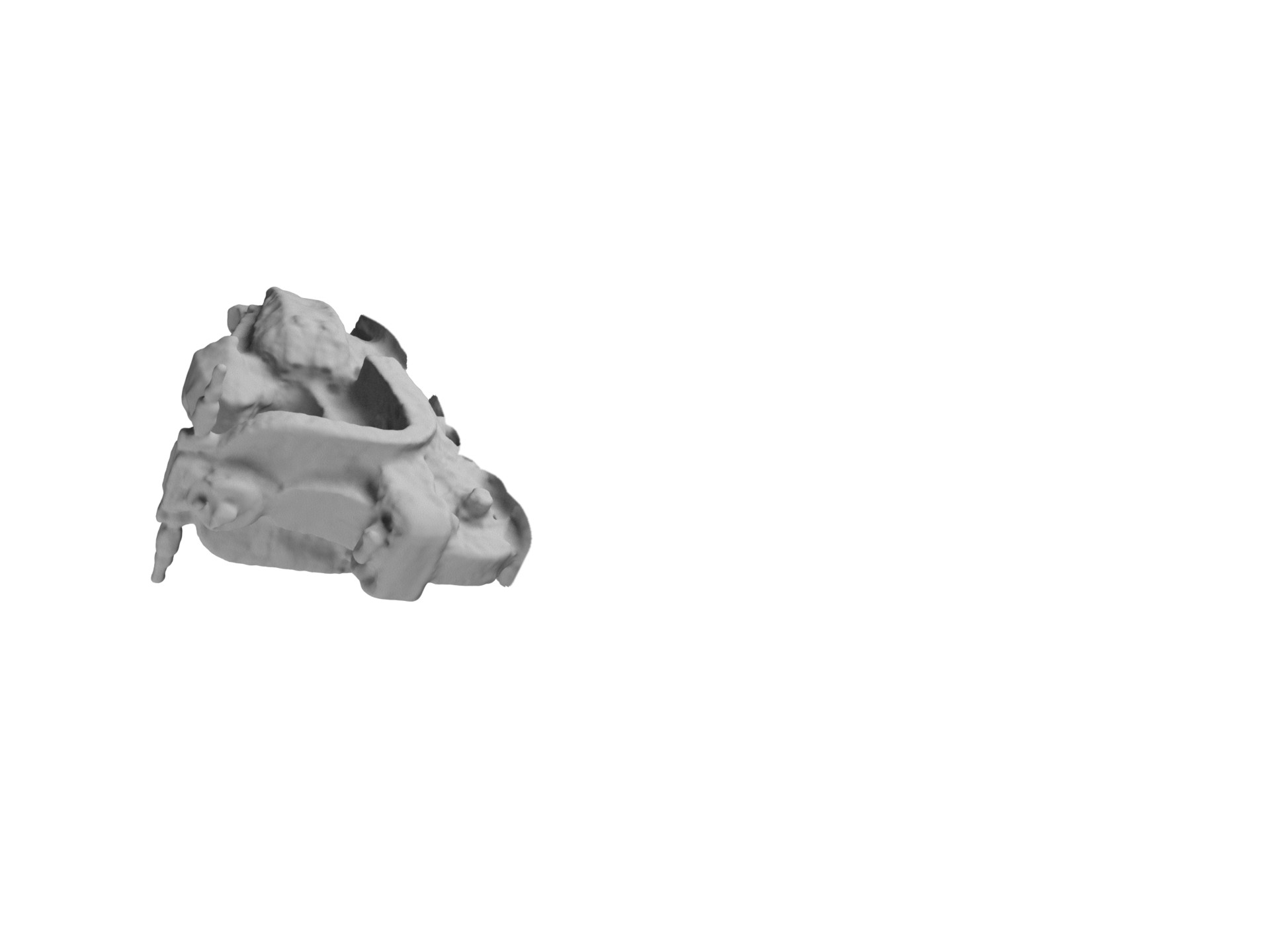}
\end{minipage}

\begin{minipage}[b]{0.245\linewidth}
\centering
\includegraphics[width=1.0\linewidth]{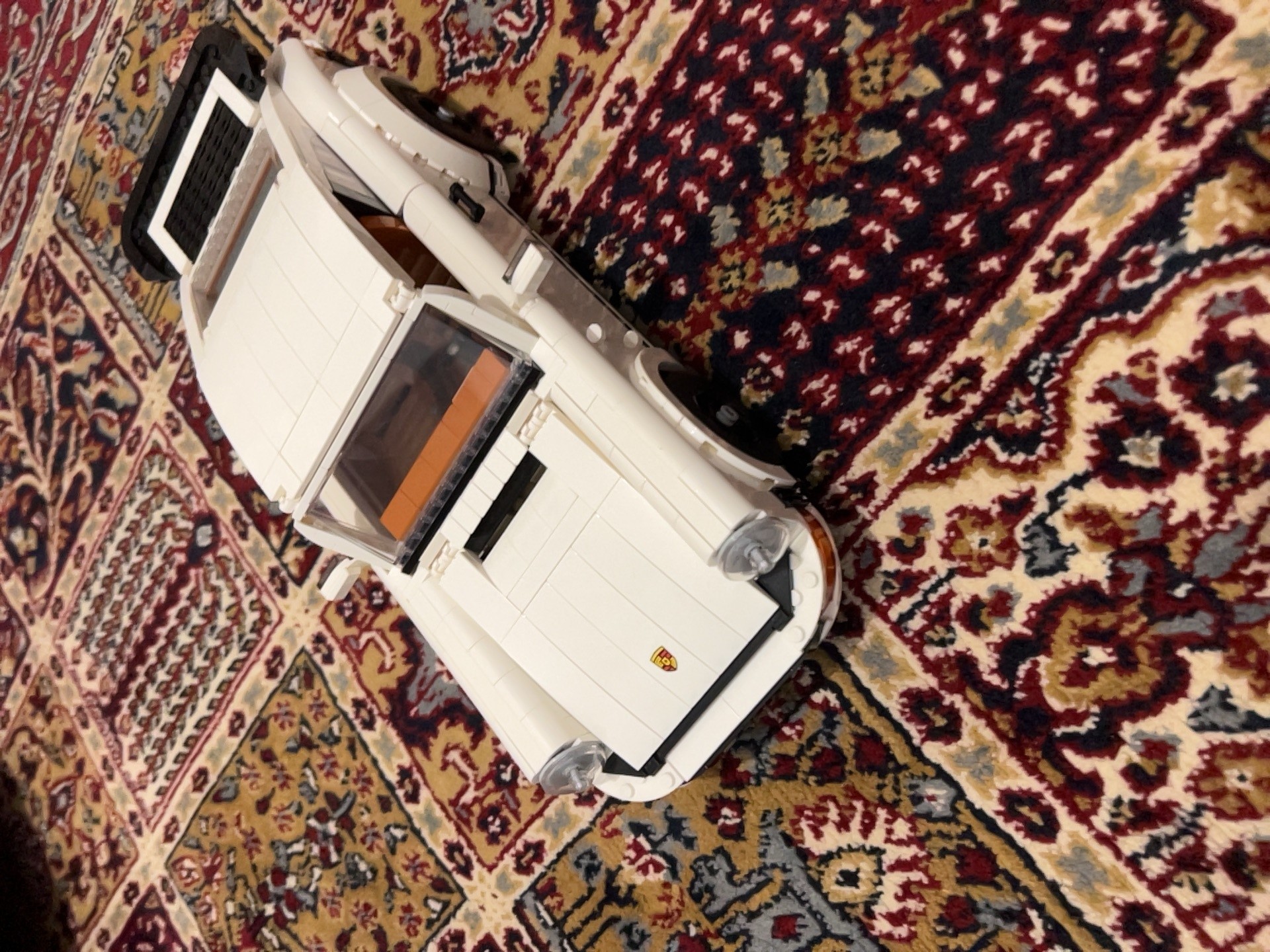}
\end{minipage}
\begin{minipage}[b]{0.245\linewidth}
\centering
\includegraphics[width=1.0\linewidth]{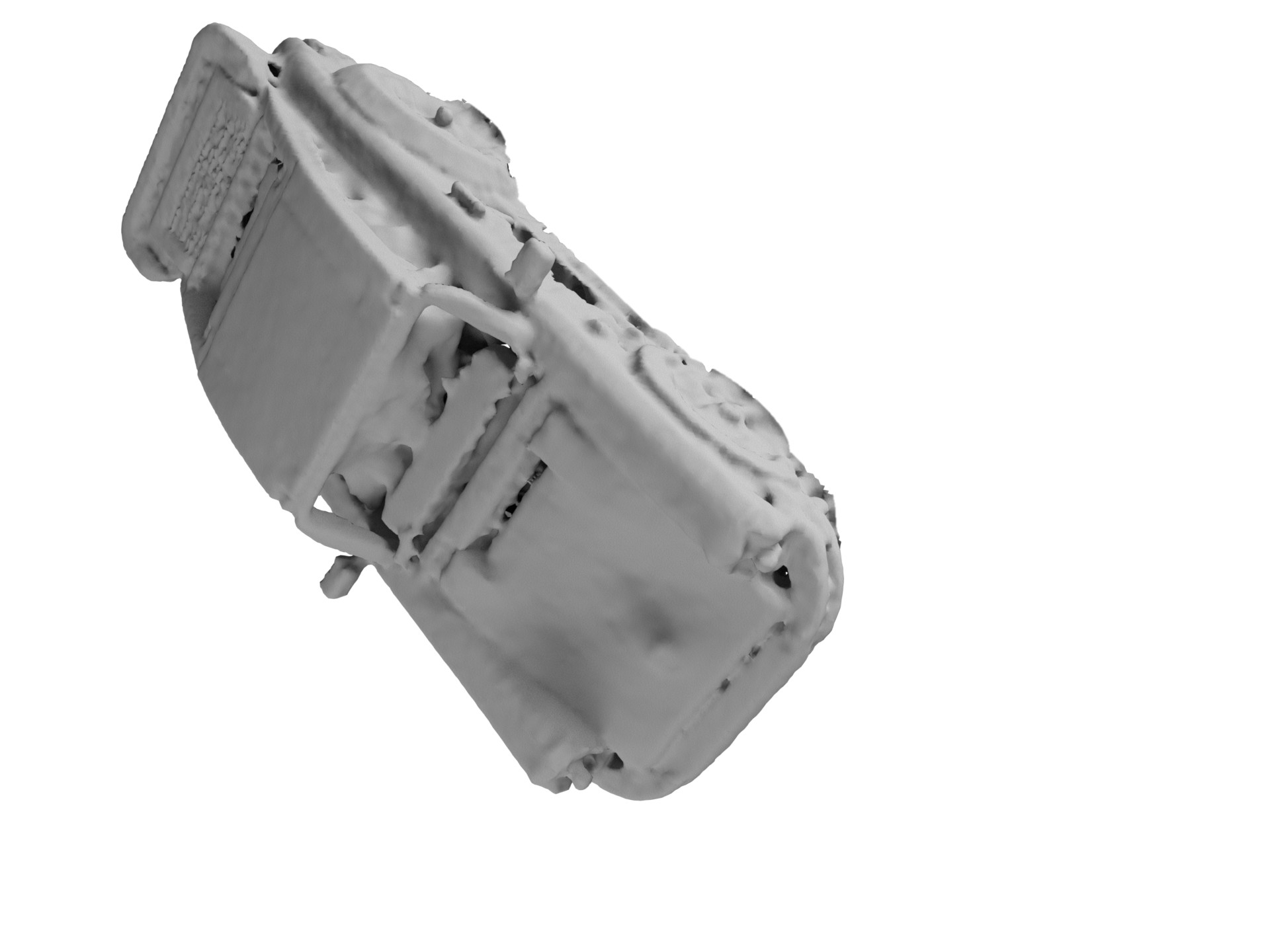}
\end{minipage}
\begin{minipage}[b]{0.245\linewidth}
\centering
\includegraphics[width=1.0\linewidth]{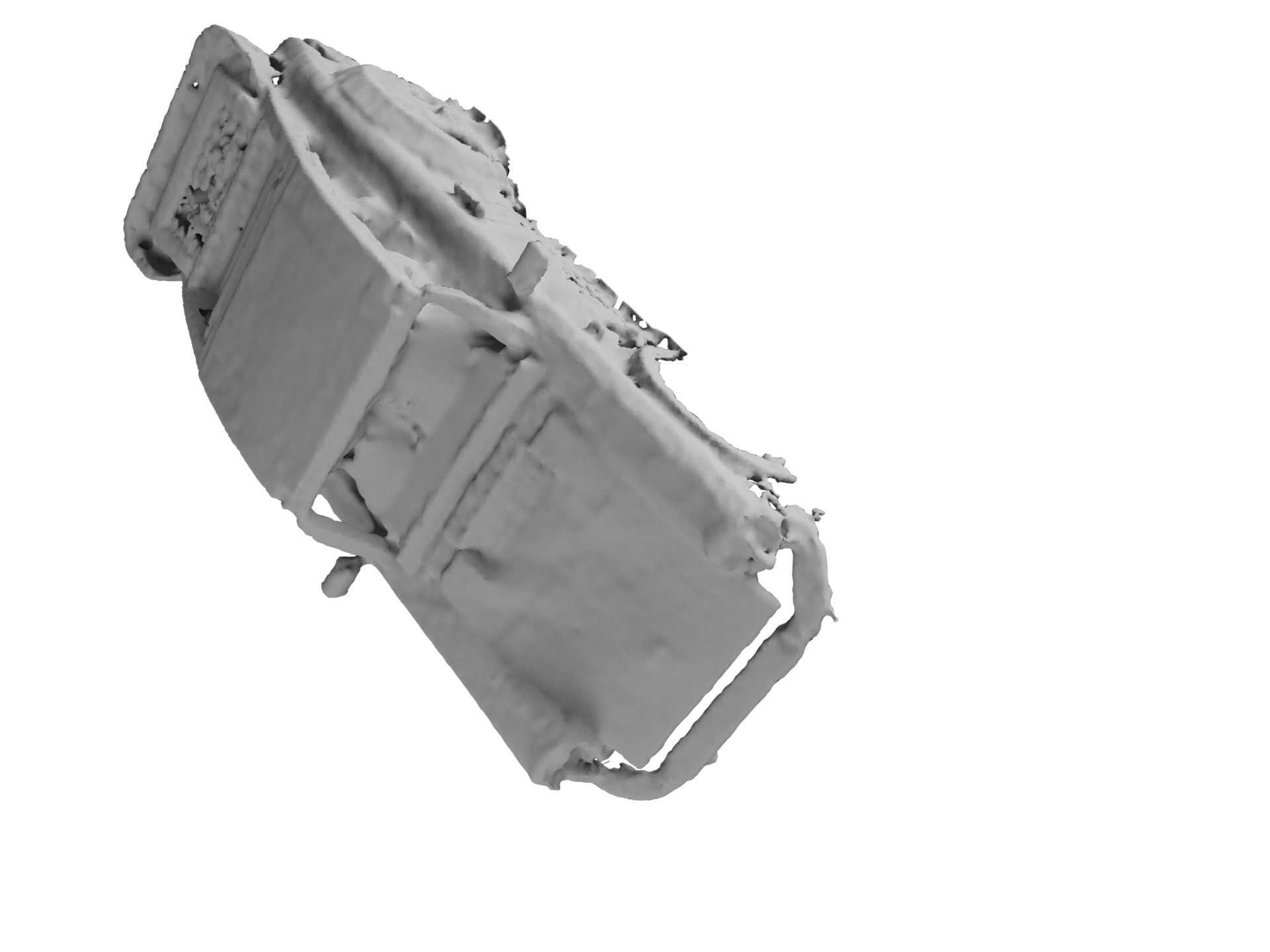}
\end{minipage}
\begin{minipage}[b]{0.245\linewidth}
\centering
\includegraphics[width=1.0\linewidth]{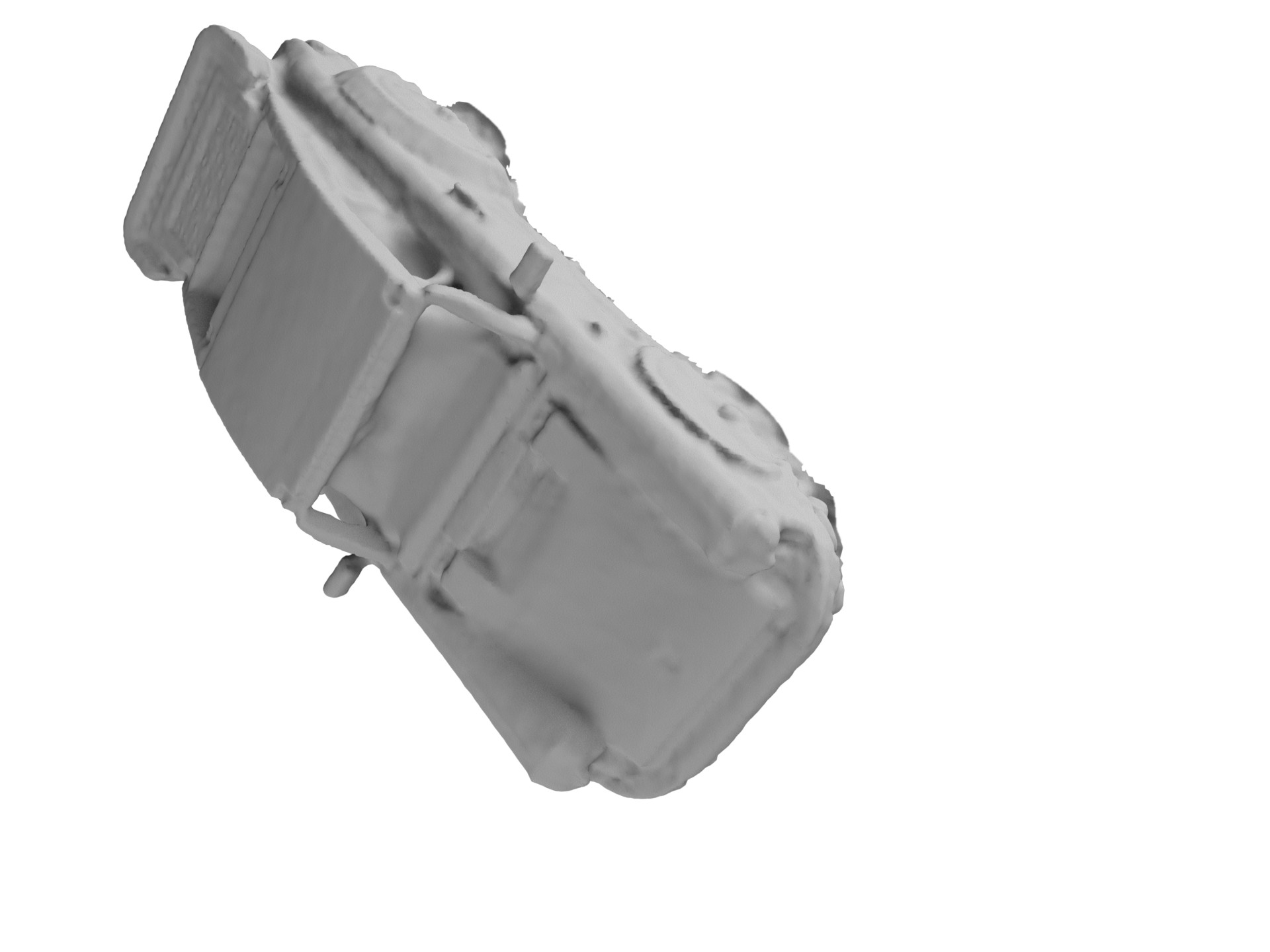}
\end{minipage}

\begin{minipage}[b]{0.245\linewidth}
\centering
\includegraphics[width=1.0\linewidth]{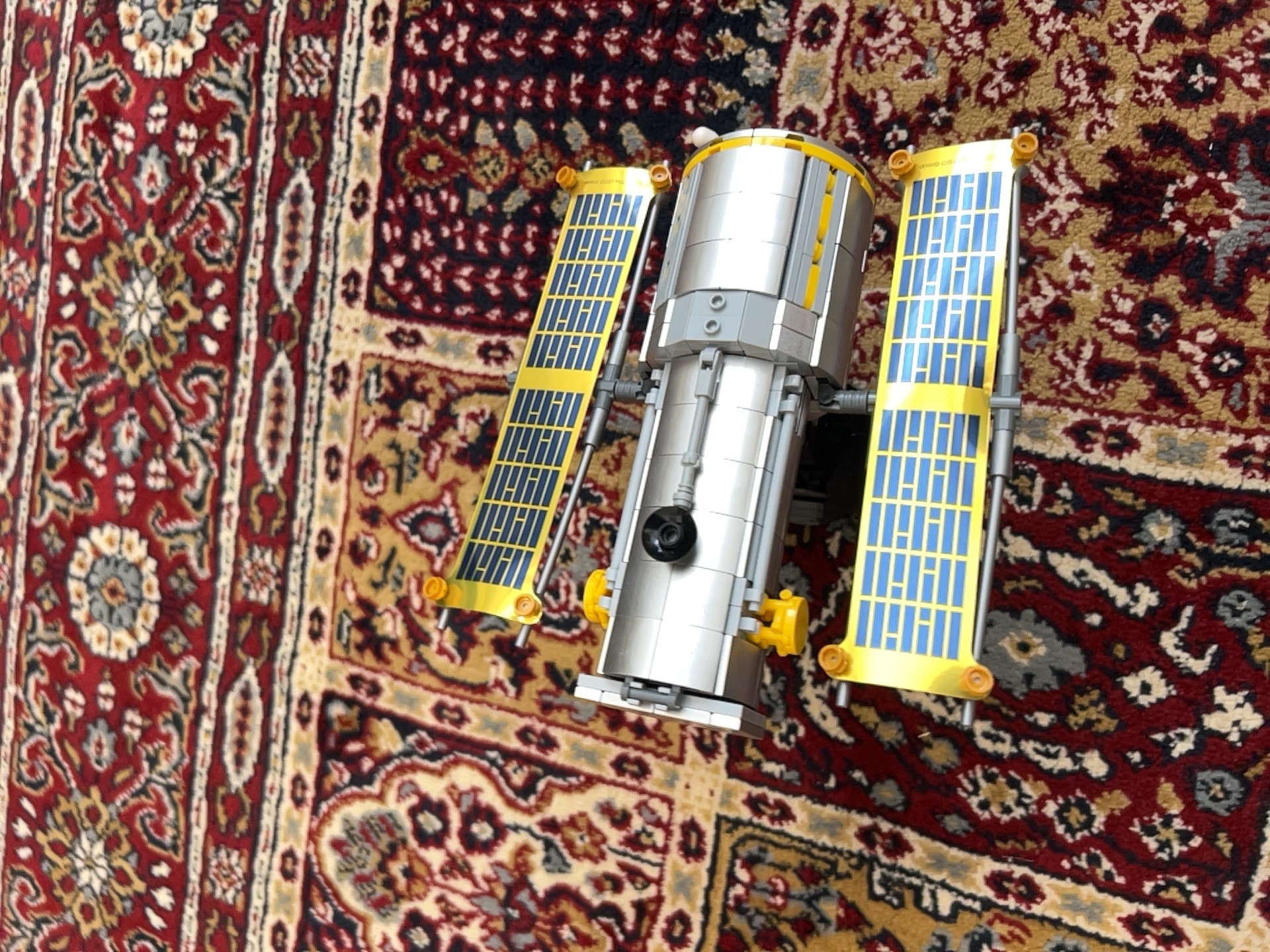}
\end{minipage}
\begin{minipage}[b]{0.245\linewidth}
\centering
\includegraphics[width=1.0\linewidth]{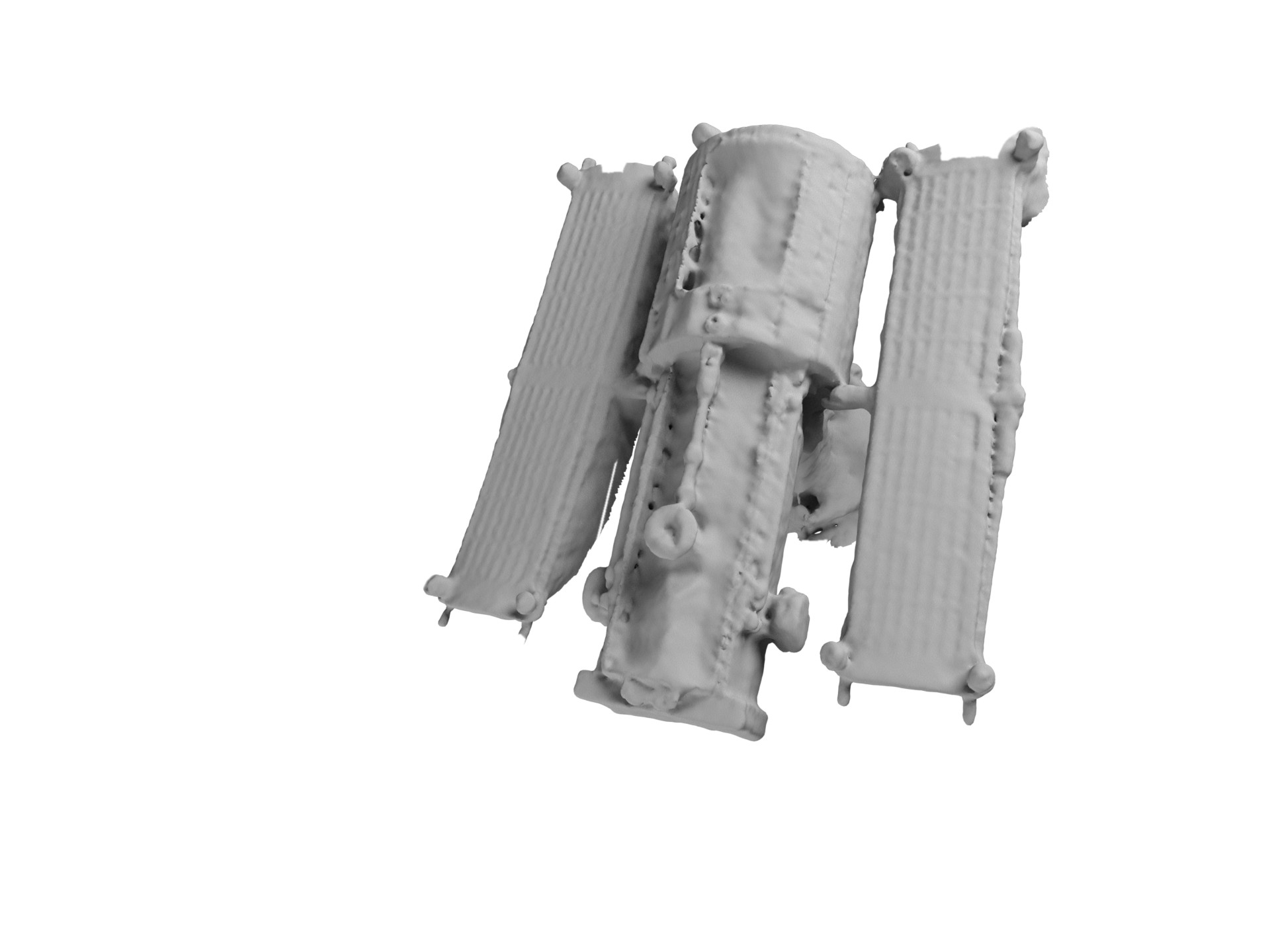}
\end{minipage}
\begin{minipage}[b]{0.245\linewidth}
\centering
\includegraphics[width=1.0\linewidth]{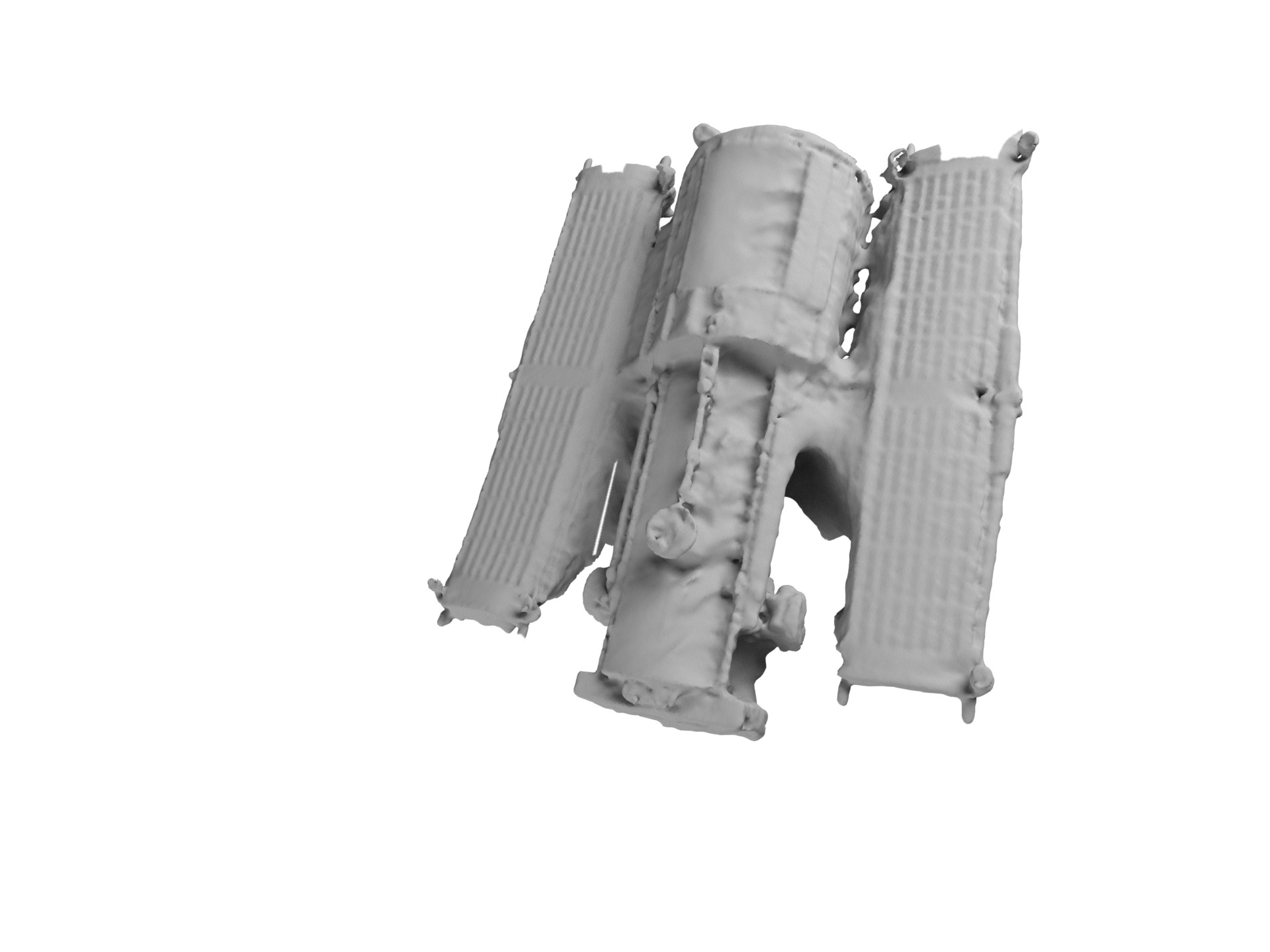}
\end{minipage}
\begin{minipage}[b]{0.245\linewidth}
\centering
\includegraphics[width=1.0\linewidth]{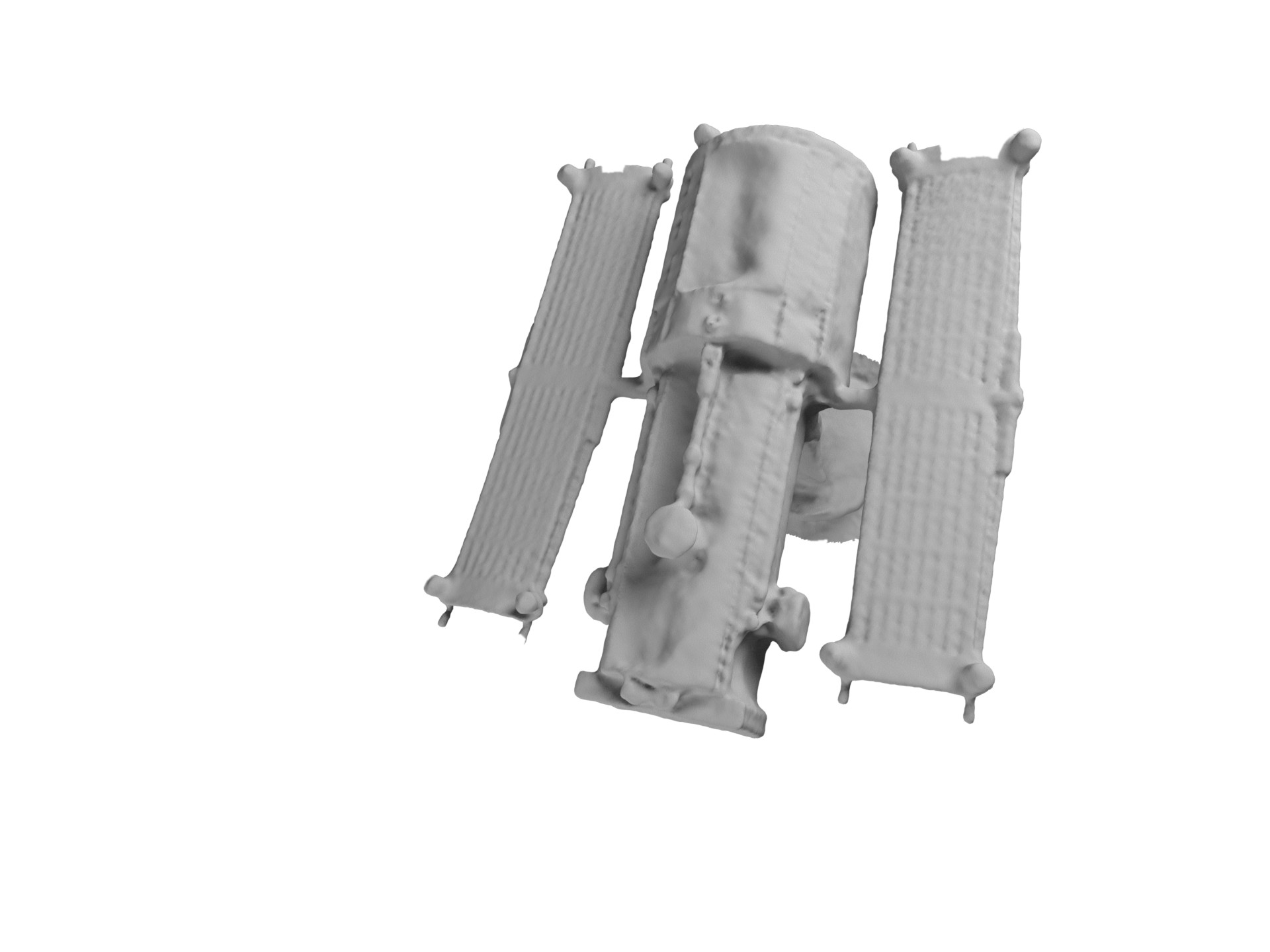}
\end{minipage}

\begin{minipage}[b]{0.245\linewidth}
\centering
\includegraphics[width=1.0\linewidth]{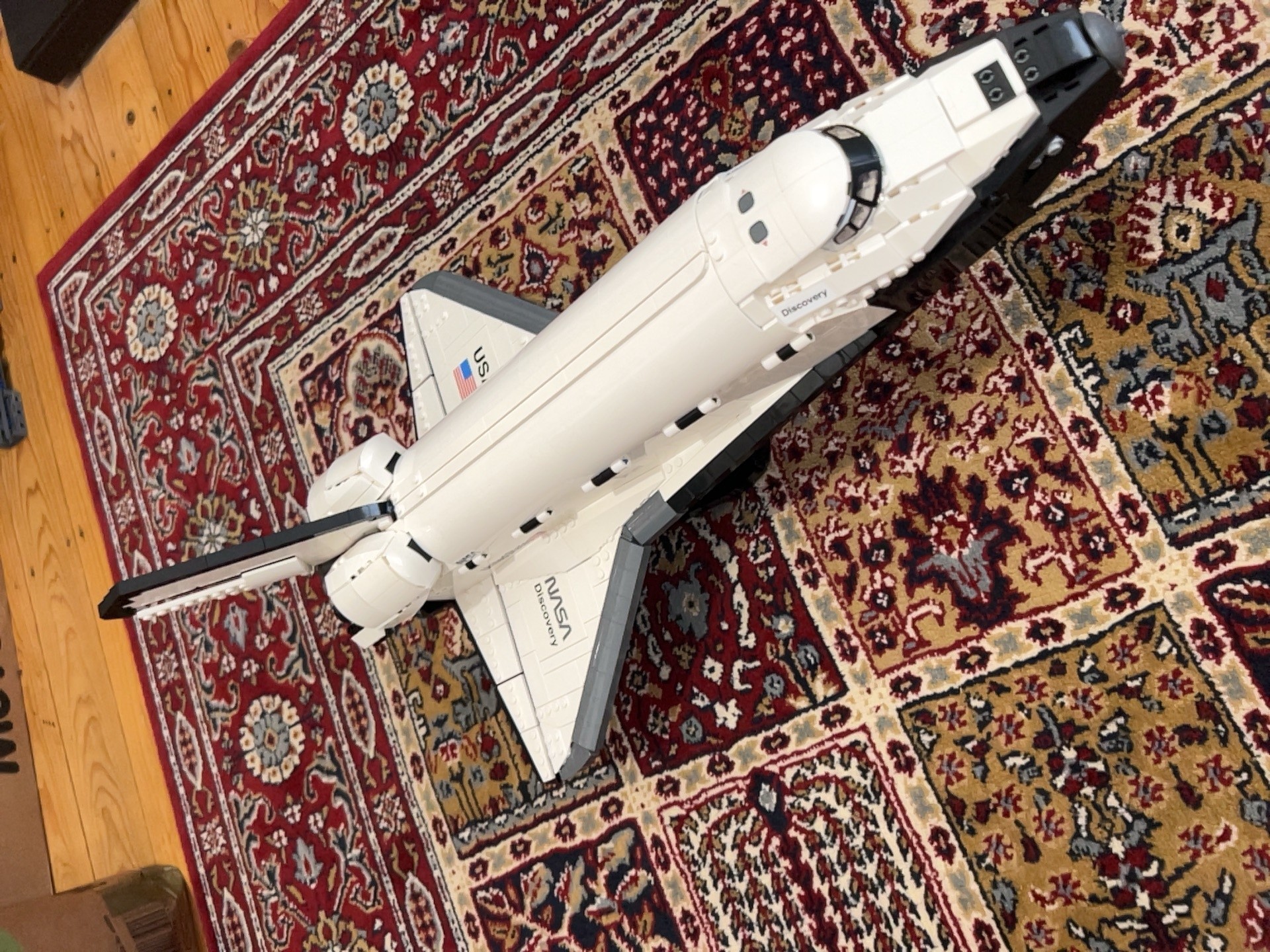}
\subcaption{Ground Truth}
\end{minipage}
\begin{minipage}[b]{0.245\linewidth}
\centering
\includegraphics[width=1.0\linewidth]{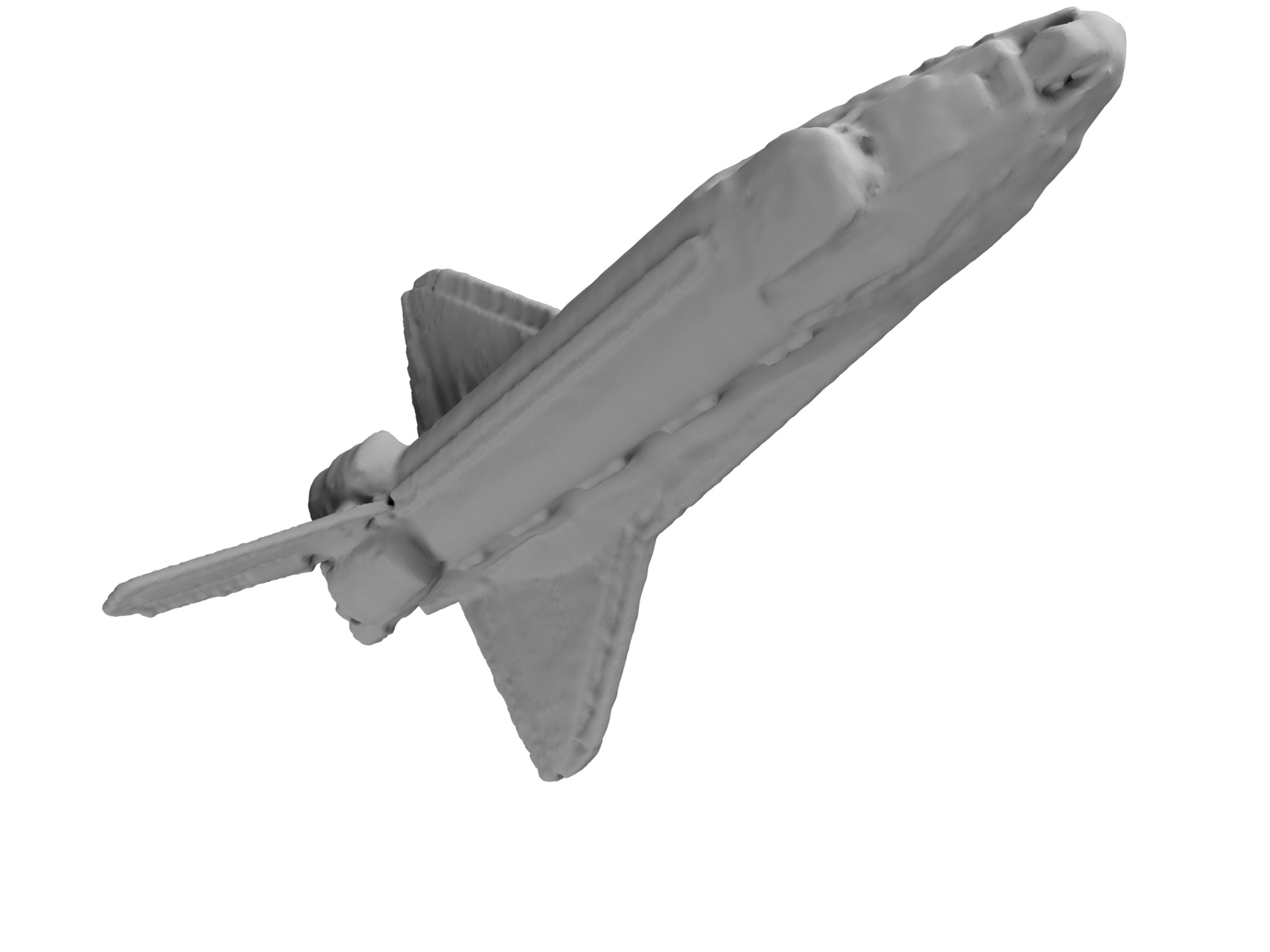}
\subcaption{NeuS-Facto}
\end{minipage}
\begin{minipage}[b]{0.245\linewidth}
\centering
\includegraphics[width=1.0\linewidth]{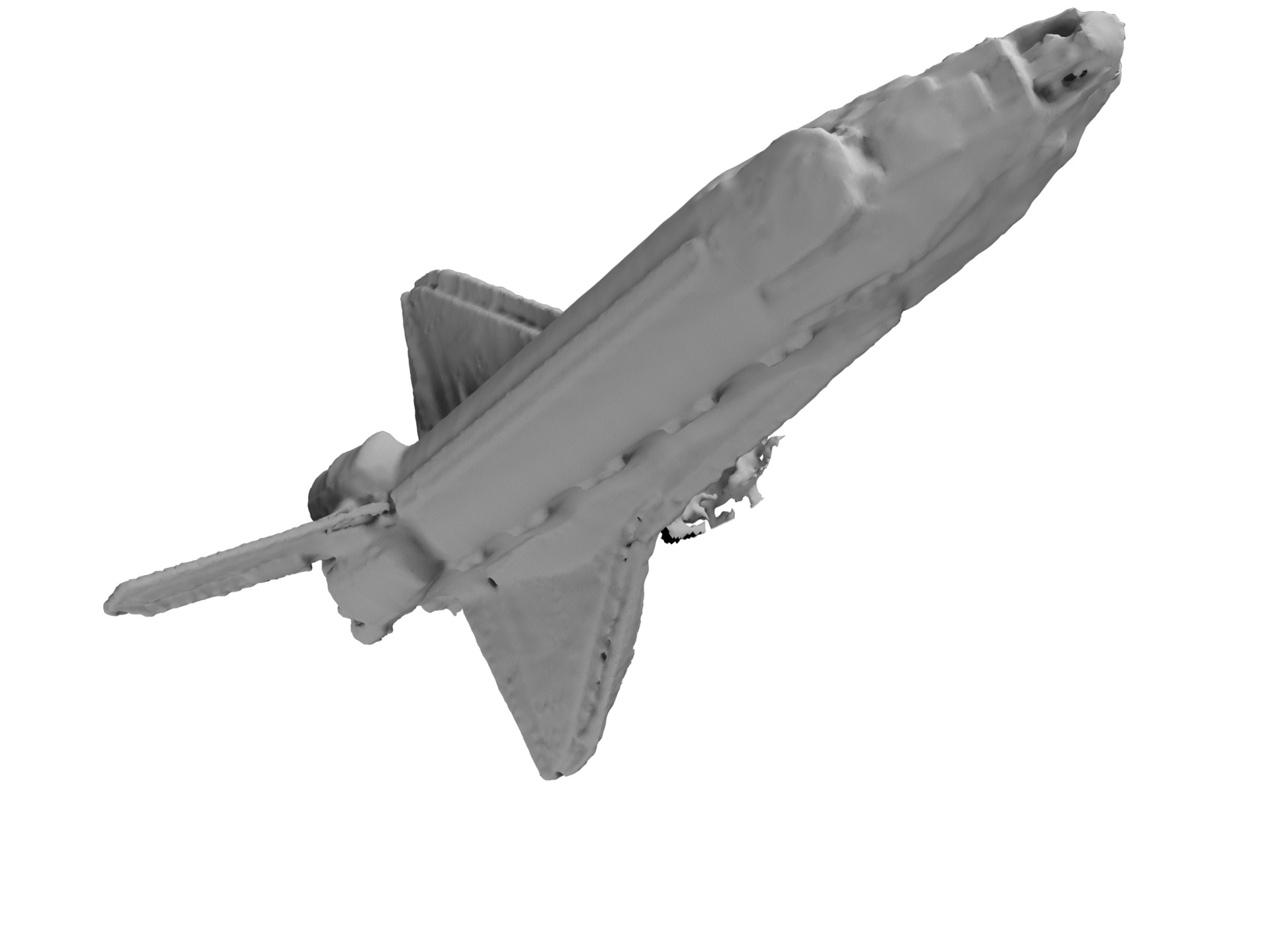}
\subcaption{OaV-Facto}
\end{minipage}
\begin{minipage}[b]{0.245\linewidth}
\centering
\includegraphics[width=1.0\linewidth]{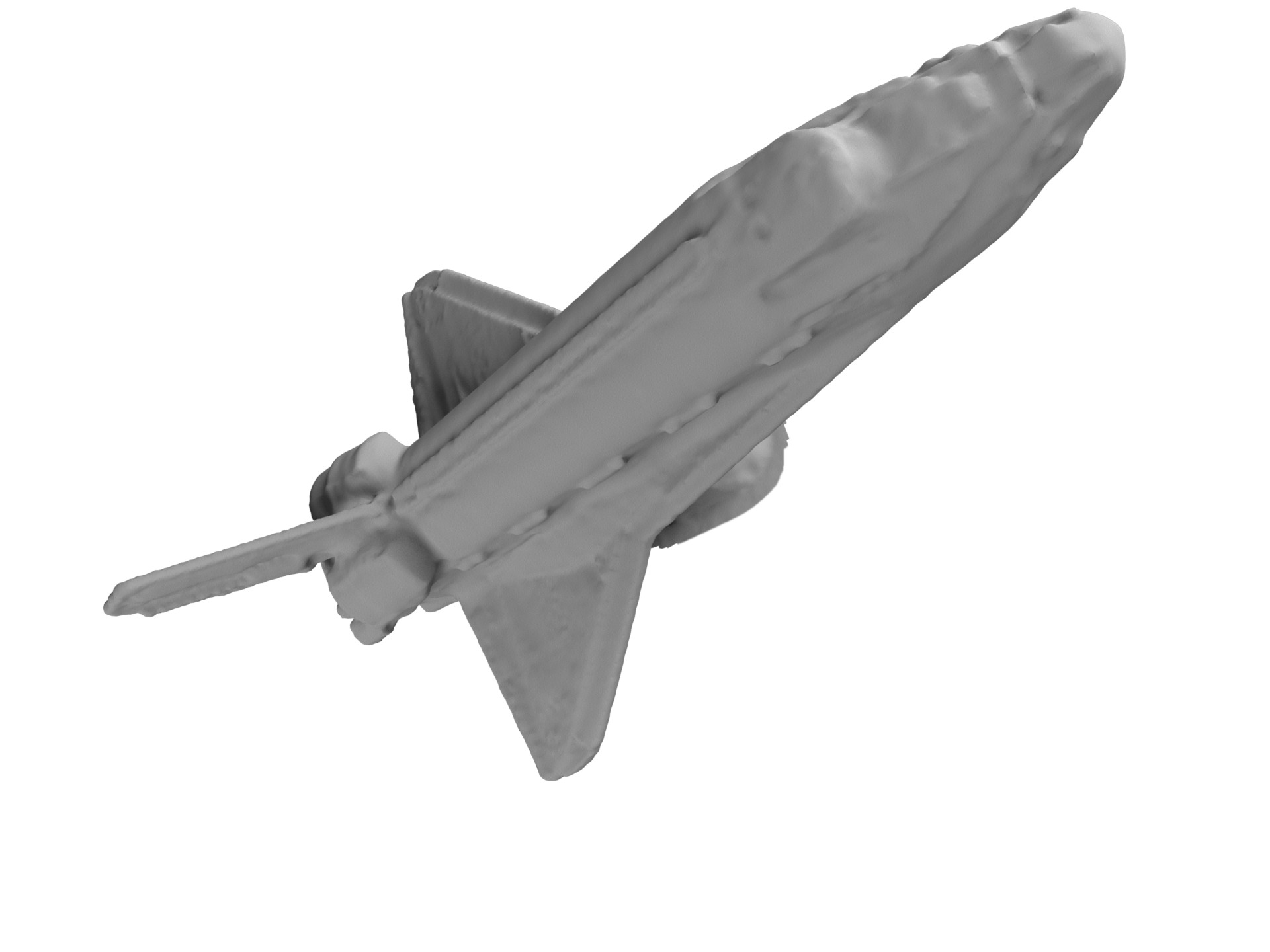}
\subcaption{SSDP-Facto}
\end{minipage}

\caption{Visualization examples on the MobileBrick dataset.}
\end{figure}

\end{document}